\documentclass[11pt, a4paper, copyright, gdm]{google}

\usepackage[authoryear, sort&compress, round]{natbib}
\usepackage{multirow}
\usepackage{graphicx}
\usepackage[utf8]{inputenc}
\usepackage{booktabs}
\usepackage{amsmath}
\usepackage{amssymb}
\usepackage{geometry}
\usepackage[table]{xcolor}
\usepackage{array} %
\usepackage{caption}
\usepackage{subcaption}
\usepackage{placeins}
\usepackage{makecell}
\usepackage{float}
\usepackage{cleveref}
\usepackage{xfrac}
\usepackage{scalerel,xparse}
\usepackage{hyperref}
\usepackage{lineno}

\hypersetup{
    colorlinks=true,
    linkcolor=blue,
    filecolor=magenta,      
    urlcolor=blue,
    pdftitle={Overleaf Example},
    pdfpagemode=FullScreen,
}

\definecolor{win_green}{RGB}{0, 128, 0}

\uselogo{} 

\title{Image Generators are Generalist Vision Learners}

\reportnumber{0001} %

\author[$\star$]{Valentin Gabeur}
\author[$\star$]{Shangbang Long}
\author[$\star$]{Songyou Peng}
\author[$\nabla$]{Paul Voigtlaender}
\author[$\nabla$]{Shuyang Sun}
\author[$\nabla$]{Yanan Bao}
\author[$\nabla$]{Karen Truong}
\author[$\nabla$]{Zhicheng Wang}
\author[$\nabla$]{Wenlei Zhou}
\author[$\nabla$]{Jonathan T. Barron}
\author[$\nabla$]{Kyle Genova}
\author[$\nabla$]{Nithish Kannen}
\author[$\nabla$]{Sherry Ben}
\author[$\nabla$]{Yandong Li}
\author[$\nabla$]{Mandy Guo}
\author[$\nabla$]{Suhas Yogin}
\author[$\dagger$]{Yiming Gu}
\author[$\dagger$]{Huizhong Chen}
\author[$\ddagger$]{Oliver Wang}
\author[$\ddagger$]{Saining Xie}
\author[$\ddagger$]{Howard Zhou}
\author[$\ddagger$]{Kaiming He}
\author[$\ddagger$]{Thomas Funkhouser}
\author[$\ddagger$]{Jean-Baptiste Alayrac}
\author[$\ddagger$]{Radu Soricut}

\affil[$\star$]{project leads and equal contributions}
\affil[$\nabla$]{core contributors}
\affil[$\dagger$]{project advisors}
\affil[$\ddagger$]{leadership sponsors}

\correspondingauthor{vision-banana@google.com}

\begin{abstract}

Recent works show that image and video generators exhibit zero-shot visual understanding behaviors, in a way reminiscent of how Large Language Models (LLMs) develop emergent capabilities of language understanding and reasoning from generative pretraining. 
While it has long been conjectured that the ability to create visual content implies an ability to understand it, 
there has been limited work demonstrating that generalist image generators can achieve state-of-the-art understanding capabilities on many different vision tasks.   
In this work, we demonstrate that image generation training serves a role similar to LLM pretraining, and lets models learn powerful and general visual representations that enable \textbf{state-of-the-art} performance on various vision tasks.
We introduce \textit{Vision Banana}, a generalist model built by instruction-tuning \textit{Nano Banana Pro} (NBP) on a mixture of its original training data alongside a small amount of vision task data.
By parameterizing the output space of vision tasks as RGB images, we seamlessly reframe perception as image generation.
Our generalist model, \textit{Vision Banana}, achieves state-of-the-art results on a variety of vision tasks involving both 2D and 3D understanding, beating or rivaling zero-shot domain-specialists, including \textit{Segment Anything Model 3} on segmentation tasks, and the \textit{Depth Anything} series on metric depth estimation.
We show that these results can be achieved with lightweight instruction-tuning without sacrificing the base model's image generation capabilities.
The superior results suggest that image generation pretraining is a generalist vision learner. 
It also shows that image generation serves as a unified and universal interface for vision tasks, similar to text generation's role in language understanding and reasoning.
We could be witnessing a major paradigm shift for computer vision, where generative vision pretraining takes a central role in building Foundational Vision Models for both generation and understanding.

\vspace{3mm}
Project Page: \href{https://vision-banana.github.io}{vision-banana.github.io}
\end{abstract}

\begin{document}

\maketitle

\newcommand{\emojivb}{\raisebox{-0.2ex}{\includegraphics[height=1.2em]{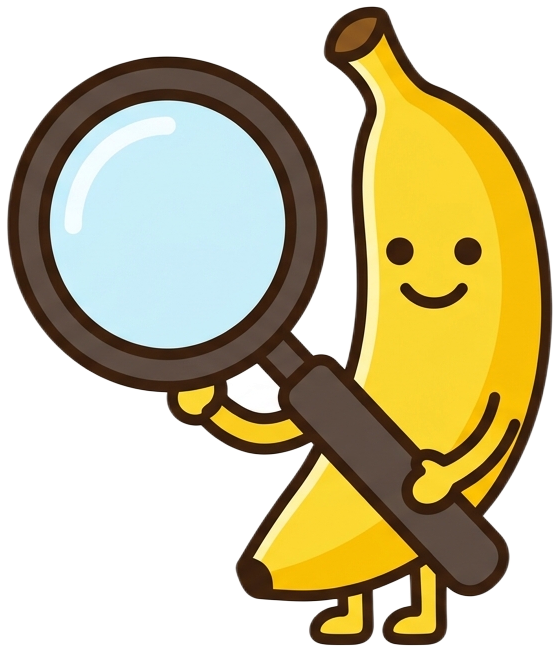}}}

 \section{Introduction}

In recent years, advanced image and video generation models \citep{nbp,veo3,flux2,seedance2,uni1,gpt1p5} have demonstrated unprecedented generation capabilities, synthesizing highly complex, high-fidelity visual context with precise semantic control. This remarkable capability for visual creation suggests that these models possess a deep, internalized comprehension of the visual world's underlying structures, semantics, and relationships. 
However, leading methods on visual representation learning in general do not belong to the family of generative modeling.
Instead, they include supervised discriminative learning \citep{krizhevsky2012imagenet_alexnet,dehghani2023scaling_vit22b,dosovitskiy2020image_vit}, 
contrastive learning  \citep{simclr,he2020momentum_moco,chen2020improved_moco2,zhai2023siglip1,tschannen2025siglip2,radford2021learning_clip},
bootstrapping \citep{caron2021emerging_dinov1,grill2020bootstrap_byol}, 
auto-encoding \citep{he2022masked_mae,bao2021beit,chen2024deconstructing_dae} among others, and their combinations \citep{oquab2023dinov2,simeoni2025dinov3,zhou2021ibot,cao2026tipsv2}.
Early efforts in generative vision pretraining \citep{chen2020generative_igpt, bai2024sequential_lvm} have shown promising scaling behaviors but their effectiveness has lagged behind non-generative models.

\newcommand{\ie}{i.e.,\ }

In this paper, we investigate \textit{whether visual generative models are secretly generalist vision learners}, \ie whether models trained for image generation develop internal representations that are suitable for visual understanding tasks. 
To achieve this, we finetune a pretrained image generator with a small amount of computer vision data (depth estimation, surface normal estimation, segmentation, etc.).
We then evaluate the resulting model on a wide variety of vision benchmarks.
If the finetuned model performs at or near SOTA on these benchmarks, while retaining its image generation capabilities, then there is strong evidence that the image generator was indeed a foundation model for visual understanding -- i.e., a generalist vision learner.

This is not the first paper to study the hidden understanding capabilities of generative models or use image and video generators as base models for visual understanding.
Early research efforts show that generative models develop some understanding capabilities hidden in their features \citep{bhattad2023stylegan,du2023generative,li2023your,ranzato2011deep,hjelm2018learning,clark2023text,baranchuk2021label,chen2016infogan,zhao2023unleashing,mukhopadhyay2023diffusion,tang2023emergent,zhang2023tale,li2024sd4match,hedlin2023unsupervised,yang2023diffusion}.
More recent research observes that state-of-the-art image and video generators can generate visual content that look like RGB visualizations of computer vision outputs for tasks such as segmentation, depth estimation, and surface normal estimation \citep{zuo2025nanobanana_low_level,wiedemer2025veo_learner}.  However, those methods do not provide state-of-the-art results on modern benchmarks.
This is partially because these models do not strictly follow the prompts to produce vision outputs in the desired formats that can be decoded back to vision outputs for computing quantitative metrics.
Other researchers \citep{he2024lotus,lotus2,marigold,ye2024stablenormal,yu2024high,zhao2025diception,wang2026thfm,wu2025qwen_image,garcia2025fine,xu2023odise} adapt the generation architectures by adding specialized modules and performing full-finetuning to achieve SOTA-level results on specific target tasks.
Although these methods successfully leverage the understanding capabilities of the pre-trained features, they sacrifice the model's generality across other understanding and generation tasks.

\begin{figure}[!t]
    \centering
    \includegraphics[width=\textwidth]{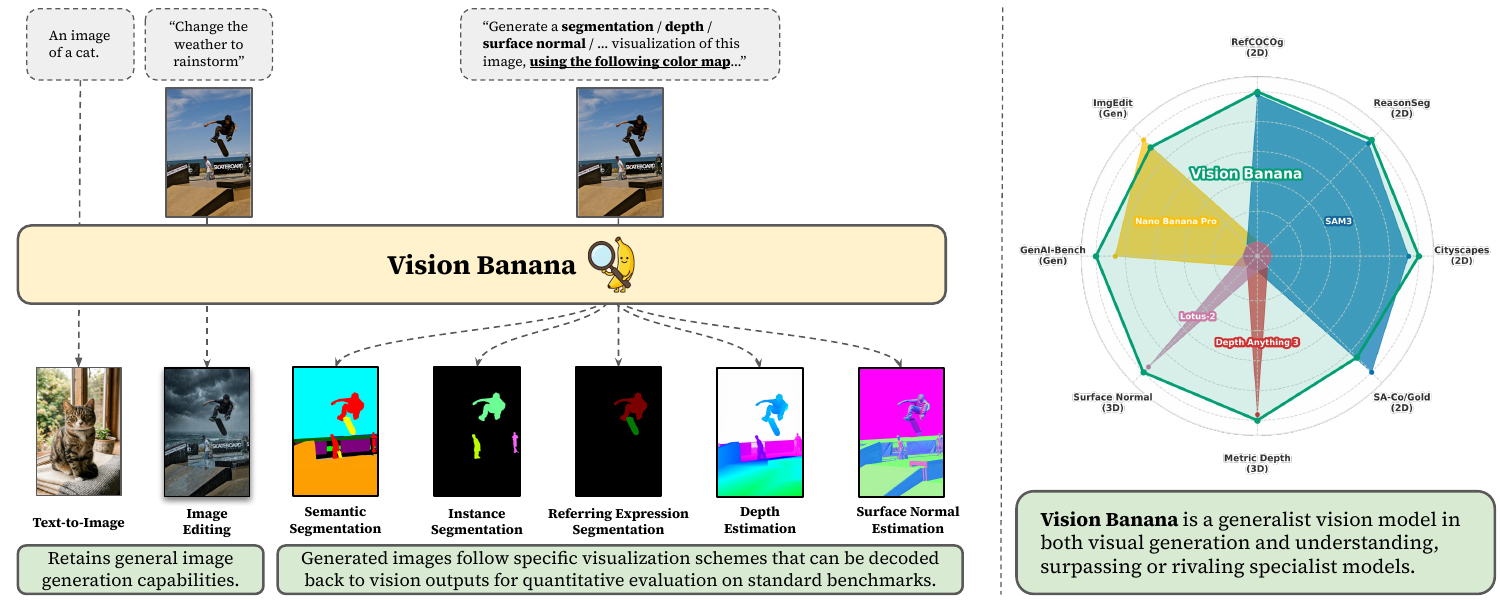}
    \vspace{1mm}
    \caption{We demonstrate the hidden visual understanding capabilities of image generators by instruction-tuning Nano Banana Pro.
    The instruction-tuned model, \textbf{Vision Banana} \emojivb, can produce visualizations in a precise format that can enable evaluation on established benchmarks. 
    }
    \label{fig:teaser_img}
    \vspace{6mm}
\end{figure}

We take an approach motivated by recent advancements in large language models (LLMs). In natural language processing (NLP), generative pretraining \citep{brown2020language_gpt3,chowdhery2023palm} is performed to produce base models, often referred to as LLMs, that are good at generating text, whereas instruction-tuning \citep{ouyang2022training_instructgpt,wei2021finetuned_flan} guides them to follow specific tasks and produce text in requested formats and stay on the task.
Analogously, we position a visual generative model as a ``base'' model and perform instruction-tuning to align the model to produce visual output in desired formats, in accordance with the prompts, as illustrated in Fig. \ref{fig:teaser_img}.
Specifically, the model is instructed to produce RGB images that can be decoded to computer vision outputs. 
Such instruction prompts and decodable visualization schemes are designed to bridge and calibrate the visual generations to formats where measurable metrics for benchmarking can be applied.
For example, by prompting the model to ``\textit{Segment the skateboard category in pure yellow (<255, 255, 0>)}'', we can easily parse the mask for \textit{skateboard} by clustering pixels whose values are close to \textit{<255, 255, 0>}.
This strategy has three main advantages.
First, it supports a wide variety of tasks with a single unified model -- after instruction tuning, the weights are shared among all tasks, and only the prompt changes.   
Second, it requires relatively little new training data, since the instruction tuning is solely teaching the model how to format computer vision outputs as RGB.
Third, it helps the model retain its original image generation capabilities, since the outputs are simply new RGB images.

\begin{table}[t]
\centering
\resizebox{\textwidth}{!}{

\begin{tabular}{@{}l|l|c|l@{}}
\toprule
\textbf{Capabilities} & \textbf{Benchmarks and Metrics} & \textbf{Vision Banana} \emojivb & \textbf{Best Counterpart} \\ \midrule

\multirow{4}{*}{2D Understanding} & Referring segmentation: RefCOCOg UMD val (cIoU $\uparrow$) & \textbf{\color{win_green}73.8} & 73.4 (\textit{SAM3 Agent}) \\ %

 & Referring segmentation: ReasonSeg val (gIoU $\uparrow$) & \textbf{\color{win_green} 79.3} & 77.0 (\textit{SAM3 Agent}) \\  %

 & Semantic segmentation: Cityscapes val (mIoU $\uparrow$) & \textbf{\color{win_green} 69.9} & 65.2 (\textit{SAM3}) \\ %

 & Instance segmentation: SA-Co/Gold ($cgF_1$ $\uparrow$) & \textbf{\color{win_green} 47.5} & 24.6 (\textit{OWLv2}) \\ \midrule

\multirow{2}{*}{3D Understanding} & Metric depth estimation: average of 4 datasets ($\delta_1$$\uparrow$) & \textbf{\color{win_green} 0.929} & 0.918 (\textit{Depth Anything 3}) \\ %

 & Surface normal estimation: average of 4 datasets (mean angle error $\downarrow$) & \textbf{\color{win_green} 18.928} & 19.642 (\textit{Lotus-2}) \\ \midrule

 \multirow{2}{*}{Visual Generation} & Text-to-image: GenAI-Bench (win rate against the other $\uparrow$) & \textbf{\color{win_green} 53.5\%} & $46.5\%$ (\textit{Nano Banana Pro}) \\ %

  & Image editing: ImgEdit (win rate against the other $\uparrow$) & $47.8\%$ & \textbf{\color{win_green} 52.2\%} (\textit{Nano Banana Pro}) \\\bottomrule

\end{tabular}
}
\raggedright
\caption{The instruction-tuned \textbf{Vision Banana} model surpasses or rivals SOTA specialists across visual generation and understanding. 
For 2D visual understanding, it beats the highly specialized \textbf{Segment Anything Model 3} \citep{sam3} on 3 segmentation datasets, and outperforms OWLv2 \citep{minderer2023gowol} on instance segmentation.
For 3D visual understanding, it surpasses the best metric depth estimation expert, \textbf{Depth Anything 3} \citep{dav3}, and the best surface normal estimation specialist, Lotus-2 \citep{lotus2}.
In visual generation, Vision Banana inherits its capabilities from Nano Banana Pro and is on par with it on text-to-image and image editing.
}
\label{tab:overall}
\end{table}

We present \textbf{Vision Banana} \emojivb, a generalist vision model trained by performing a lightweight instruction-tuning to Nano Banana Pro on a mixture of its original image generation data and our additional vision task data.    During evaluation across several benchmarks, we find that Visual Banana excels at both visual understanding and generation, as summarized in Tab. \ref{tab:overall}. On the understanding side, Vision Banana surpasses or matches state-of-the-art results on both 2D and 3D tasks. For example, it beats the highly specialized segmentation model, \textit{SAM 3} \citep{sam3}, on various segmentation tasks, and the 3D expert, \textit{Depth Anything 3} \citep{dav3}, on metric depth estimation. On the generation side, it performs on par with its base model on image generation and editing benchmarks. On GenAI-Bench \citep{li2024genaibench}, Vision Banana scores a $53.5\%$ win rate against its base model. On ImgEdit \citep{ye2025imgedit} for image editing, Vision Banana's win rate is $47.8\%$. Since these results are achieved with a single unified model built by a lightweight instruction-tuning on its base model, there is strong evidence that Nano Banana Pro already possessed internal representations for visual understanding, which only needed to be unlocked with instruction tuning. 

The implications of this study are two-fold.
First, it suggests that image generators are indeed generalist vision learners under the hood, with
generative vision pretraining playing a foundational role similar to language model pretraining.
Second, it suggests that image generation can serve as a universal interface for unified visual understanding, mirroring the role of text generation in language understanding and reasoning.
We could be witnessing a major paradigm shift for computer vision, where generative vision pretraining takes a central role in building \textbf{Foundational Vision Models} for both generation and understanding.

\section{Method}

\paragraph{Instruction-tuning Nano Banana Pro}
Recent image and video generators have demonstrated zero-shot capabilities in generating visualizations of visual understanding tasks \citep{wiedemer2025veo_learner,zuo2025nanobanana_low_level}. 
To rigorously investigate and benchmark these capabilities, we need to \textit{align} the models to generate visualizations that can be decoded back to visual task outputs for quantitative evaluation. 
For example, in metric depth estimation, a generated depth heatmap must be invertible back to physical depth values for quantitative assessment. 
Therefore, we create \textbf{Vision Banana} by instruction-tuning our base model, Nano Banana Pro, on a selection of vision tasks formatted in such invertible manners.
Specifically, we mix vision task data into Nano Banana Pro's own training mixture at a very low ratio.
This process allows us to align the model's emergent generative representations into measurable physical geometry and semantic labels, allowing our single generalist model to be evaluated and compared against task-specific specialists.

Mixing the vision data at a low ratio serves as a lightweight instruction-tuning strategy, ensuring that our vision task alignment does not degrade the model's original generative priors.
This strategy distinguishes our work from previous approaches that perform full fine-tuning on generative models without retaining the image generation data mixture \citep{gan2023instructcv,marigold,zhao2025diception}.
We validate the preservation of image generation capabilities by benchmarking Vision Banana against the base Nano Banana Pro on two tasks: text-to-image generation (GenAI-Bench~\citep{li2024genaibench}) and image editing (ImgEdit~\citep{ye2025imgedit}).
In human evaluations, we obtain win rates of 53.5\% and 47.8\% respectively, indicating that Vision Banana successfully maintains the generative power of its base model.
We provide a detailed discussion of these generative capabilities in \cref{sec:generation_capabilities}.
Qualitative comparisons in \cref{fig:t2i-demo} (text-to-image generation) and \cref{fig:imgedit-demo} (image editing) in the appendix confirm that the outputs remain highly similar between Vision Banana and Nano Banana Pro.
These results verify that Vision Banana does not forget its generative nature.

\paragraph{Vision Tasks and Data.}
We evaluate our framework on two fundamental categories of visual understanding: $2$D scene understanding and $3$D structure inference. 
The $2$D suite consists of referring expression, semantic, and instance segmentation, which collectively test the model's capability to ground natural language and segment the corresponding objects. 
For $3$D understanding, we focus on monocular metric depth and surface normal estimation, which demand geometric reasoning and internal knowledge about object scales.
To collect data for instruction tuning, we utilize in-house model annotations for web-crawled 2D images, as well as synthetic data from rendering engines for 3D tasks. 
Crucially, no training data from our evaluation benchmarks is included in the instruction-tuning mixture, ensuring that our results reflect true generalist capability.


\section{Vision Banana - Generalist Vision Model from Image Generator}

In this section, we present qualitative and quantitative assessments compared to task-specific specialist models.
Built upon an image generator,
Vision Banana achieves SOTA-level results across a broad range of visual understanding tasks, without specialized architectures or custom training losses.

\begin{table}[t]
    \centering
    \setlength{\tabcolsep}{3pt}
    
    \begin{minipage}[t]{0.36\textwidth}
        \centering
        \scriptsize
        \begin{tabular}[t]{@{}lc@{}}
        \toprule
        Model & mIoU $\uparrow$ \\
        \midrule
        \multicolumn{2}{l}{\color{gray} \textit{Non Zero-Shot Transfer}} \\
        \midrule
        \color{gray}SegMan-L~\citep{fu2025segman} & \color{gray}84.2 \\
        \midrule
        \multicolumn{2}{l}{\textit{Zero-Shot Transfer}} \\
        \midrule
        APE-D~\citep{shen2024_aped} & 44.2 \\
        OpenSeeD~\citep{zhang2023simple_openseed} & 47.8 \\
        X-Decoder~\citep{zou2023generalized_xdecoder} & 52.0 \\
        SAM 3~\citep{sam3} & 65.2 \\
        \textbf{Vision Banana} & \textbf{\color{win_green} 69.9} \\
        \bottomrule
        \end{tabular}
        \caption{\textbf{Semantic segmentation results on Cityscapes val.}}
        \label{tab-seg-cityscapes}
    \end{minipage}
    \hfill
    \begin{minipage}[t]{0.60\textwidth}
        \centering
        \scriptsize
        \begin{tabular}[t]{@{}lccc@{}}
        \toprule
        Model & $cgF_1$ $\uparrow$ & IL\_MCC $\uparrow$ & $pmF_1$ $\uparrow$ \\
        \midrule
        \multicolumn{4}{l}{\color{gray} \textit{Non Zero-Shot Transfer}} \\
        \midrule
        \color{gray} SAM 3~\citep{sam3} & \color{gray} 54.1 & \color{gray} 0.82 & \color{gray} 66.1 \\
        \color{gray} SAM 3~\citep{sam3} + Llama 3.2 (ft) & \color{gray} 61.2 & \color{gray} 0.86 & \color{gray} 70.8 \\
        \midrule
        \multicolumn{4}{l}{\textit{Zero-Shot Transfer}} \\
        \midrule
        gDino-T~\citep{liu2024grounding_dino} & 3.3 & 0.15 & 16.2 \\
        LLMDet-L~\citep{fu2025llmdet} & 6.5 & 0.21 & 27.3 \\
        Gemini 2.5~\citep{gemini25}  & 13.0 & 0.29 & 46.1 \\
        APE-D~\citep{shen2024_aped}  & 16.4 & 0.40 & 36.9 \\
        DINO-X~\citep{ren2024dinox} & 21.3 & 0.38 & 55.2 \\
        OWLv2~\citep{minderer2023gowol}  & 24.6 & 0.57 & 42.0 \\
        \textbf{Vision Banana} + Gemini 3.1 Flash-Lite & \textbf{\color{win_green} 47.5} & \textbf{\color{win_green} 0.84} & \textbf{\color{win_green} 56.0} \\
        \bottomrule
        \end{tabular}
        \caption{\textbf{Instance segmentation results on SA-Co/Gold.}}
        \label{tab-seg-saco}
    \end{minipage}
    
    \vspace{2em}
    
    \begin{minipage}[t]{0.48\textwidth}
        \centering
        \scriptsize
        \begin{tabular}[t]{@{}lc@{}}
        \toprule
        Model & cIoU $\uparrow$ \\
        \midrule
        \multicolumn{2}{l}{\color{gray} \textit{Non Zero-Shot Transfer}} \\
        \midrule
        \color{gray}HyperSeg-Phi2-2.7B~\citep{wei2024hyperseg}  & \color{gray}79.4 \\
        \color{gray}X-SAM-Phi3-3.8B~\citep{wang2026xsam}  & \color{gray}83.8 \\
        \midrule
        \multicolumn{2}{l}{\textit{Zero-Shot Transfer}} \\
        \midrule
        HybridGL~\citep{liu2025hybrid} & 51.3 \\
        LocalizationHeads-LLaVA-1.5-13B~\citep{kang2025your}  & 67.7 \\
        SAM 3~\citep{sam3} + Gemini 2.5 Pro  & 73.4 \\
        \textbf{Vision Banana}  & \textbf{\color{win_green} 73.8} \\
        \bottomrule
        \end{tabular}
        \caption{\textbf{Referring expression segmentation results on RefCOCOg val (UMD).}}
        \label{tab-seg-refcocog}
    \end{minipage}
    \hfill
    \begin{minipage}[t]{0.48\textwidth}
        \centering
        \scriptsize
        \begin{tabular}[t]{@{}llc@{}}
        \toprule
        Model & gIoU $\uparrow$ \\
        \midrule
        \multicolumn{2}{l}{\color{gray} \textit{Non Zero-Shot Transfer}} \\
        \midrule
        \color{gray} X-SAM-Phi-3-3.8B~\citep{wang2026xsam} & \color{gray} 56.6 \\
        \color{gray} LISA-13B-LLaVA1.5~\citep{reasonseg} & \color{gray} 65.0 \\
        \midrule
        \multicolumn{2}{l}{\textit{Zero-Shot Transfer}} \\
        \midrule
        SegZero-Qwen2.5-VL-7B~\citep{liu2025segzero} & 62.6 \\
        RSVP-GPT-4o~\citep{lu2025rsvp} & 64.7 \\
        SAM 3~\citep{sam3} + Gemini 2.5 Pro & 77.0 \\
        \textbf{Vision Banana} + Gemini 2.5 Pro & \textbf{\color{win_green} 79.3} \\
        \bottomrule
        \end{tabular}
        \caption{\textbf{Referring expression segmentation results on ReasonSeg val.}}
        \label{tab-seg-reasonseg}
    \end{minipage}
\end{table}

\subsection{2D Semantic Understanding}

Image segmentation stands as a cornerstone of visual understanding, traditionally requiring complex, task-specific models to classify pixels into semantic categories or object instances. 
Current leading methods such as the Segment Anything series \citep{kirillov2023segment,ravi2024sam2,sam3} tackle this through heavy architectural specialization and large volumes of expensive, human-annotated mask data. 
Vision Banana challenges this prevailing paradigm by demonstrating that SOTA segmentation can naturally emerge from image generation pretraining. 
Rather than training on vast amounts of meticulously crafted segmentation examples, we tap into the rich representations learned by the base image generation model. 
By instructing the model to generate multi-colored images of segmentation masks, we obtain dense segmentation maps from which individual masks can be decoded, therefore enabling segmentation through image generation. 
As detailed in Tab. \ref{tab-seg-cityscapes}, \ref{tab-seg-saco}, \ref{tab-seg-refcocog} and \ref{tab-seg-reasonseg}, this elegant generative approach outperforms highly tuned specialist models, achieving SOTA zero-shot transfer performance on all evaluated segmentation benchmarks. 
We compare with other methods that have not been trained on in-domain data, i.e., the training splits of these benchmarks.
We denote them as ``Zero-Shot Transfer'' in the tables.
The usage of this term follows Segment Anything \citep{kirillov2023segment} and CLIP \citep{radford2021learning_clip}.
Non zero-shot transfer methods are marked in \textcolor{gray}{gray}.

\paragraph{Semantic Segmentation.}

\begin{figure}[t]
    \begin{subfigure}[t]{0.32\textwidth}
        \captionsetup{justification=raggedright, singlelinecheck=false,font=small}
        \includegraphics[width=\textwidth]{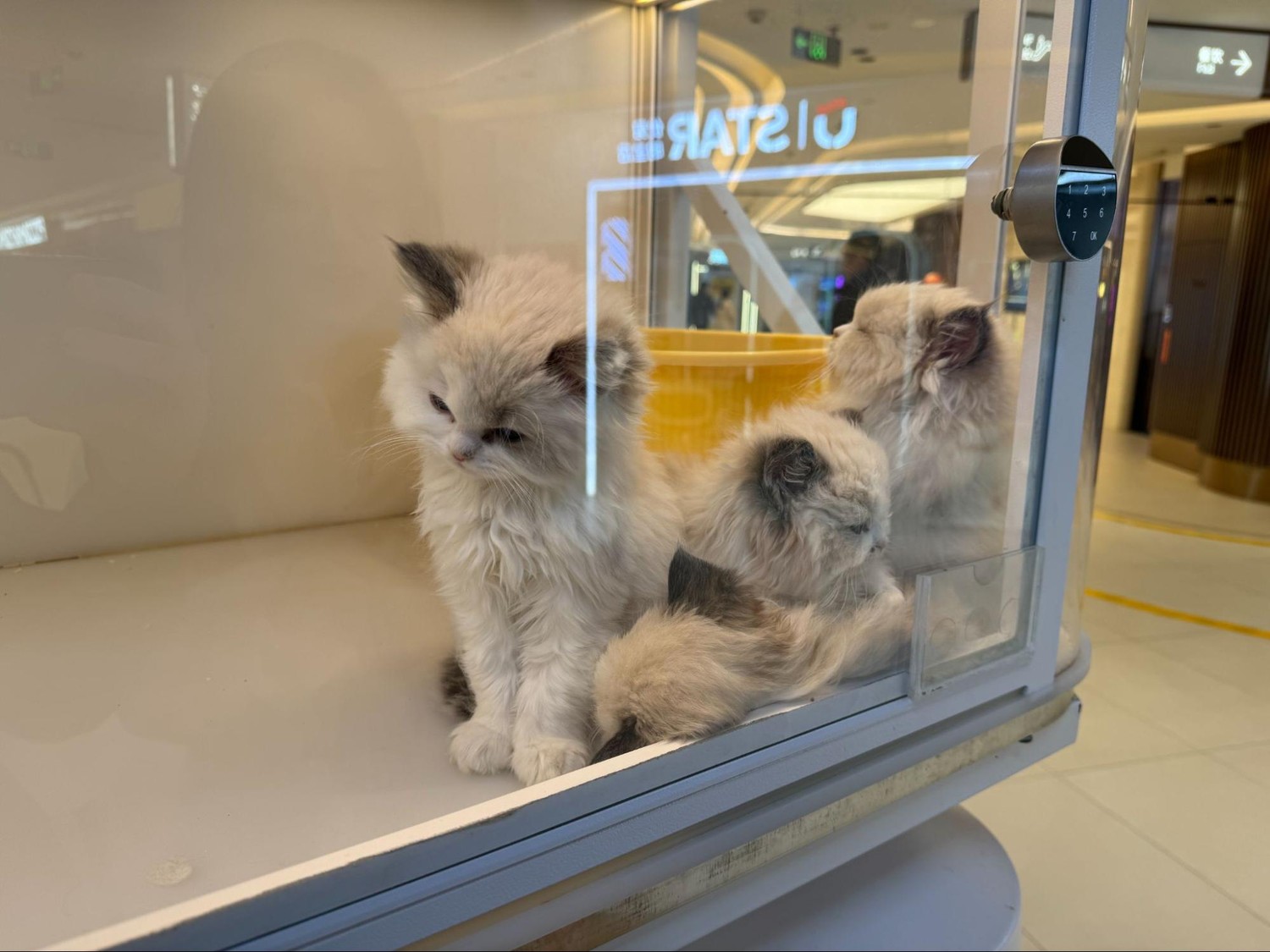}
        \caption{Example 1}
        \label{fig:demo-semseg1}
    \end{subfigure}
    \hfill
    \begin{subfigure}[t]{0.32\textwidth}
        \captionsetup{font=tiny, justification=raggedright, singlelinecheck=false}
        \includegraphics[width=\textwidth]{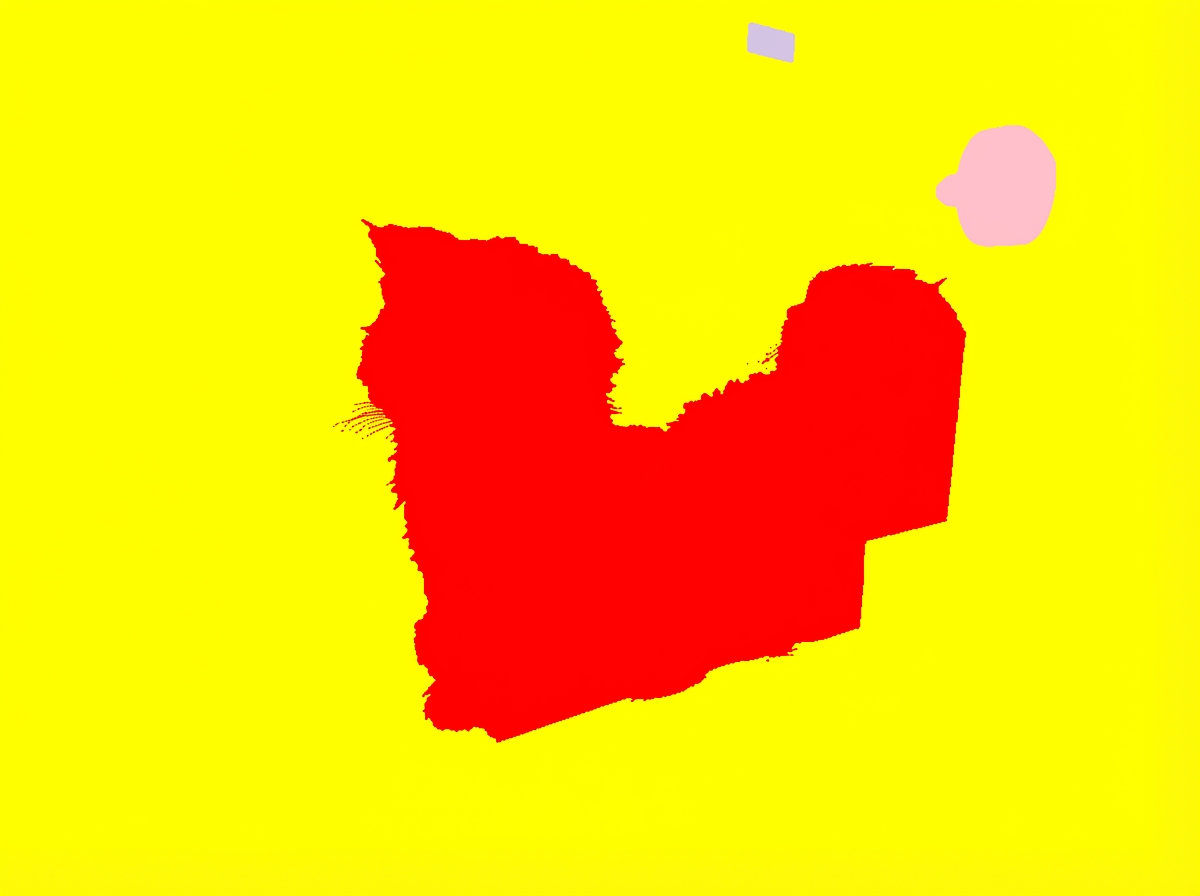}
        \caption*{``Generate a semantic segmentation visualization image, using this color mapping: \{"cat": "red", "lock": "pink", "exit sign": "light purple", "background": yellow\}.''}
    \end{subfigure}
    \hfill
    \begin{subfigure}[t]{0.32\textwidth}
        \captionsetup{font=tiny, justification=raggedright, singlelinecheck=false}
        \includegraphics[width=\textwidth]{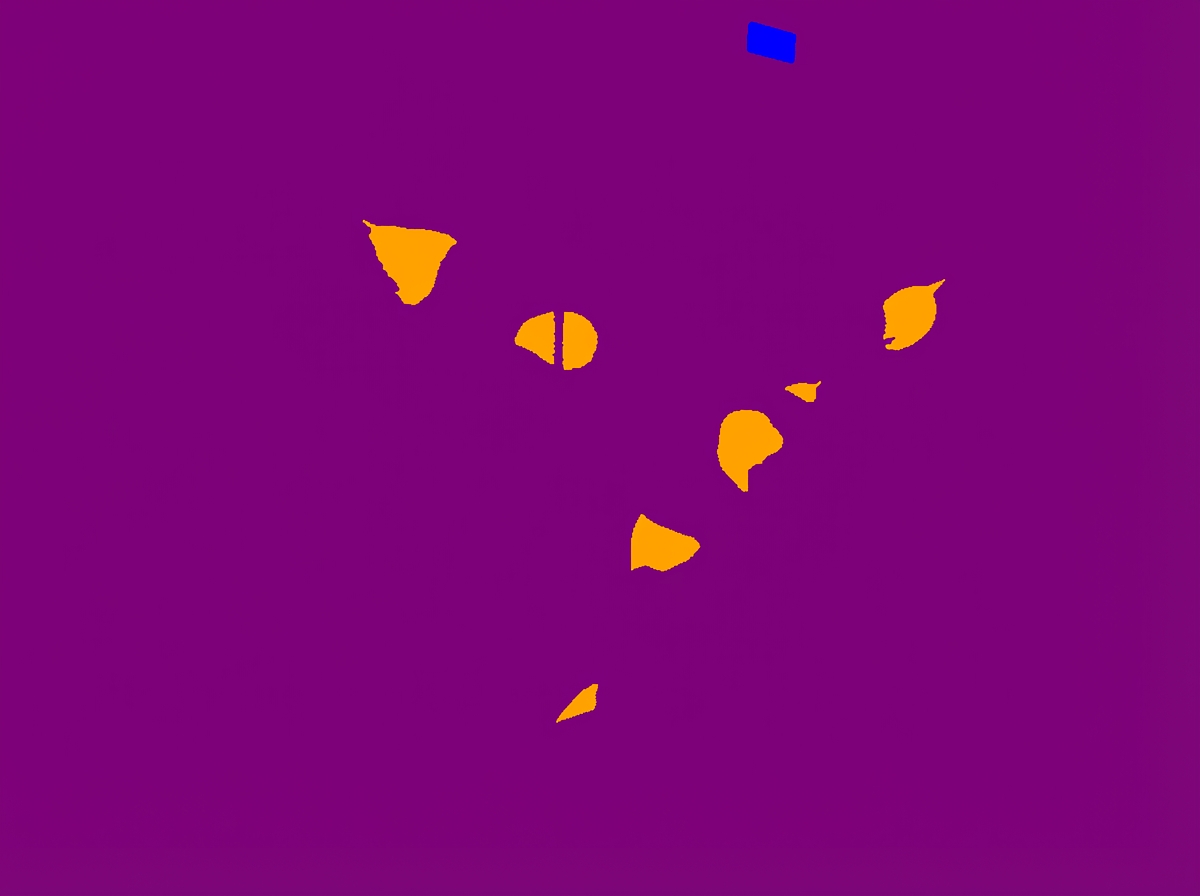}
        \caption*{``Generate a visualization image of semantic segmentation, using this color mapping: \{"cat ears": <255, 165, 0>, "exit sign": <0, 0, 255>, "background":<125,0, 125>\}''}
    \end{subfigure}

    \vspace{3mm}
    \begin{subfigure}[t]{0.32\textwidth}
        \captionsetup{justification=raggedright, singlelinecheck=false,font=small}
        \includegraphics[width=\textwidth]{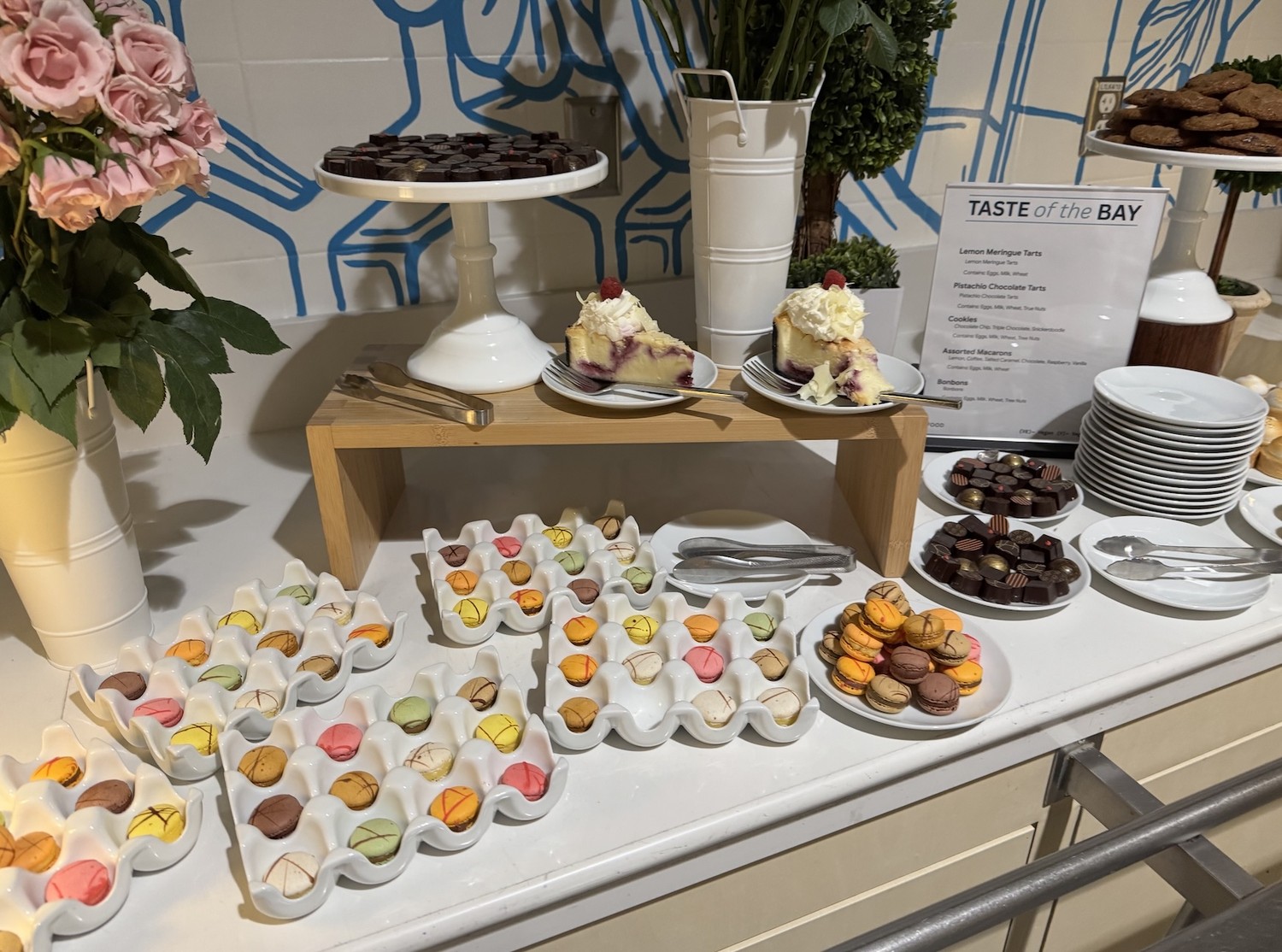}
        \caption{Example 2}
        \label{fig:demo-semseg2}
    \end{subfigure}
    \hfill
    \begin{subfigure}[t]{0.32\textwidth}
        \captionsetup{font=tiny, justification=raggedright, singlelinecheck=false}
        \includegraphics[width=\textwidth]{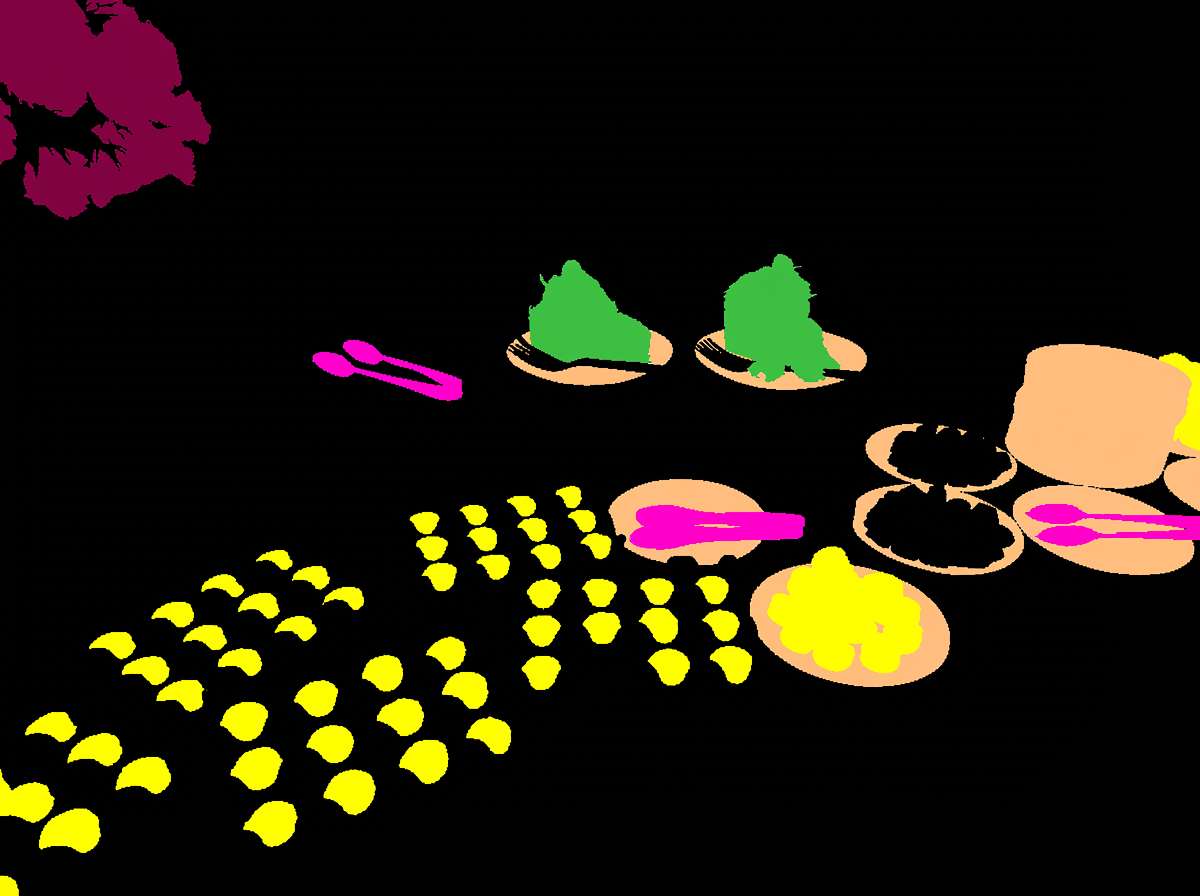}
        \caption*{``This image is a per-pixel class labeling of the input. The macaron cakes are represented by (255, 255, 0). The round plates are represented by (255, 192, 128). The slice cakes are depicted in (64, 192, 64). The flowers are shown in (128, 0, 64). The tongs are (255, 0, 192).''}
    \end{subfigure}
    \hfill
    \begin{subfigure}[t]{0.32\textwidth}
        \captionsetup{font=tiny, justification=raggedright, singlelinecheck=false}
        \includegraphics[width=\textwidth]{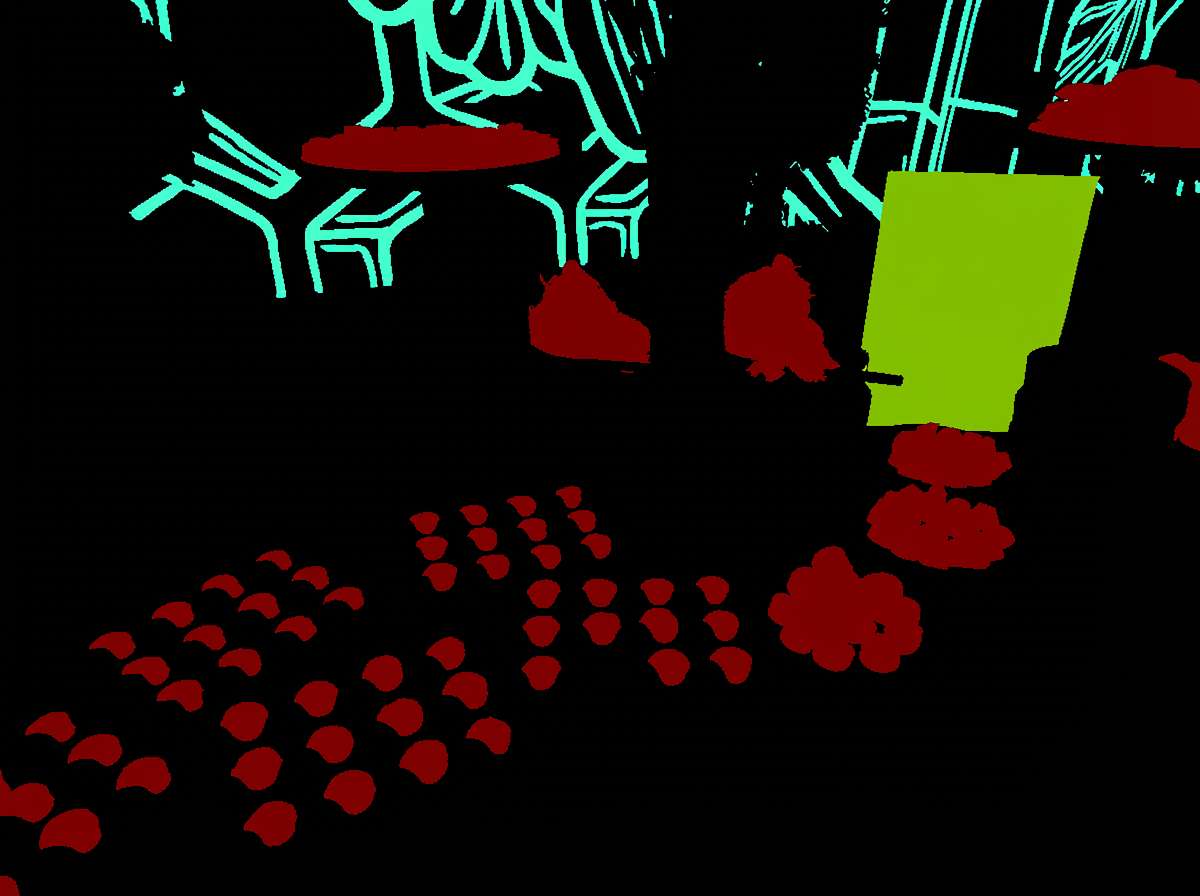}
        \caption*{``Generate a semantic segmentation visualization of the input. The menu is \#80C000. The dessert is \#800000. The patterns on the wall is \#40FFC0''}
    \end{subfigure}

    \caption{Vision Banana can perform semantic segmentation, following the instruction prompts.
    It handles various prompting styles.
    It can also segment anything specified via text prompts, from single-word nouns to phrases.
    It is able to produce segmentation masks with fine details, such as the cat whiskers in Example 1 (middle).
    }
    \label{fig:demo-semantic-segmentation}
\end{figure}

Semantic segmentation involves classifying each pixel into a predefined category without distinguishing between individual instances.
For example, the Cityscapes benchmark \citep{cityscapes} defines $19$ classes, including \textit{road}, \textit{person}, and \textit{sky}.
While instance and referring expression segmentation also convey semantic information, we use the term ``semantic segmentation'' here strictly in this instance-agnostic, category-level sense.
This nature of the classical semantic segmentation task can be specified via a text prompt, and we train the model to follow such instructions.
We prompt the model to generate a visualization image where each pixel is colored according to its class, as shown in Fig. \ref{fig:demo-semantic-segmentation}. 

Crucially, our approach is open-vocabulary: the target categories are not limited to a fixed set and can be specified dynamically in the prompt along with their corresponding color mappings.
We support various prompting styles, including natural language descriptions (e.g., ``the macaron cakes are represented by yellow'') and structured JSON mappings, with colors specified as named colors, hex codes, or RGB tuples.
For quantitative evaluation, we post-process the generated image by assigning each pixel to the class whose target color is closest in the RGB space.

We compare Vision Banana with existing methods on the Cityscapes validation set in \cref{tab-seg-cityscapes}.
During evaluation, we use the same text prompt for each example, providing the full class-to-color mapping for the 19 classes, including the ones not present in the image.
As shown in \cref{tab-seg-cityscapes}, Vision Banana outperforms SAM 3 by $4.7$ points in mIoU and achieves the best performance among open-vocabulary models, narrowing the gap with closed-set, non-zero-shot specialists like SegMan \citep{fu2025segman}.

\begin{figure}[!t]
    \begin{subfigure}[t]{0.32\textwidth}
        \captionsetup{justification=raggedright, singlelinecheck=false,font=small}
        \includegraphics[width=\textwidth]{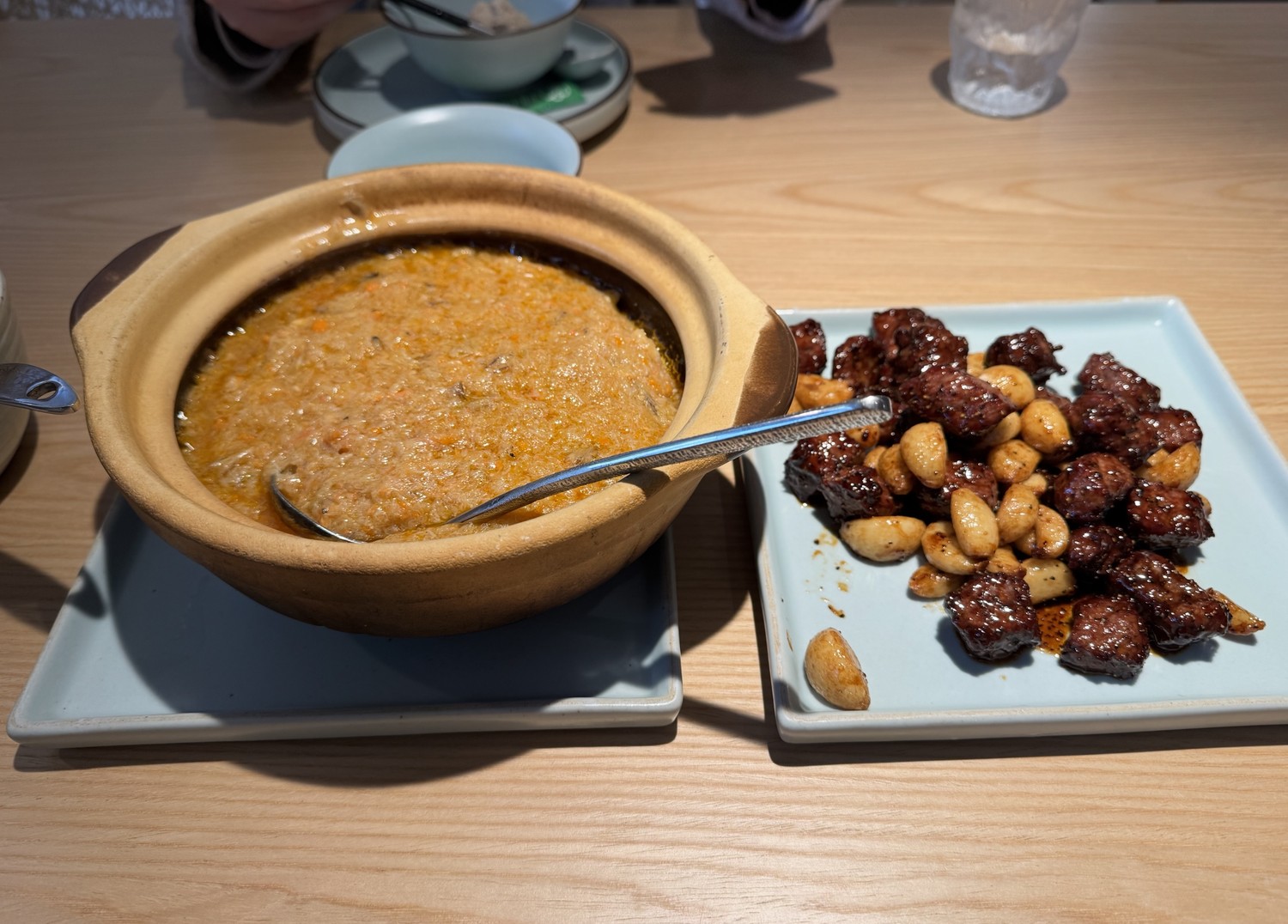}
        \caption{Example 1}
        \label{fig:demo-insseg1}
    \end{subfigure}
    \hfill
    \begin{subfigure}[t]{0.32\textwidth}
        \captionsetup{justification=raggedright, singlelinecheck=false,font=tiny}
        \includegraphics[width=\textwidth]{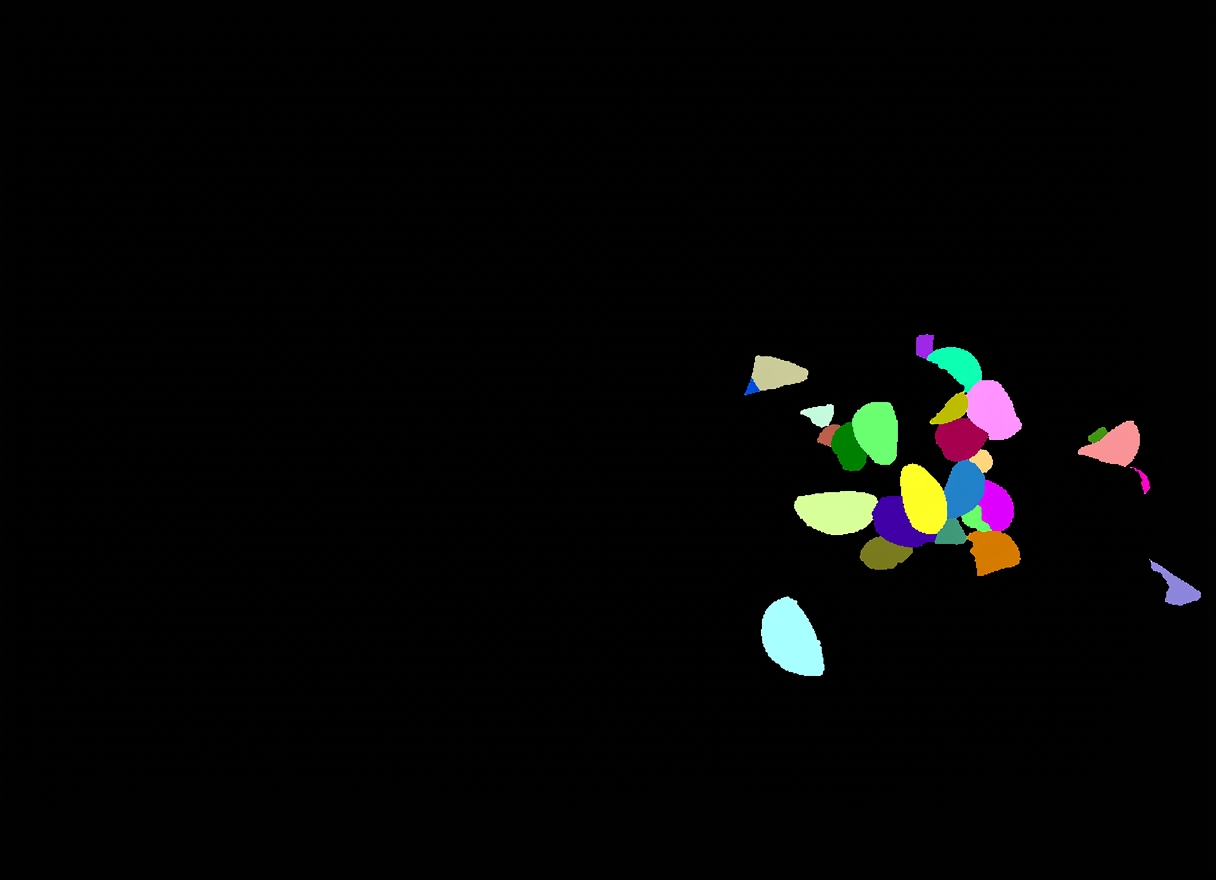}
        \caption*{``Generate an instance segmentation visualization of this image. Each piece of garlic is colored differently.''}
        \label{fig:demo-insseg1_pred1}
    \end{subfigure}
    \hfill
    \begin{subfigure}[t]{0.32\textwidth}
        \captionsetup{justification=raggedright, singlelinecheck=false,font=tiny}
        \includegraphics[width=\textwidth]{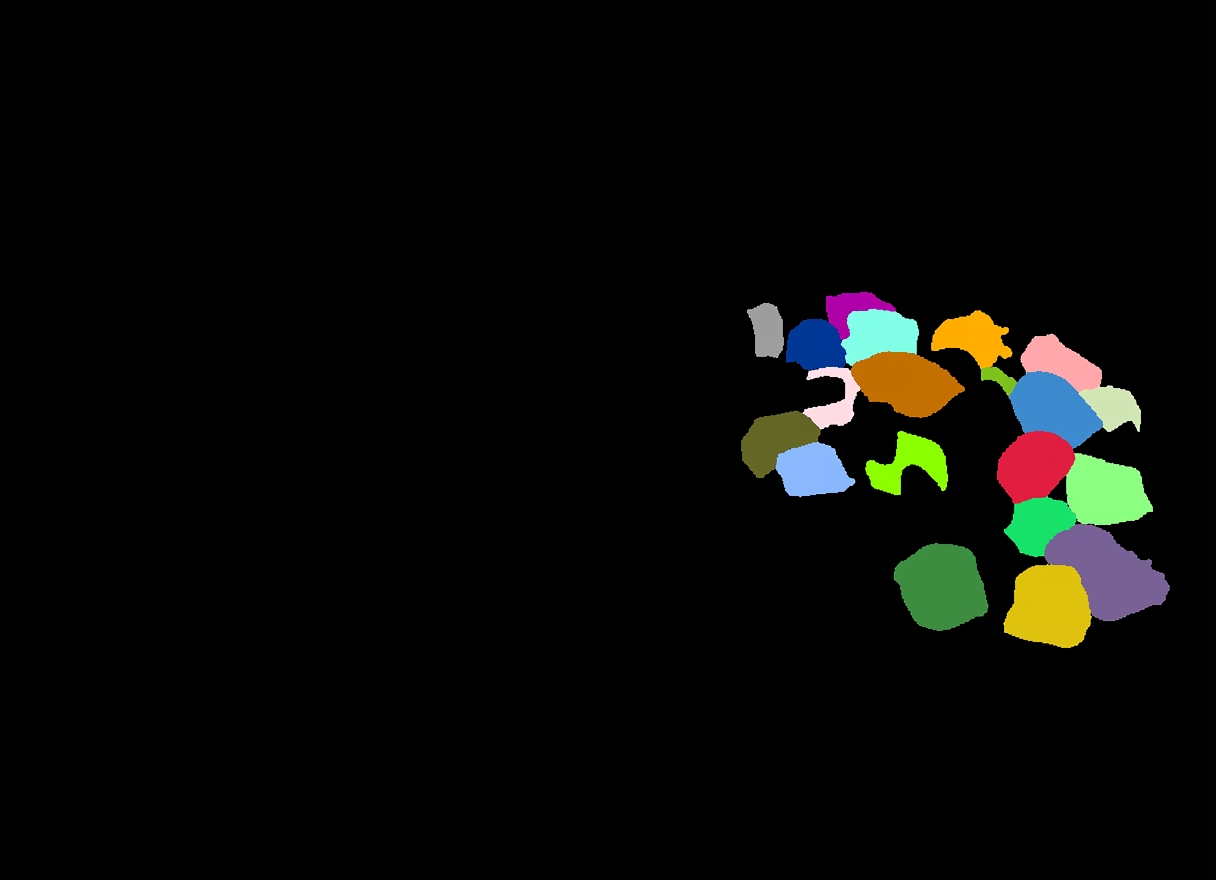}
        \caption*{``Generate an instance segmentation visualization of this image.  Each piece of beef is colored differently.''}
        \label{fig:demo-insseg1_pred2}
    \end{subfigure}

    \vspace{3mm}
    \begin{subfigure}[t]{0.32\textwidth}
        \captionsetup{justification=raggedright, singlelinecheck=false,font=small}
        \includegraphics[width=\textwidth]{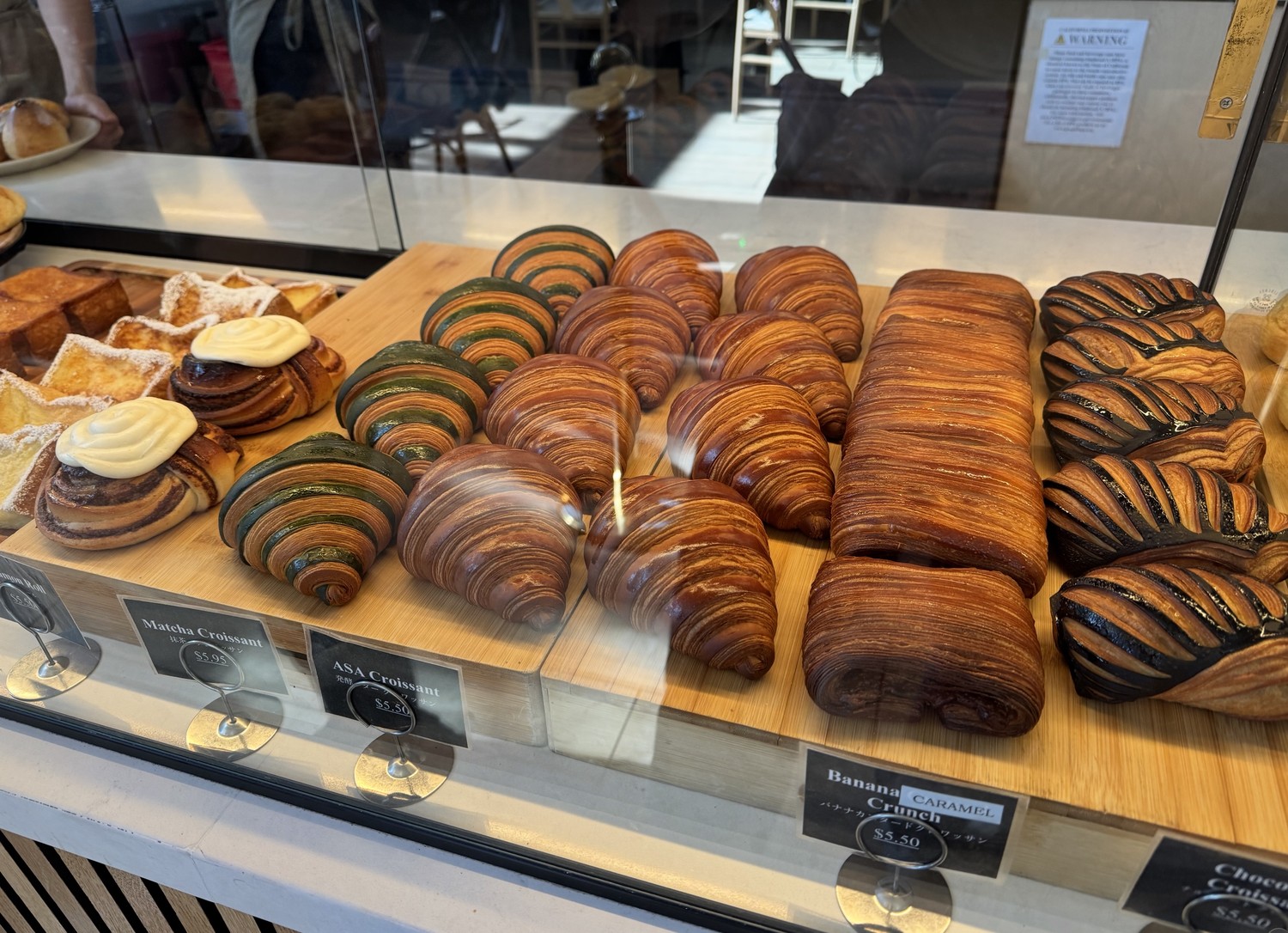}
        \caption{Example 2}
        \label{fig:demo-insseg2_in}
    \end{subfigure}
    \hfill
    \begin{subfigure}[t]{0.32\textwidth}
        \captionsetup{justification=raggedright, singlelinecheck=false,font=tiny}
        \includegraphics[width=\textwidth]{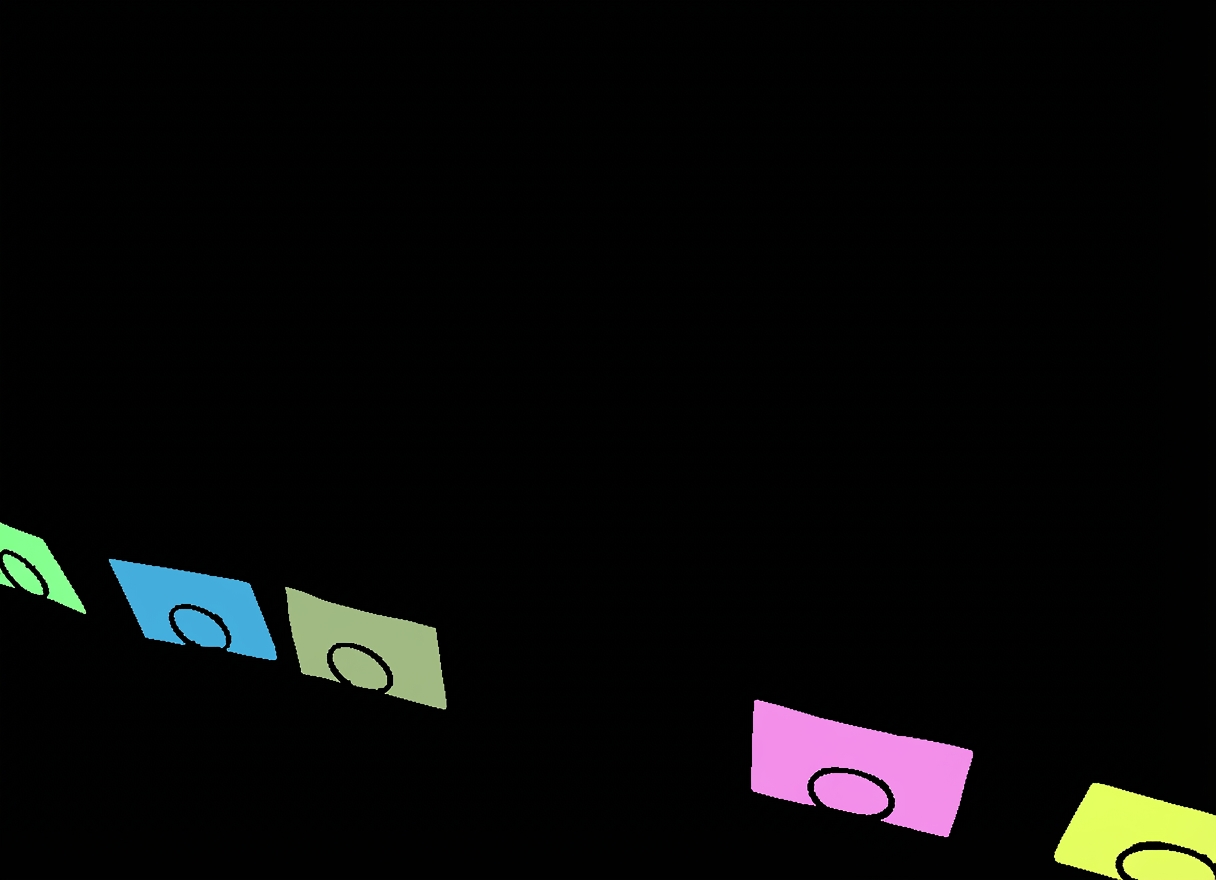}
        \caption*{``Generate an instance segmentation visualization of this image. Each price tag is colored differently. ''}
        \label{fig:demo-insseg2_pred_1}
    \end{subfigure}
    \hfill
    \begin{subfigure}[t]{0.32\textwidth}
        \captionsetup{justification=raggedright, singlelinecheck=false,font=tiny}
        \includegraphics[width=\textwidth]{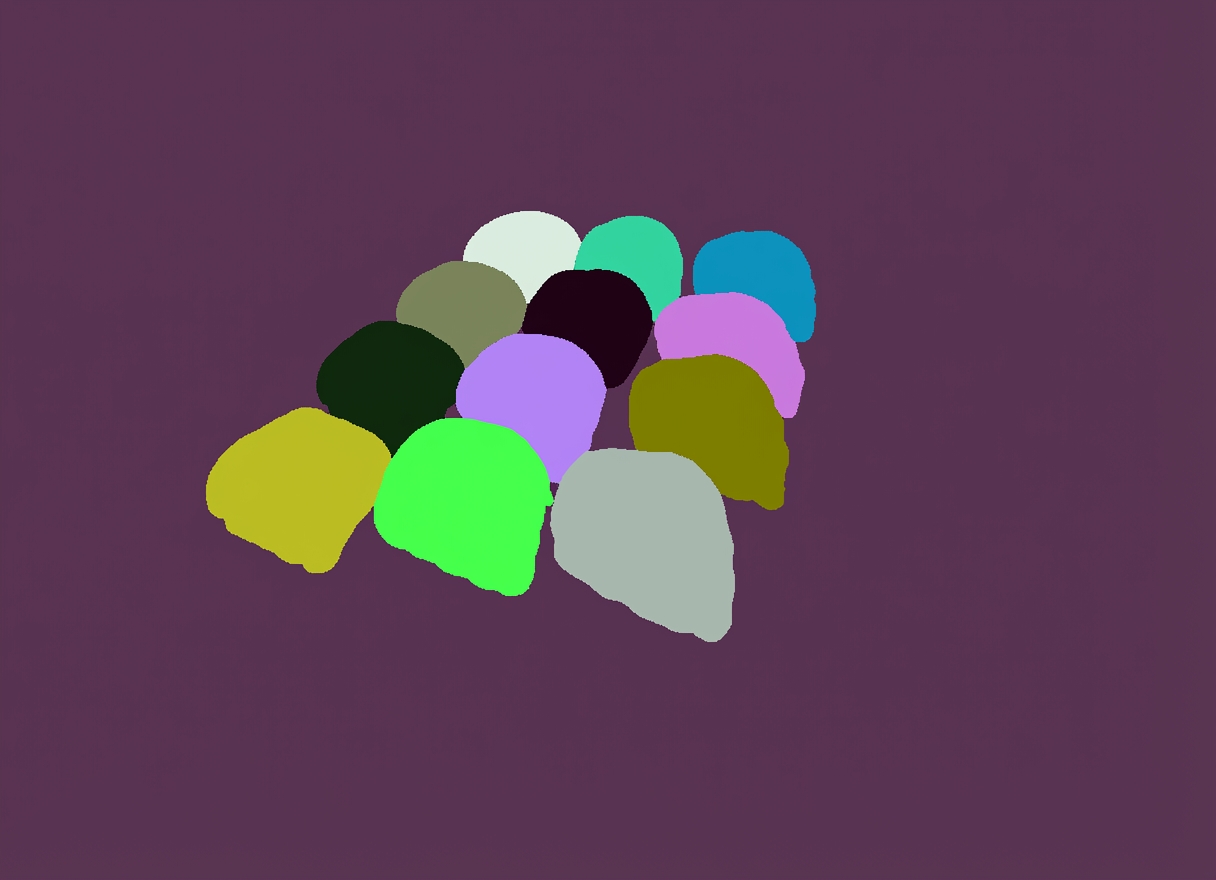}
        \caption*{``This image is a segmentation task derived from the input. The "crescent"-shaped croissant instances are each represented by a unique, solid color. Background is RGB(88, 50, 82).''}
        \label{fig:demo-insseg2_pred_2}
    \end{subfigure}

    \vspace{3mm}
    \begin{subfigure}[t]{0.32\textwidth}
        \captionsetup{justification=raggedright, singlelinecheck=false,font=small}
        \includegraphics[width=\textwidth]{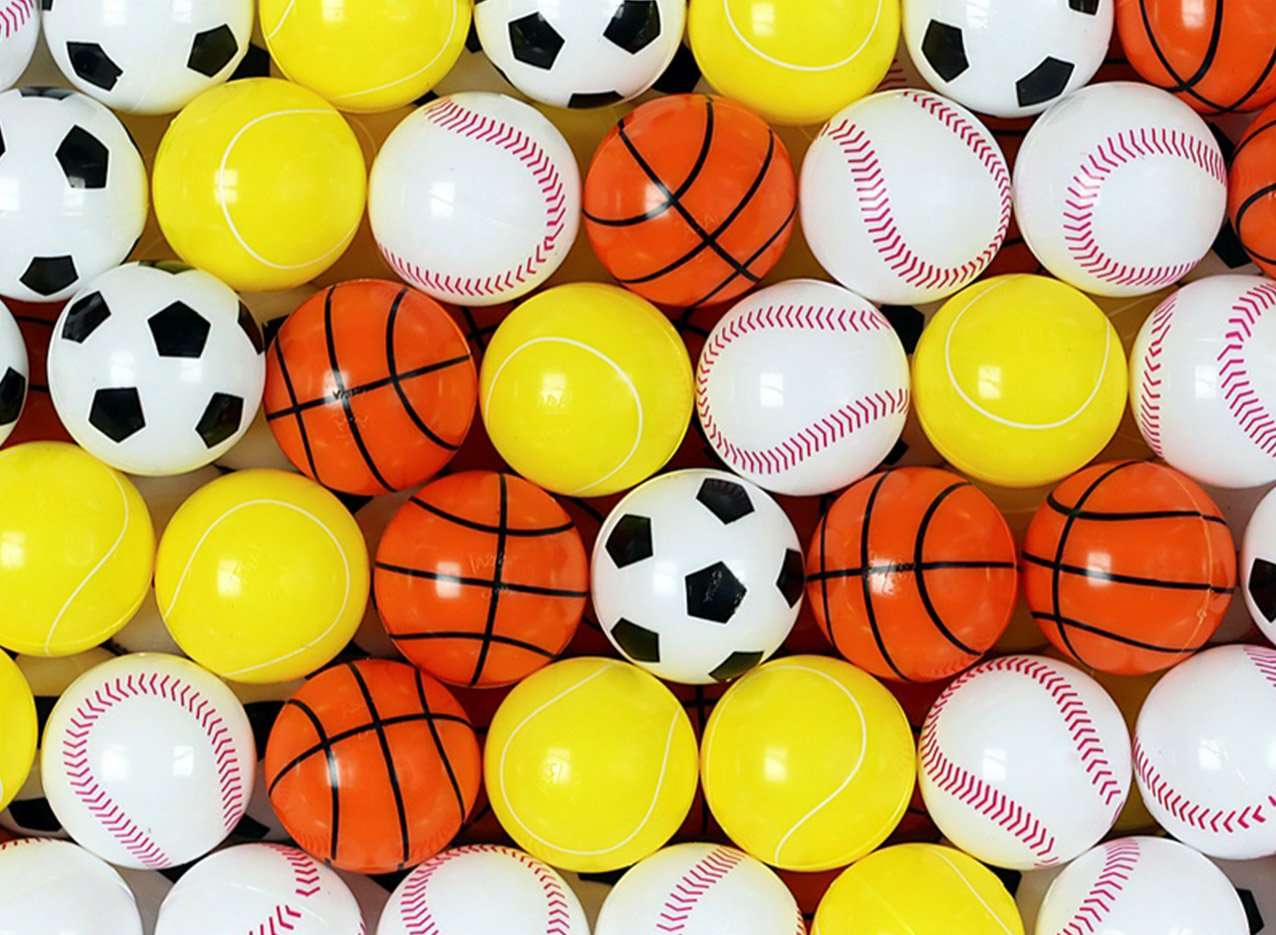}
        \caption{Example 3}
        \label{fig:demo-insseg3_in}
    \end{subfigure}
    \hfill
    \begin{subfigure}[t]{0.32\textwidth}
        \captionsetup{justification=raggedright, singlelinecheck=false,font=tiny}
        \includegraphics[width=\textwidth]{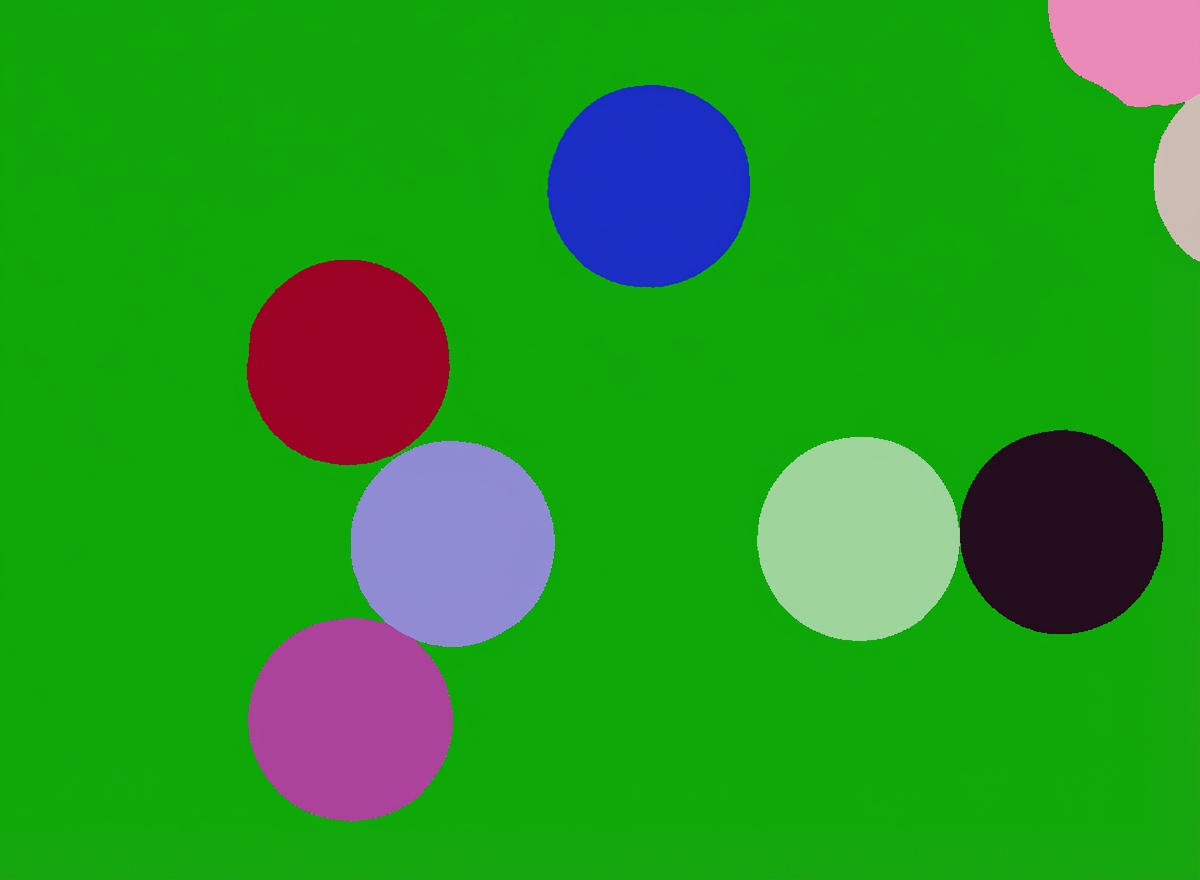}
        \caption*{``This image shows segmentation masks for the basketballs from the input image. The background is set to \#10aa05. Each basketball instance is represented by a solid circular mask, and a different colora is used for each mask.''}
        \label{fig:demo-insseg3_pred-1}
    \end{subfigure}
    \hfill
    \begin{subfigure}[t]{0.32\textwidth}
        \captionsetup{justification=raggedright, singlelinecheck=false,font=tiny}
        \includegraphics[width=\textwidth]{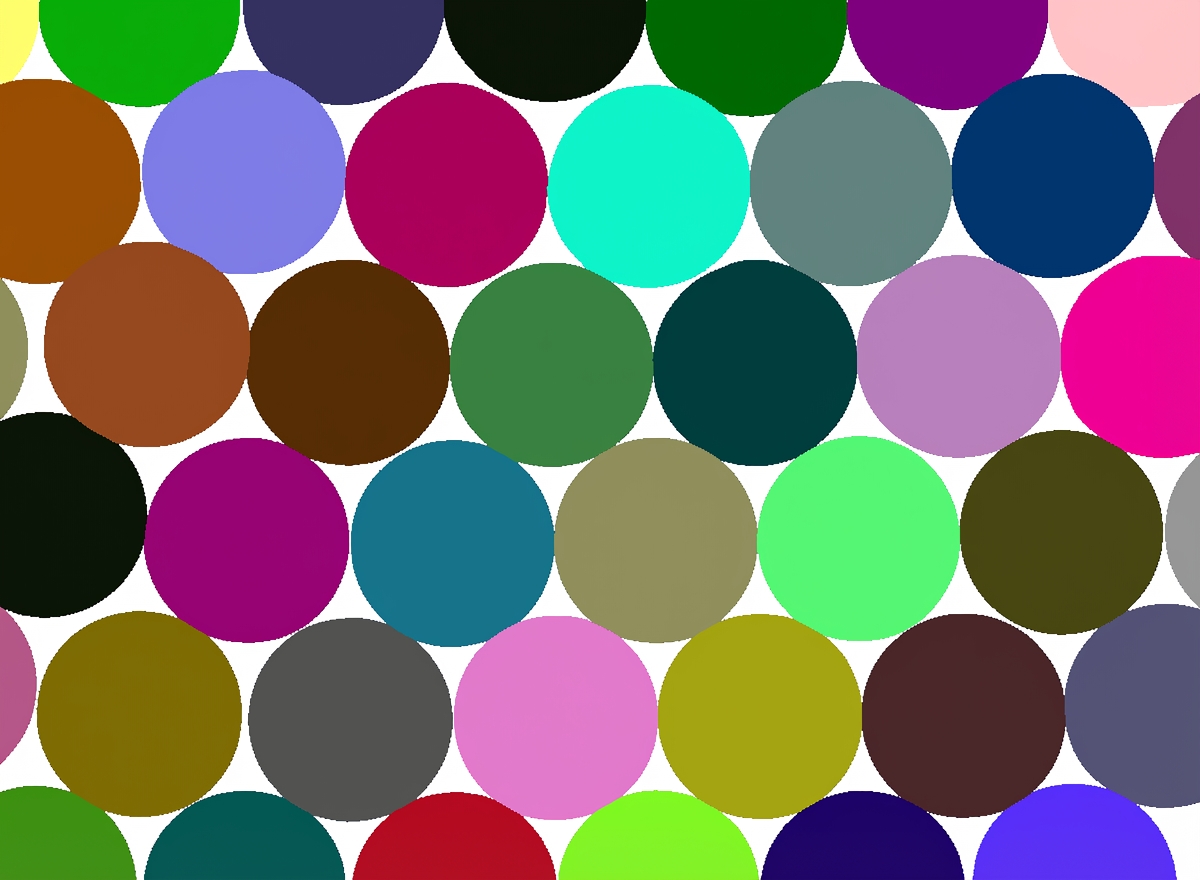}
        \caption*{``This image shows segmentation masks for the balls from the input image. The background is set to white color. Each ball is represented by a different color.''}
        \label{fig:demo-insseg3_pred-2}
    \end{subfigure}

    \caption{Vision Banana can perform instance segmentation, one class at a time.
    It renders different instances with different colors.
    It can understand nuanced language concepts as well.
    }
    \label{fig:demo-instance-segmentation}
\end{figure}

\newpage

\paragraph{Instance Segmentation.}

Unlike semantic segmentation, instance segmentation requires the model to distinguish between individual objects that belong to the same class.
For example, if an image contains five dogs, we expect the model to produce an individual mask for each animal.
This poses a unique challenge for Vision Banana: since the number of instances is unknown a priori, we cannot assign specific colors in the prompt beforehand.
To address this challenge, we prompt the model with only the target class and the background color, instructing it to assign a unique, distinct color to each individual instance.
We let the model dynamically assign distinct colors to different instances of that class.
Qualitative examples are shown in \cref{fig:demo-instance-segmentation}.
Individual instance masks can be extracted from the generated RGB images using a multi-stage clustering algorithm, detailed in \cref{sec:mask_parsing} of the appendix.

We evaluate our model on the open-vocabulary noun-phrase (NP) instance segmentation benchmark SA-Co/Gold \citep{sam3}.
We summarize the main results in \cref{tab-seg-saco} and provide a comprehensive category breakdown in \cref{tab-seg-saco_full}, \cref{sec:saco_details} of the appendix.
We also perform a qualitative evaluation in \cref{sec:saco_qualitative} of the appendix.
The SA-Co/Gold benchmark consists of 168k Image-NP pairs, where a large majority are negative queries (i.e., the target NP is absent from the image).
While Vision Banana could theoretically handle these negative queries by generating a solid black image, we did not tune the model to generate such empty mask images. We instead defer the task of classifying the Image-NP pairs as positives or negatives to a MLLM. To do so, we prompt Gemini 3.1 Flash-Lite with ``Is there an instance of <NP> visible in this image? Choose your answer from the following options: (A): Yes, (B): No.''. We then only process the positive-predicted examples with Vision Banana to generate an image of the masks. This approach is related to the "SAM 3 + Llama 3.2 (ft)" method, presented as "SAM 3 + EV" in \cite{sam3} where they fine-tuned Llama 3.2 to produce a presence score for SAM 3.

Under the zero-shot transfer setting, Vision Banana (paired with Gemini 3.1 Flash-Lite) achieves state-of-the-art performance, outperforming existing models including Gemini 2.5 \citep{gemini25}, APE-D \citep{shen2024_aped}, DINO-X \citep{ren2024dinox}, and OWLv2 \citep{minderer2023gowol}.
While Vision Banana still lags behind the SAM 3 specialist on SA-Co/Gold, we emphasize that unlike SAM 3, we did not include the SA-Co dataset in our training data mixture. 

\paragraph{Referring Expression Segmentation.}

Unlike traditional fixed-class segmentation, referring expression segmentation evaluates a model's ability to segment objects described by long, free-form natural language queries.
This task requires models to comprehend and reason about nuanced natural language expressions, as well as capture complex relationships between objects.
As summarized in \cref{tab-seg-refcocog,tab-seg-reasonseg}, our model achieves state-of-the-art performance under the zero-shot transfer setting, obtaining a cIoU of $73.8\%$ on RefCOCOg UMD \citep{refcoco} and a gIoU of $79.3\%$ on ReasonSeg \citep{reasonseg}.
It consistently outperforms SAM 3 Agent \citep{sam3} (which pairs SAM 3 with Gemini 2.5 Pro) and other recent zero-shot methods, including HybridGL \citep{liu2025hybrid}, LocalizationHeads \citep{kang2025your}, SegZero \citep{liu2025segzero}, and RSVP \citep{lu2025rsvp}.
On RefCOCOg, a performance gap remains compared to methods that are trained on the training split like HyperSeg \citep{wei2024hyperseg} and X-SAM \citep{wang2026xsam}.

\begin{figure}[!t]
    \centering
    \captionsetup[subfigure]{justification=raggedright, singlelinecheck=false,font=tiny}

    \begin{subfigure}[t]{0.24\textwidth}
    \captionsetup{font=small}
        \includegraphics[width=\textwidth]{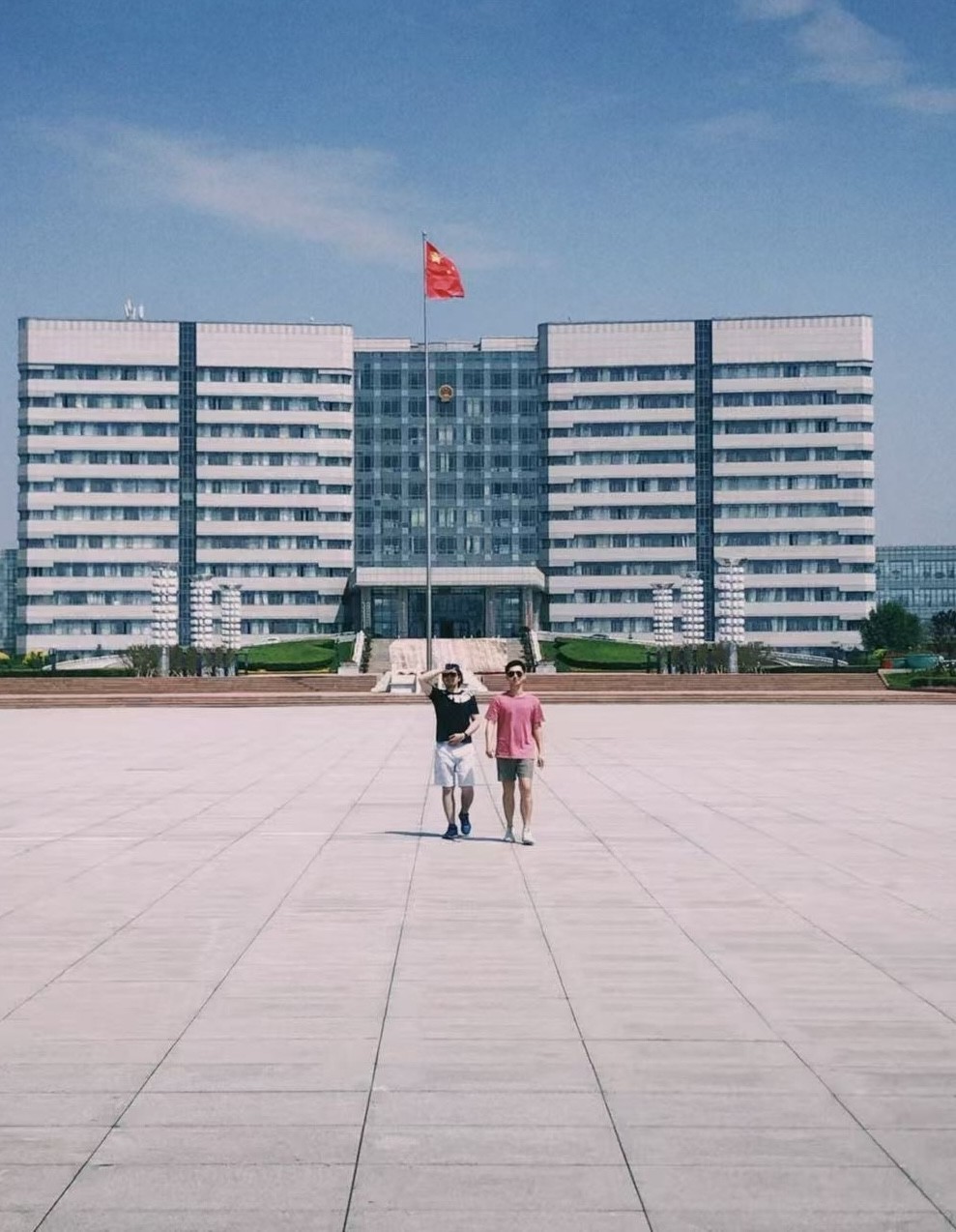}
        \caption{Input image}
    \end{subfigure}
    \hfill
    \begin{subfigure}[t]{0.24\textwidth}
        \includegraphics[width=\textwidth]{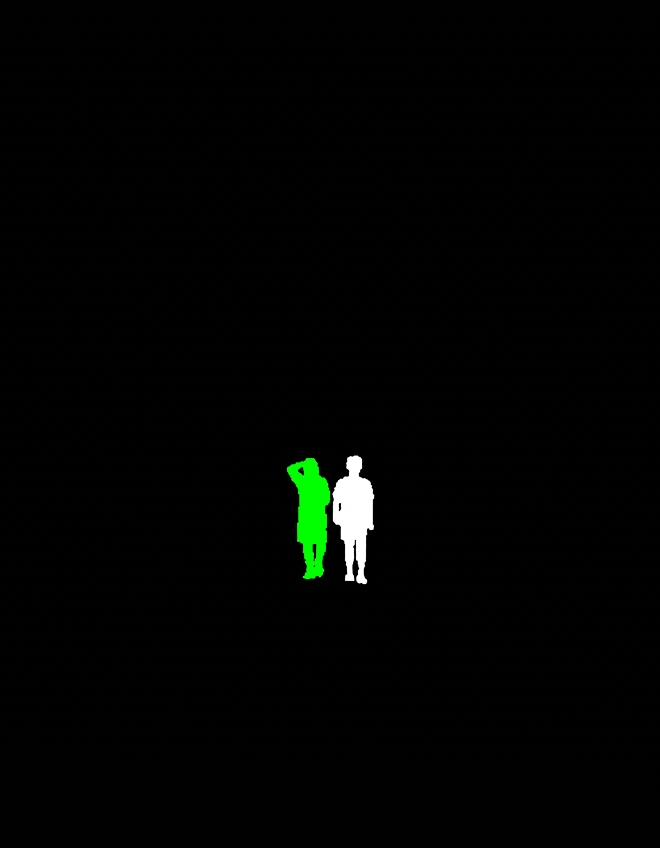}
        \caption*{A segmentation map image. The area that corresponds to the man in pink t shirt is rendered solid white; the other man is rendered in green.}
    \end{subfigure}
    \hfill
    \begin{subfigure}[t]{0.24\textwidth}
    \captionsetup{font=small}
        \includegraphics[width=\textwidth]{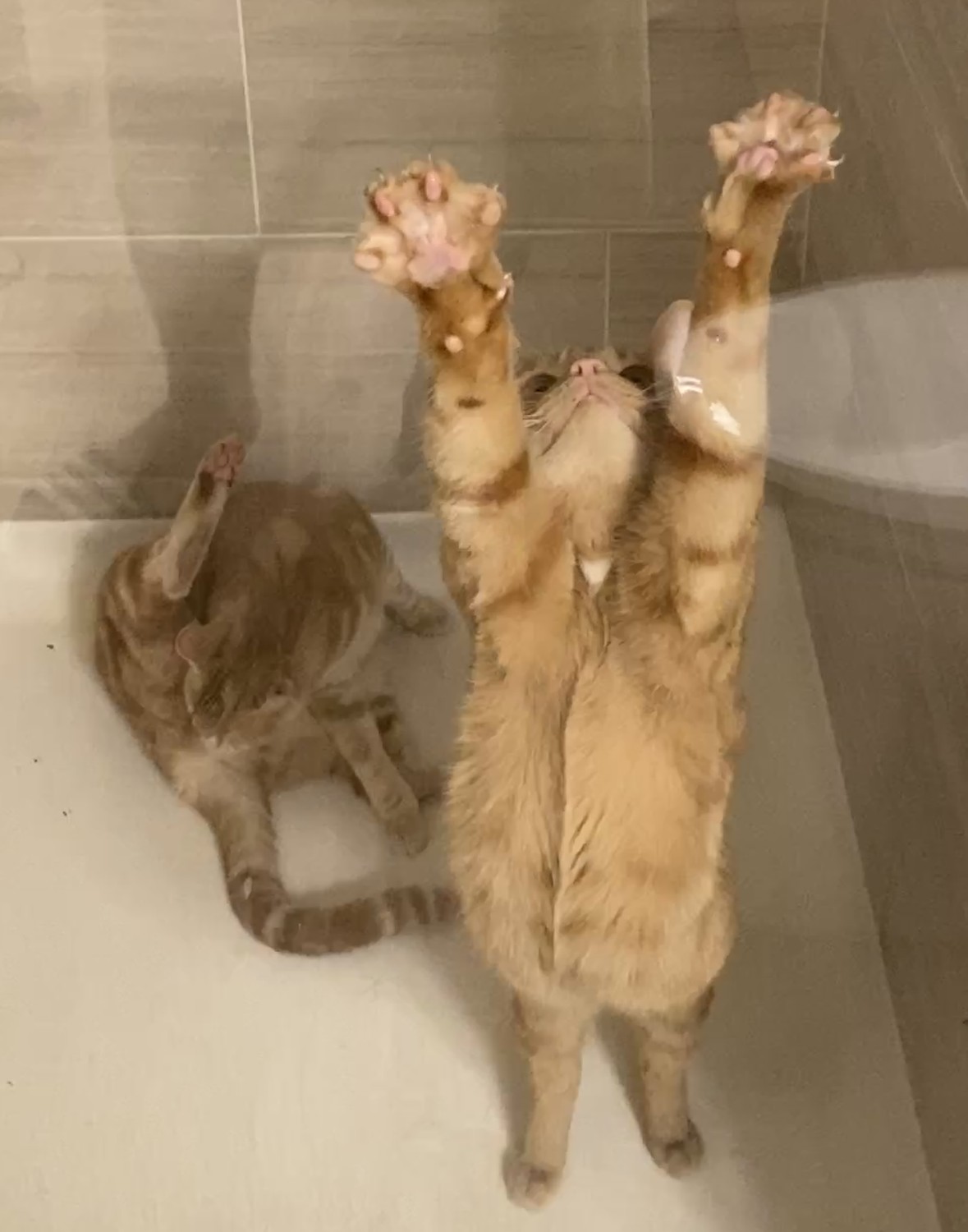}
        \caption{Input image}
    \end{subfigure}
    \hfill
    \begin{subfigure}[t]{0.24\textwidth}
        \includegraphics[width=\textwidth]{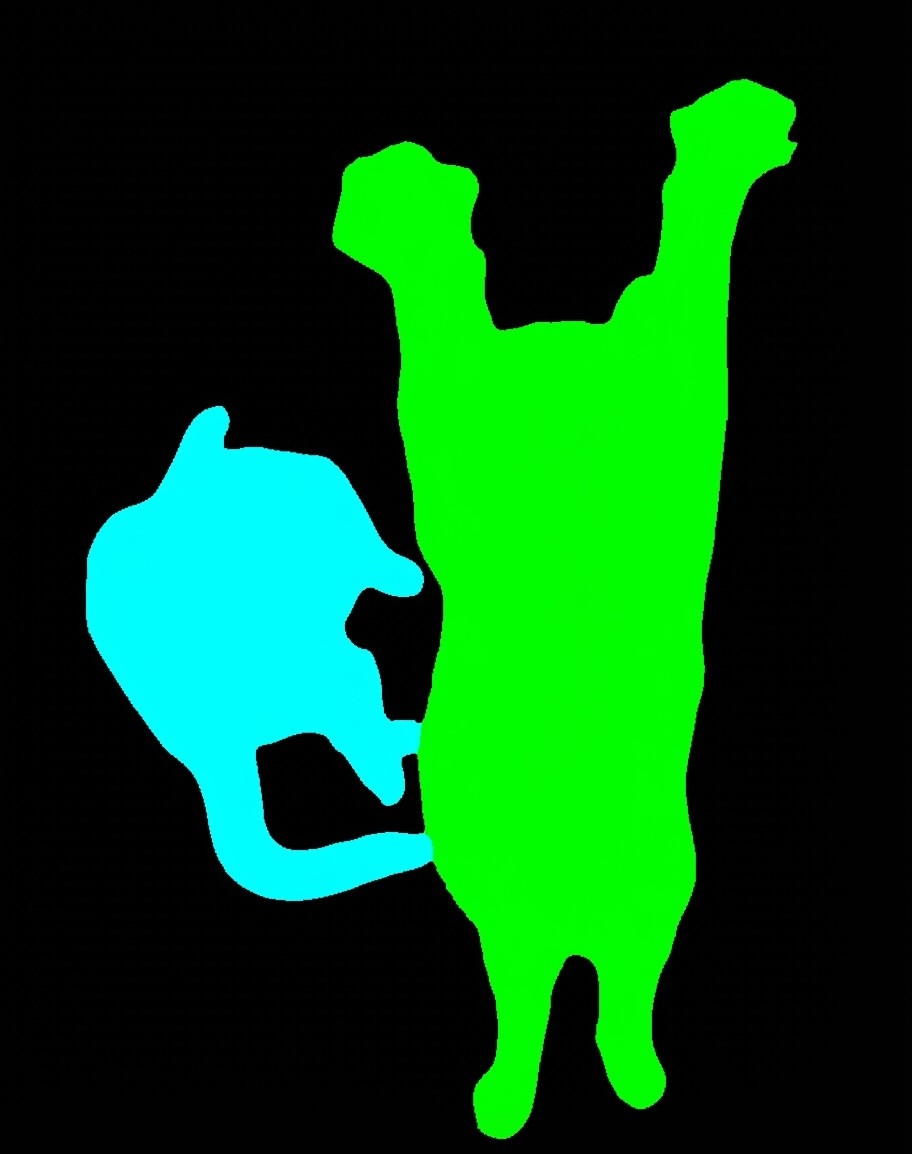}
        \caption*{A segmentation map image. The stretching cat is rendered in green, the cat that is cleaning itself is in cyan.}
    \end{subfigure}

    \vspace{3mm}
    \begin{subfigure}[t]{0.24\textwidth}
    \captionsetup{font=small}
        \includegraphics[width=\textwidth]{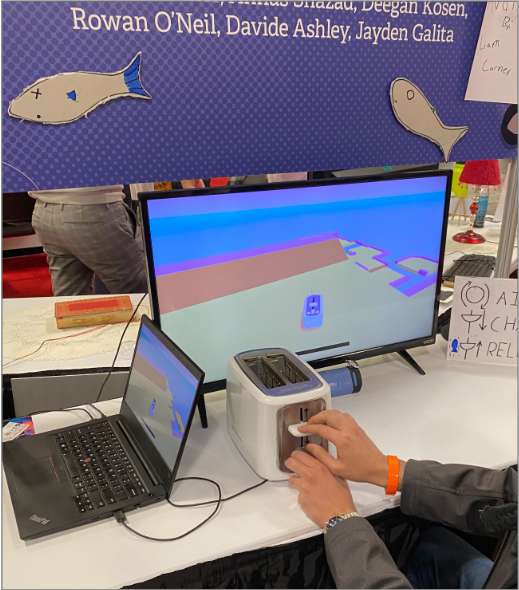}
        \caption{Input image}
    \end{subfigure}
    \hfill
    \begin{subfigure}[t]{0.24\textwidth}
        \includegraphics[width=\textwidth]{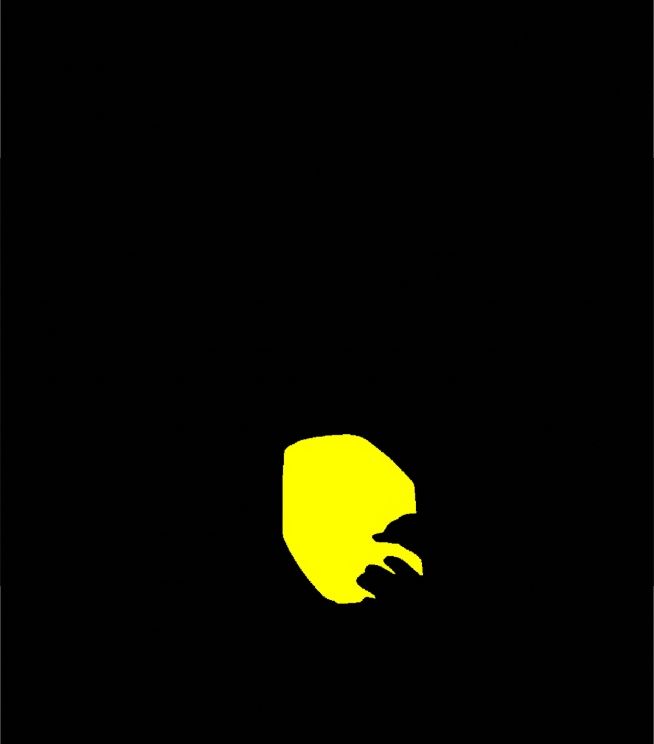}
        \caption*{This image shows segmentation masks from the given image. The background is black color. The game control device is represented by a solid yellow.}
    \end{subfigure}
    \hfill
    \begin{subfigure}[t]{0.24\textwidth}
    \captionsetup{font=small}
        \includegraphics[width=\textwidth]{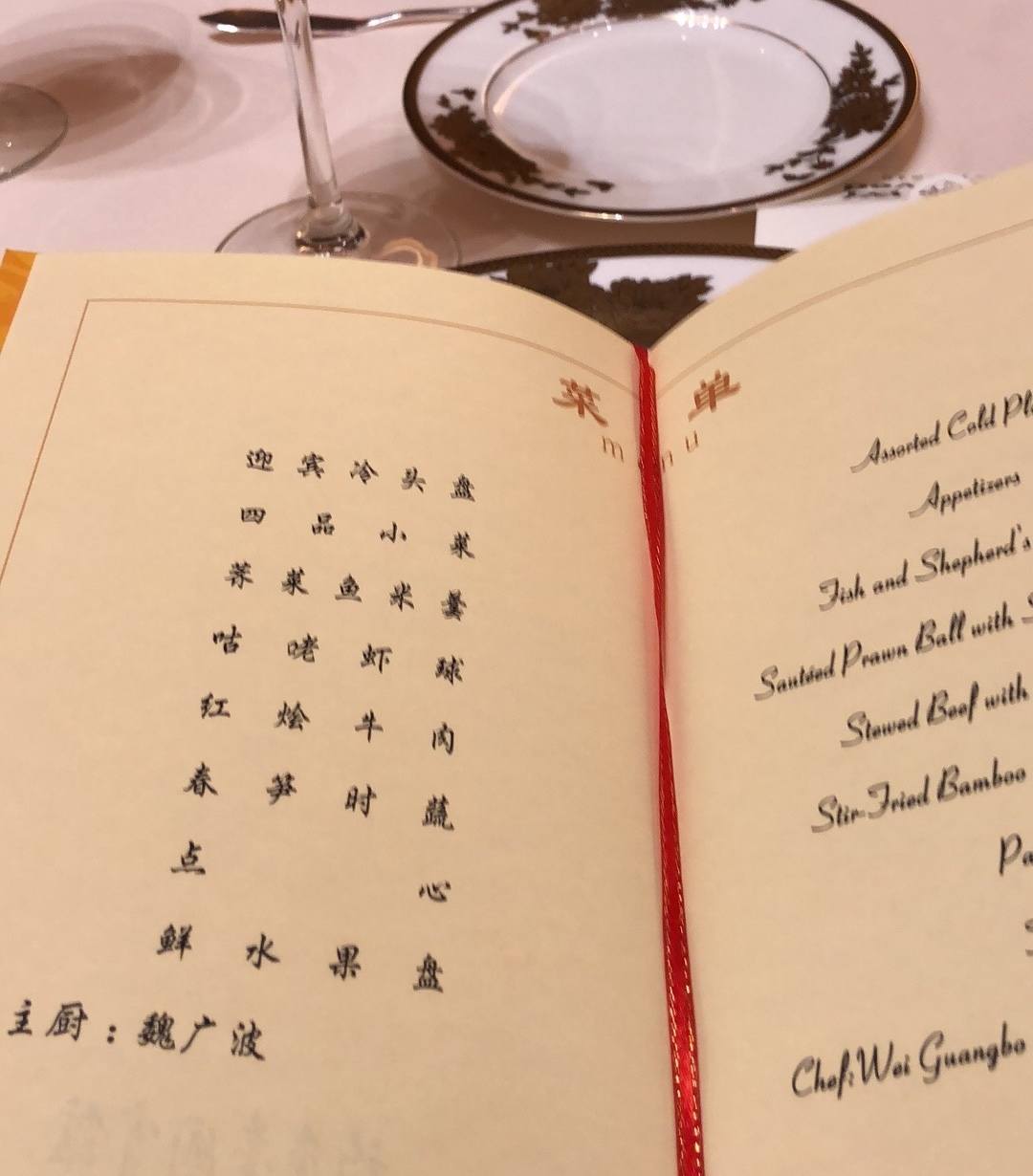}
        \caption{Input image}
    \end{subfigure}
    \hfill
    \begin{subfigure}[t]{0.24\textwidth}
        \includegraphics[width=\textwidth]{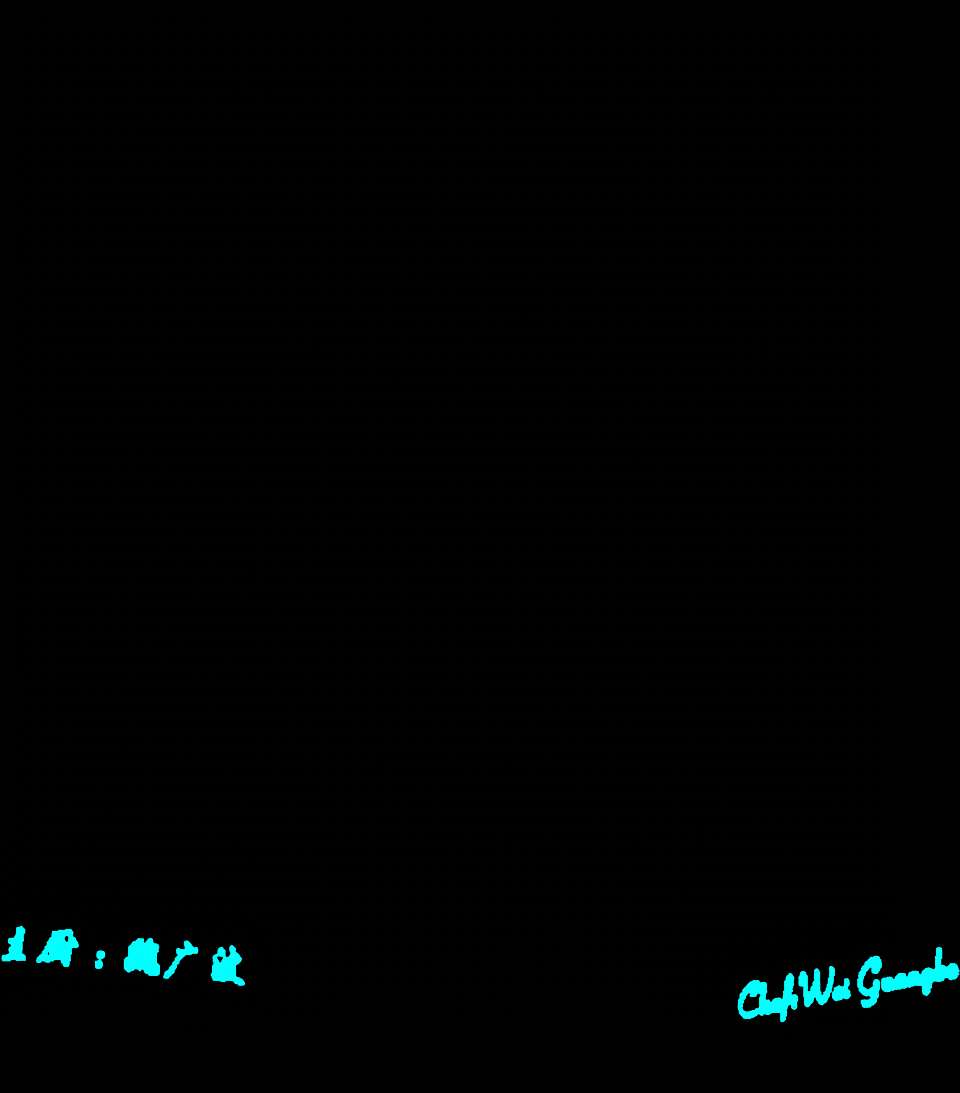}
        \caption*{This image shows segmentation masks from the given image. The background is black color. The chef's names in both Chinese and English are rendered as cyan color.}
    \end{subfigure}

    \caption{Vision Banana can understand natural language prompts and reason about them, including but not limited to: (a) description of objects' appearances (``\textit{man in pink t shirt}''); (b) description of actions (``\textit{stretching}'' and ``\textit{cleaning}''); (c) objects that have uncommon usage (\textit{toaster as a game controller}); and (d) multilingual text content (\textit{text on the menu in Chinese and English}).
    This requires strong and comprehensive visual understanding capability.
    }
    \label{fig:refseg-demo}
\end{figure}

For the complex reasoning queries in ReasonSeg, we follow standard practice by delegating the reasoning step to a multimodal LLM.
Specifically, we utilize Gemini 2.5 Pro to translate the reasoning query into a descriptive reference, which then serves as the prompt for Vision Banana.
We evaluate this pipeline in a single-turn inference setup, where both Gemini and Vision Banana are queried exactly once.
In this setting, Vision Banana paired with Gemini 2.5 Pro outperforms several non-zero-shot methods that were trained directly on ReasonSeg, including X-SAM \citep{wang2026xsam} and LISA \citep{reasonseg}.
Qualitative results in \cref{fig:refseg-demo} illustrate Vision Banana's ability to ground diverse language cues, from physical actions (``stretching cat'') and atypical object roles (``toaster as a game controller'') to multilingual text on signage.
This highlights a key advantage of our approach: the rich multimodal priors inherited from generative pre-training allow Vision Banana to reason about 'what' to segment more effectively than specialized segmentation models.

Intriguingly, Vision Banana also exhibits strong cross-task transfer by demonstrating a similar grasp of referring expressions combined with standard semantic and instance segmentation tasks, despite not being explicitly trained on free-form queries for those tasks.
For example, in Fig. \ref{fig:demo-semseg2} (right), the model understands what ``patterns on the wall'' is referring to.
In Fig. \ref{fig:demo-insseg2_in} (right), the model successfully distinguishes crescent-shaped croissants from other variations of croissants.
These findings suggest that our generative pre-training yields highly robust, transferable representations across distinct visual grounding paradigms.

\subsection{3D Understanding from Monocular Images}
Vision Banana demonstrates a strong ability to infer 3D structures from 2D monocular images. We evaluate this capability on two classical tasks: monocular metric depth estimation and surface normal estimation.
As summarized in Tab. \ref{tab:overall}, Vision Banana achieves SOTA performance on both tasks, surpassing specialists such as Depth Anything V3 \citep{dav3} and Lotus-2 \citep{lotus2}.

\paragraph{Metric Depth Estimation.}
The goal of depth estimation is to produce a depth map from a monocular image, where each pixel's value represents the physical metric distance from the camera plane to the observed object~\citep{eigen2014depth}. This is a fundamental computer vision task that benefits a wide range of applications such as robotics, augmented/virtual reality, and autonomous driving. 
However, depth estimation is inherently ill-posed, as 2D projections inherently discard critical 3D geometric information.
Furthermore, monocular depth estimation is particularly challenging due to the absence of parallax cues available in multi-view setups, even when camera intrinsic parameters are known.

\newpage

In the deep-learning era, the research community has largely framed depth estimation as a dense per-pixel supervised regression problem, employing specialized architectures and domain-specific loss functions. Most recent SOTA methods rely on camera intrinsics during training, inference, or both~\citep{yang2024dav2,bochkovskii2024depthpro,wang2024moge,wang2025moge2,lotus2,he2024lotus,hu2024metric3dv2,depthlm,dav3,piccinelli2025unidepthv2,unik3d}.
While using intrinsics mitigates the inherent ambiguity of depth estimation, it also necessitates specialized model designs. 
In contrast, our work is predicated on the hypothesis that the mode-seeking nature of generative modeling naturally resolves training target ambiguities, thereby eliminating the need for such specialized techniques.
Furthermore, the broad world knowledge acquired during pretraining endows the model with stronger priors on object sizes and distances compared to narrowly targeted models.
To enable Nano Banana Pro to estimate depth in metric units, we instruct the model to output a carefully constructed false-color visualization of depth values.

\newcommand\lft{\mathopen{}\left}
\newcommand\rgt{\aftergroup\mathclose\aftergroup{\aftergroup}\right}
\newcommand{\power}{\lambda}

\begin{figure}[htbp]
  \centering
  \begin{minipage}[r]{0.35\textwidth}
    \includegraphics[width=\linewidth]{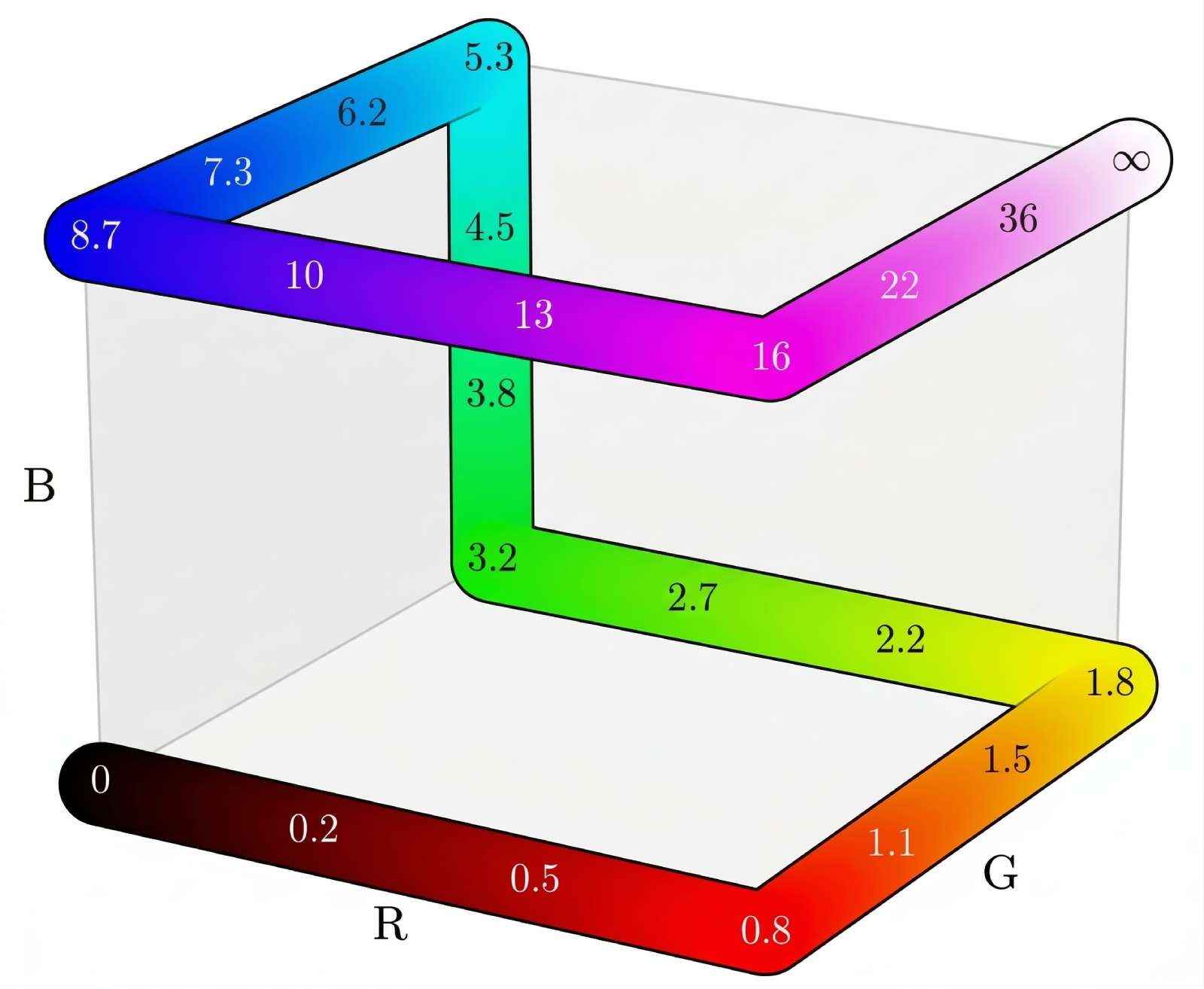}
  \end{minipage}
  \begin{minipage}[l]{0.5\textwidth}
    \caption{A visualization of our bijection between scalar metric distances $d \geq 0$ and RGB color values in $[0, 1]^3$, which is achieved by curving metric depth with a power transform, and then interpolating along the edges of the color cube according to that curved metric depth. The metric depth values (in meters) corresponding to various RGB colors are overlaid. }

    \label{fig:color_tubes_minipage}
  \end{minipage}
\end{figure}

To visualize depth maps as RGB images, we establish a mapping between unbounded depth values in $[0, \infty)$ and bounded RGB values in $[0, 1]^3$. 
Because the utility of accurate metric depth for nearby image content is generally higher than that of distant content (e.g.,\ graspable objects matter more for robotics tasks, stereo/monodepth benchmarks usually measure accuracy terms of disparity or relative/log-depth) we ``curve'' metric depth prior to RGB encoding.
Specifically, this is achieved by first applying the power transform of \citet{barron2025power}  to warp the depth values, and then using those curved distances to produce a false-color visualization. 
We constrain the power transform to $\power < -1$ and rescale it to map metric distances $d \in [0, \infty)$ to normalized distances in $[0, 1)$:
\begin{equation}
  f(d, \power, c) = 1 - \lft( 1 - \sfrac{d}{\power c}  \rgt)^{\power + 1}
\end{equation}
In all experiments, we set the shape parameter to $\power = -3$ and the scale parameter to $c = \sfrac{10}{3}$. These curved and normalized distances $f(d, \power, c)$ are then used to interpolate along a piecewise-linear function that follows the edges of the RGB cube, traversing along its edges from black to white, similarly to the first iteration of a 3D Hilbert curve. A visualization of this process is provided in Fig.~\ref{fig:color_tubes_minipage}. 

\vspace{3mm}

\vspace{4mm}

This mapping from normalized distance to RGB color can be inverted by simply projecting the RGB values onto the nearest line segment and then inverting the linear interpolation along the cube's edges. Because both the false-color visualization and the power transform are strictly invertible, their composition forms a bijection between metric depth in $[0, \infty]$ and RGB space in $[0, 1]^3$. During training, we apply this mapping to ground-truth metric depths to generate RGB training targets. At inference, we apply the inverse mapping to decode the model's generated RGB images back into metric depth, enabling direction evaluation on standard depth benchmarks. To enhance the model's robustness across diverse color representations, we augment our training data with alternative color maps, such as \textit{Plasma}, \textit{Inferno}, \textit{Viridis}, and \textit{grayscale}.

\begin{table}[t]
\centering
\resizebox{\textwidth}{!}{
\begin{tabular}{|l|c|c|c|c|c|c|c|}

\hline
\multicolumn{2}{|c|}{} & \makecell{\textbf{DepthLM-7B} \\ \tiny{\citep{depthlm}}} & \makecell{\textbf{Depth Any. v3}\\ \tiny{\citep{dav3}}} & \makecell{\textbf{Depth Pro}\\ \tiny{\citep{bochkovskii2024depthpro}}} & \makecell{\textbf{UniK3D}\\ \tiny{\citep{unik3d}}} & \makecell{\textbf{MoGe-2}\\ \tiny{\citep{wang2025moge2}}} &  \textbf{\makecell{Vision \\ Banana}} \emojivb \\ \hline

\multirow{2}{*}{\makecell[l]{\textbf{Camera}\\\textbf{Intrinsics}}} & \textbf{Inference} & \checkmark & \checkmark & & & &  \\ \cline{2-8}
 & \textbf{Training} & \checkmark & \checkmark & \checkmark & \checkmark & \checkmark &  \\ \hline

\textbf{Average} & $\delta_1 \uparrow$ & \textit{partial}* & \textit{partial}* & 0.715 & 0.823 & 0.802 & \textbf{\color{win_green}0.882} \\ \cline{2-8}
\textbf{Benchmarks} & AbsRel $\downarrow$ & & & & 0.156 & 0.144 & \textbf{\color{win_green}0.116} \\ \hline

\textbf{NYU} & $\delta_1 \uparrow$ & 0.915 & 0.963 & 0.961 & \textbf{\color{win_green}0.965} & 0.961 & 0.948 \\ \cline{2-8}
{\tiny \citep{nyuv2}} & AbsRel $\downarrow$ & & \textbf{\color{win_green}0.07} & & 0.074 & 0.0733 & 0.081 \\ \hline

\textbf{iBims1} & $\delta_1 \uparrow$ & 0.92 & & 0.913 & 0.919 & 0.830 & \textbf{\color{win_green}0.934} \\ \cline{2-8}
{\tiny \citep{koch2018evaluation_ibims1}} & AbsRel $\downarrow$ & & & & 0.104 & 0.136 & \textbf{\color{win_green}0.078} \\ \hline

\textbf{ETH3D} & $\delta_1 \uparrow$ & 0.718 & 0.917 & 0.415 & 0.687 & 0.908 & \textbf{\color{win_green}0.935} \\ \cline{2-8}
{\tiny \citep{eth3d}} & AbsRel $\downarrow$ & & 0.104 & 0.327 & 0.236 & 0.104 & \textbf{\color{win_green}0.103} \\ \hline

\textbf{DIODE-Indoor} & $\delta_1 \uparrow$ & & 0.838 & 0.671 & 0.713 & 0.664 & \textbf{\color{win_green}0.917} \\ \cline{2-8}
{\tiny \citep{diode_dataset}} & AbsRel $\downarrow$ & & 0.123 & 0.199 & 0.161 & 0.175 & \textbf{\color{win_green}0.108} \\ \hline

\textbf{KITTI} & $\delta_1 \uparrow$ & & \textbf{\color{win_green}0.953} & 0.843$\ddagger$ & 0.812 & 0.629 & 0.915 \\ \cline{2-8}
{\tiny \citep{Uhrig2017THREEDV_kitti}} & AbsRel $\downarrow$ & & \textbf{\color{win_green}0.086} & 0.121$\ddagger$ & 0.174 & 0.181 & 0.107 \\ \hline

\textbf{nuScenes} & $\delta_1 \uparrow$ & \textit{\color{win_green}0.865}$\dagger$ & & 0.491 & 0.840 & 0.820 & 0.643 \\ \cline{2-8}
{\tiny \citep{caesar2020nuscenes}} & AbsRel $\downarrow$ & & & 0.287 & \textbf{\color{win_green}0.189} & 0.195 & 0.219 \\ \hline

\end{tabular}
}
\raggedright
\footnotesize{
* The average $\delta_1$ of DepthLM-7B on the 4 datasets it evaluated on (NYU + iBims1 + ETH3D + nuScenes) is $0.855$; our average $\delta_1$ on the same 4 datasets is $0.865$. The average $\delta_1$ of Depth-Anything V3 on the 4 datasets it evaluated on (NYU + ETH3D + DIODE + KITTI) is $0.918$; our average $\delta_1$ on the same 4 datasets is $0.929$. \\
$\dagger$ DepthLM is trained on nuScenes so it’s not zero-shot. \\
$\ddagger$ Numbers reported by Depth-Anything V3 \citep{dav3}. }

\caption{Monocular metric depth estimation under the zero-shot transfer setting. Vision Banana achieves superior results on public datasets without using camera intrinsics in neither training of inference. 
Metrics marked with $\uparrow$ are better if higher; metrics marked with $\downarrow$ are better if lower.
}
\label{tab:depth}
\end{table}

\vspace{3mm}

Tab. \ref{tab:depth} presents the empirical results of Vision Banana compared to specialist models across six major academic benchmarks. Vision Banana achieves an average $\delta_1$ accuracy of 0.882, outperforming Unik3D~\citep{unik3d} by nearly 6 points, while achieving a 20\% lower absolute relative error (AbsRel) compared to MoGe-2~\citep{wang2025moge2}. 
Notably, Vision Banana outperforms Depth Anything V3 \citep{dav3} on average across the four datasets (NYU, ETH3D, DIODE, KITTI) on which it was evaluated ($0.929$ v.s. $0.918$), demonstrating robust performance in both near-field and distant scenes.
Our model is trained entirely on synthetic depth data created from simulation engines --- we use zero real-world depth data, and exclude training data from any of the depth datasets we evaluate on.
Note that this result is achieved without relying on camera parameters (neither intrinsics nor extrinsics) during \emph{both} training or inference. 
By leveraging the immense geometric priors embedded in its foundation model, Vision Banana infers absolute scale solely from visual cues and object relationships, enabling zero-shot generalization to any arbitrary input image.

\begin{figure}[!p]
    \centering
    \begin{tabular}{@{} c @{\hspace{2pt}} c @{\hspace{2pt}} c @{\hspace{2pt}} c @{}}
        
        \includegraphics[width=0.24\textwidth]{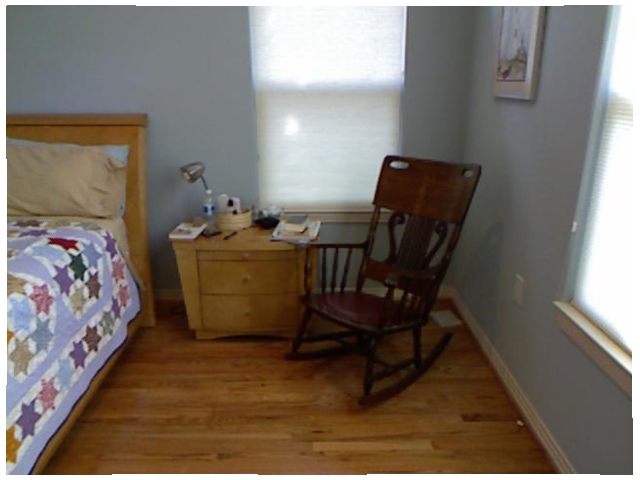} &
        \includegraphics[width=0.24\textwidth]{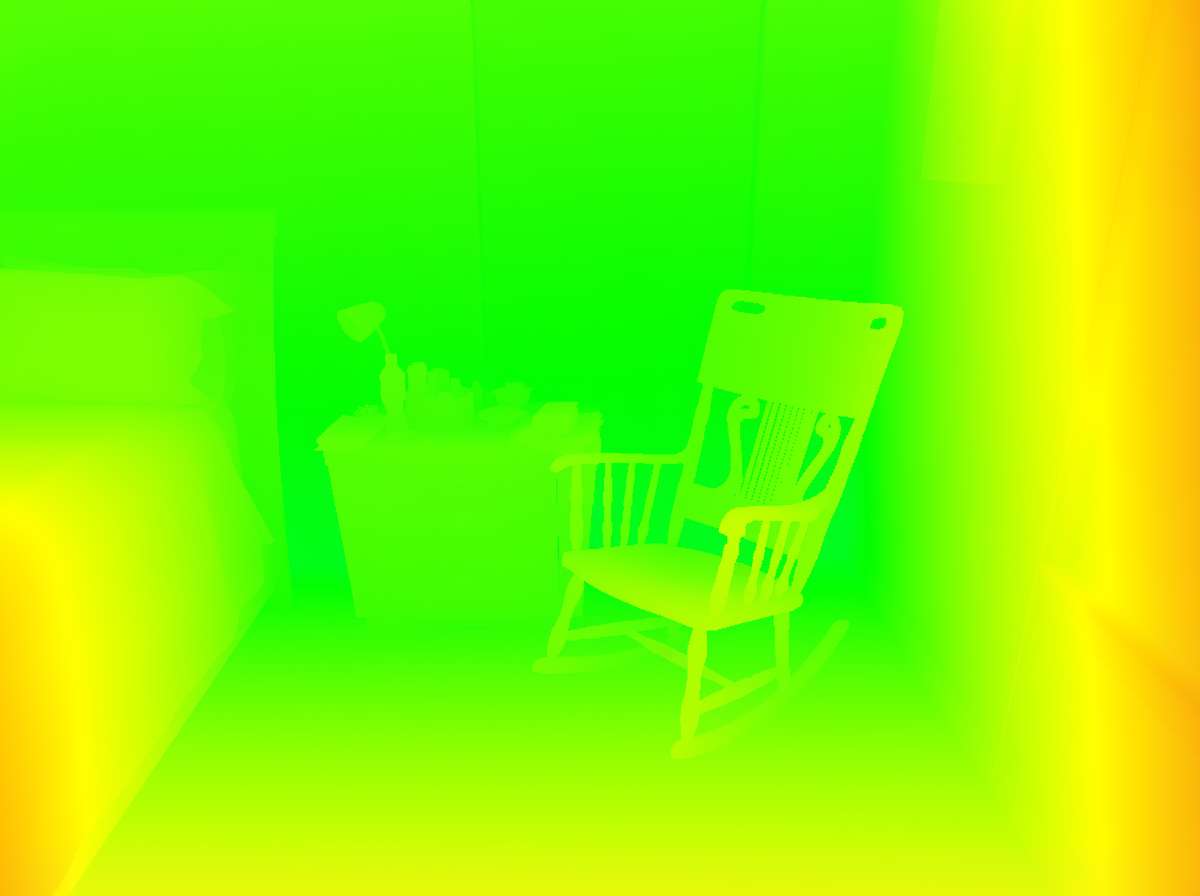} &
        \includegraphics[width=0.24\textwidth]{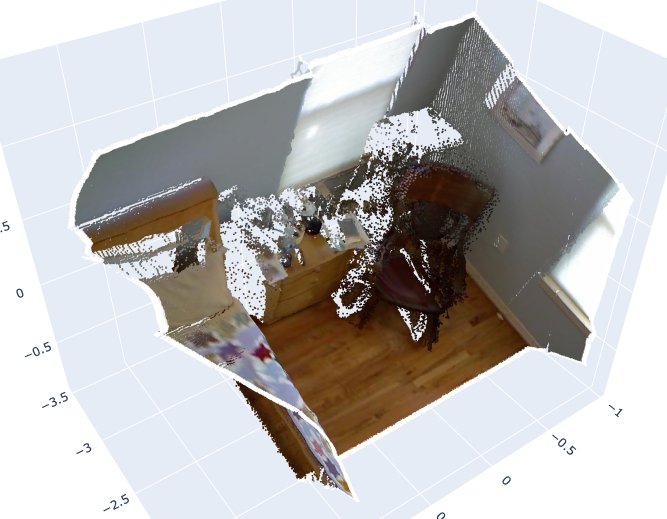} &
        \includegraphics[width=0.24\textwidth]{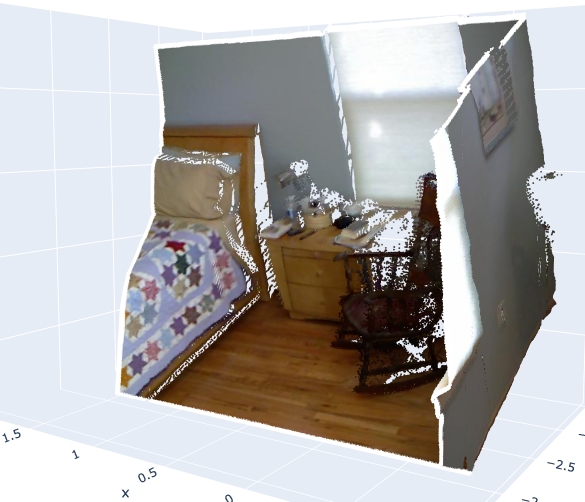} \\[2pt]
        
        \includegraphics[width=0.24\textwidth]{assets/depth_demo/depth_162_0.jpg} &
        \includegraphics[width=0.24\textwidth]{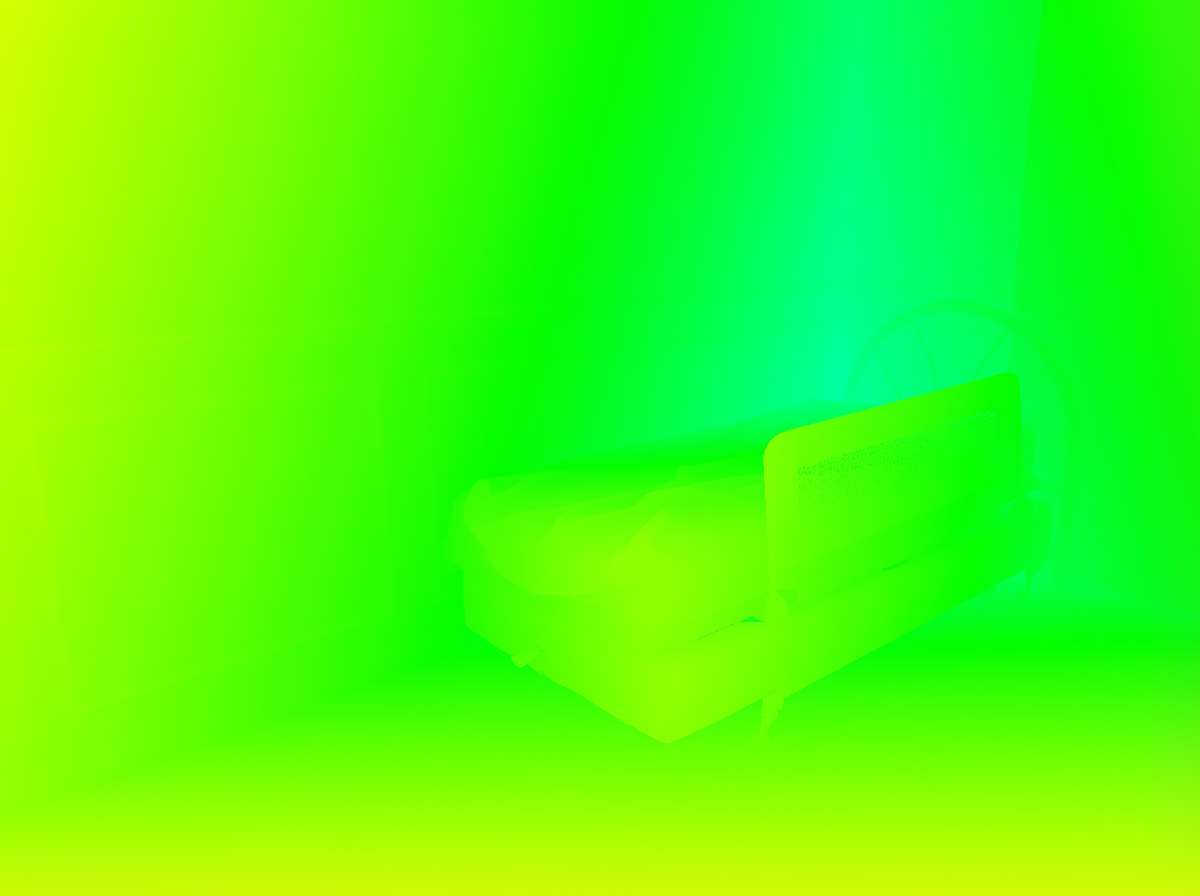} &
        \includegraphics[width=0.24\textwidth]{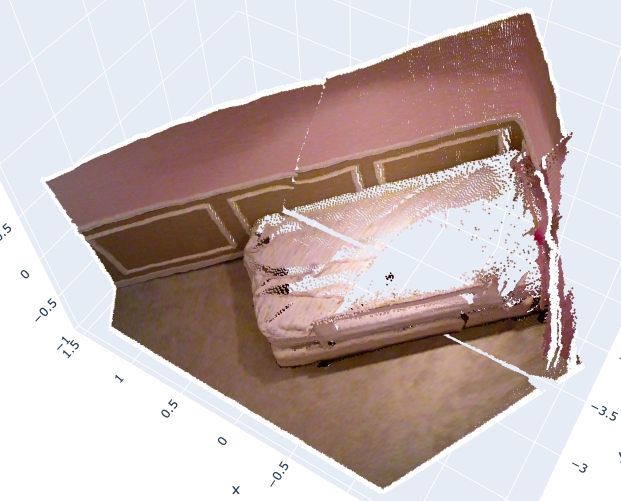} &
        \includegraphics[width=0.24\textwidth]{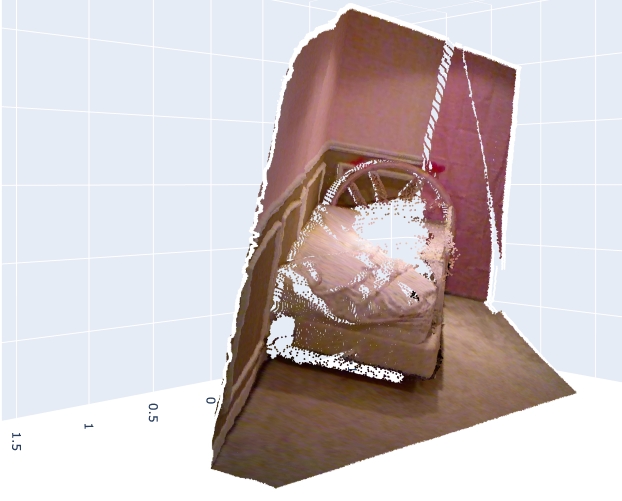} \\[2pt]
        
        \includegraphics[width=0.24\textwidth]{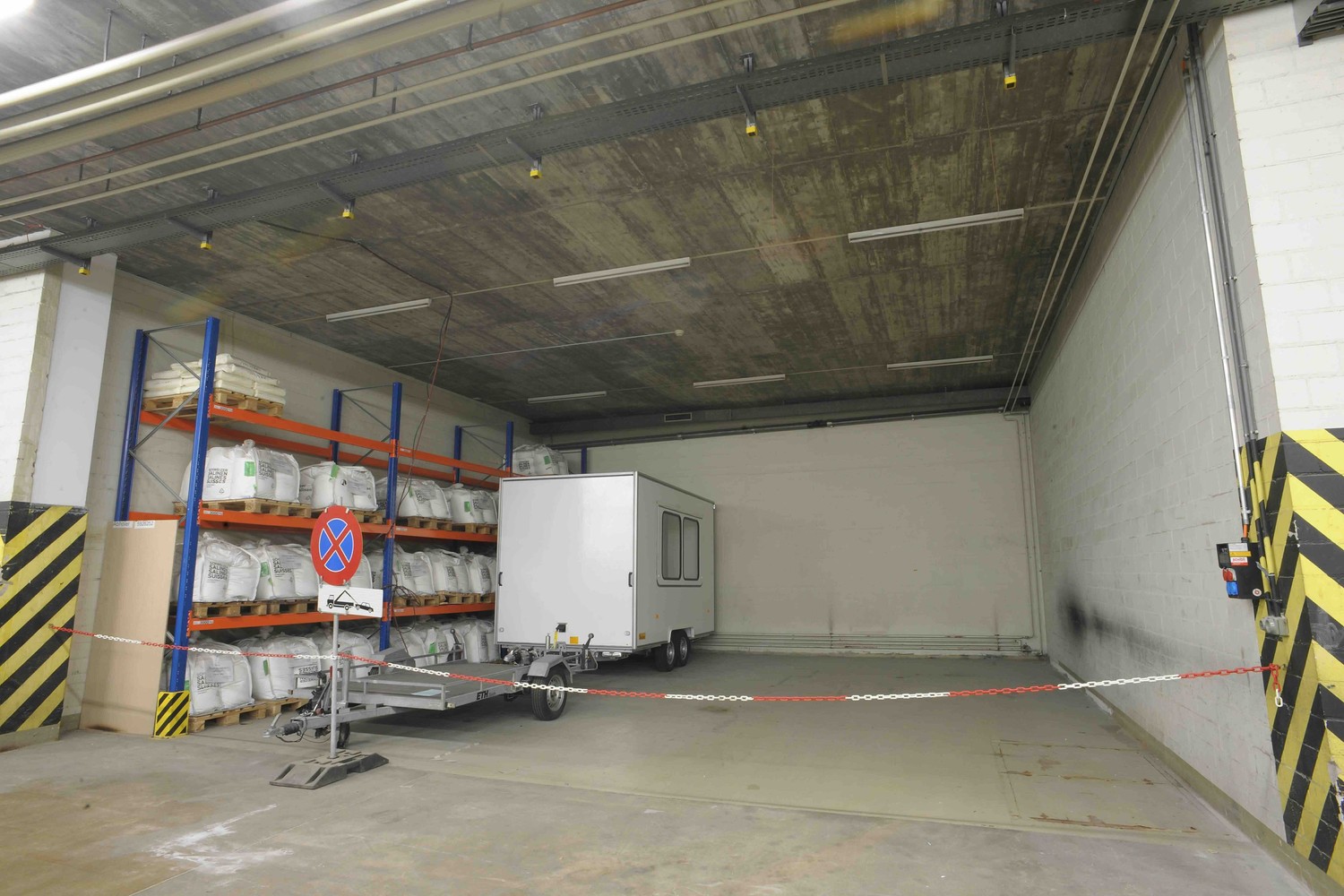} &
        \includegraphics[width=0.24\textwidth]{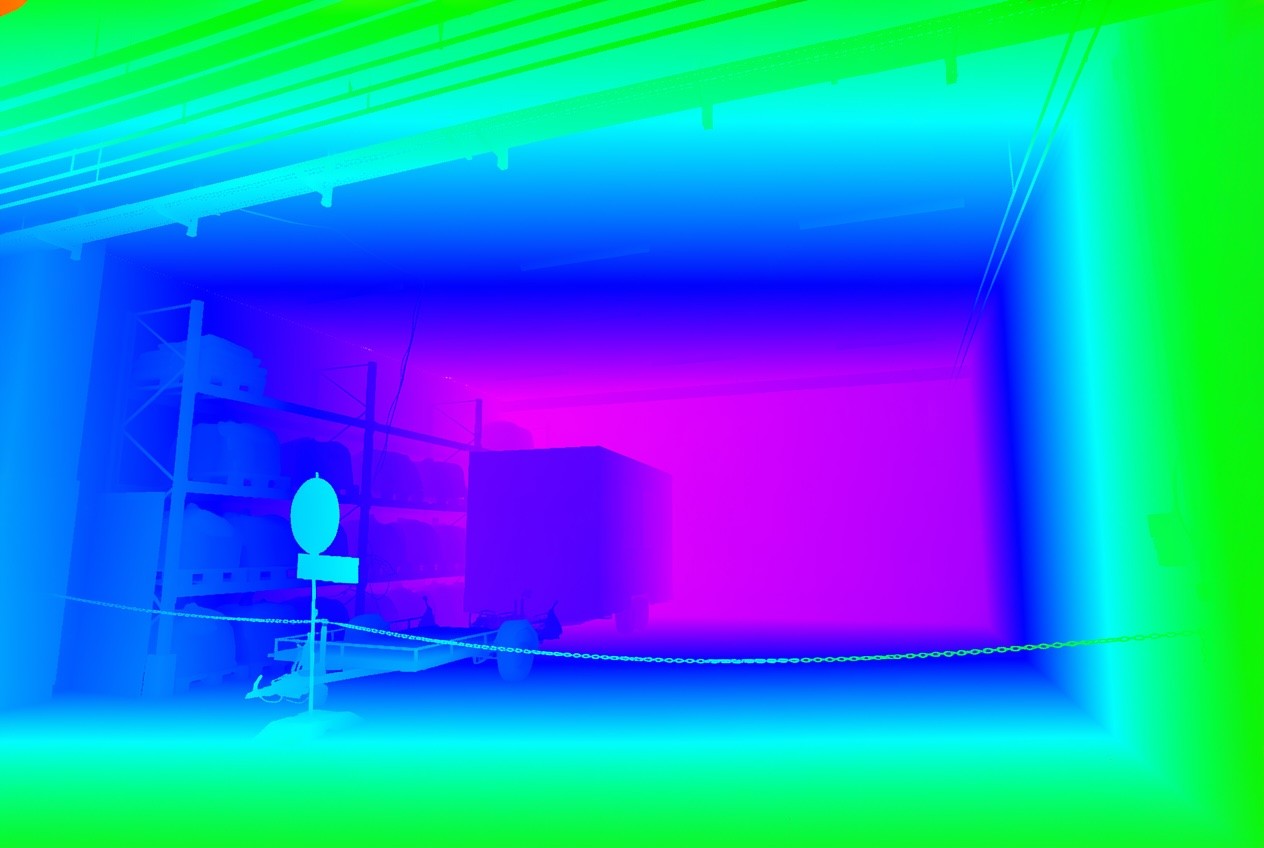} &
        \includegraphics[width=0.24\textwidth]{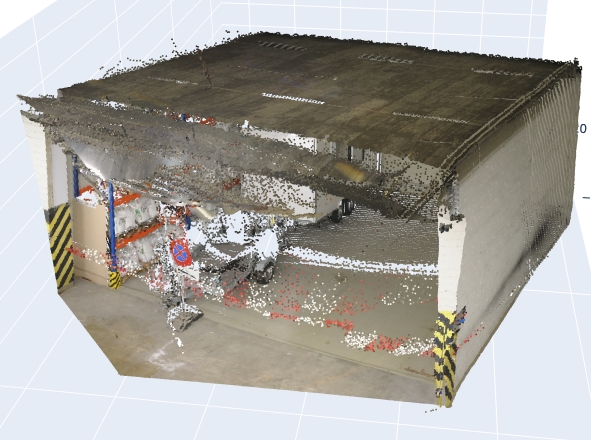} &
        \includegraphics[width=0.24\textwidth]{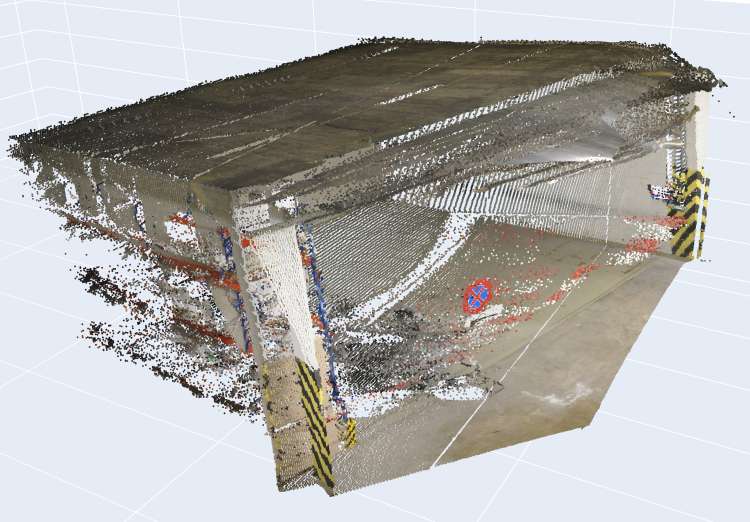} \\[2pt]
        
        \includegraphics[width=0.24\textwidth]{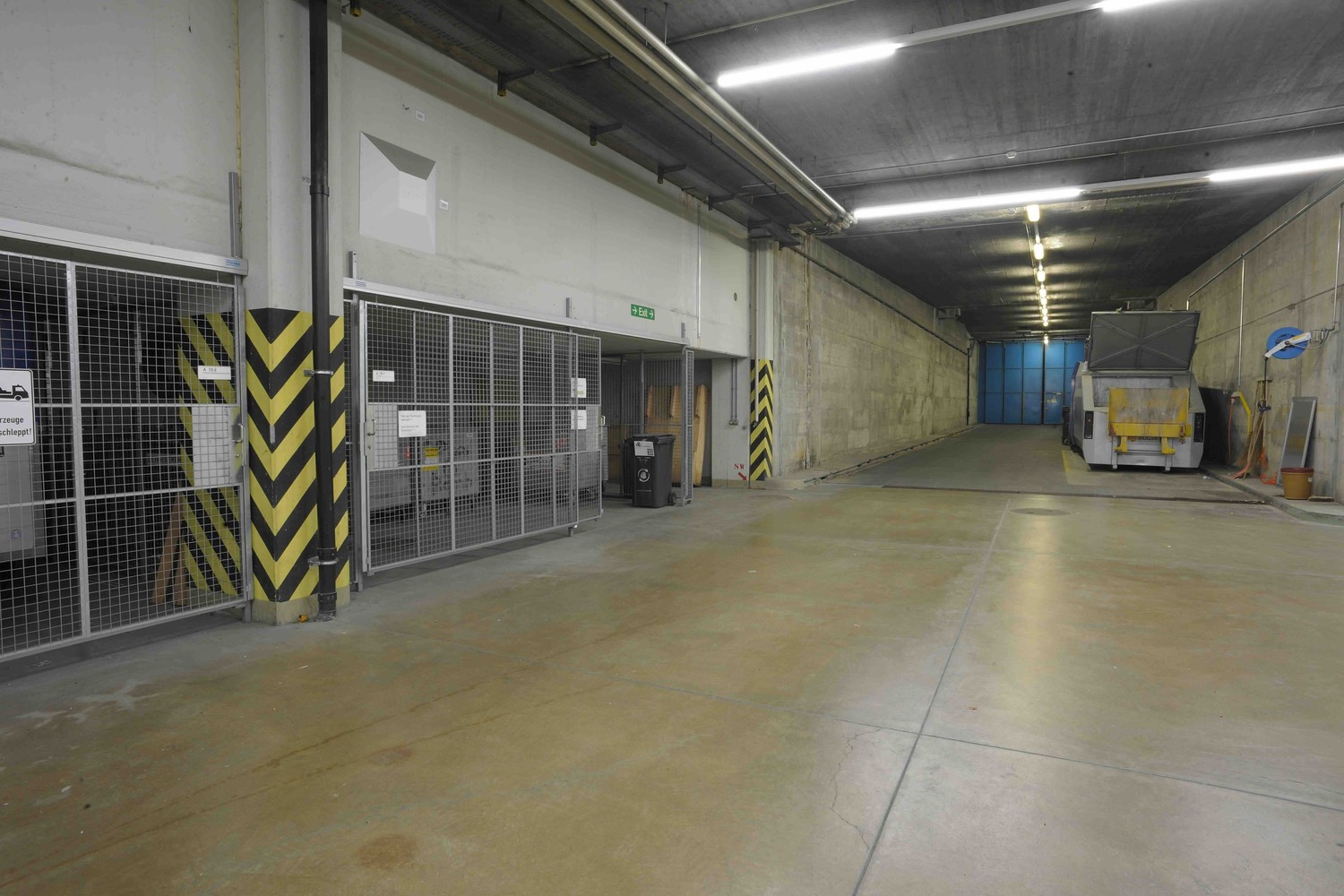} &
        \includegraphics[width=0.24\textwidth]{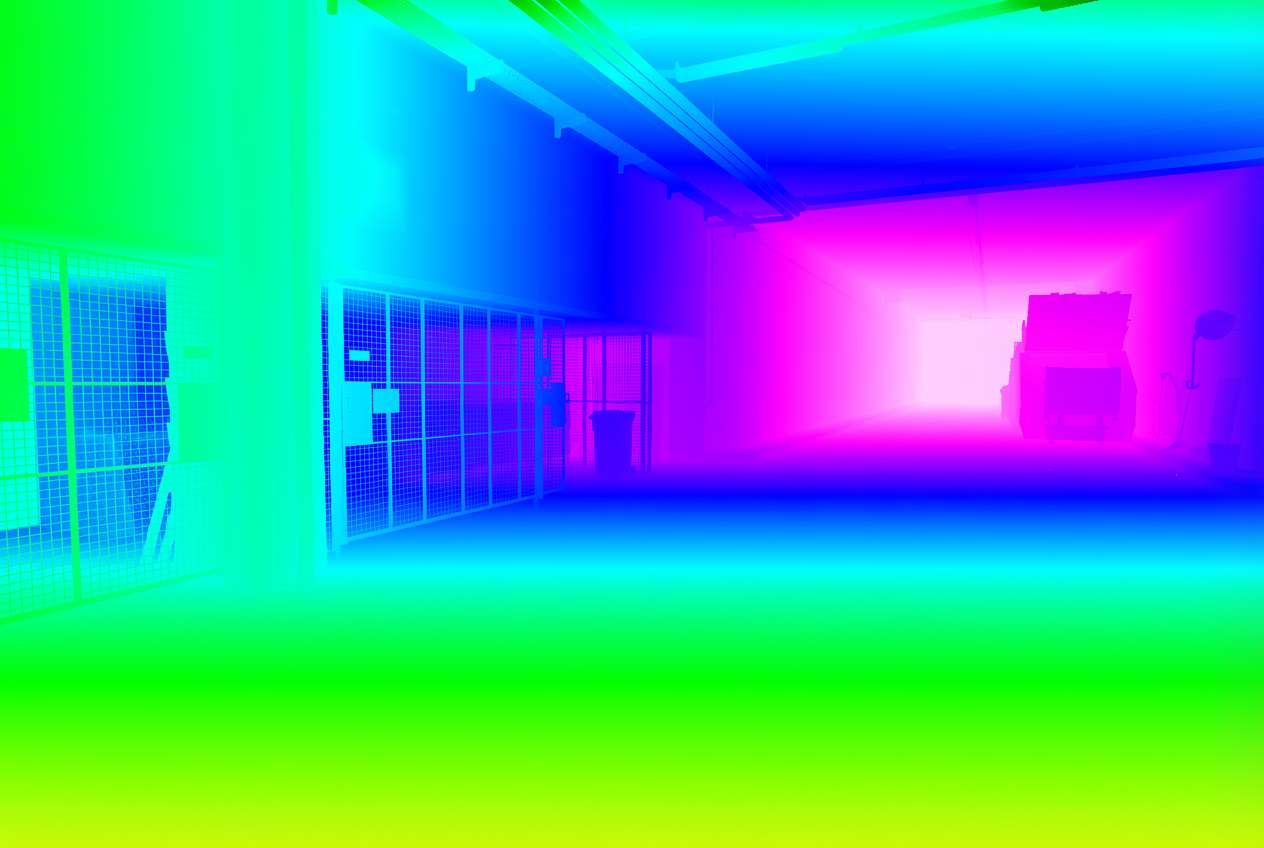} &
        \includegraphics[width=0.24\textwidth]{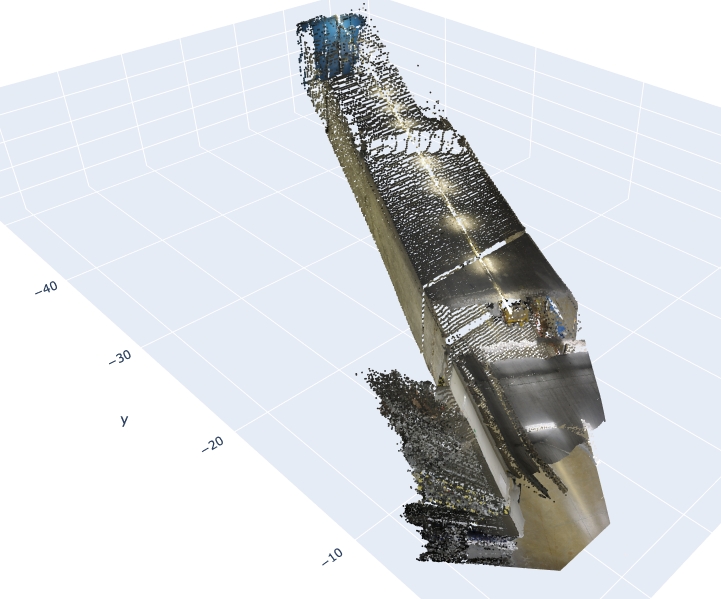} &
        \includegraphics[width=0.24\textwidth]{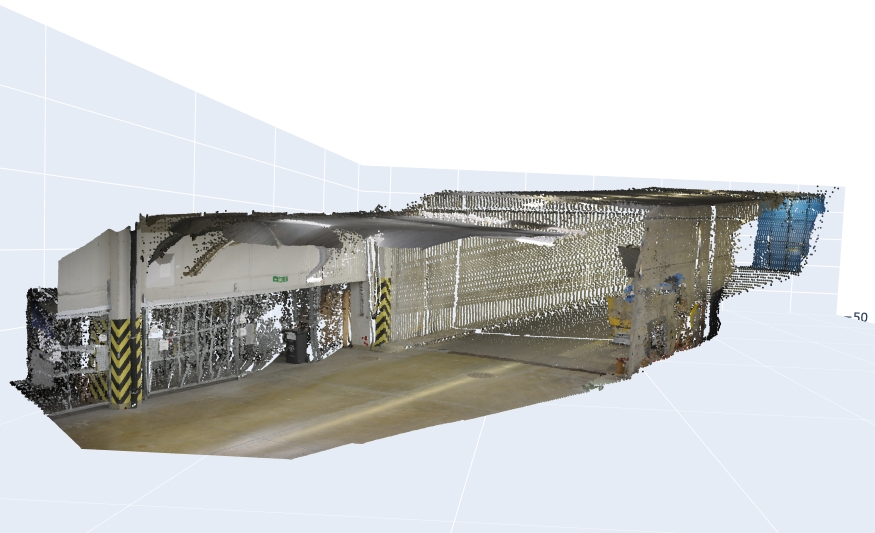} \\[2pt]
        
        \includegraphics[width=0.24\textwidth]{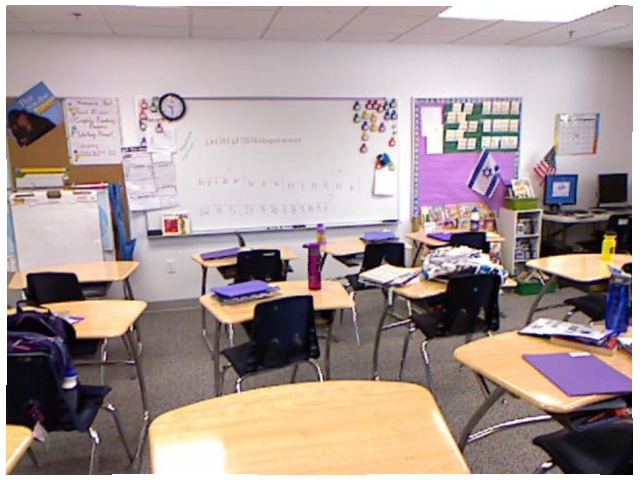} &
        \includegraphics[width=0.24\textwidth]{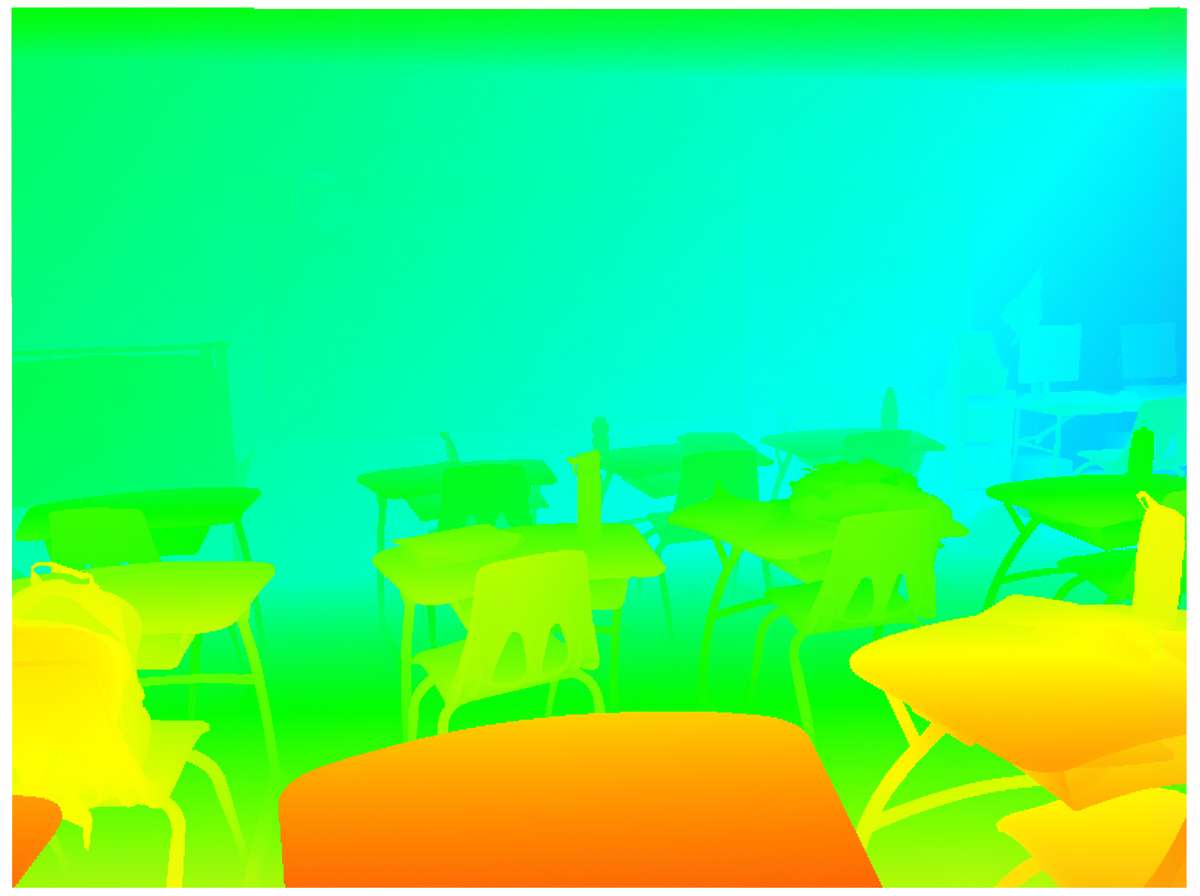} &
        \includegraphics[width=0.24\textwidth]{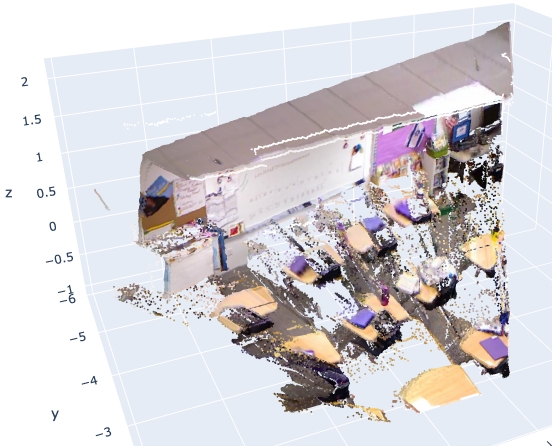} &
        \includegraphics[width=0.24\textwidth]{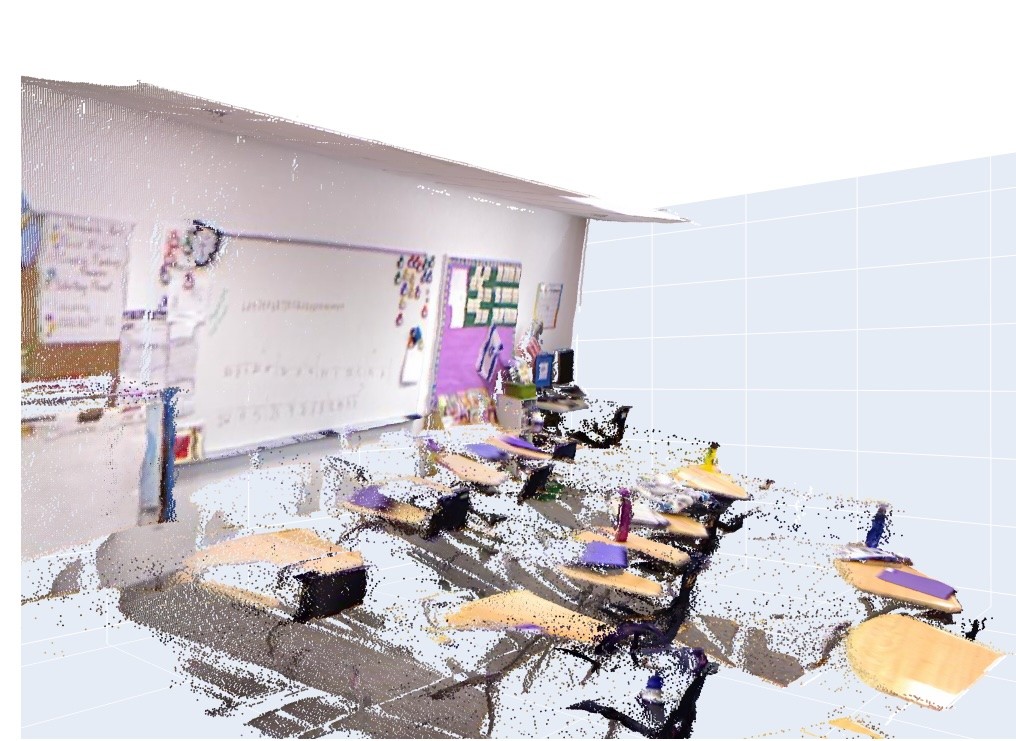} \\
        \small Input image & \small Generated depth image & \small Vis. view 1 & \small Vis. view 2
    \end{tabular}
    \caption{
    Demonstration of Vision Banana's metric depth estimation capabilities. 
    The two columns to the left are the input images and the depth visualization image generated by Vision Banana.
    The depth images are then decoded back to metric depth values. Combining them with the camera intrinsics, we can reconstruct the 3D scene accurately. The two columns on the right are random views of the reconstructed scenes.
    Note that camera intrinsics are not needed in predicting the depth itself.
    Samples taken from NYU v2 \citep{nyuv2} and ETH 3D \citep{eth3d}.
    }
    \label{fig:depth-demo1}
\end{figure}

\begin{figure}[!t]
    \centering
    \begin{subfigure}{0.32\textwidth}
        \centering
        \includegraphics[width=\textwidth]{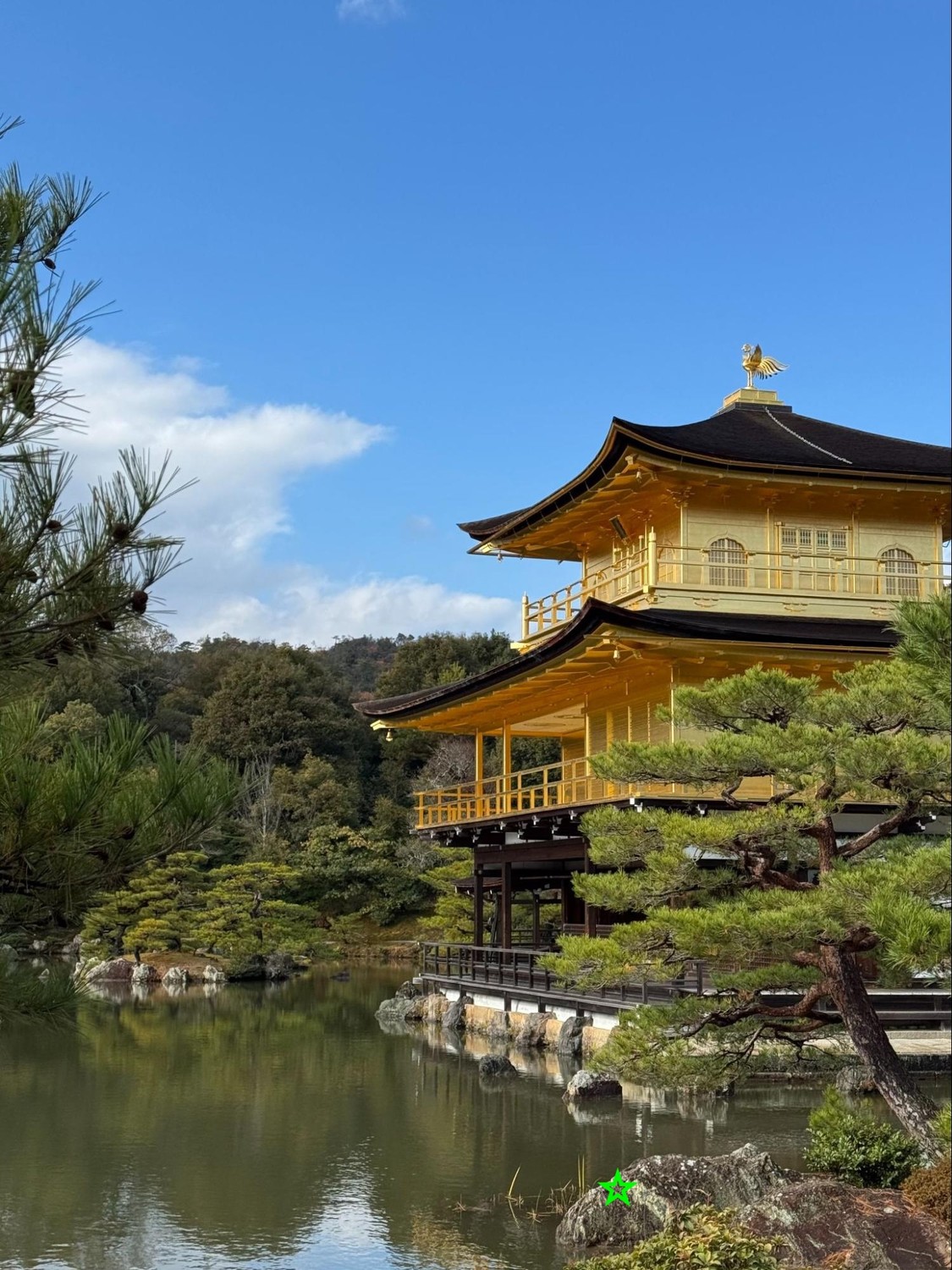}
        \caption{Photo taken at Kinkaku-Ji}
    \end{subfigure}
    \hfill
    \begin{subfigure}{0.32\textwidth}
        \centering
        \includegraphics[width=\textwidth]{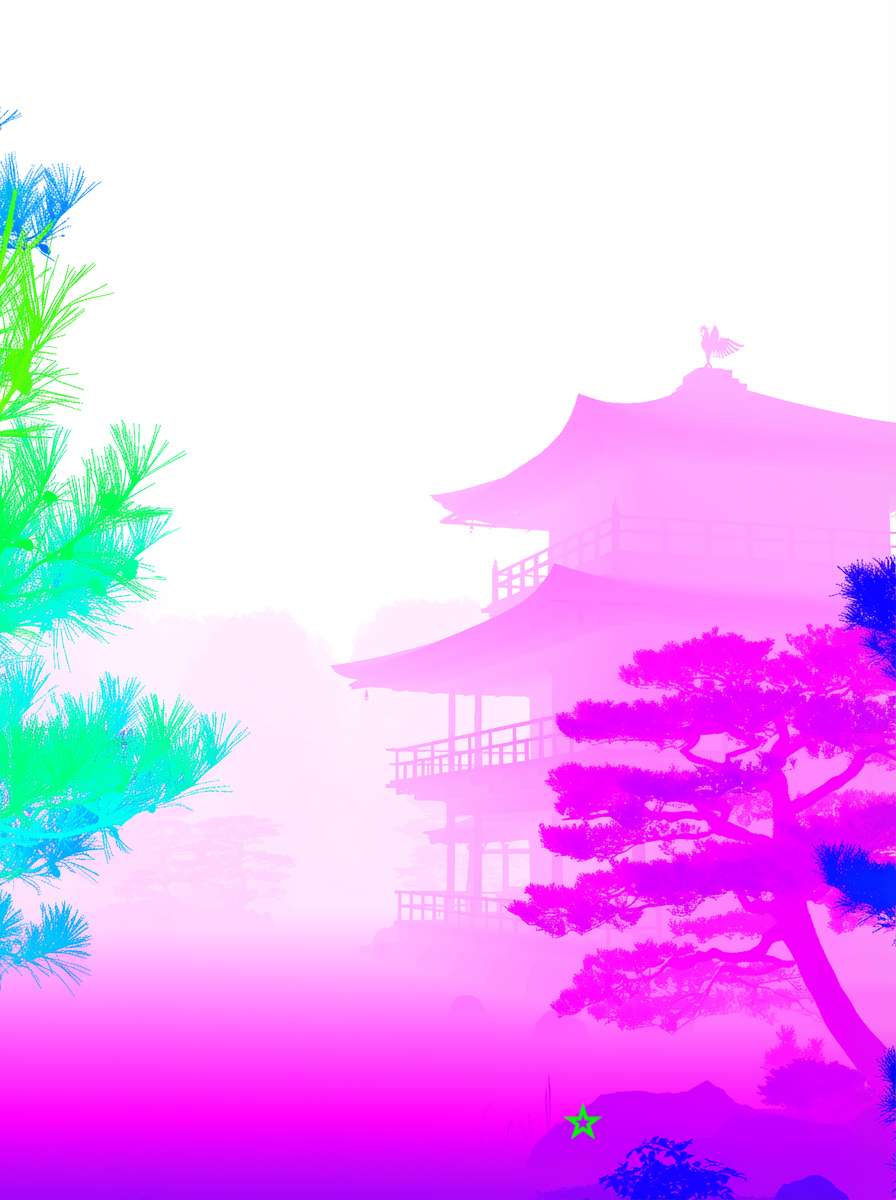}
        \caption{Vision Banana estimated depth}
    \end{subfigure}
    \hfill
    \begin{subfigure}{0.32\textwidth}
        \centering
        \includegraphics[width=\textwidth]{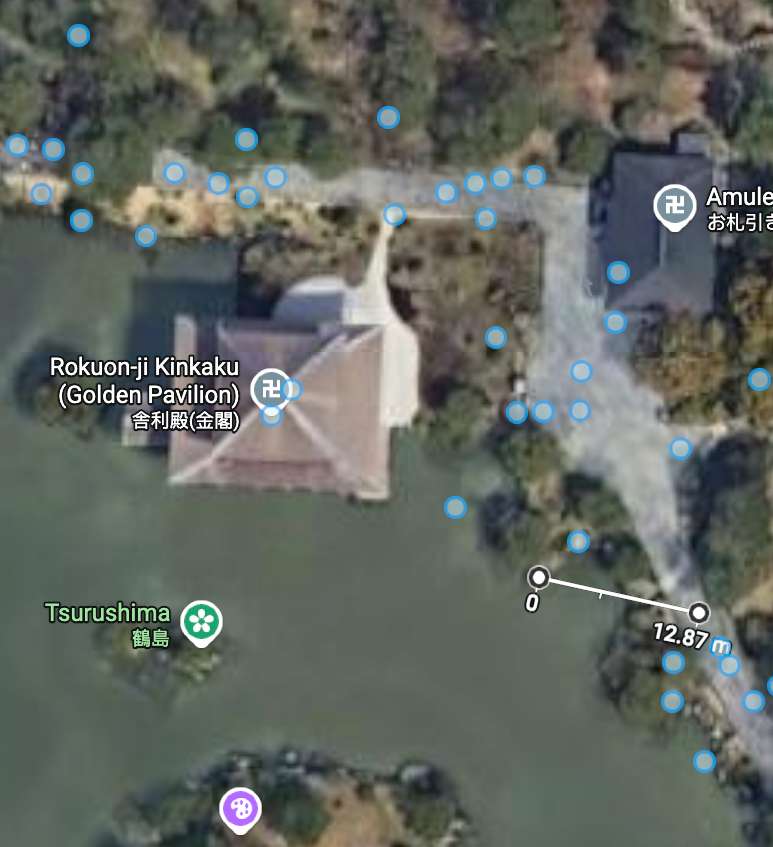}
        \caption{Measurement from Google Maps}
    \end{subfigure}

    \caption{Vision Banana depth estimation in the wild. (a) Author of this paper takes a picture near Kinkaku-Ji with a consumer cell-phone. (b) Vision Banana generates a depth estimation image. The depth value at the position marked by a green star is decoded to be $13.71$ meters. (c) Author then measures the actual distance using Google Map, which turns out to be 12.87 meters. The AbsRel error at this point is around $0.065$.}
    \label{fig:depth-kinkakuji}
\end{figure}

Qualitative inspections further validate the model's capabilities. As illustrated in Fig. ~\ref{fig:depth-demo1}, Vision Banana generates highly precise depth maps that preserve crisp geometric details, even in cluttered environments like classrooms. 
When these 2D predictions are unprojected into 3D point clouds, they exhibit global consistency across diverse scenes, maintaining accurate planar surfaces and correct geometry. 
In addition to common academic benchmarks, we also conducted a ``vibe test'' using a casual smartphone photograph, as shown in Fig. \ref{fig:depth-kinkakuji}. Crossed validated by depth measured on Google Maps, Vision Banana successfully produced an accurate depth estimation on this photo captured by a consumer device unseen during training.

\newpage
\paragraph{Surface Normal Estimation.}

Surface normal estimation represents another critical vision task. Surface normals, which are unit vectors $(x, y, z)$ with values ranging from $-1.0$ to $1.0$, serve as a critical proxy for local geometry and scene structures. 
Unlike the complex color mapping required for metric depth, the visualization of surface normals is intrinsically aligned with the RGB color space, allowing straightforward integration into our model.

We specifically utilize a camera-space normal formulation using the standard right-handed coordinate system (+x right, +y up, +z pointing out of the image plane). In this representation, the directional vector components map directly to RGB channels, i.e. $R=trunc( (1-x)/2, min=0, max=1)\times 255$, $G=trunc( (1+y)/2, min=0, max=1)\times 255$, $B=trunc( (1+z)/2, min=0, max=1)\times 255$:

\begin{itemize}
    \item Facing Left $(-1, 0, 0)$: Encoded as  Pinkish Red.
    \item Facing Up $(0, 1, 0)$: Encoded as Light Green.
    \item Facing the Camera $(0, 0, 1)$: Encoded as Light Blue.
\end{itemize}

Table~\ref{tab:surface_normal} compares Vision Banana against SOTA specialist methods on four public benchmarks. When averaged across the three indoor datasets, Vision Banana achieves the lowest mean and median angular errors. It also demonstrates competitive accuracy on outdoor scenes.

\begin{table}[!t]
\centering
\resizebox{\textwidth}{!}{
\begin{tabular}{l cc cc cc cc cc}
\toprule
\multirow{3}{*}{Methods} & \multicolumn{2}{c}{Indoor} & \multicolumn{6}{c}{Indoor} & \multicolumn{2}{c}{Outdoor} \\
\cmidrule(lr){2-3} \cmidrule(lr){4-9} \cmidrule(lr){10-11}
 & \multicolumn{2}{c}{Average} & \multicolumn{2}{c}{\makecell{NYUv2 \\ {\scriptsize \citep{nyuv2}}}} & \multicolumn{2}{c}{\makecell{DIODE-indoor \\ {\scriptsize \citep{diode_dataset}}}} & \multicolumn{2}{c}{\makecell{ScanNet \\ {\scriptsize \citep{dai2017scannet}}}} & \multicolumn{2}{c}{\makecell{VKitti \\ {\scriptsize \citep{cabon2020virtualkitty}}}} \\
\cmidrule(lr){2-3} \cmidrule(lr){4-5} \cmidrule(lr){6-7} \cmidrule(lr){8-9} \cmidrule(lr){10-11}
 & mean $\downarrow$ & median $\downarrow$ & mean $\downarrow$ & median $\downarrow$ & mean $\downarrow$ & median $\downarrow$ & mean $\downarrow$ & median $\downarrow$ & mean $\downarrow$ & median $\downarrow$ \\
\midrule
Marigold {\scriptsize \citep{marigold}} & 19.606 & 11.828 & 20.864 & 11.134 & 16.671 & 12.084 & 21.284 & 12.268 & -- & -- \\
DSINE {\scriptsize \citep{DSINE}} & 17.017 & 10.190 & \textbf{\color{win_green}16.4} & \textbf{\color{win_green}8.4} & 18.453 & 13.871 & 16.2 & 8.3 & 28.9 & 9.9 \\
StableNormal {\scriptsize \citep{ye2024stablenormal}} & 17.168 & 10.028 & 19.707 & 10.527 & \textbf{\color{win_green}13.701} & \textbf{\color{win_green}9.46} & 18.098 & 10.097 & -- & -- \\
Lotus-2-Normal {\scriptsize \citep{lotus2}} & 16.558 & -- & 16.9 & N/A & 18.575 & N/A & \textbf{\color{win_green}14.2} & N/A & \textbf{\color{win_green}28.894} & \textbf{\color{win_green}9.677} \\ \hline
 \textbf{Vision Banana} \emojivb & \textbf{\color{win_green}15.549} & \textbf{\color{win_green}9.300} & 17.778 & 8.876 & 13.818 & 11.556 & 15.052 & \textbf{\color{win_green}7.468} & 29.063 & 10.699 \\
\bottomrule
\end{tabular}
}
\caption{Surface normal estimation results. Vision Banana achieves the lowest mean and median angle errors on the indoor datasets on average, and is on par with previous SOTA on outdoor scenes.}
\label{tab:surface_normal}
\end{table}

\begin{figure}[!t]
    \centering
    \begin{tabular}{@{} c @{\hspace{2pt}} c @{\hspace{2pt}} c @{}}

        \includegraphics[width=0.32\textwidth]{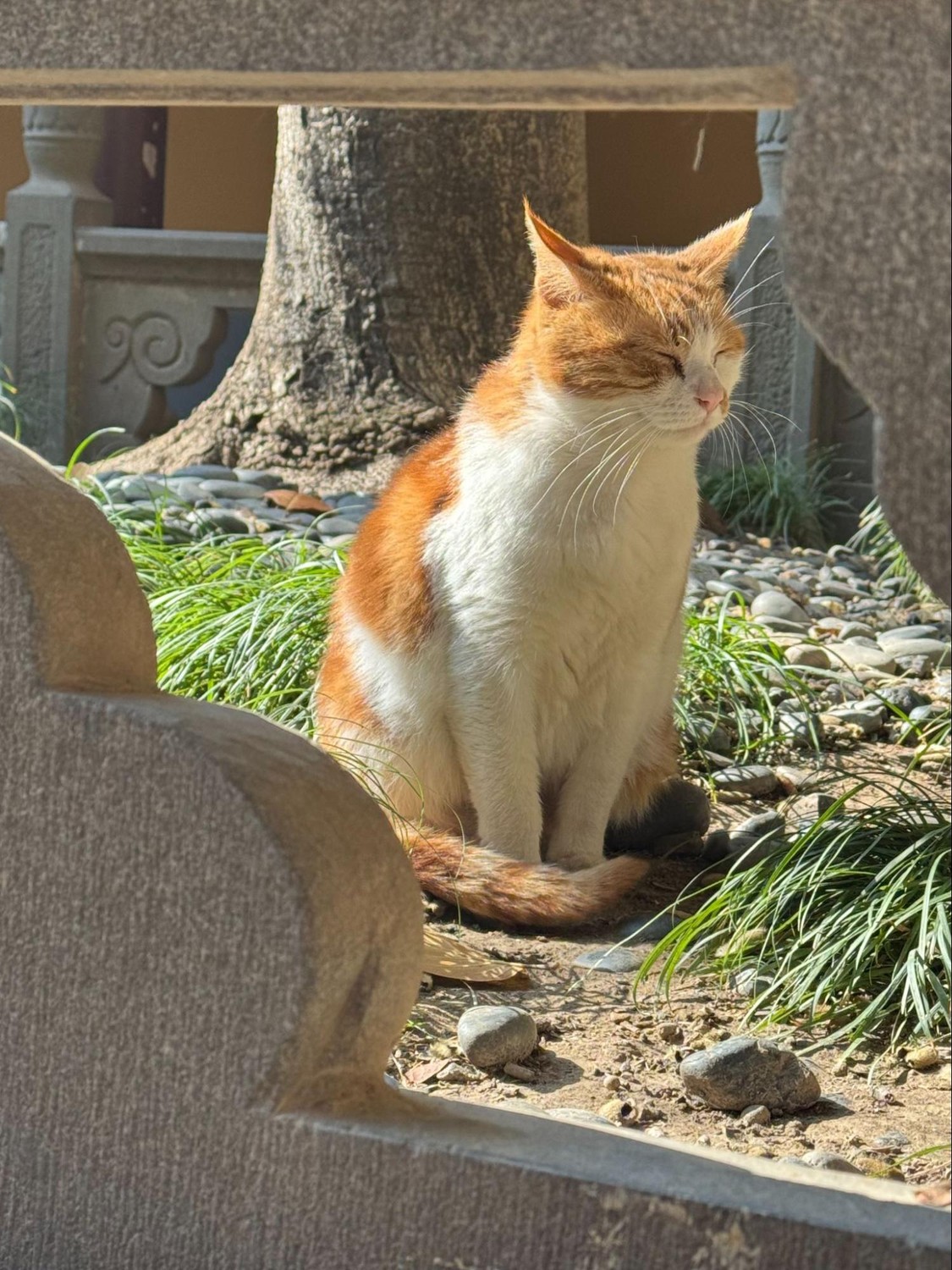} &
        \includegraphics[width=0.32\textwidth]{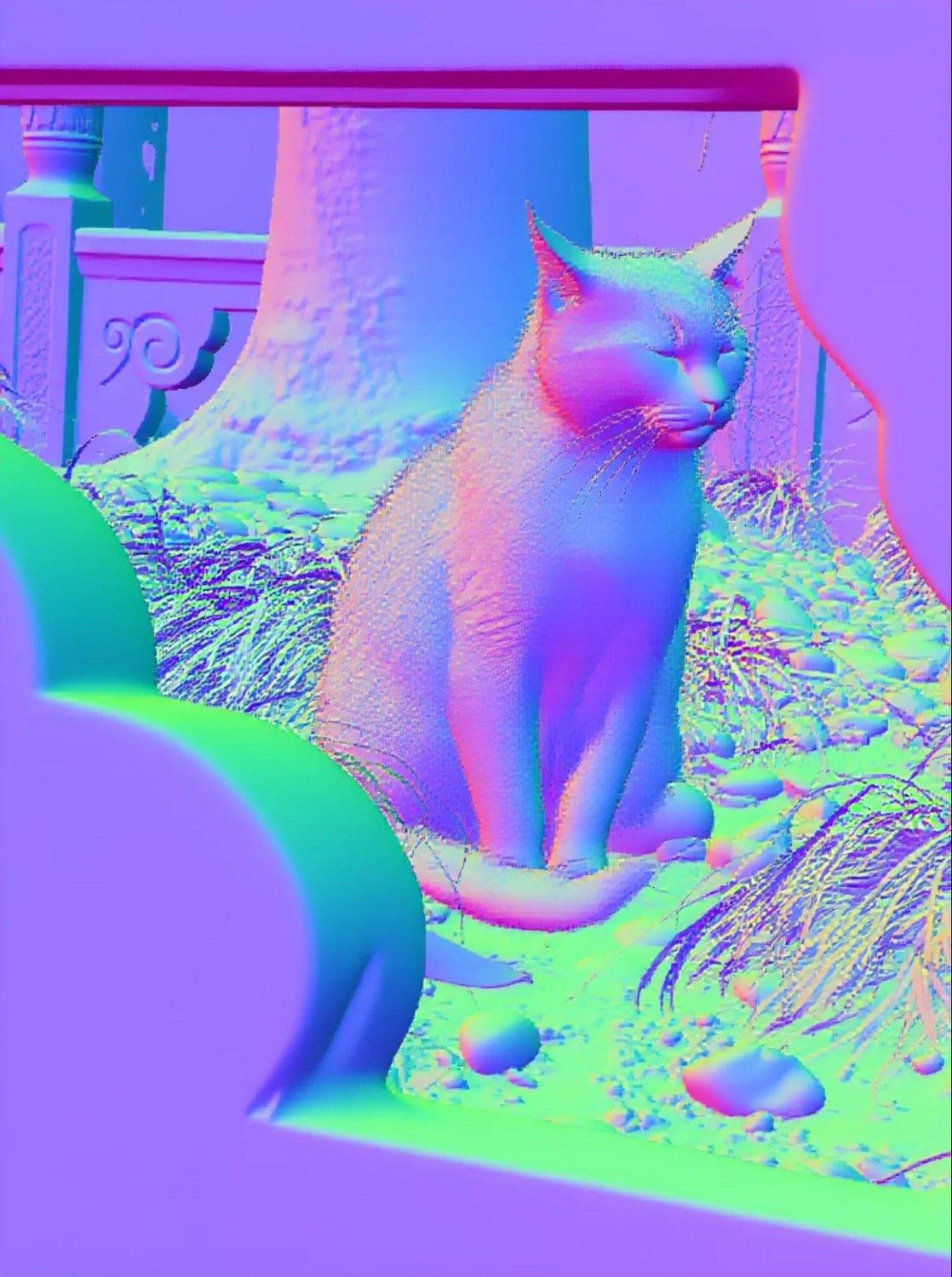} &
        \includegraphics[width=0.32\textwidth]{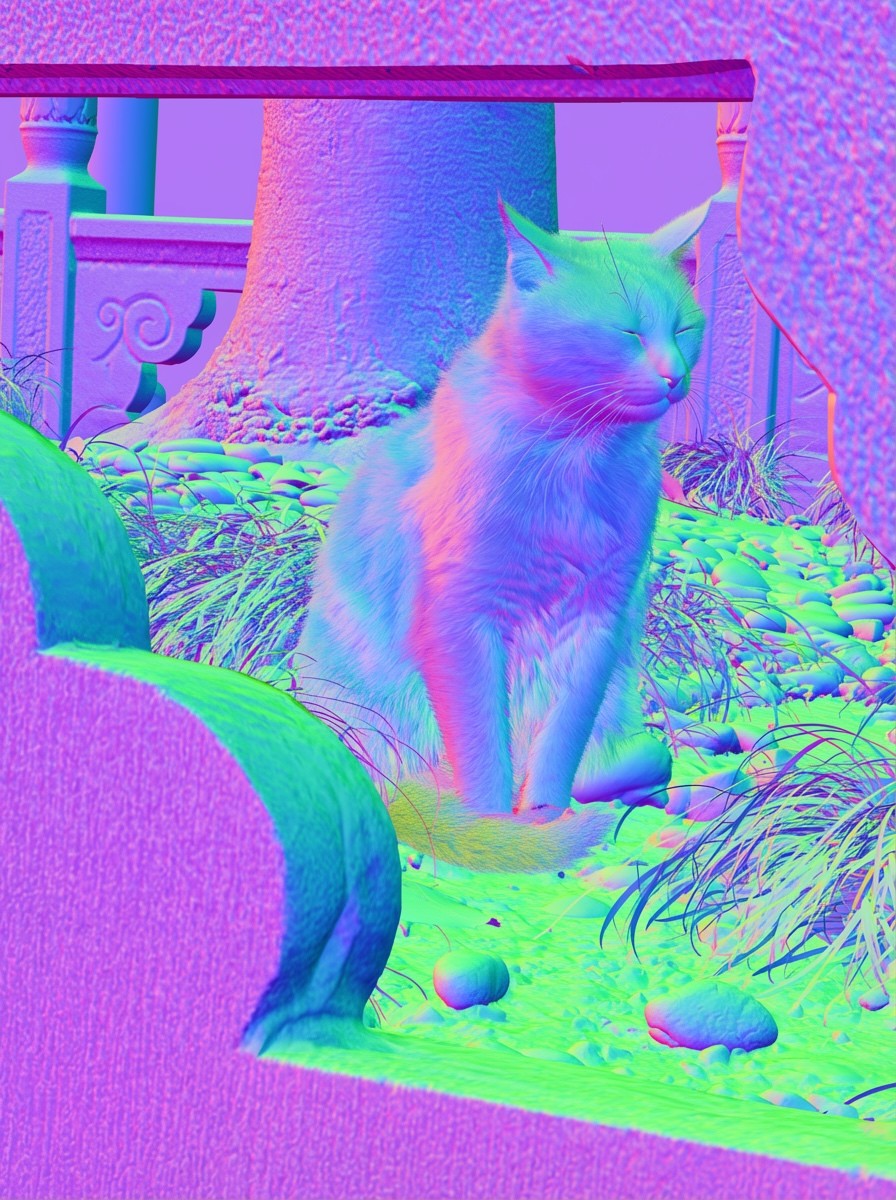} \\[2pt]
        
        \includegraphics[width=0.32\textwidth]{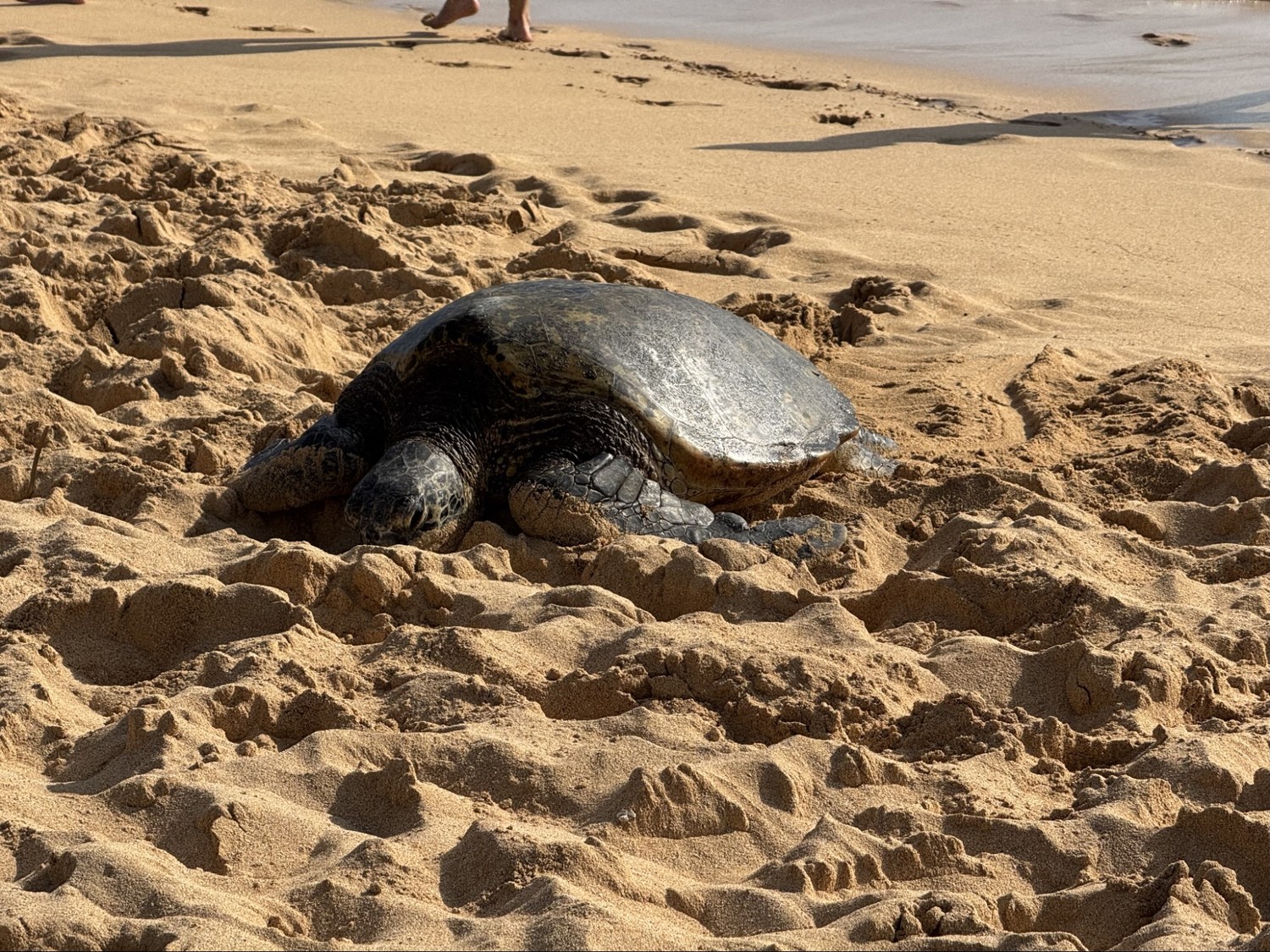} &
        \includegraphics[width=0.32\textwidth]{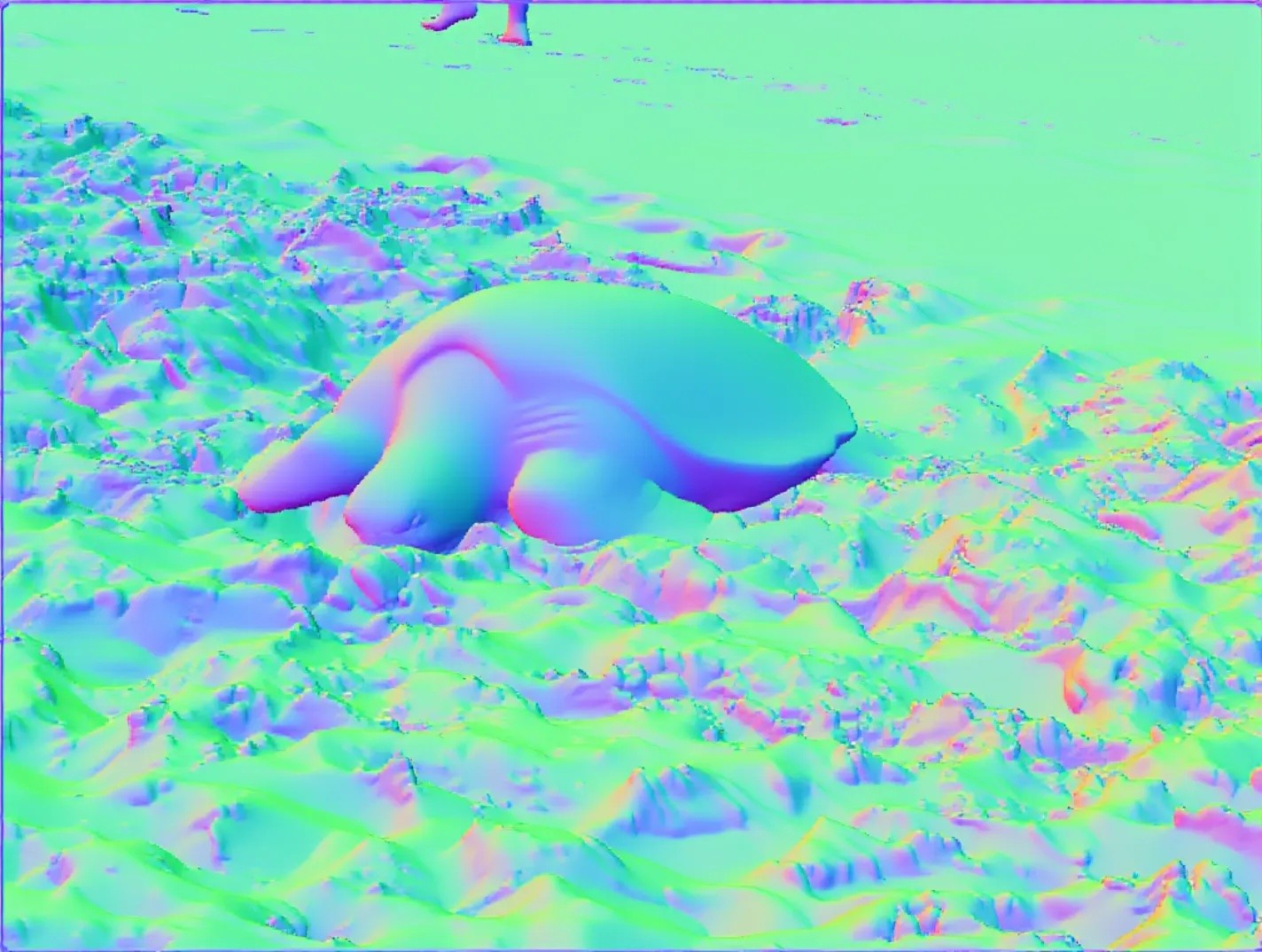} &
        \includegraphics[width=0.32\textwidth]{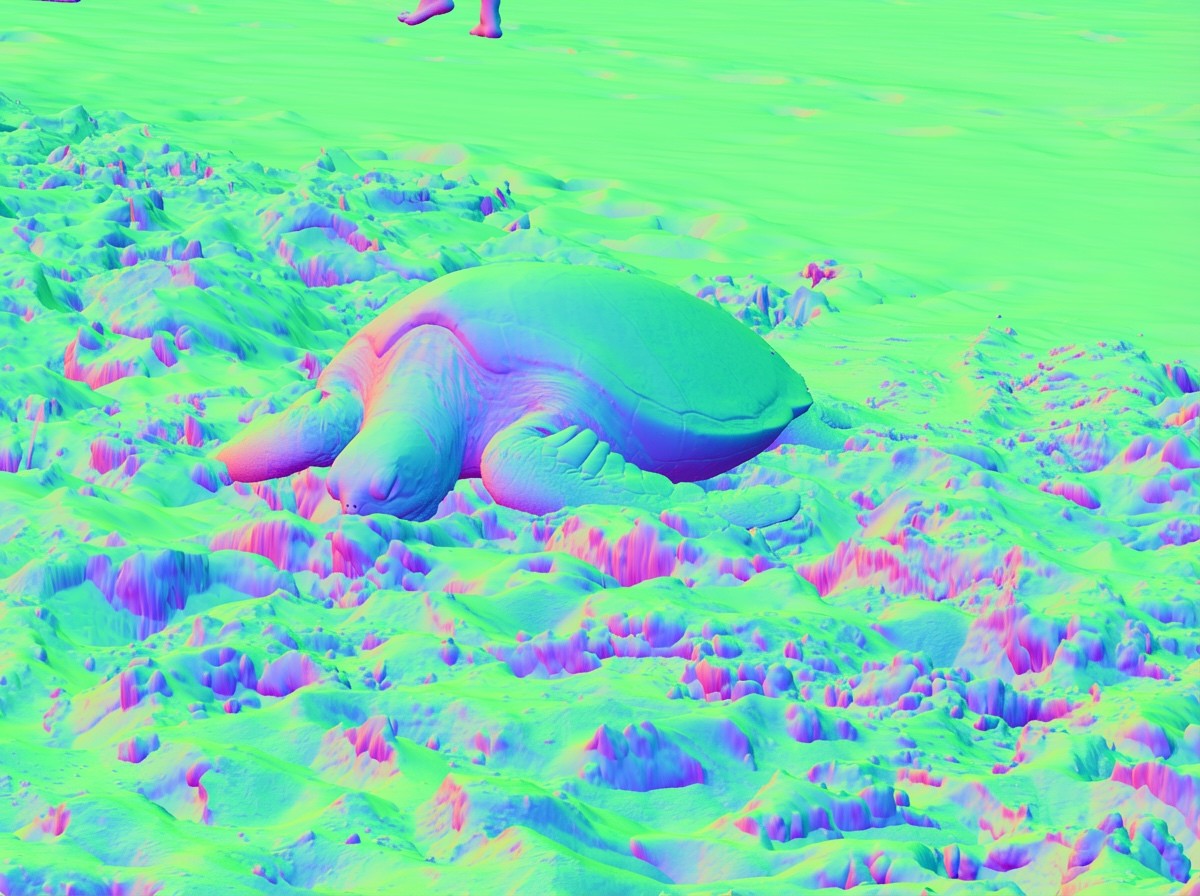} \\[2pt]
      
        \includegraphics[width=0.32\textwidth]{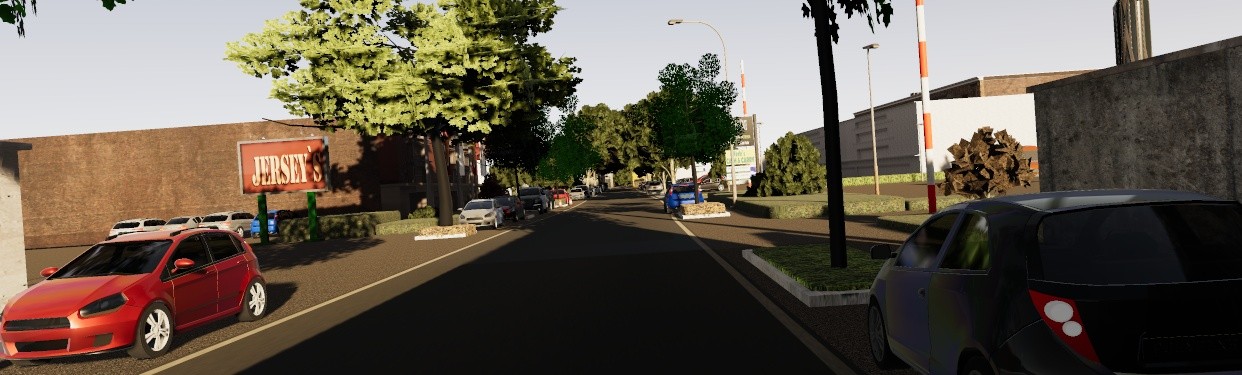} &
        \includegraphics[width=0.32\textwidth]{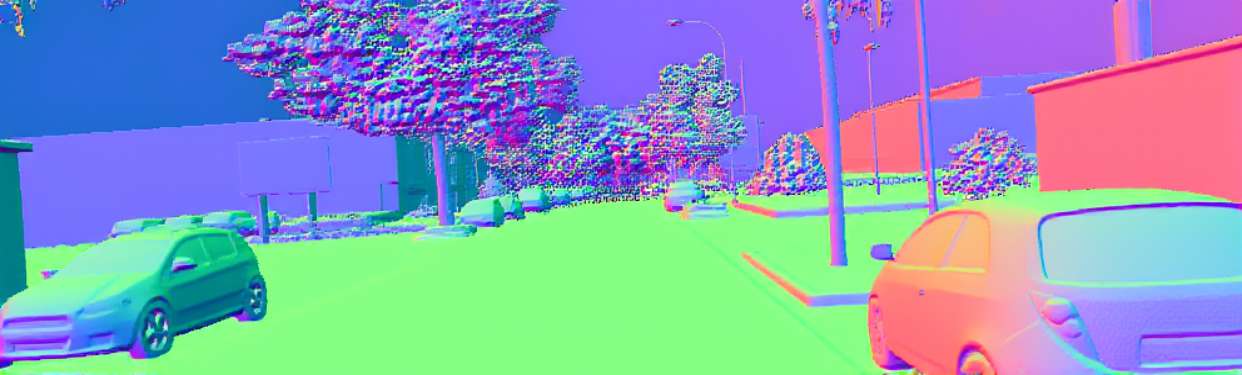} &
        \includegraphics[width=0.32\textwidth]{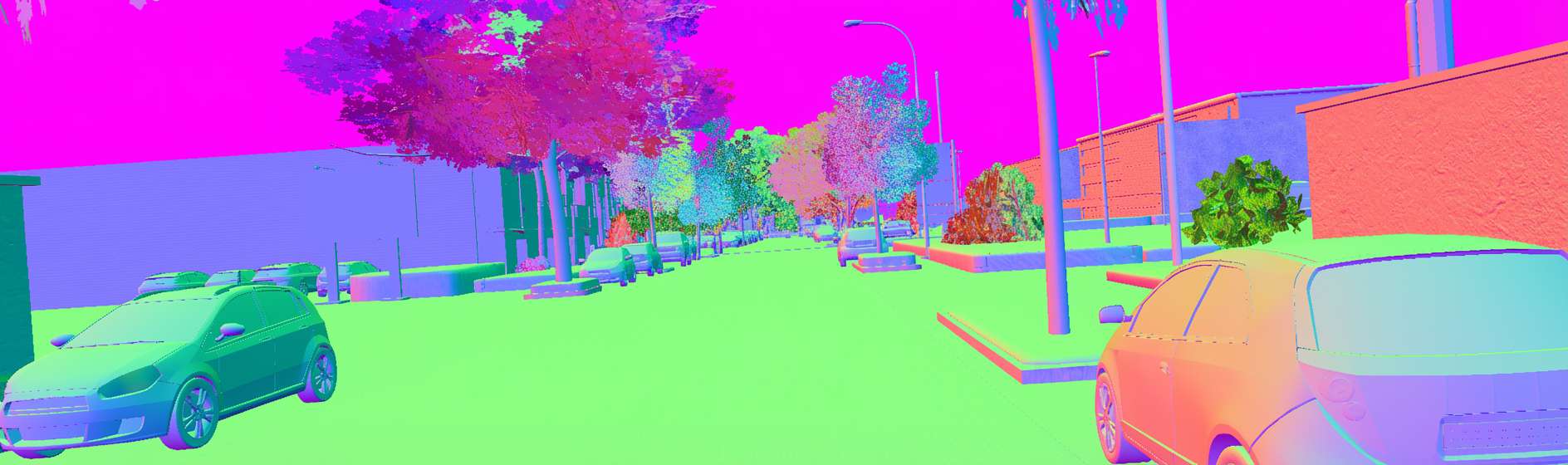} \\
        \small Input image & \small Lotus-2-Normal & \small \textbf{Vision Banana} \emojivb
    \end{tabular}
    \caption{
    Comparison with SOTA surface normal estimation method Lotus-2~\citep{lotus2}.
    Results of Lotus-2 are obtained using its Hugging-Face demo: \href{https://huggingface.co/spaces/haodongli/Lotus-2_Normal}{\url{https://huggingface.co/spaces/haodongli/Lotus-2_Normal}}.
    Vision Banana can produce surface normal map with much higher visual quality and better fine-grained details.
    Zoom-in for the details.
    }
    \label{fig:surface_normal_demo}
\end{figure}

Figure~\ref{fig:surface_normal_demo} visually compares the output from Vision Banana with the leading external method, Lotus-2 \citep{lotus2}. Vision Banana consistently produces surface normal maps with significantly higher fidelity and finer granular details. 
The bottom row of Fig. \ref{fig:surface_normal_demo} highlights a sample from Virtual KITTI 2 \citep{cabon2020virtualkitty}. Although Vision Banana registers slightly higher quantitative errors on this benchmark compared to Lotus-2, it yields demonstrably superior visual quality. Also note that Lotus-2 is trained on Virtual KITTI 2 for surface normal estimation, whereas Vision Banana maintains a strict zero-shot transfer protocol, having never seen the training sets of any evaluated benchmarks.

\section{Discussion}

\paragraph{Image Generators are Generalist Vision Learners.}
Generative pretraining \citep{radford2018improving_gpt1,radford2019language_gpt2,brown2020language_gpt3} has fundamentally transformed language understanding and reasoning. 
In the meantime, recent observations of emergent vision capabilities \citep{wiedemer2025veo_learner,zuo2025nanobanana_low_level} have ignited speculation that computer vision is approaching a similar paradigm shift.
By instruction-tuning a leading image generator, Nano Banana Pro, into a state-of-the-art visual generation and understanding model, we confirm that \textbf{this shift is already underway}.
Models pretrained on large-scale image generation naturally acquire robust visual understanding capabilities.
These generative priors surpass the specialized architectures and dedicated training paradigms traditionally employed by specialist vision models.
We are witnessing a paradigm shift for computer vision that will be fueled by generative vision pretraining, which we believe paves the way for true \textbf{Foundational Vision Models} and \textbf{Artificial General Intelligence from Vision} (AGI-V).

\paragraph{Image Generation as a Universal Interface.}
As a byproduct of this study, we show that image generation can serve as the universal interface for computer vision, analogous to how text generation acts as the unifying interface for many tasks embedded in natural language, including language understanding, generation, reasoning, math, coding, agentic tasks, etc.. 
By representing vision task outputs as RGB images, we can use natural language prompts to seamlessly instruct the model. While we are not the first to encode vision outputs as RGB \citep{marigold,zhao2025diception,gan2023instructcv,wang2023images,lu2022unified,lu2024unified,xie2024show,ai2025ming}, we demonstrate that when combined with powerful pretrained visual generators, this simple design is sufficient to outperform modern domain-specific specialist models.

In addition to the unification of vision task outputs as RGB images, generative modeling naturally provides a workaround for the ambiguity in vision tasks where a single input can correspond to several modes of the output distribution. 
In order to prevent the collapse of the output to a blurry mean, expert discriminative models \citep{sam3,dav3} usually resort to custom architectures and training losses. 
For example, the Segment Anything models \citep{kirillov2023segment,ravi2024sam2,sam3} return several segmentation masks but only apply the loss to a single one.
Generative models, however, inherently learn the full data distribution, gracefully managing ambiguity by design.
By eliminating the need for bespoke architectural designs, this formulation could lead to a truly unified ``omni'' multimodal model.

\paragraph{Future Work.}
While Vision Banana achieves SOTA results on fundamental tasks for 2D semantic understanding and 3D understanding from monocular images, several exciting avenues remain for future exploration. 
First, scaling the diversity of instruction-tuned tasks may unlock further emergent cross-task generalization, similar to behaviors observed in LLMs \citep{wei2021finetuned_flan}.
Second, our current evaluation focuses on monocular image inputs.
In the future, we can extend this framework to process multi-view inputs \citep{wang2025vggt} and video inputs \citep{zhang2025d4rt}.
Similarly, investigating whether video generators yield even richer, temporally-aware visual representations presents a highly promising research direction.
Another important next step is exploring the synergistic integration of foundational vision models with large language models to enhance cross-modality reasoning.
Finally, utilizing image generators like Nano Banana Pro currently incurs a significantly higher computational overhead than running lightweight specialist models.
Developing acceleration and cost-reduction strategies will be an essential hurdle to overcome for the deployment of generative vision framework.

\section*{Acknowledgment}
We thank 
Xi Chen, Fei Xia, 
Kaushik Shivakumar, Abhishek Sinha, Phillip Lippe, Yilin Gao, Javier Rey, Sanghyun Woo,
Renshen Wang, Wentao Yuan, Keran Rong, Rundi Wu,
Manoj Kumar, Manli Shu, Francesco Piccinno, Ishita Dasgupta, 
Benigno Uria, Miki Rubinstein, Aäron van den Oord, Jon Shlens
for their helpful discussions, advice, and technical guidance.

\bibliography{main}

\begin{thebibliography}{110}
\providecommand{\natexlab}[1]{#1}
\providecommand{\url}[1]{\texttt{#1}}
\expandafter\ifx\csname urlstyle\endcsname\relax
  \providecommand{\doi}[1]{doi: #1}\else
  \providecommand{\doi}{doi: \begingroup \urlstyle{rm}\Url}\fi

\bibitem[Bae and Davison(2024)]{DSINE}
G.~Bae and A.~J. Davison.
\newblock Rethinking inductive biases for surface normal estimation.
\newblock In \emph{Proceedings of the IEEE/CVF Conference on Computer Vision and Pattern Recognition}, pages 9535--9545, 2024.

\bibitem[Bai et~al.(2024)Bai, Geng, Mangalam, Bar, Yuille, Darrell, Malik, and Efros]{bai2024sequential_lvm}
Y.~Bai, X.~Geng, K.~Mangalam, A.~Bar, A.~L. Yuille, T.~Darrell, J.~Malik, and A.~A. Efros.
\newblock Sequential modeling enables scalable learning for large vision models.
\newblock In \emph{Proceedings of the IEEE/CVF Conference on Computer Vision and Pattern Recognition}, pages 22861--22872, 2024.

\bibitem[Bao et~al.(2021)Bao, Dong, Piao, and Wei]{bao2021beit}
H.~Bao, L.~Dong, S.~Piao, and F.~Wei.
\newblock Beit: Bert pre-training of image transformers.
\newblock \emph{arXiv preprint arXiv:2106.08254}, 2021.

\bibitem[Baranchuk et~al.(2021)Baranchuk, Rubachev, Voynov, Khrulkov, and Babenko]{baranchuk2021label}
D.~Baranchuk, I.~Rubachev, A.~Voynov, V.~Khrulkov, and A.~Babenko.
\newblock Label-efficient semantic segmentation with diffusion models.
\newblock \emph{arXiv preprint arXiv:2112.03126}, 2021.

\bibitem[Barron(2025)]{barron2025power}
J.~T. Barron.
\newblock A power transform.
\newblock \emph{arXiv preprint arXiv:2502.10647}, 2025.

\bibitem[Bhattad et~al.(2023)Bhattad, McKee, Hoiem, and Forsyth]{bhattad2023stylegan}
A.~Bhattad, D.~McKee, D.~Hoiem, and D.~Forsyth.
\newblock Stylegan knows normal, depth, albedo, and more.
\newblock \emph{Advances in Neural Information Processing Systems}, 36:\penalty0 73082--73103, 2023.

\bibitem[{Black Forest Labs}(2025)]{flux2}
{Black Forest Labs}.
\newblock {FLUX.2: Frontier Visual Intelligence}.
\newblock \url{https://bfl.ai/blog/flux-2}, 2025.

\bibitem[Bochkovskii et~al.(2024)Bochkovskii, Delaunoy, Germain, Santos, Zhou, Richter, and Koltun]{bochkovskii2024depthpro}
A.~Bochkovskii, A.~Delaunoy, H.~Germain, M.~Santos, Y.~Zhou, S.~R. Richter, and V.~Koltun.
\newblock Depth pro: Sharp monocular metric depth in less than a second.
\newblock \emph{arXiv preprint arXiv:2410.02073}, 2024.

\bibitem[Brown et~al.(2020)Brown, Mann, Ryder, Subbiah, Kaplan, Dhariwal, Neelakantan, Shyam, Sastry, Askell, et~al.]{brown2020language_gpt3}
T.~Brown, B.~Mann, N.~Ryder, M.~Subbiah, J.~D. Kaplan, P.~Dhariwal, A.~Neelakantan, P.~Shyam, G.~Sastry, A.~Askell, et~al.
\newblock Language models are few-shot learners.
\newblock \emph{Advances in neural information processing systems}, 33:\penalty0 1877--1901, 2020.

\bibitem[ByteDance(2026)]{seedance2}
ByteDance.
\newblock {Seedance 2.0}.
\newblock \url{https://seed.bytedance.com/en/seedance2_0/}, 2026.
\newblock Accessed: 2026-03-18.

\bibitem[Cabon et~al.(2020)Cabon, Murray, and Humenberger]{cabon2020virtualkitty}
Y.~Cabon, N.~Murray, and M.~Humenberger.
\newblock Virtual kitti 2.
\newblock \emph{arXiv preprint arXiv:2001.10773}, 2020.

\bibitem[Caesar et~al.(2020)Caesar, Bankiti, Lang, Vora, Liong, Xu, Krishnan, Pan, Baldan, and Beijbom]{caesar2020nuscenes}
H.~Caesar, V.~Bankiti, A.~H. Lang, S.~Vora, V.~E. Liong, Q.~Xu, A.~Krishnan, Y.~Pan, G.~Baldan, and O.~Beijbom.
\newblock nuscenes: A multimodal dataset for autonomous driving.
\newblock In \emph{Proceedings of the IEEE/CVF conference on computer vision and pattern recognition}, pages 11621--11631, 2020.

\bibitem[Cai et~al.(2025)Cai, Yeh, Xu, Liu, Meyer, Lei, Zhao, Li, Chandra, and Shi]{depthlm}
Z.~Cai, C.-F. Yeh, H.~Xu, Z.~Liu, G.~Meyer, X.~Lei, C.~Zhao, S.-W. Li, V.~Chandra, and Y.~Shi.
\newblock Depthlm: Metric depth from vision language models.
\newblock \emph{arXiv preprint arXiv:2509.25413}, 2025.

\bibitem[Cao et~al.(2026)Cao, Chen, Maninis, Chen, Karpur, Xia, Dua, Dabral, Han, Han, et~al.]{cao2026tipsv2}
B.~Cao, K.~Chen, K.-K. Maninis, K.~Chen, A.~Karpur, Y.~Xia, S.~Dua, T.~Dabral, G.~Han, B.~Han, et~al.
\newblock Tipsv2: Advancing vision-language pretraining with enhanced patch-text alignment.
\newblock \emph{arXiv preprint arXiv:2604.12012}, 2026.

\bibitem[Carion et~al.(2025)Carion, Gustafson, Hu, Debnath, Hu, Suris, Ryali, Alwala, Khedr, Huang, et~al.]{sam3}
N.~Carion, L.~Gustafson, Y.-T. Hu, S.~Debnath, R.~Hu, D.~Suris, C.~Ryali, K.~V. Alwala, H.~Khedr, A.~Huang, et~al.
\newblock Sam 3: Segment anything with concepts.
\newblock \emph{arXiv preprint arXiv:2511.16719}, 2025.

\bibitem[Caron et~al.(2021)Caron, Touvron, Misra, J{\'e}gou, Mairal, Bojanowski, and Joulin]{caron2021emerging_dinov1}
M.~Caron, H.~Touvron, I.~Misra, H.~J{\'e}gou, J.~Mairal, P.~Bojanowski, and A.~Joulin.
\newblock Emerging properties in self-supervised vision transformers.
\newblock In \emph{Proceedings of the IEEE/CVF international conference on computer vision}, pages 9650--9660, 2021.

\bibitem[Chen et~al.(2020{\natexlab{a}})Chen, Radford, Child, Wu, Jun, Luan, and Sutskever]{chen2020generative_igpt}
M.~Chen, A.~Radford, R.~Child, J.~Wu, H.~Jun, D.~Luan, and I.~Sutskever.
\newblock Generative pretraining from pixels.
\newblock In \emph{International conference on machine learning}, pages 1691--1703. PMLR, 2020{\natexlab{a}}.

\bibitem[Chen et~al.(2020{\natexlab{b}})Chen, Kornblith, Norouzi, and Hinton]{simclr}
T.~Chen, S.~Kornblith, M.~Norouzi, and G.~Hinton.
\newblock A simple framework for contrastive learning of visual representations.
\newblock In \emph{International conference on machine learning}, pages 1597--1607. PmLR, 2020{\natexlab{b}}.

\bibitem[Chen et~al.(2016)Chen, Duan, Houthooft, Schulman, Sutskever, and Abbeel]{chen2016infogan}
X.~Chen, Y.~Duan, R.~Houthooft, J.~Schulman, I.~Sutskever, and P.~Abbeel.
\newblock Infogan: Interpretable representation learning by information maximizing generative adversarial nets.
\newblock \emph{Advances in neural information processing systems}, 29, 2016.

\bibitem[Chen et~al.(2020{\natexlab{c}})Chen, Fan, Girshick, and He]{chen2020improved_moco2}
X.~Chen, H.~Fan, R.~Girshick, and K.~He.
\newblock Improved baselines with momentum contrastive learning.
\newblock \emph{arXiv preprint arXiv:2003.04297}, 2020{\natexlab{c}}.

\bibitem[Chen et~al.(2024)Chen, Liu, Xie, and He]{chen2024deconstructing_dae}
X.~Chen, Z.~Liu, S.~Xie, and K.~He.
\newblock Deconstructing denoising diffusion models for self-supervised learning.
\newblock \emph{arXiv preprint arXiv:2401.14404}, 2024.

\bibitem[Chowdhery et~al.(2023)Chowdhery, Narang, Devlin, Bosma, Mishra, Roberts, Barham, Chung, Sutton, Gehrmann, et~al.]{chowdhery2023palm}
A.~Chowdhery, S.~Narang, J.~Devlin, M.~Bosma, G.~Mishra, A.~Roberts, P.~Barham, H.~W. Chung, C.~Sutton, S.~Gehrmann, et~al.
\newblock Palm: Scaling language modeling with pathways.
\newblock \emph{Journal of machine learning research}, 24\penalty0 (240):\penalty0 1--113, 2023.

\bibitem[Clark and Jaini(2023)]{clark2023text}
K.~Clark and P.~Jaini.
\newblock Text-to-image diffusion models are zero shot classifiers.
\newblock \emph{Advances in Neural Information Processing Systems}, 36:\penalty0 58921--58937, 2023.

\bibitem[Cordts et~al.(2016)Cordts, Omran, Ramos, Rehfeld, Enzweiler, Benenson, Franke, Roth, and Schiele]{cityscapes}
M.~Cordts, M.~Omran, S.~Ramos, T.~Rehfeld, M.~Enzweiler, R.~Benenson, U.~Franke, S.~Roth, and B.~Schiele.
\newblock The cityscapes dataset for semantic urban scene understanding.
\newblock In \emph{Proceedings of the IEEE conference on computer vision and pattern recognition}, pages 3213--3223, 2016.

\bibitem[Dai et~al.(2017)Dai, Chang, Savva, Halber, Funkhouser, and Nie{\ss}ner]{dai2017scannet}
A.~Dai, A.~X. Chang, M.~Savva, M.~Halber, T.~Funkhouser, and M.~Nie{\ss}ner.
\newblock Scannet: Richly-annotated 3d reconstructions of indoor scenes.
\newblock In \emph{Proceedings of the IEEE conference on computer vision and pattern recognition}, pages 5828--5839, 2017.

\bibitem[Dehghani et~al.(2023)Dehghani, Djolonga, Mustafa, Padlewski, Heek, Gilmer, Steiner, Caron, Geirhos, Alabdulmohsin, et~al.]{dehghani2023scaling_vit22b}
M.~Dehghani, J.~Djolonga, B.~Mustafa, P.~Padlewski, J.~Heek, J.~Gilmer, A.~P. Steiner, M.~Caron, R.~Geirhos, I.~Alabdulmohsin, et~al.
\newblock Scaling vision transformers to 22 billion parameters.
\newblock In \emph{International conference on machine learning}, pages 7480--7512. PMLR, 2023.

\bibitem[Dosovitskiy et~al.(2020)Dosovitskiy, Beyer, Kolesnikov, Weissenborn, Zhai, Unterthiner, Dehghani, Minderer, Heigold, Gelly, et~al.]{dosovitskiy2020image_vit}
A.~Dosovitskiy, L.~Beyer, A.~Kolesnikov, D.~Weissenborn, X.~Zhai, T.~Unterthiner, M.~Dehghani, M.~Minderer, G.~Heigold, S.~Gelly, et~al.
\newblock An image is worth 16x16 words: Transformers for image recognition at scale.
\newblock \emph{arXiv preprint arXiv:2010.11929}, 2020.

\bibitem[Du et~al.(2023)Du, Kolkin, Shakhnarovich, and Bhattad]{du2023generative}
X.~Du, N.~Kolkin, G.~Shakhnarovich, and A.~Bhattad.
\newblock Generative models: What do they know? do they know things? let's find out!
\newblock \emph{arXiv preprint arXiv:2311.17137}, 2023.

\bibitem[Eigen et~al.(2014)Eigen, Puhrsch, and Fergus]{eigen2014depth}
D.~Eigen, C.~Puhrsch, and R.~Fergus.
\newblock Depth map prediction from a single image using a multi-scale deep network.
\newblock In \emph{Advances in Neural Information Processing Systems (NeurIPS)}, volume~27, 2014.

\bibitem[Fu et~al.(2025{\natexlab{a}})Fu, Yang, Mo, Yan, Wei, Meng, Xie, and Zheng]{fu2025llmdet}
S.~Fu, Q.~Yang, Q.~Mo, J.~Yan, X.~Wei, J.~Meng, X.~Xie, and W.-S. Zheng.
\newblock Llmdet: Learning strong open-vocabulary object detectors under the supervision of large language models.
\newblock In \emph{Proceedings of the Computer Vision and Pattern Recognition Conference}, pages 14987--14997, 2025{\natexlab{a}}.

\bibitem[Fu et~al.(2025{\natexlab{b}})Fu, Lou, and Yu]{fu2025segman}
Y.~Fu, M.~Lou, and Y.~Yu.
\newblock Segman: Omni-scale context modeling with state space models and local attention for semantic segmentation.
\newblock In \emph{Proceedings of the Computer Vision and Pattern Recognition Conference}, pages 19077--19087, 2025{\natexlab{b}}.

\bibitem[Gan et~al.(2023)Gan, Park, Schubert, Philippakis, and Alaa]{gan2023instructcv}
Y.~Gan, S.~Park, A.~Schubert, A.~Philippakis, and A.~M. Alaa.
\newblock Instructcv: Instruction-tuned text-to-image diffusion models as vision generalists.
\newblock \emph{arXiv preprint arXiv:2310.00390}, 2023.

\bibitem[Garcia et~al.(2025)Garcia, Zeid, Schmidt, De~Geus, Hermans, and Leibe]{garcia2025fine}
G.~M. Garcia, K.~A. Zeid, C.~Schmidt, D.~De~Geus, A.~Hermans, and B.~Leibe.
\newblock Fine-tuning image-conditional diffusion models is easier than you think.
\newblock In \emph{Proceedings of the Winter Conference on Applications of Computer Vision}, pages 753--762, 2025.

\bibitem[{Gemini Team}(2025)]{gemini25}
{Gemini Team}.
\newblock Gemini 2.5: Pushing the frontier with advanced reasoning, multimodality, long context, and next generation agentic capabilities.
\newblock \emph{arXiv preprint}, 2025.

\bibitem[{Google}(2025{\natexlab{a}})]{nbp}
{Google}.
\newblock Introducing nano banana pro.
\newblock \url{https://blog.google/innovation-and-ai/products/nano-banana-pro/}, 2025{\natexlab{a}}.
\newblock Accessed: 2026-03-15.

\bibitem[{Google}(2025{\natexlab{b}})]{veo3}
{Google}.
\newblock Veo 3 announcement.
\newblock \url{https://blog.google/innovation-and-ai/products/generative-media-models-io-2025/}, 2025{\natexlab{b}}.
\newblock Accessed: 2026-03-15.

\bibitem[Grill et~al.(2020)Grill, Strub, Altch{\'e}, Tallec, Richemond, Buchatskaya, Doersch, Avila~Pires, Guo, Gheshlaghi~Azar, et~al.]{grill2020bootstrap_byol}
J.-B. Grill, F.~Strub, F.~Altch{\'e}, C.~Tallec, P.~Richemond, E.~Buchatskaya, C.~Doersch, B.~Avila~Pires, Z.~Guo, M.~Gheshlaghi~Azar, et~al.
\newblock Bootstrap your own latent-a new approach to self-supervised learning.
\newblock \emph{Advances in neural information processing systems}, 33:\penalty0 21271--21284, 2020.

\bibitem[He et~al.(2024)He, Li, Yin, Liang, Li, Zhou, Zhang, Liu, and Chen]{he2024lotus}
J.~He, H.~Li, W.~Yin, Y.~Liang, L.~Li, K.~Zhou, H.~Zhang, B.~Liu, and Y.-C. Chen.
\newblock Lotus: Diffusion-based visual foundation model for high-quality dense prediction.
\newblock \emph{arXiv preprint arXiv:2409.18124}, 2024.

\bibitem[He et~al.(2025)He, Li, Sheng, and Chen]{lotus2}
J.~He, H.~Li, M.~Sheng, and Y.-C. Chen.
\newblock Lotus-2: Advancing geometric dense prediction with powerful image generative model.
\newblock \emph{arXiv preprint arXiv:2512.01030}, 2025.

\bibitem[He et~al.(2020)He, Fan, Wu, Xie, and Girshick]{he2020momentum_moco}
K.~He, H.~Fan, Y.~Wu, S.~Xie, and R.~Girshick.
\newblock Momentum contrast for unsupervised visual representation learning.
\newblock In \emph{Proceedings of the IEEE/CVF conference on computer vision and pattern recognition}, pages 9729--9738, 2020.

\bibitem[He et~al.(2022)He, Chen, Xie, Li, Doll{\'a}r, and Girshick]{he2022masked_mae}
K.~He, X.~Chen, S.~Xie, Y.~Li, P.~Doll{\'a}r, and R.~Girshick.
\newblock Masked autoencoders are scalable vision learners.
\newblock In \emph{Proceedings of the IEEE/CVF conference on computer vision and pattern recognition}, pages 16000--16009, 2022.

\bibitem[Hedlin et~al.(2023)Hedlin, Sharma, Mahajan, Isack, Kar, Tagliasacchi, and Yi]{hedlin2023unsupervised}
E.~Hedlin, G.~Sharma, S.~Mahajan, H.~Isack, A.~Kar, A.~Tagliasacchi, and K.~M. Yi.
\newblock Unsupervised semantic correspondence using stable diffusion.
\newblock \emph{Advances in Neural Information Processing Systems}, 36:\penalty0 8266--8279, 2023.

\bibitem[Hjelm et~al.(2018)Hjelm, Fedorov, Lavoie-Marchildon, Grewal, Bachman, Trischler, and Bengio]{hjelm2018learning}
R.~D. Hjelm, A.~Fedorov, S.~Lavoie-Marchildon, K.~Grewal, P.~Bachman, A.~Trischler, and Y.~Bengio.
\newblock Learning deep representations by mutual information estimation and maximization.
\newblock \emph{arXiv preprint arXiv:1808.06670}, 2018.

\bibitem[Hu et~al.(2024)Hu, Yin, Zhang, Cai, Long, Chen, Wang, Yu, Shen, and Shen]{hu2024metric3dv2}
M.~Hu, W.~Yin, C.~Zhang, Z.~Cai, X.~Long, H.~Chen, K.~Wang, G.~Yu, C.~Shen, and S.~Shen.
\newblock Metric3d v2: A versatile monocular geometric foundation model for zero-shot metric depth and surface normal estimation.
\newblock \emph{IEEE Transactions on Pattern Analysis and Machine Intelligence}, 2024.

\bibitem[{Inclusion AI}(2025)]{ai2025ming}
{Inclusion AI}.
\newblock Ming-flash-omni: A sparse, unified architecture for multimodal perception and generation.
\newblock \emph{arXiv preprint arXiv:2510.24821}, 2025.

\bibitem[Kang et~al.(2025)Kang, Kim, Kim, and Hwang]{kang2025your}
S.~Kang, J.~Kim, J.~Kim, and S.~J. Hwang.
\newblock Your large vision-language model only needs a few attention heads for visual grounding.
\newblock In \emph{Proceedings of the Computer Vision and Pattern Recognition Conference}, pages 9339--9350, 2025.

\bibitem[Kazemzadeh et~al.(2014)Kazemzadeh, Ordonez, Matten, and Berg]{refcoco}
S.~Kazemzadeh, V.~Ordonez, M.~Matten, and T.~Berg.
\newblock Referitgame: Referring to objects in photographs of natural scenes.
\newblock In \emph{Proceedings of the 2014 conference on empirical methods in natural language processing (EMNLP)}, pages 787--798, 2014.

\bibitem[Ke et~al.(2024)Ke, Obukhov, Huang, Metzger, Daudt, and Schindler]{marigold}
B.~Ke, A.~Obukhov, S.~Huang, N.~Metzger, R.~C. Daudt, and K.~Schindler.
\newblock Repurposing diffusion-based image generators for monocular depth estimation.
\newblock In \emph{Proceedings of the IEEE/CVF conference on computer vision and pattern recognition}, pages 9492--9502, 2024.

\bibitem[Kirillov et~al.(2023)Kirillov, Mintun, Ravi, Mao, Rolland, Gustafson, Xiao, Whitehead, Berg, Lo, Doll{\'a}r, and Girshick]{kirillov2023segment}
A.~Kirillov, E.~Mintun, N.~Ravi, H.~Mao, C.~Rolland, L.~Gustafson, T.~Xiao, S.~Whitehead, A.~C. Berg, W.-Y. Lo, P.~Doll{\'a}r, and R.~Girshick.
\newblock Segment anything.
\newblock In \emph{IEEE/CVF International Conference on Computer Vision (ICCV)}, pages 4015--4026, 2023.

\bibitem[Koch et~al.(2018)Koch, Liebel, Fraundorfer, and Korner]{koch2018evaluation_ibims1}
T.~Koch, L.~Liebel, F.~Fraundorfer, and M.~Korner.
\newblock Evaluation of cnn-based single-image depth estimation methods.
\newblock In \emph{Proceedings of the European Conference on Computer Vision (ECCV) Workshops}, pages 0--0, 2018.

\bibitem[Krizhevsky et~al.(2012)Krizhevsky, Sutskever, and Hinton]{krizhevsky2012imagenet_alexnet}
A.~Krizhevsky, I.~Sutskever, and G.~E. Hinton.
\newblock Imagenet classification with deep convolutional neural networks.
\newblock \emph{Advances in neural information processing systems}, 25, 2012.

\bibitem[Lai et~al.(2024)Lai, Tian, Chen, Li, Yuan, Liu, and Jia]{reasonseg}
X.~Lai, Z.~Tian, Y.~Chen, Y.~Li, Y.~Yuan, S.~Liu, and J.~Jia.
\newblock Lisa: Reasoning segmentation via large language model.
\newblock In \emph{Proceedings of the IEEE/CVF conference on computer vision and pattern recognition}, pages 9579--9589, 2024.

\bibitem[Li et~al.(2023)Li, Prabhudesai, Duggal, Brown, and Pathak]{li2023your}
A.~C. Li, M.~Prabhudesai, S.~Duggal, E.~Brown, and D.~Pathak.
\newblock Your diffusion model is secretly a zero-shot classifier.
\newblock In \emph{Proceedings of the IEEE/CVF International Conference on Computer Vision}, pages 2206--2217, 2023.

\bibitem[Li et~al.(2024{\natexlab{a}})Li, Lin, Pathak, Li, Fei, Wu, Ling, Xia, Zhang, Neubig, et~al.]{li2024genaibench}
B.~Li, Z.~Lin, D.~Pathak, J.~Li, Y.~Fei, K.~Wu, T.~Ling, X.~Xia, P.~Zhang, G.~Neubig, et~al.
\newblock Genai-bench: Evaluating and improving compositional text-to-visual generation.
\newblock \emph{arXiv preprint arXiv:2406.13743}, 2024{\natexlab{a}}.

\bibitem[Li et~al.(2024{\natexlab{b}})Li, Lu, Han, and Prisacariu]{li2024sd4match}
X.~Li, J.~Lu, K.~Han, and V.~A. Prisacariu.
\newblock Sd4match: Learning to prompt stable diffusion model for semantic matching.
\newblock In \emph{Proceedings of the IEEE/CVF Conference on Computer Vision and Pattern Recognition}, pages 27558--27568, 2024{\natexlab{b}}.

\bibitem[Lin et~al.(2025)Lin, Chen, Liew, Chen, Li, Shi, Feng, and Kang]{dav3}
H.~Lin, S.~Chen, J.~Liew, D.~Y. Chen, Z.~Li, G.~Shi, J.~Feng, and B.~Kang.
\newblock Depth anything 3: Recovering the visual space from any views.
\newblock \emph{arXiv preprint arXiv:2511.10647}, 2025.

\bibitem[Liu et~al.(2024)Liu, Zeng, Ren, Li, Zhang, Yang, Jiang, Li, Yang, Su, Zhu, and Zhang]{liu2024grounding_dino}
S.~Liu, Z.~Zeng, T.~Ren, F.~Li, H.~Zhang, J.~Yang, Q.~Jiang, C.~Li, J.~Yang, H.~Su, J.~Zhu, and L.~Zhang.
\newblock Grounding {DINO:} marrying {DINO} with grounded pre-training for open-set object detection.
\newblock In \emph{{ECCV} {(47)}}, Lecture Notes in Computer Science, pages 38--55. Springer, 2024.

\bibitem[Liu and Li(2025)]{liu2025hybrid}
T.~Liu and S.~Li.
\newblock Hybrid global-local representation with augmented spatial guidance for zero-shot referring image segmentation.
\newblock In \emph{Proceedings of the IEEE/CVF Conference on Computer Vision and Pattern Recognition}, pages 29634--29643, 2025.

\bibitem[Liu et~al.(2025)Liu, Peng, Zhong, Yue, Lu, Yu, and Jia]{liu2025segzero}
Y.~Liu, B.~Peng, Z.~Zhong, Z.~Yue, F.~Lu, B.~Yu, and J.~Jia.
\newblock Seg-zero: Reasoning-chain guided segmentation via cognitive reinforcement.
\newblock \emph{arXiv preprint arXiv:2503.06520}, 2025.

\bibitem[Lu et~al.(2022)Lu, Clark, Zellers, Mottaghi, and Kembhavi]{lu2022unified}
J.~Lu, C.~Clark, R.~Zellers, R.~Mottaghi, and A.~Kembhavi.
\newblock Unified-io: A unified model for vision, language, and multi-modal tasks.
\newblock \emph{arXiv preprint arXiv:2206.08916}, 2022.

\bibitem[Lu et~al.(2024)Lu, Clark, Lee, Zhang, Khosla, Marten, Hoiem, and Kembhavi]{lu2024unified}
J.~Lu, C.~Clark, S.~Lee, Z.~Zhang, S.~Khosla, R.~Marten, D.~Hoiem, and A.~Kembhavi.
\newblock Unified-io 2: Scaling autoregressive multimodal models with vision language audio and action.
\newblock In \emph{Proceedings of the IEEE/CVF Conference on Computer Vision and Pattern Recognition}, pages 26439--26455, 2024.

\bibitem[Lu et~al.(2025)Lu, Cao, Wu, Li, Tang, Ji, Wu, Wu, and Zhu]{lu2025rsvp}
Y.~Lu, J.~Cao, Y.~Wu, B.~Li, L.~Tang, Y.~Ji, C.~Wu, J.~Wu, and W.~Zhu.
\newblock Rsvp: Reasoning segmentation via visual prompting and multi-modal chain-of-thought.
\newblock In \emph{Proceedings of the 63rd Annual Meeting of the Association for Computational Linguistics (Volume 1: Long Papers)}, pages 14699--14716, 2025.

\bibitem[Luma(2026)]{uni1}
Luma.
\newblock {UNI-1}.
\newblock \url{https://lumalabs.ai/uni-1/}, 2026.
\newblock Accessed: 2026-03-19.

\bibitem[Minderer et~al.(2023)Minderer, Gritsenko, and Houlsby]{minderer2023gowol}
M.~Minderer, A.~Gritsenko, and N.~Houlsby.
\newblock Scaling open-vocabulary object detection.
\newblock \emph{Advances in Neural Information Processing Systems}, 36:\penalty0 72983--73007, 2023.

\bibitem[Mukhopadhyay et~al.(2023)Mukhopadhyay, Gwilliam, Agarwal, Padmanabhan, Swaminathan, Hegde, Zhou, and Shrivastava]{mukhopadhyay2023diffusion}
S.~Mukhopadhyay, M.~Gwilliam, V.~Agarwal, N.~Padmanabhan, A.~Swaminathan, S.~Hegde, T.~Zhou, and A.~Shrivastava.
\newblock Diffusion models beat gans on image classification.
\newblock \emph{arXiv preprint arXiv:2307.08702}, 2023.

\bibitem[OpenAI(2026)]{gpt1p5}
OpenAI.
\newblock {GPT-Image-1.5}.
\newblock \url{https://openai.com/index/new-chatgpt-images-is-here/}, 2026.
\newblock Accessed: 2026-03-19.

\bibitem[Oquab et~al.(2023)Oquab, Darcet, Moutakanni, Vo, Szafraniec, Khalidov, Fernandez, Haziza, Massa, El-Nouby, et~al.]{oquab2023dinov2}
M.~Oquab, T.~Darcet, T.~Moutakanni, H.~Vo, M.~Szafraniec, V.~Khalidov, P.~Fernandez, D.~Haziza, F.~Massa, A.~El-Nouby, et~al.
\newblock Dinov2: Learning robust visual features without supervision.
\newblock \emph{arXiv preprint arXiv:2304.07193}, 2023.

\bibitem[Ouyang et~al.(2022)Ouyang, Wu, Jiang, Almeida, Wainwright, Mishkin, Zhang, Agarwal, Slama, Ray, et~al.]{ouyang2022training_instructgpt}
L.~Ouyang, J.~Wu, X.~Jiang, D.~Almeida, C.~Wainwright, P.~Mishkin, C.~Zhang, S.~Agarwal, K.~Slama, A.~Ray, et~al.
\newblock Training language models to follow instructions with human feedback.
\newblock \emph{Advances in neural information processing systems}, 35:\penalty0 27730--27744, 2022.

\bibitem[Piccinelli et~al.(2025{\natexlab{a}})Piccinelli, Sakaridis, Segu, Yang, Li, Abbeloos, and Van~Gool]{unik3d}
L.~Piccinelli, C.~Sakaridis, M.~Segu, Y.-H. Yang, S.~Li, W.~Abbeloos, and L.~Van~Gool.
\newblock Unik3d: Universal camera monocular 3d estimation.
\newblock In \emph{Proceedings of the Computer Vision and Pattern Recognition Conference}, pages 1028--1039, 2025{\natexlab{a}}.

\bibitem[Piccinelli et~al.(2025{\natexlab{b}})Piccinelli, Sakaridis, Yang, Segu, Li, Abbeloos, and Van~Gool]{piccinelli2025unidepthv2}
L.~Piccinelli, C.~Sakaridis, Y.-H. Yang, M.~Segu, S.~Li, W.~Abbeloos, and L.~Van~Gool.
\newblock Unidepthv2: Universal monocular metric depth estimation made simpler.
\newblock \emph{IEEE Transactions on Pattern Analysis and Machine Intelligence}, 2025{\natexlab{b}}.

\bibitem[Radford et~al.(2018)Radford, Narasimhan, Salimans, Sutskever, et~al.]{radford2018improving_gpt1}
A.~Radford, K.~Narasimhan, T.~Salimans, I.~Sutskever, et~al.
\newblock Improving language understanding by generative pre-training.
\newblock 2018.

\bibitem[Radford et~al.(2019)Radford, Wu, Child, Luan, Amodei, Sutskever, et~al.]{radford2019language_gpt2}
A.~Radford, J.~Wu, R.~Child, D.~Luan, D.~Amodei, I.~Sutskever, et~al.
\newblock Language models are unsupervised multitask learners.
\newblock \emph{OpenAI blog}, 1\penalty0 (8):\penalty0 9, 2019.

\bibitem[Radford et~al.(2021)Radford, Kim, Hallacy, Ramesh, Goh, Agarwal, Sastry, Askell, Mishkin, Clark, et~al.]{radford2021learning_clip}
A.~Radford, J.~W. Kim, C.~Hallacy, A.~Ramesh, G.~Goh, S.~Agarwal, G.~Sastry, A.~Askell, P.~Mishkin, J.~Clark, et~al.
\newblock Learning transferable visual models from natural language supervision.
\newblock In \emph{International conference on machine learning}, pages 8748--8763. PmLR, 2021.

\bibitem[Ranzato et~al.(2011)Ranzato, Susskind, Mnih, and Hinton]{ranzato2011deep}
M.~Ranzato, J.~Susskind, V.~Mnih, and G.~Hinton.
\newblock On deep generative models with applications to recognition.
\newblock In \emph{CVPR 2011}, pages 2857--2864. IEEE, 2011.

\bibitem[Ravi et~al.(2024)Ravi, Gabeur, Hu, Hu, Ryali, Ma, Khedr, R{\"a}dle, Rolland, Gustafson, Mintun, Pan, Alwala, Carion, Wu, Girshick, Doll{\'a}r, and Feichtenhofer]{ravi2024sam2}
N.~Ravi, V.~Gabeur, Y.-T. Hu, R.~Hu, C.~Ryali, T.~Ma, H.~Khedr, R.~R{\"a}dle, C.~Rolland, L.~Gustafson, E.~Mintun, J.~Pan, K.~V. Alwala, N.~Carion, C.-Y. Wu, R.~Girshick, P.~Doll{\'a}r, and C.~Feichtenhofer.
\newblock Sam 2: Segment anything in images and videos.
\newblock \emph{arXiv preprint arXiv:2408.00714}, 2024.

\bibitem[Ren et~al.(2024)Ren, Chen, Jiang, Zeng, Xiong, Liu, Ma, Shen, Gao, Jiang, et~al.]{ren2024dinox}
T.~Ren, Y.~Chen, Q.~Jiang, Z.~Zeng, Y.~Xiong, W.~Liu, Z.~Ma, J.~Shen, Y.~Gao, X.~Jiang, et~al.
\newblock Dino-x: A unified vision model for open-world object detection and understanding.
\newblock \emph{arXiv preprint arXiv:2411.14347}, 2024.

\bibitem[Schops et~al.(2019)Schops, Sattler, and Pollefeys]{eth3d}
T.~Schops, T.~Sattler, and M.~Pollefeys.
\newblock Bad slam: Bundle adjusted direct rgb-d slam.
\newblock In \emph{Proceedings of the IEEE/CVF Conference on Computer Vision and Pattern Recognition}, pages 134--144, 2019.

\bibitem[Shen et~al.(2024)Shen, Fu, Chen, Zhang, Li, Sun, Wu, Lin, and Ji]{shen2024_aped}
Y.~Shen, C.~Fu, P.~Chen, M.~Zhang, K.~Li, X.~Sun, Y.~Wu, S.~Lin, and R.~Ji.
\newblock Aligning and prompting everything all at once for universal visual perception.
\newblock In \emph{Proceedings of the IEEE/CVF Conference on Computer Vision and Pattern Recognition}, pages 13193--13203, 2024.

\bibitem[Silberman et~al.(2012)Silberman, Hoiem, Kohli, and Fergus]{nyuv2}
N.~Silberman, D.~Hoiem, P.~Kohli, and R.~Fergus.
\newblock Indoor segmentation and support inference from rgbd images.
\newblock In \emph{European conference on computer vision}, pages 746--760. Springer, 2012.

\bibitem[Sim{\'e}oni et~al.(2025)Sim{\'e}oni, Vo, Seitzer, Baldassarre, Oquab, Jose, Khalidov, Szafraniec, Yi, Ramamonjisoa, et~al.]{simeoni2025dinov3}
O.~Sim{\'e}oni, H.~V. Vo, M.~Seitzer, F.~Baldassarre, M.~Oquab, C.~Jose, V.~Khalidov, M.~Szafraniec, S.~Yi, M.~Ramamonjisoa, et~al.
\newblock Dinov3.
\newblock \emph{arXiv preprint arXiv:2508.10104}, 2025.

\bibitem[Tang et~al.(2023)Tang, Jia, Wang, Phoo, and Hariharan]{tang2023emergent}
L.~Tang, M.~Jia, Q.~Wang, C.~P. Phoo, and B.~Hariharan.
\newblock Emergent correspondence from image diffusion.
\newblock \emph{Advances in neural information processing systems}, 36:\penalty0 1363--1389, 2023.

\bibitem[Tschannen et~al.(2025)Tschannen, Gritsenko, Wang, Naeem, Alabdulmohsin, Parthasarathy, Evans, Beyer, Xia, Mustafa, et~al.]{tschannen2025siglip2}
M.~Tschannen, A.~Gritsenko, X.~Wang, M.~F. Naeem, I.~Alabdulmohsin, N.~Parthasarathy, T.~Evans, L.~Beyer, Y.~Xia, B.~Mustafa, et~al.
\newblock Siglip 2: Multilingual vision-language encoders with improved semantic understanding, localization, and dense features.
\newblock \emph{arXiv preprint arXiv:2502.14786}, 2025.

\bibitem[Uhrig et~al.(2017)Uhrig, Schneider, Schneider, Franke, Brox, and Geiger]{Uhrig2017THREEDV_kitti}
J.~Uhrig, N.~Schneider, L.~Schneider, U.~Franke, T.~Brox, and A.~Geiger.
\newblock Sparsity invariant cnns.
\newblock In \emph{International Conference on 3D Vision (3DV)}, 2017.

\bibitem[Vasiljevic et~al.(2019)Vasiljevic, Kolkin, Zhang, Luo, Wang, Dai, Daniele, Mostajabi, Basart, Walter, and Shakhnarovich]{diode_dataset}
I.~Vasiljevic, N.~Kolkin, S.~Zhang, R.~Luo, H.~Wang, F.~Z. Dai, A.~F. Daniele, M.~Mostajabi, S.~Basart, M.~R. Walter, and G.~Shakhnarovich.
\newblock {DIODE}: {A} {D}ense {I}ndoor and {O}utdoor {DE}pth {D}ataset.
\newblock \emph{CoRR}, abs/1908.00463, 2019.
\newblock URL \url{http://arxiv.org/abs/1908.00463}.

\bibitem[Wang et~al.(2026{\natexlab{a}})Wang, Qiao, Jie, Huang, Feng, Zheng, Ma, Lan, and Liang]{wang2026xsam}
H.~Wang, L.~Qiao, Z.~Jie, Z.~Huang, C.~Feng, Q.~Zheng, L.~Ma, X.~Lan, and X.~Liang.
\newblock X-sam: From segment anything to any segmentation.
\newblock In \emph{Proceedings of the AAAI Conference on Artificial Intelligence}, volume~40, pages 26187--26196, 2026{\natexlab{a}}.

\bibitem[Wang et~al.(2025{\natexlab{a}})Wang, Chen, Karaev, Vedaldi, Rupprecht, and Novotny]{wang2025vggt}
J.~Wang, M.~Chen, N.~Karaev, A.~Vedaldi, C.~Rupprecht, and D.~Novotny.
\newblock Vggt: Visual geometry grounded transformer.
\newblock In \emph{Proceedings of the Computer Vision and Pattern Recognition Conference}, pages 5294--5306, 2025{\natexlab{a}}.

\bibitem[Wang et~al.(2026{\natexlab{b}})Wang, Zanfir, Bazavan, Andriluka, and Sminchisescu]{wang2026thfm}
L.~Wang, A.~Zanfir, E.~G. Bazavan, M.~Andriluka, and C.~Sminchisescu.
\newblock Thfm: A unified video foundation model for 4d human perception and beyond.
\newblock \emph{arXiv preprint arXiv:2603.25892}, 2026{\natexlab{b}}.

\bibitem[Wang et~al.(2025{\natexlab{b}})Wang, Xu, Dai, Xiang, Deng, Tong, and Yang]{wang2024moge}
R.~Wang, S.~Xu, C.~Dai, J.~Xiang, Y.~Deng, X.~Tong, and J.~Yang.
\newblock Moge: Unlocking accurate monocular geometry estimation for open-domain images with optimal training supervision.
\newblock In \emph{IEEE/CVF Conference on Computer Vision and Pattern Recognition (CVPR)}, 2025{\natexlab{b}}.

\bibitem[Wang et~al.(2025{\natexlab{c}})Wang, Xu, Dong, Deng, Xiang, Lv, Sun, Tong, and Yang]{wang2025moge2}
R.~Wang, S.~Xu, Y.~Dong, Y.~Deng, J.~Xiang, Z.~Lv, G.~Sun, X.~Tong, and J.~Yang.
\newblock Moge-2: Accurate monocular geometry with metric scale and sharp details.
\newblock \emph{arXiv preprint arXiv:2507.02546}, 2025{\natexlab{c}}.

\bibitem[Wang et~al.(2023)Wang, Wang, Cao, Shen, and Huang]{wang2023images}
X.~Wang, W.~Wang, Y.~Cao, C.~Shen, and T.~Huang.
\newblock Images speak in images: A generalist painter for in-context visual learning.
\newblock In \emph{Proceedings of the IEEE/CVF Conference on Computer Vision and Pattern Recognition}, pages 6830--6839, 2023.

\bibitem[Wei et~al.(2024)Wei, Zhong, Tan, Liu, Zhao, Hu, and Yang]{wei2024hyperseg}
C.~Wei, Y.~Zhong, H.~Tan, Y.~Liu, Z.~Zhao, J.~Hu, and Y.~Yang.
\newblock Hyperseg: Towards universal visual segmentation with large language model.
\newblock \emph{arXiv preprint arXiv:2411.17606}, 2024.

\bibitem[Wei et~al.(2021)Wei, Bosma, Zhao, Guu, Yu, Lester, Du, Dai, and Le]{wei2021finetuned_flan}
J.~Wei, M.~Bosma, V.~Y. Zhao, K.~Guu, A.~W. Yu, B.~Lester, N.~Du, A.~M. Dai, and Q.~V. Le.
\newblock Finetuned language models are zero-shot learners.
\newblock \emph{arXiv preprint arXiv:2109.01652}, 2021.

\bibitem[Wiedemer et~al.(2025)Wiedemer, Li, Vicol, Gu, Matarese, Swersky, Kim, Jaini, and Geirhos]{wiedemer2025veo_learner}
T.~Wiedemer, Y.~Li, P.~Vicol, S.~S. Gu, N.~Matarese, K.~Swersky, B.~Kim, P.~Jaini, and R.~Geirhos.
\newblock Video models are zero-shot learners and reasoners.
\newblock \emph{arXiv preprint arXiv:2509.20328}, 2025.

\bibitem[Wu et~al.(2025)Wu, Li, Zhou, Lin, Gao, Yan, Yin, Bai, Xu, Chen, et~al.]{wu2025qwen_image}
C.~Wu, J.~Li, J.~Zhou, J.~Lin, K.~Gao, K.~Yan, S.-m. Yin, S.~Bai, X.~Xu, Y.~Chen, et~al.
\newblock Qwen-image technical report.
\newblock \emph{arXiv preprint arXiv:2508.02324}, 2025.

\bibitem[Xie et~al.(2024)Xie, Mao, Bai, Zhang, Wang, Lin, Gu, Chen, Yang, and Shou]{xie2024show}
J.~Xie, W.~Mao, Z.~Bai, D.~J. Zhang, W.~Wang, K.~Q. Lin, Y.~Gu, Z.~Chen, Z.~Yang, and M.~Z. Shou.
\newblock Show-o: One single transformer to unify multimodal understanding and generation.
\newblock \emph{arXiv preprint arXiv:2408.12528}, 2024.

\bibitem[Xu et~al.(2023)Xu, Liu, Vahdat, Byeon, Wang, and De~Mello]{xu2023odise}
J.~Xu, S.~Liu, A.~Vahdat, W.~Byeon, X.~Wang, and S.~De~Mello.
\newblock Open-vocabulary panoptic segmentation with text-to-image diffusion models.
\newblock In \emph{Proceedings of the IEEE/CVF conference on computer vision and pattern recognition}, pages 2955--2966, 2023.

\bibitem[Yang et~al.(2024)Yang, Kang, Huang, Zhao, Xu, Feng, and Zhao]{yang2024dav2}
L.~Yang, B.~Kang, Z.~Huang, Z.~Zhao, X.~Xu, J.~Feng, and H.~Zhao.
\newblock Depth anything v2.
\newblock \emph{Advances in Neural Information Processing Systems}, 37:\penalty0 21875--21911, 2024.

\bibitem[Yang and Wang(2023)]{yang2023diffusion}
X.~Yang and X.~Wang.
\newblock Diffusion model as representation learner.
\newblock In \emph{Proceedings of the IEEE/CVF International Conference on Computer Vision}, pages 18938--18949, 2023.

\bibitem[Ye et~al.(2024)Ye, Qiu, Gu, Zuo, Wu, Dong, Bo, Xiu, and Han]{ye2024stablenormal}
C.~Ye, L.~Qiu, X.~Gu, Q.~Zuo, Y.~Wu, Z.~Dong, L.~Bo, Y.~Xiu, and X.~Han.
\newblock Stablenormal: Reducing diffusion variance for stable and sharp normal.
\newblock \emph{ACM Transactions on Graphics (ToG)}, 43\penalty0 (6):\penalty0 1--18, 2024.

\bibitem[Ye et~al.(2025)Ye, He, Li, Lin, Yuan, Yan, Hou, and Yuan]{ye2025imgedit}
Y.~Ye, X.~He, Z.~Li, B.~Lin, S.~Yuan, Z.~Yan, B.~Hou, and L.~Yuan.
\newblock Imgedit: A unified image editing dataset and benchmark.
\newblock \emph{arXiv preprint arXiv:2505.20275}, 2025.

\bibitem[Yu et~al.(2024)Yu, Jiang, Zhang, Chen, Li, Zhang, and Lu]{yu2024high}
Q.~Yu, P.-T. Jiang, H.~Zhang, J.~Chen, B.~Li, L.~Zhang, and H.~Lu.
\newblock High-precision dichotomous image segmentation via probing diffusion capacity.
\newblock \emph{arXiv preprint arXiv:2410.10105}, 2024.

\bibitem[Zhai et~al.(2023)Zhai, Mustafa, Kolesnikov, and Beyer]{zhai2023siglip1}
X.~Zhai, B.~Mustafa, A.~Kolesnikov, and L.~Beyer.
\newblock Sigmoid loss for language image pre-training.
\newblock In \emph{Proceedings of the IEEE/CVF international conference on computer vision}, pages 11975--11986, 2023.

\bibitem[Zhang et~al.(2025)Zhang, Moing, Koppula, Rocco, Momeni, Xie, Sun, Sukthankar, Barral, Hadsell, et~al.]{zhang2025d4rt}
C.~Zhang, G.~L. Moing, S.~Koppula, I.~Rocco, L.~Momeni, J.~Xie, S.~Sun, R.~Sukthankar, J.~K. Barral, R.~Hadsell, et~al.
\newblock Efficiently reconstructing dynamic scenes one d4rt at a time.
\newblock \emph{arXiv preprint arXiv:2512.08924}, 2025.

\bibitem[Zhang et~al.(2023{\natexlab{a}})Zhang, Li, Zou, Liu, Li, Yang, and Zhang]{zhang2023simple_openseed}
H.~Zhang, F.~Li, X.~Zou, S.~Liu, C.~Li, J.~Yang, and L.~Zhang.
\newblock A simple framework for open-vocabulary segmentation and detection.
\newblock In \emph{Proceedings of the IEEE/CVF International Conference on Computer Vision}, pages 1020--1031, 2023{\natexlab{a}}.

\bibitem[Zhang et~al.(2023{\natexlab{b}})Zhang, Herrmann, Hur, Polania~Cabrera, Jampani, Sun, and Yang]{zhang2023tale}
J.~Zhang, C.~Herrmann, J.~Hur, L.~Polania~Cabrera, V.~Jampani, D.~Sun, and M.-H. Yang.
\newblock A tale of two features: Stable diffusion complements dino for zero-shot semantic correspondence.
\newblock \emph{Advances in Neural Information Processing Systems}, 36:\penalty0 45533--45547, 2023{\natexlab{b}}.

\bibitem[Zhao et~al.(2025)Zhao, Sun, Liu, Zheng, Zhu, Zhao, Chen, He, and Shen]{zhao2025diception}
C.~Zhao, Y.~Sun, M.~Liu, H.~Zheng, M.~Zhu, Z.~Zhao, H.~Chen, T.~He, and C.~Shen.
\newblock Diception: A generalist diffusion model for visual perceptual tasks.
\newblock \emph{arXiv preprint arXiv:2502.17157}, 2025.

\bibitem[Zhao et~al.(2023)Zhao, Rao, Liu, Liu, Zhou, and Lu]{zhao2023unleashing}
W.~Zhao, Y.~Rao, Z.~Liu, B.~Liu, J.~Zhou, and J.~Lu.
\newblock Unleashing text-to-image diffusion models for visual perception.
\newblock In \emph{IEEE/CVF International Conference on Computer Vision (ICCV)}, pages 5729--5739, 2023.

\bibitem[Zhou et~al.(2021)Zhou, Wei, Wang, Shen, Xie, Yuille, and Kong]{zhou2021ibot}
J.~Zhou, C.~Wei, H.~Wang, W.~Shen, C.~Xie, A.~Yuille, and T.~Kong.
\newblock ibot: Image bert pre-training with online tokenizer.
\newblock \emph{arXiv preprint arXiv:2111.07832}, 2021.

\bibitem[Zou et~al.(2023)Zou, Dou, Yang, Gan, Li, Li, Dai, Behl, Wang, Yuan, et~al.]{zou2023generalized_xdecoder}
X.~Zou, Z.-Y. Dou, J.~Yang, Z.~Gan, L.~Li, C.~Li, X.~Dai, H.~Behl, J.~Wang, L.~Yuan, et~al.
\newblock Generalized decoding for pixel, image, and language.
\newblock In \emph{Proceedings of the IEEE/CVF conference on computer vision and pattern recognition}, pages 15116--15127, 2023.

\bibitem[Zuo et~al.(2025)Zuo, Deng, Zhou, Zhu, Zhang, Zhang, Yan, Huang, Chen, Deng, Jin, Sang, and Gao]{zuo2025nanobanana_low_level}
J.~Zuo, H.~Deng, H.~Zhou, J.~Zhu, Y.~Zhang, Y.~Zhang, Y.~Yan, K.~Huang, W.~Chen, Y.~Deng, R.~Jin, N.~Sang, and C.~Gao.
\newblock Is nano banana pro a low-level vision all-rounder? {A} comprehensive evaluation on 14 tasks and 40 datasets.
\newblock \emph{arXiv preprint}, 2025.

\end{thebibliography}

\newpage

\section*{Appendix}

\appendix

\section{Multi-stage clustering algorithm for parsing instance masks}
\label{sec:mask_parsing}
While the postprocessing of RGB images into binary masks is straightforward for referring expression segmentation and semantic segmentation, the task is more complex for instance segmentation, for which the number of masks to predict is unknown a priori.
We therefore instruct the model to dynamically determine the number of masks and assign a unique color to each individual instance.
However, because the output is a generated image, it is subject to high-frequency generation noise, slight color drift across a single object mask, and blending artifacts at boundaries.
Standard connected-component algorithms or simple color thresholding fail in the presence of these artifacts.
Our proposed algorithm acts as a robust post-processing, ensuring that continuous generative outputs are cleanly mapped back to discrete instance masks.

To parse these individual instance masks from the generated multi-colored segmentation images, we design a multi-stage connected component grouping algorithm.  
For any generated RGB mask image $I \in [0, 255]^{H \times W \times 3}$ with a predefined background color $C_{\text{bg}}$, the algorithm proceeds as follows:
\begin{enumerate}
    \item \textbf{Background Initialization}: We first identify and label all background pixels.
    A pixel $p$ is classified as background (and initialized with label $0$) if its color is close to the predefined background color $C_{\text{bg}}$:
    \begin{equation}
        \|C(p) - C_{\text{bg}}\|^2 < \tau^2
    \end{equation}
    where $C(p)$ denotes the RGB vector of pixel $p$, and $\tau = 14$ is the color tolerance threshold.
    Background pixels are excluded from subsequent steps.
    
    \item \textbf{Color Similarity Grouping}: We group the remaining unlabeled pixels into initial components using a seed-based floodfill algorithm with 16-connectivity.
    Two adjacent pixels $p_i, p_j$ are merged into the same component if their RGB color distance to the component's seed pixel $p_{\text{seed}}$ satisfies:
    \begin{equation}
        \|C(p_i) - C(p_{\text{seed}})\|^2 \leq \tau^2
    \end{equation}
    
    \item \textbf{Noise Pruning}: To filter out minor noise, any component $S$ is discarded (re-labeled as background $0$) if its total pixel area is smaller than a fraction $\theta_{\text{size}}$ of the image size:
    \begin{equation}
        |S| < \theta_{\text{size}} \cdot (H \cdot W)
    \end{equation}
    where $\theta_{\text{size}} = 2 \times 10^{-4}$ (representing 0.02\% of the image area).
    
    \item \textbf{Boundary Artifact Elimination}: Generative models often produce thin colored boundary halos between distinct solid-colored objects.
    To prevent these from forming separate spurious components, we apply a binary erosion $\ominus$ using a square structuring element $K$ of size $3 \times 3$.
    A component $S$ is pruned if its preserved area ratio after erosion is below $\theta_{\text{erosion}} = 0.1$:
    \begin{equation}
        \frac{|S \ominus K|}{|S|} < \theta_{\text{erosion}}
    \end{equation}
    
    \item \textbf{Spatially Constrained Merging}: Spatially disjoint components $S_a, S_b$ that could represent parts of the same physical object (e.g., due to foreground occlusion) are merged if their average colors $\mu_a, \mu_b$ are similar ($\|\mu_a - \mu_b\|^2 \leq \tau^2$) and if their combined bounding box area $A(S_a \cup S_b)$ does not expand excessively compared to the sum of their individual areas:
    \begin{equation}
        A(S_a \cup S_b) \leq \gamma \cdot (A(S_a) + A(S_b))
    \end{equation}
    where $\gamma = 5.0$ is the maximum bounding box expansion factor.
\end{enumerate}

\clearpage

\section{Detailed SA-Co/Gold Quantitative Evaluation}
\label{sec:saco_details}
To provide a deeper understanding of our model's instance segmentation capabilities, we report the per-category performance breakdown on the SA-Co/Gold \citep{sam3} benchmark in \cref{tab-seg-saco_full}.
This benchmark evaluates models on diverse noun-phrase queries across seven distinct subsets: Metaclip, SA-1B, Crowded, Food\&Drink, Sports Equip., Attributes, and Wiki-Common.

\begin{table}[htbp]
	\centering
	\resizebox{\textwidth}{!}{
		\renewcommand{\arraystretch}{1.2}
		\setlength{\tabcolsep}{4pt}
		
		\begin{tabular}{l|ccc|ccc|ccc|ccc|ccc|ccc|ccc|ccc}
			\hline
			& \multicolumn{3}{c|}{\textbf{Average}}
			& \multicolumn{3}{c|}{\textbf{Metaclip}} 
			& \multicolumn{3}{c|}{\textbf{SA-1B}} 
			& \multicolumn{3}{c|}{\textbf{Crowded}} 
			& \multicolumn{3}{c|}{\textbf{Food\&Drink}} 
			& \multicolumn{3}{c|}{\textbf{Sports Equip.}} 
			& \multicolumn{3}{c|}{\textbf{Attributes}} 
			& \multicolumn{3}{c}{\textbf{Wiki-Common}} \\
			\hline
			& cgF\textsubscript{1} & IL\_MCC & pmF\textsubscript{1} 
			& cgF\textsubscript{1} & IL\_MCC & pmF\textsubscript{1} 
			& cgF\textsubscript{1} & IL\_MCC & pmF\textsubscript{1} 
			& cgF\textsubscript{1} & IL\_MCC & pmF\textsubscript{1} 
			& cgF\textsubscript{1} & IL\_MCC & pmF\textsubscript{1} 
			& cgF\textsubscript{1} & IL\_MCC & pmF\textsubscript{1} 
			& cgF\textsubscript{1} & IL\_MCC & pmF\textsubscript{1} 
			& cgF\textsubscript{1} & IL\_MCC & pmF\textsubscript{1} \\
            \hline
            
            \rowcolor{gray!20}
            \multicolumn{25}{l}{\textbf{Trained on SA-Co}} \\
			SAM 3~\citep{sam3} & 54.1 & 0.82 & 66.1 & 47.3 & 0.81 & 58.6 & 53.7 & 0.86 & 62.6 & 61.1 & \textbf{0.9} & 67.7 & 53.4 & 0.79 & 67.3 & 65.5 & \textbf{0.89} & 73.8 & 54.9 & 0.76 & 72.0 & 42.5 & 0.70 & 60.9 \\
			SAM 3 + Llama 3.2 (ft)~\citep{sam3}  & \textbf{61.2} & \textbf{0.86} & \textbf{70.8} & \textbf{54.2} & \textbf{0.85} & \textbf{64.0} & \textbf{56.0} & \textbf{0.89} & \textbf{62.9} & \textbf{61.3} & 0.88 & \textbf{69.8} & \textbf{67.6} & \textbf{0.86} & \textbf{78.5} & \textbf{67.5} & \textbf{0.89} & \textbf{75.6} & \textbf{71.1} & \textbf{0.91} & \textbf{77.8} & \textbf{51.1} & \textbf{0.76} & \textbf{67.1} \\
			\hline
			
            \rowcolor{gray!20}
            \multicolumn{25}{l}{\textbf{Zero-shot}} \\
			gDino-T~\citep{liu2024grounding_dino} & 3.3 & 0.15 & 16.2 & 2.9 & 0.21 & 13.9 & 3.1 & 0.20 & 15.4 & 0.28 & 0.08 & 3.4 & 0.96 & 0.10 & 9.8 & 1.1 & 0.10 & 11.2 & 13.8 & 0.29 & 47.3 & 0.70 & 0.06 & 12.1 \\
			OWLv2~\citep{minderer2023gowol} & 24.6 & 0.57 & 42.0 & 17.7 & 0.52 & 34.3 & 13.3 & 0.50 & 26.8 & 15.8 & 0.51 & 30.7 & 32.0 & 0.65 & 49.4 & 36.0 & 0.64 & 56.2 & 35.6 & 0.63 & 56.2 & 21.7 & 0.54 & 40.3 \\
			LLMDet-L~\citep{fu2025llmdet} & 6.5 & 0.21 & 27.3 & 4.5 & 0.23 & 19.4 & 5.3 & 0.23 & 22.8 & 2.4 & 0.18 & 13.7 & 5.5 & 0.19 & 29.1 & 4.4 & 0.17 & 25.3 & 22.2 & 0.39 & 57.1 & 1.2 & 0.05 & 23.3 \\
			APE-D~\citep{shen2024_aped} & 16.4 & 0.40 & 36.9 & 12.6 & 0.42 & 30.1 & 2.2 & 0.22 & 10.0 & 7.2 & 0.35 & 20.3 & 22.7 & 0.51 & 45.0 & 31.8 & 0.56 & 56.5 & 26.7 & 0.47 & 57.3 & 11.6 & 0.29 & 39.5 \\
			DINO-X~\citep{ren2024dinox} & 21.3 & 0.38 & 55.2 & 17.2 & 0.35 & \textbf{49.2} & 19.7 & 0.48 & \textbf{40.9} & 12.9 & 0.34 & 37.5 & 30.1 & 0.49 & 61.7 & 28.4 & 0.41 & \textbf{69.4} & 31.0 & 0.42 & \textbf{74.0} & 9.7 & 0.18 & 53.5 \\
			Gemini 2.5~\citep{gemini25} & 13.0 & 0.29 & 46.1 & 9.9 & 0.29 & 33.8 & 13.1 & 0.41 & 32.1 & 8.2 & 0.27 & 30.3 & 19.6 & 0.33 & 59.5 & 15.1 & 0.28 & 53.5 & 18.8 & 0.30 & 63.1 & 6.5 & 0.13 & 50.3 \\
			\textbf{Vision Banana} + Gemini 3.1 Flash-Lite & \textbf{47.5} & \textbf{0.84} & \textbf{56.0} & \textbf{38.0} & \textbf{0.81} & 46.7 & \textbf{33.5} & \textbf{0.84} & 39.8 & \textbf{34.4} & \textbf{0.83} & \textbf{41.2} & \textbf{57.2} & \textbf{0.88} & \textbf{64.8} & \textbf{63.0} & \textbf{0.92} & 68.2 & \textbf{58.8} & \textbf{0.86} & 68.7 & \textbf{47.4} & \textbf{0.76} & \textbf{62.3} \\
			\hline
		\end{tabular}
	}
	\caption{Instance segmentation results on SA-Co/Gold. The evaluation results reported in this table come from the SAM3 paper \citep{sam3}, except for our method which we evaluated using the official SA-Co evaluation codebase.}
	\label{tab-seg-saco_full}
\end{table}

We adopt the official metrics established for SA-Co evaluation:
\begin{itemize}
    \item \textbf{Positive Micro-$F_1$ ($pmF_1$)}: The micro-averaged $F_1$ score computed across all positive queries (where the target noun-phrase is present in the image), evaluating the precision and recall of the predicted masks at 10 IoU thresholds in $[0.5 : 0.05 : 0.95]$.
    \item \textbf{Image-Level Matthews Correlation Coefficient ($IL\_MCC$)}: A metric evaluating the binary classification accuracy of the model in predicting whether a noun phrase query is present or absent in the image.
    \item \textbf{Classification-gated $F_1$ ($cgF_1$)}: The primary unified metric, combining $pmF_1$ and $IL\_MCC$:
    \begin{equation}
        cgF_1 = 100 \cdot pmF_1 \cdot IL\_MCC
    \end{equation}
\end{itemize}

Under the zero-shot transfer setting, Vision Banana (paired with Gemini 3.1 Flash-Lite for positive/negative query filtering) demonstrates state-of-the-art performance on the benchmark, outperforming specialized models such as OWLv2, DINO-X, and APE-D. 
Arguably, our strong $IL\_MCC$ score can be attributed to Gemini's discriminative capabilities, and prior works would also benefit from being paired with a MLLM. 
Nonetheless, Vision Banana also achieves state-of-the-art results on the $pmF_1$ metric, which directly measures open-vocabulary segmentation quality and is the focus of this work.

\section{Qualitative Instance Segmentation Analysis}
\label{sec:saco_qualitative}
\begin{figure*}[t]
  \centering
  \small
  \newcommand{\sacoimgw}{\dimexpr(\textwidth-17pt)/6\relax}
  \leftline{%
    \makebox[\sacoimgw][c]{\textbf{Original Image}}%
    \hspace{5pt}%
    \makebox[\sacoimgw][c]{\textbf{GT a}}%
    \hspace{1pt}%
    \makebox[\sacoimgw][c]{\textbf{GT b}}%
    \hspace{1pt}%
    \makebox[\sacoimgw][c]{\textbf{GT c}}%
    \hspace{5pt}%
    \makebox[\sacoimgw][c]{\makecell{\textbf{SAM 3} \\ \textbf{Predictions}}}%
    \hspace{5pt}%
    \makebox[\sacoimgw][c]{\makecell{\textbf{Vision Banana \emojivb} \\ \textbf{Predictions}}}%
  }
  \vspace{4pt}
  \includegraphics[width=\sacoimgw]{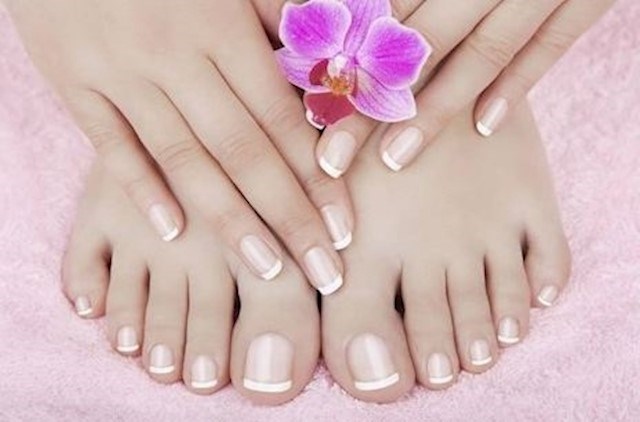}%
  \hspace{5pt}%
  \includegraphics[width=\sacoimgw]{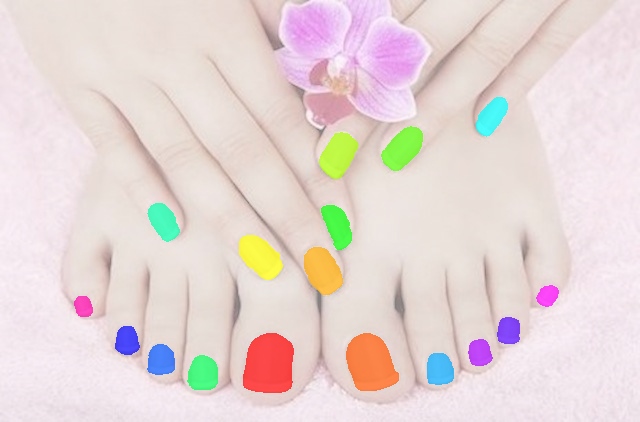}%
  \hspace{1pt}%
  \includegraphics[width=\sacoimgw]{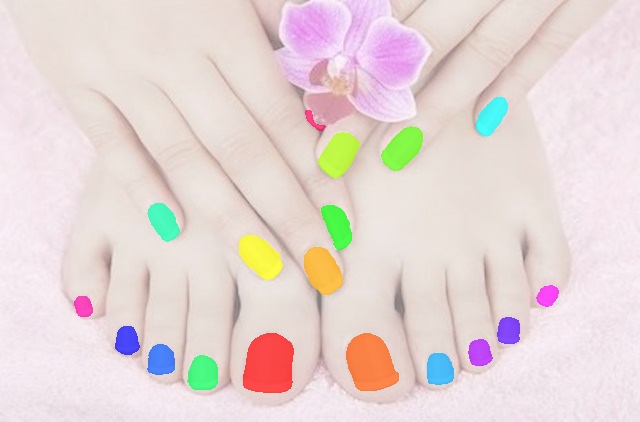}%
  \hspace{1pt}%
  \includegraphics[width=\sacoimgw]{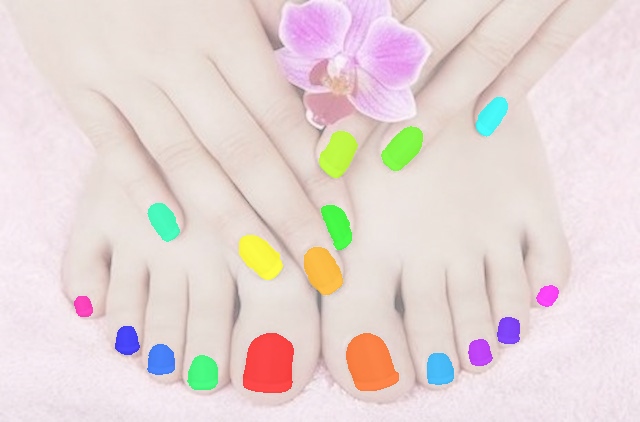}%
  \hspace{5pt}%
  \includegraphics[width=\sacoimgw]{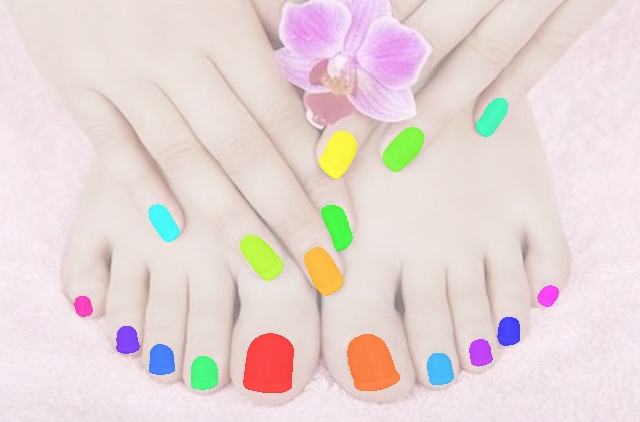}%
  \hspace{5pt}%
  \includegraphics[width=\sacoimgw]{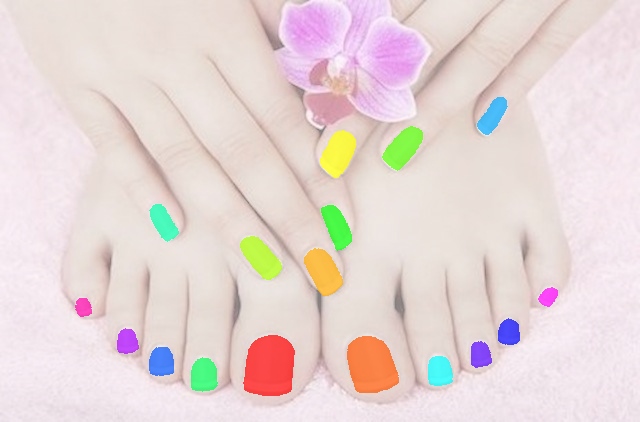}
  \\[-4pt]
  \leftline{\textbf{(\ref*{fig:saco_qualitative_successes}.a)} ``\textit{pale pink and white manicured and pedicured nail}``%
    \hfill%
    \makebox[\sacoimgw][c]{\tiny F1=0.90 (GT a), 17 masks}%
    \hspace{5pt}%
    \makebox[\sacoimgw][c]{\tiny F1=0.72 (GT c), 17 masks}}
  \includegraphics[width=\sacoimgw]{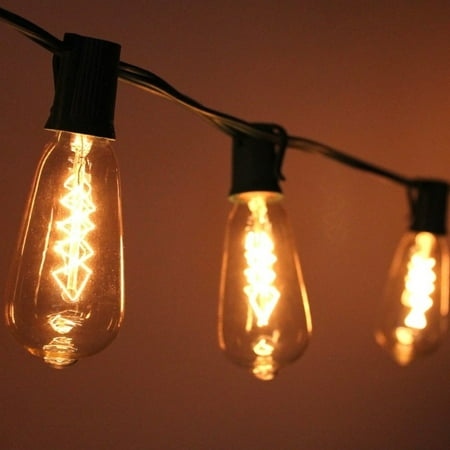}%
  \hspace{5pt}%
  \includegraphics[width=\sacoimgw]{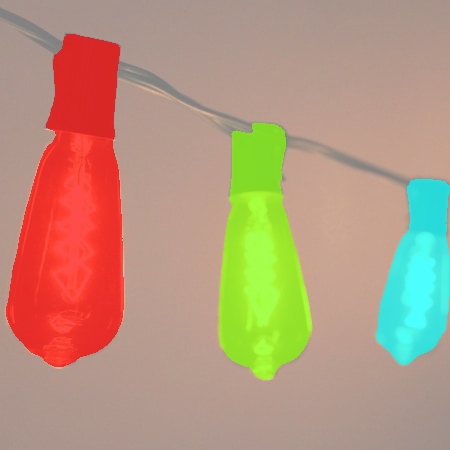}%
  \hspace{1pt}%
  \includegraphics[width=\sacoimgw]{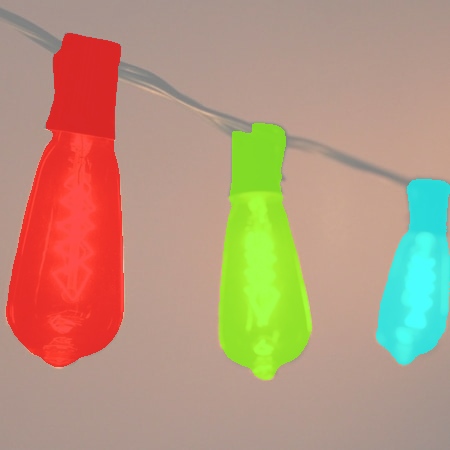}%
  \hspace{1pt}%
  \includegraphics[width=\sacoimgw]{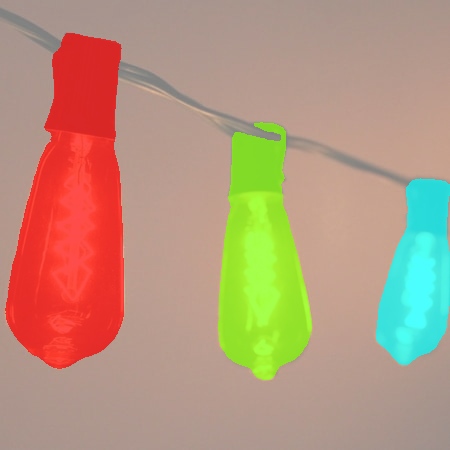}%
  \hspace{5pt}%
  \includegraphics[width=\sacoimgw]{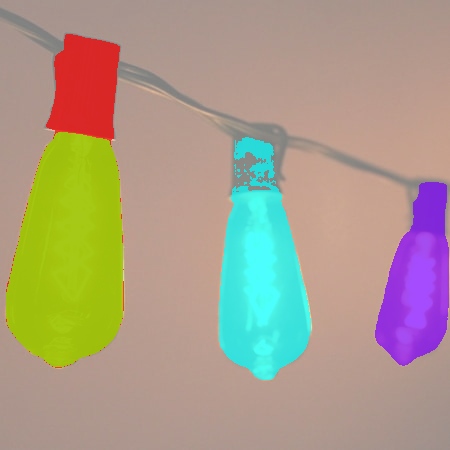}%
  \hspace{5pt}%
  \includegraphics[width=\sacoimgw]{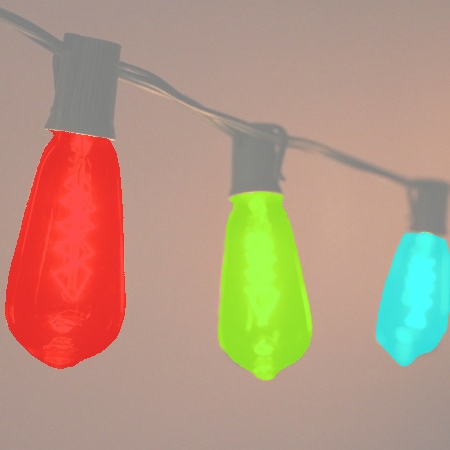}
  \\[-4pt]
  \leftline{\textbf{(\ref*{fig:saco_qualitative_successes}.b)} ``\textit{incandescent lightbulb}``%
    \hfill%
    \makebox[\sacoimgw][c]{\tiny F1=0.77 (GT a), 4 masks}%
    \hspace{5pt}%
    \makebox[\sacoimgw][c]{\tiny F1=0.60 (GT a), 3 masks}}
  \includegraphics[width=\sacoimgw]{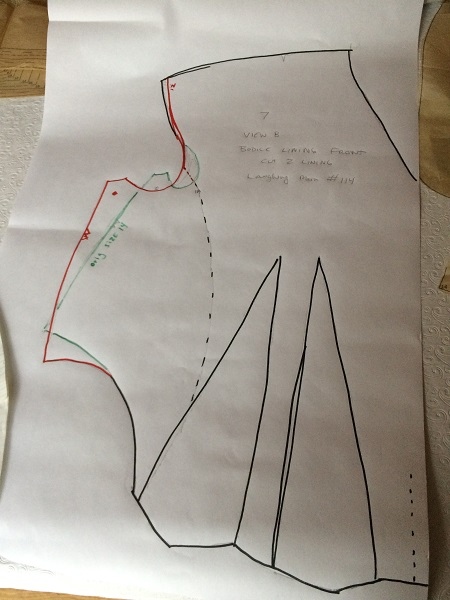}%
  \hspace{5pt}%
  \includegraphics[width=\sacoimgw]{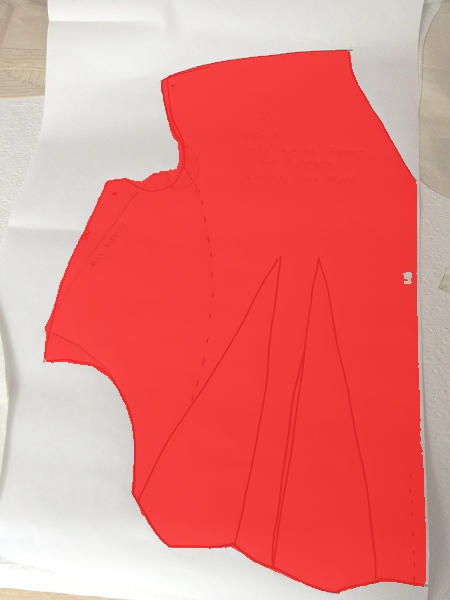}%
  \hspace{1pt}%
  \includegraphics[width=\sacoimgw]{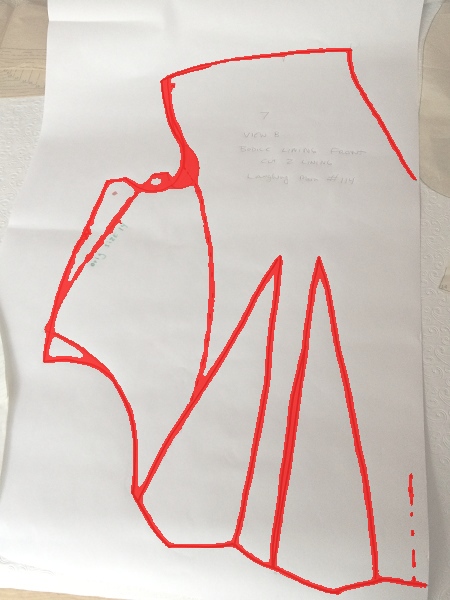}%
  \hspace{1pt}%
  \includegraphics[width=\sacoimgw]{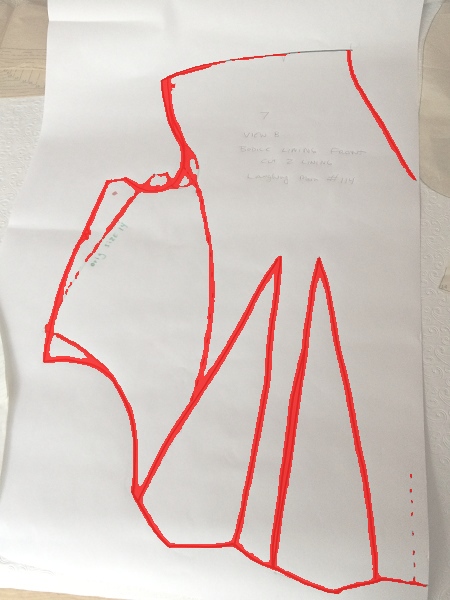}%
  \hspace{5pt}%
  \includegraphics[width=\sacoimgw]{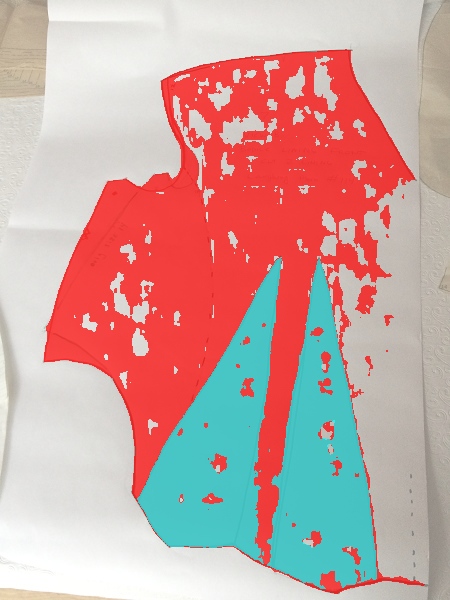}%
  \hspace{5pt}%
  \includegraphics[width=\sacoimgw]{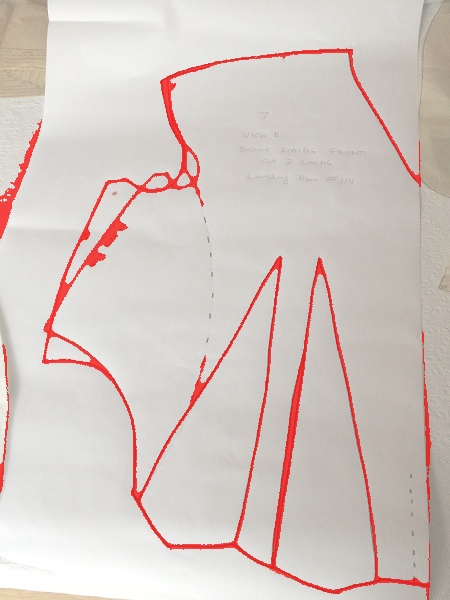}
  \\[-4pt]
  \leftline{\textbf{(\ref*{fig:saco_qualitative_successes}.c)} ``\textit{hand-drawn design}``%
    \hfill%
    \makebox[\sacoimgw][c]{\tiny F1=0.40 (GT a), 2 masks}%
    \hspace{5pt}%
    \makebox[\sacoimgw][c]{\tiny F1=0.10 (GT b), 1 mask}}
  \includegraphics[width=\sacoimgw]{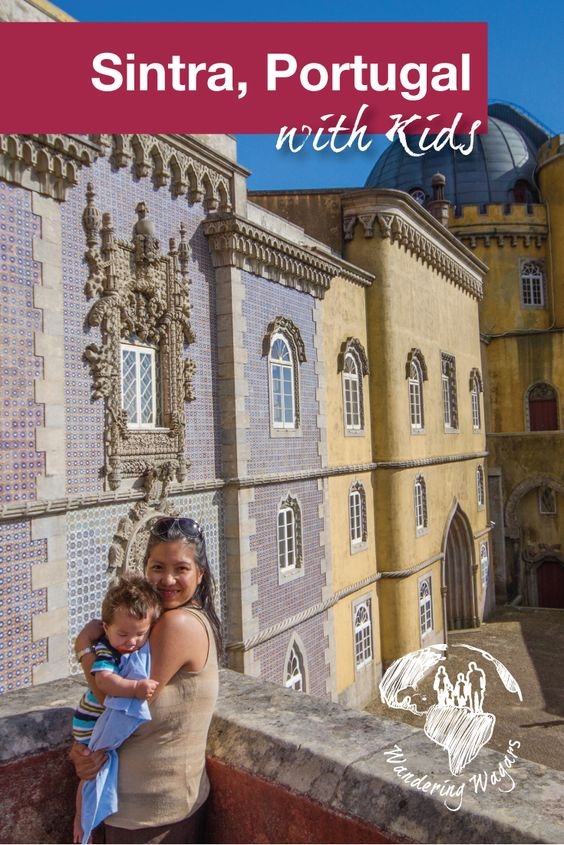}%
  \hspace{5pt}%
  \includegraphics[width=\sacoimgw]{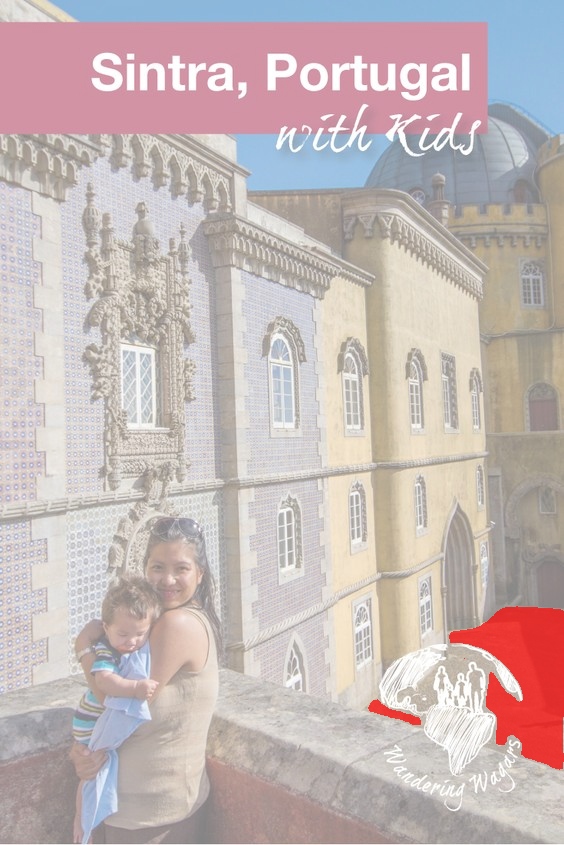}%
  \hspace{1pt}%
  \includegraphics[width=\sacoimgw]{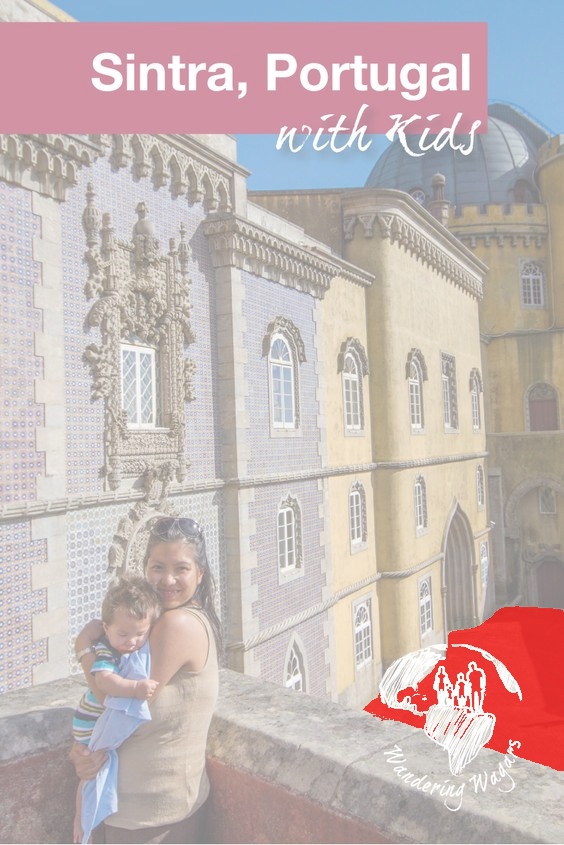}%
  \hspace{1pt}%
  \includegraphics[width=\sacoimgw]{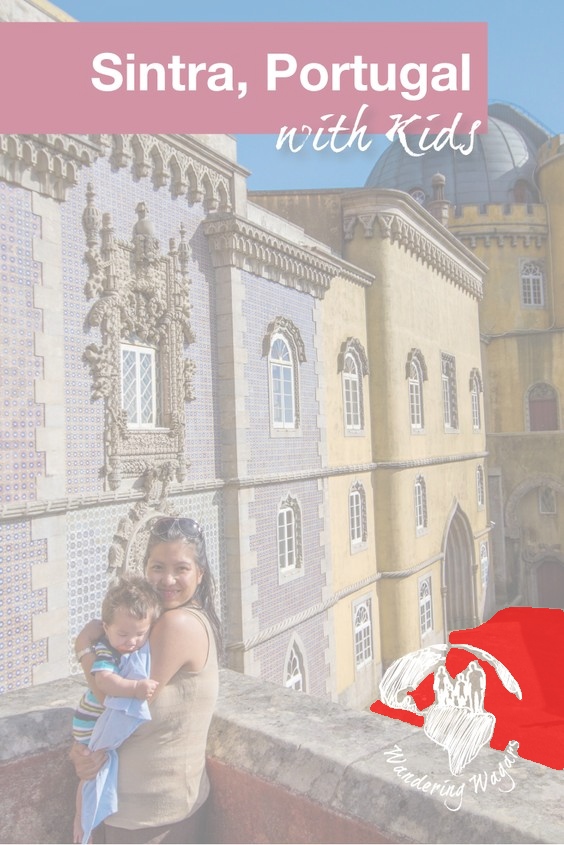}%
  \hspace{5pt}%
  \includegraphics[width=\sacoimgw]{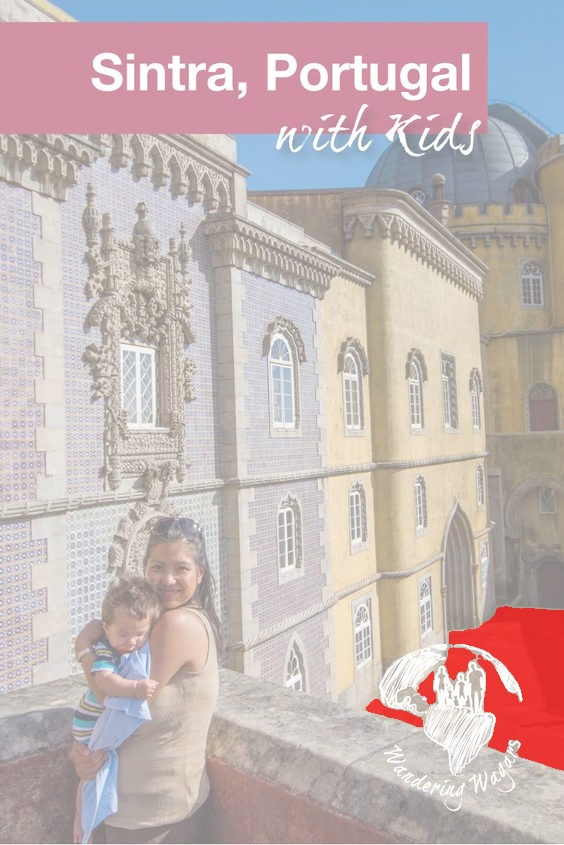}%
  \hspace{5pt}%
  \includegraphics[width=\sacoimgw]{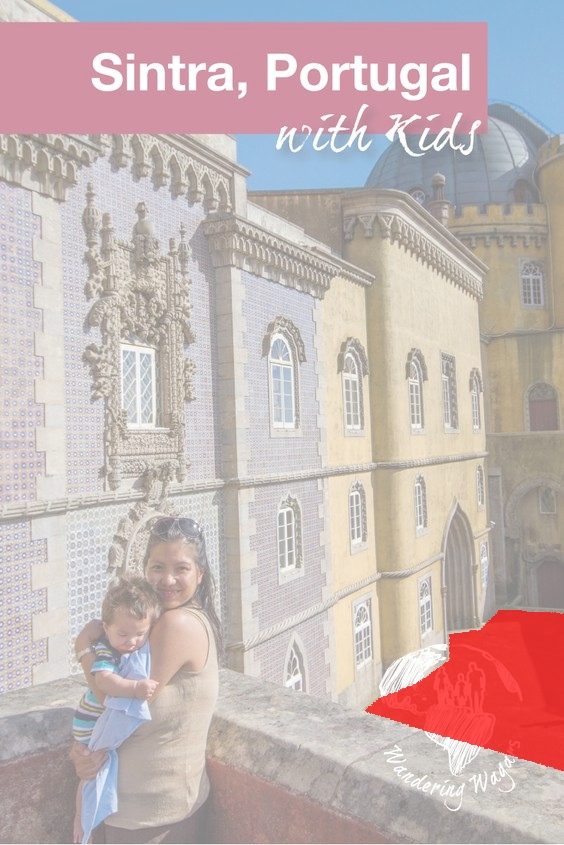}
  \\[-4pt]
  \leftline{\textbf{(\ref*{fig:saco_qualitative_successes}.d)} ``\textit{concrete floor}``%
    \hfill%
    \makebox[\sacoimgw][c]{\tiny F1=0.90 (GT c), 1 mask}%
    \hspace{5pt}%
    \makebox[\sacoimgw][c]{\tiny F1=0.50 (GT b), 1 mask}}
  \includegraphics[width=\sacoimgw]{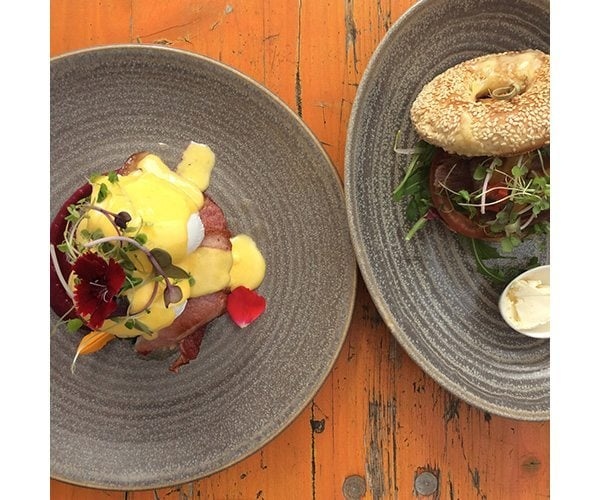}%
  \hspace{5pt}%
  \includegraphics[width=\sacoimgw]{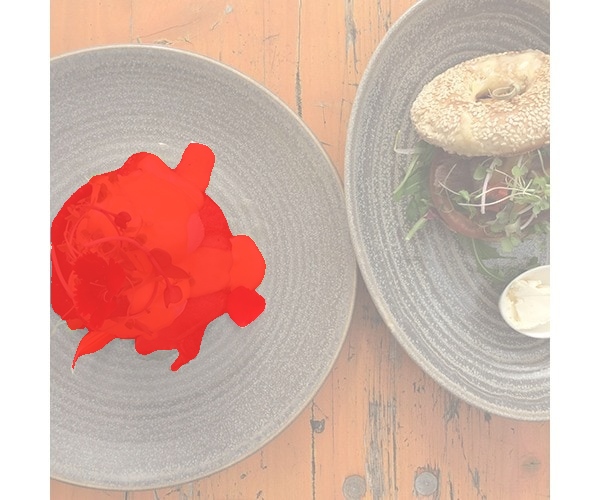}%
  \hspace{1pt}%
  \includegraphics[width=\sacoimgw]{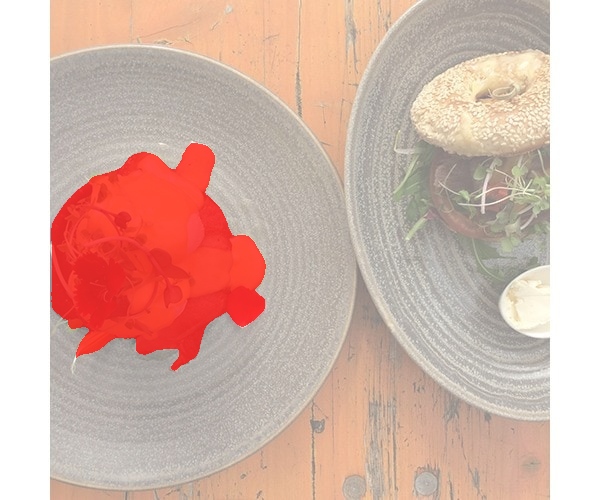}%
  \hspace{1pt}%
  \includegraphics[width=\sacoimgw]{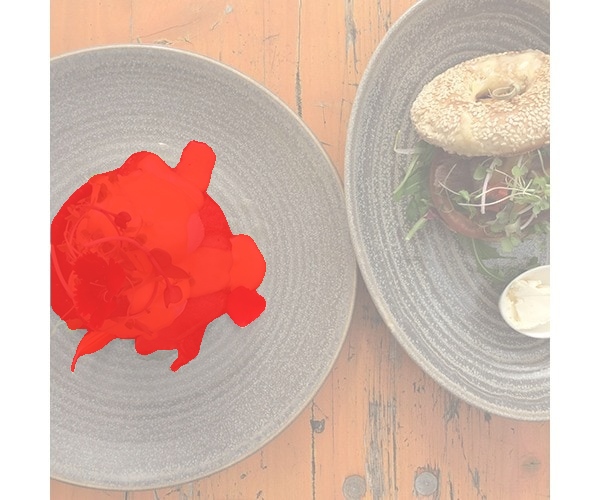}%
  \hspace{5pt}%
  \includegraphics[width=\sacoimgw]{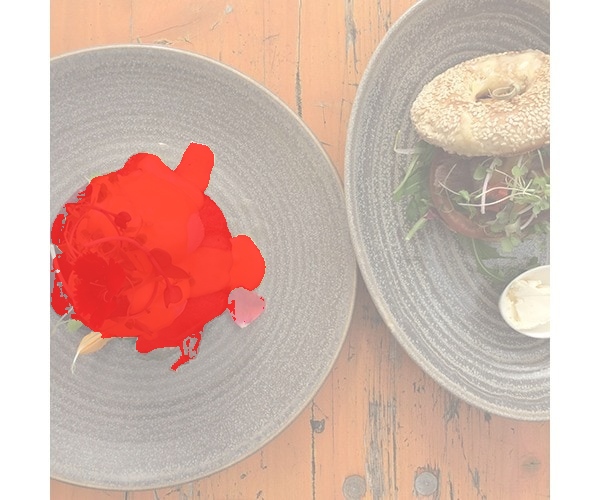}%
  \hspace{5pt}%
  \includegraphics[width=\sacoimgw]{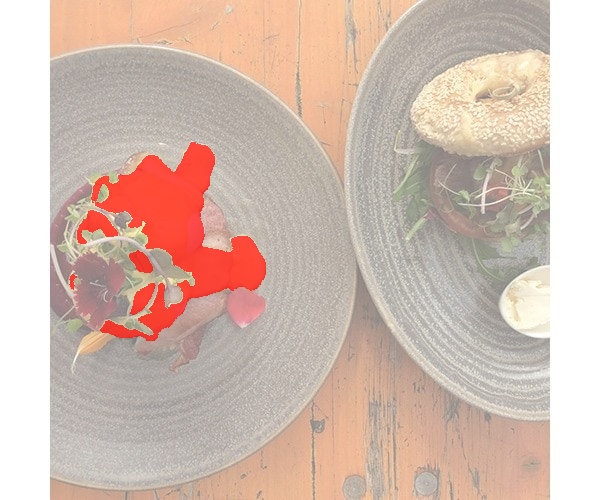}
  \\[-4pt]
  \leftline{\textbf{(\ref*{fig:saco_qualitative_successes}.e)} ``\textit{eggs Benedict}``%
    \hfill%
    \makebox[\sacoimgw][c]{\tiny F1=0.90 (GT a), 1 mask}%
    \hspace{5pt}%
    \makebox[\sacoimgw][c]{\tiny F1=0.00 (GT a), 1 mask}}
  \includegraphics[width=\sacoimgw]{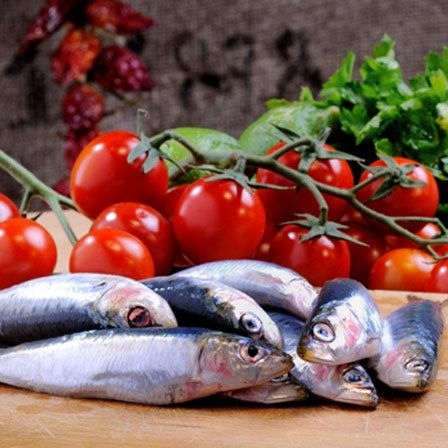}%
  \hspace{5pt}%
  \includegraphics[width=\sacoimgw]{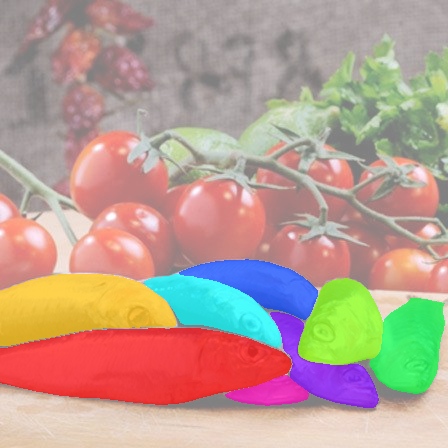}%
  \hspace{1pt}%
  \includegraphics[width=\sacoimgw]{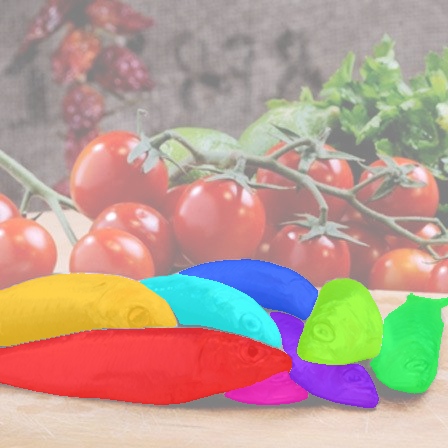}%
  \hspace{1pt}%
  \includegraphics[width=\sacoimgw]{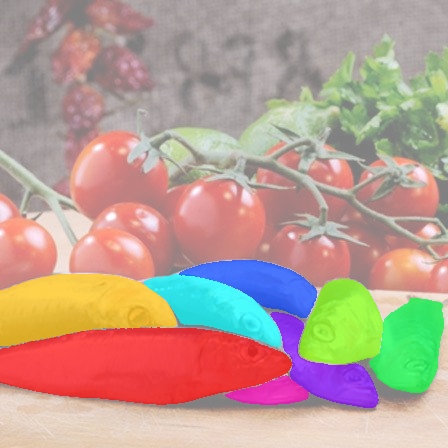}%
  \hspace{5pt}%
  \includegraphics[width=\sacoimgw]{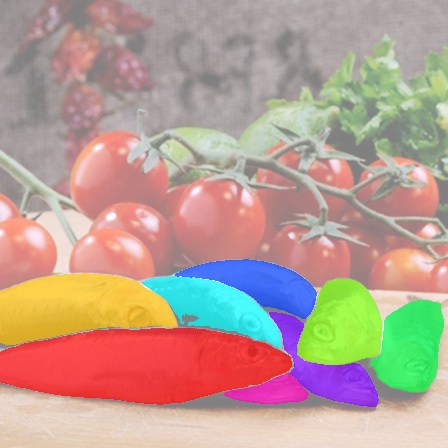}%
  \hspace{5pt}%
  \includegraphics[width=\sacoimgw]{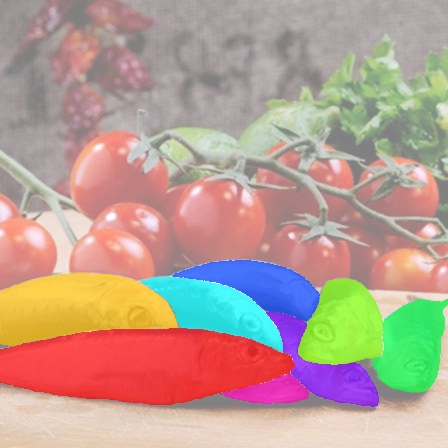}
  \\[-4pt]
  \leftline{\textbf{(\ref*{fig:saco_qualitative_successes}.f)} ``\textit{sardine}``%
    \hfill%
    \makebox[\sacoimgw][c]{\tiny F1=0.95 (GT c), 8 masks}%
    \hspace{5pt}%
    \makebox[\sacoimgw][c]{\tiny F1=0.91 (GT a), 8 masks}}
  \caption{\textbf{Success cases.}
  Qualitative evaluation of instance segmentation results on SA-Co/Gold.
  For each example, we display the noun-phrase query and the original image, followed by visualizations of the three independent ground truth annotations, the prediction from SAM 3, and the prediction from Vision Banana.
  Under each prediction visualization, we report the best F1 score along with its associated GT and the number of masks in the prediction.
  Best seen zoomed-in on screen.
  }
  \label{fig:saco_qualitative_successes}
\end{figure*}

\begin{figure*}[t]
  \centering
  \small
  \newcommand{\sacoimgw}{\dimexpr(\textwidth-17pt)/6\relax}
  \leftline{%
    \makebox[\sacoimgw][c]{\textbf{Original Image}}%
    \hspace{5pt}%
    \makebox[\sacoimgw][c]{\textbf{GT a}}%
    \hspace{1pt}%
    \makebox[\sacoimgw][c]{\textbf{GT b}}%
    \hspace{1pt}%
    \makebox[\sacoimgw][c]{\textbf{GT c}}%
    \hspace{5pt}%
    \makebox[\sacoimgw][c]{\makecell{\textbf{SAM 3} \\ \textbf{Predictions}}}%
    \hspace{5pt}%
    \makebox[\sacoimgw][c]{\makecell{\textbf{Vision Banana \emojivb} \\ \textbf{Predictions}}}%
  }
  \vspace{4pt}
  \includegraphics[width=\sacoimgw]{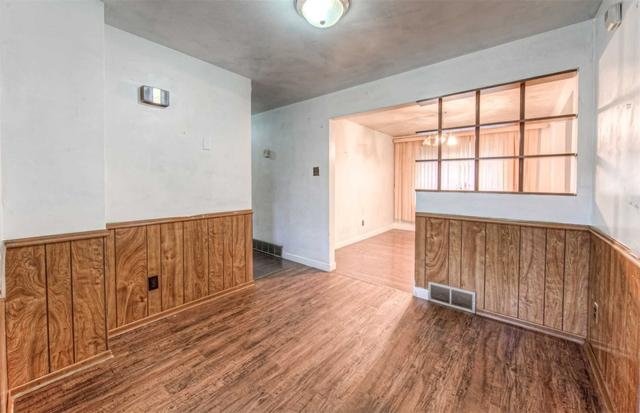}%
  \hspace{5pt}%
  \includegraphics[width=\sacoimgw]{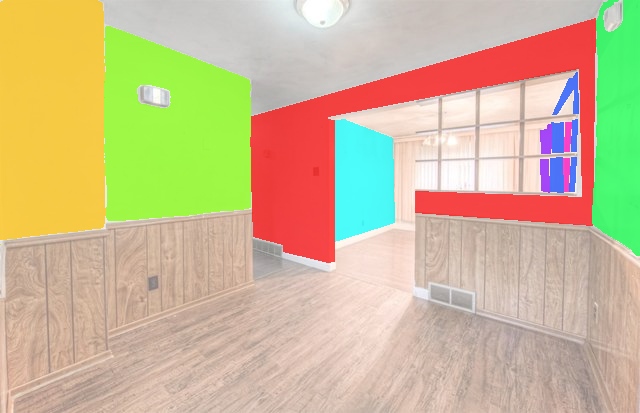}%
  \hspace{1pt}%
  \includegraphics[width=\sacoimgw]{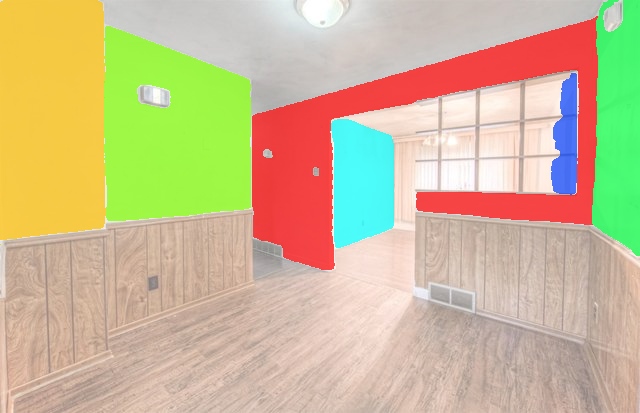}%
  \hspace{1pt}%
  \includegraphics[width=\sacoimgw]{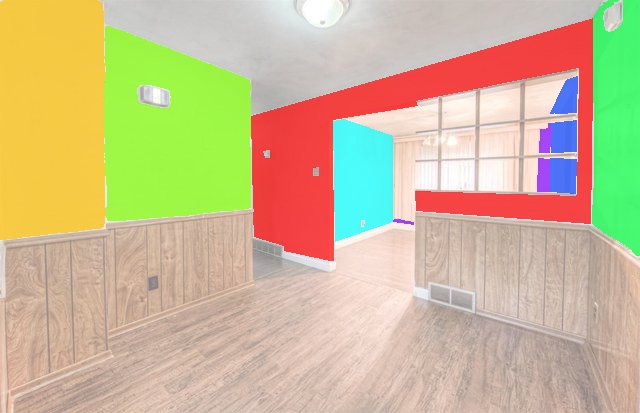}%
  \hspace{5pt}%
  \includegraphics[width=\sacoimgw]{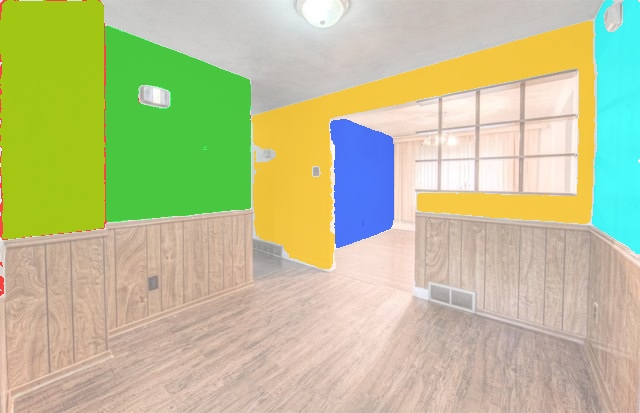}%
  \hspace{5pt}%
  \includegraphics[width=\sacoimgw]{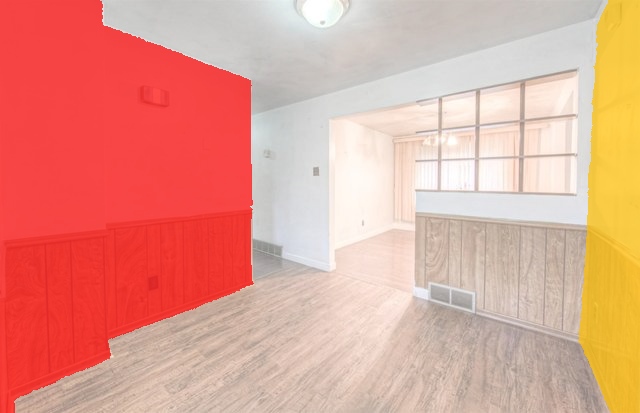}
  \\[-4pt]
  \leftline{\textbf{(\ref*{fig:saco_qualitative_failures}.a)} ``\textit{a wall}``%
    \hfill%
    \makebox[\sacoimgw][c]{\tiny F1=0.82 (GT b), 6 masks}%
    \hspace{5pt}%
    \makebox[\sacoimgw][c]{\tiny F1=0.02 (GT c), 2 masks}}
  \includegraphics[width=\sacoimgw]{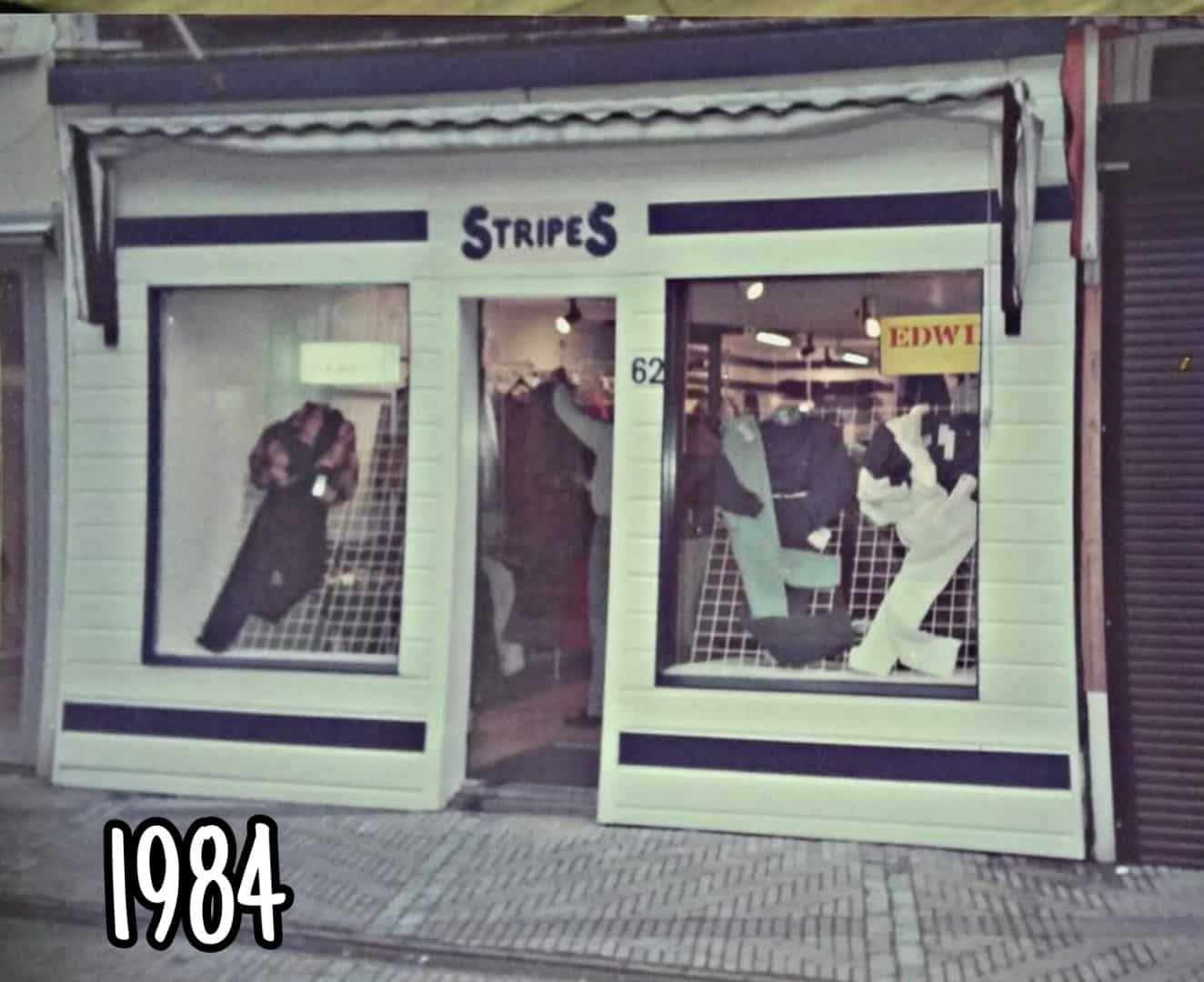}%
  \hspace{5pt}%
  \includegraphics[width=\sacoimgw]{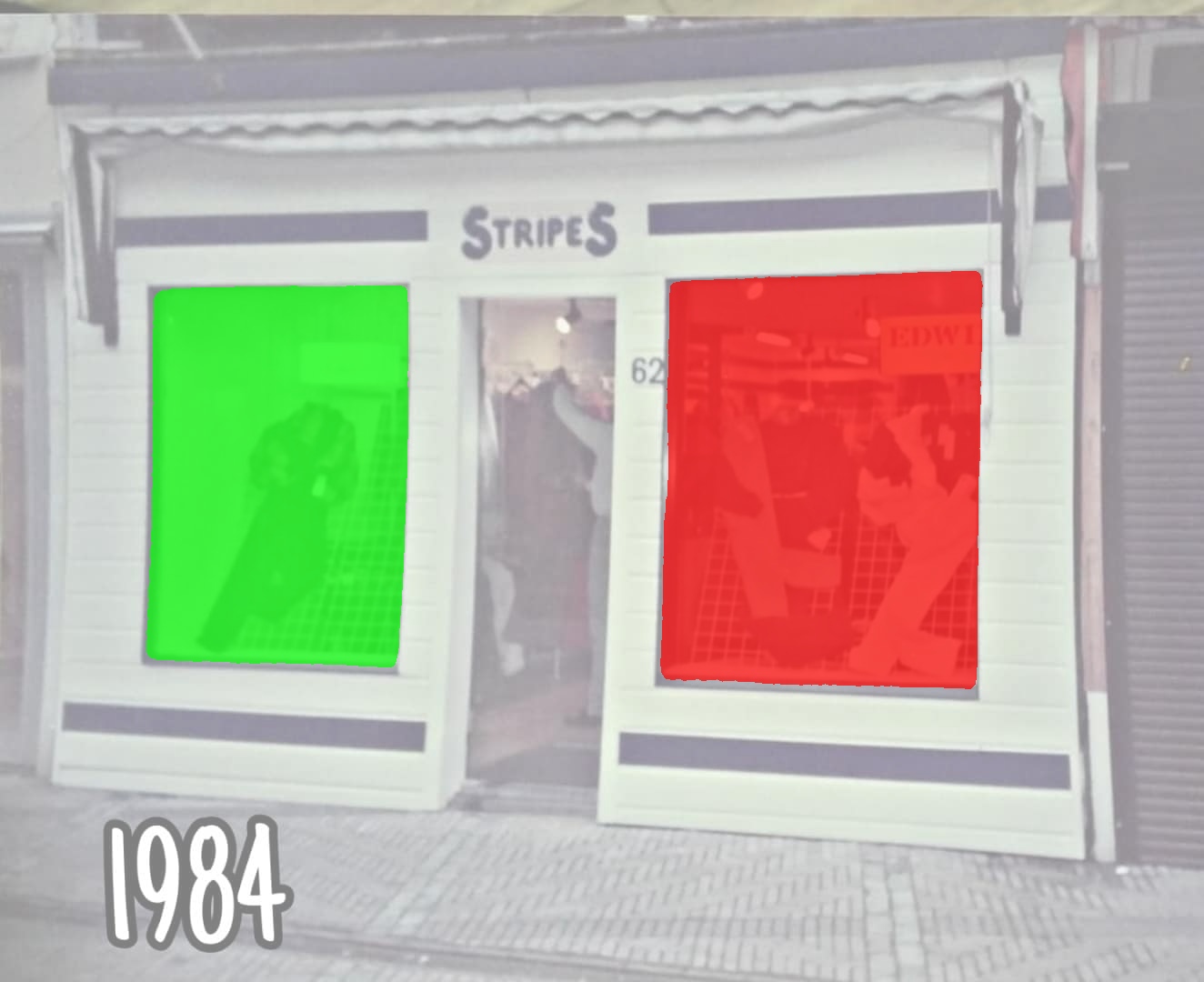}%
  \hspace{1pt}%
  \includegraphics[width=\sacoimgw]{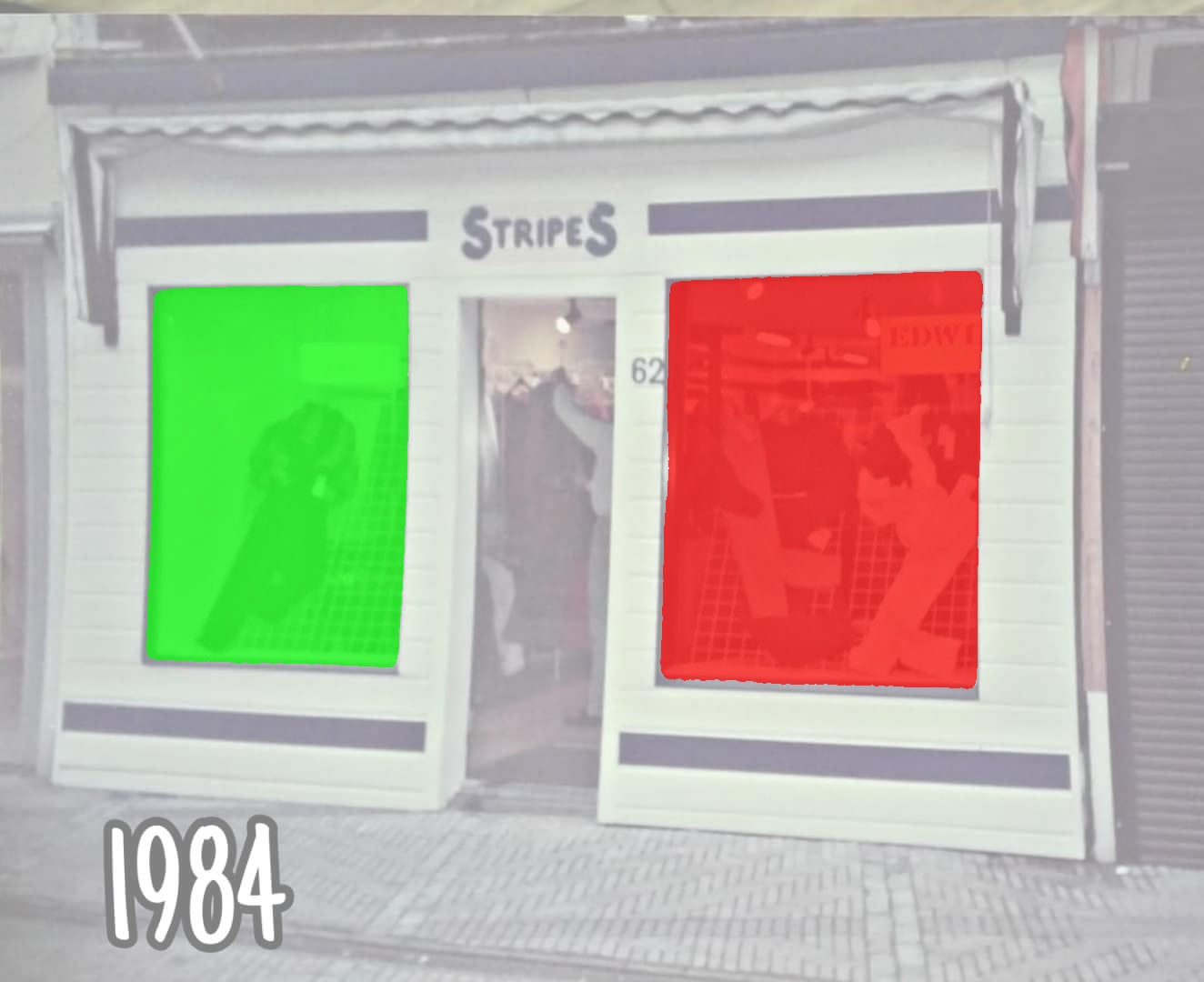}%
  \hspace{1pt}%
  \includegraphics[width=\sacoimgw]{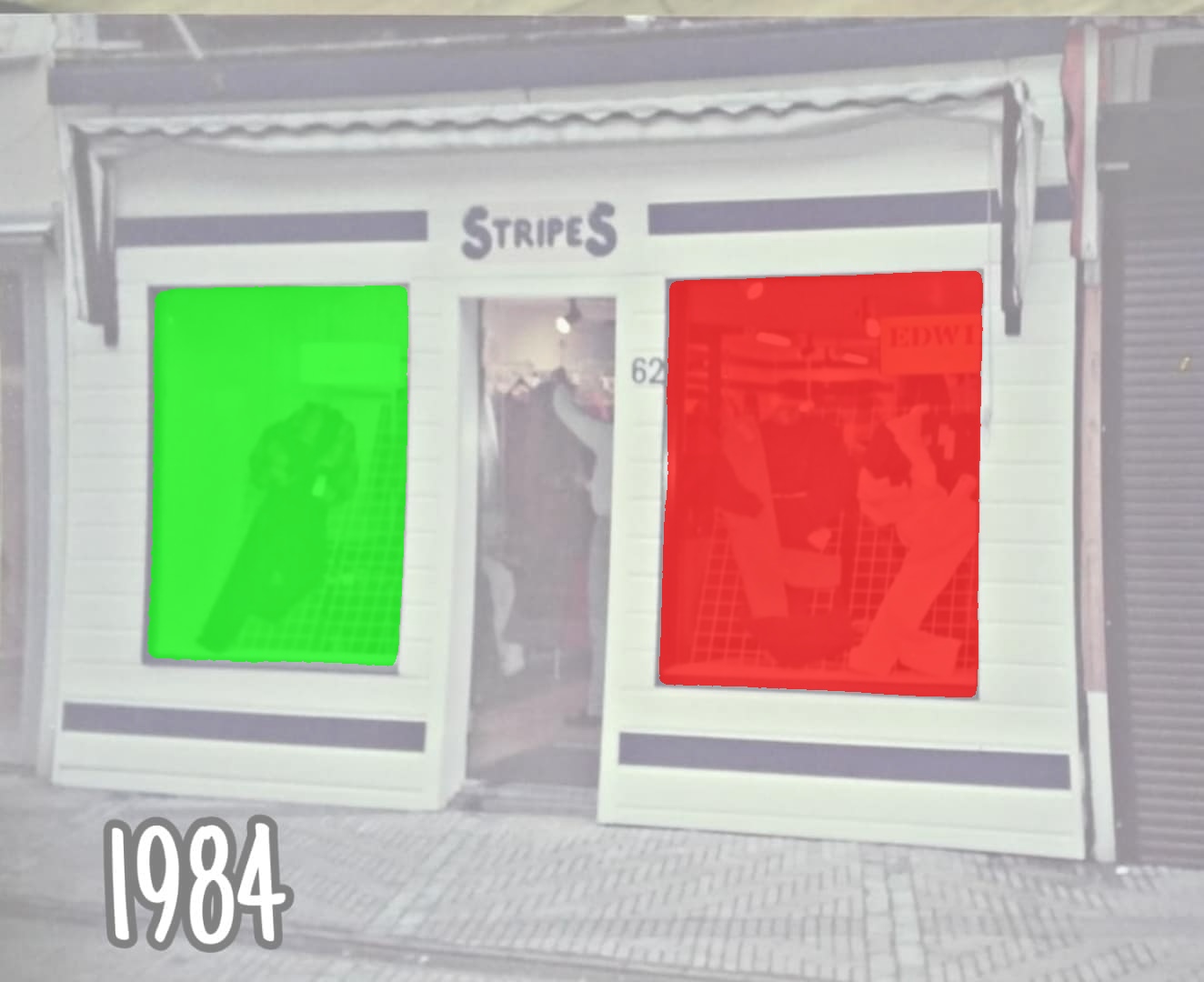}%
  \hspace{5pt}%
  \includegraphics[width=\sacoimgw]{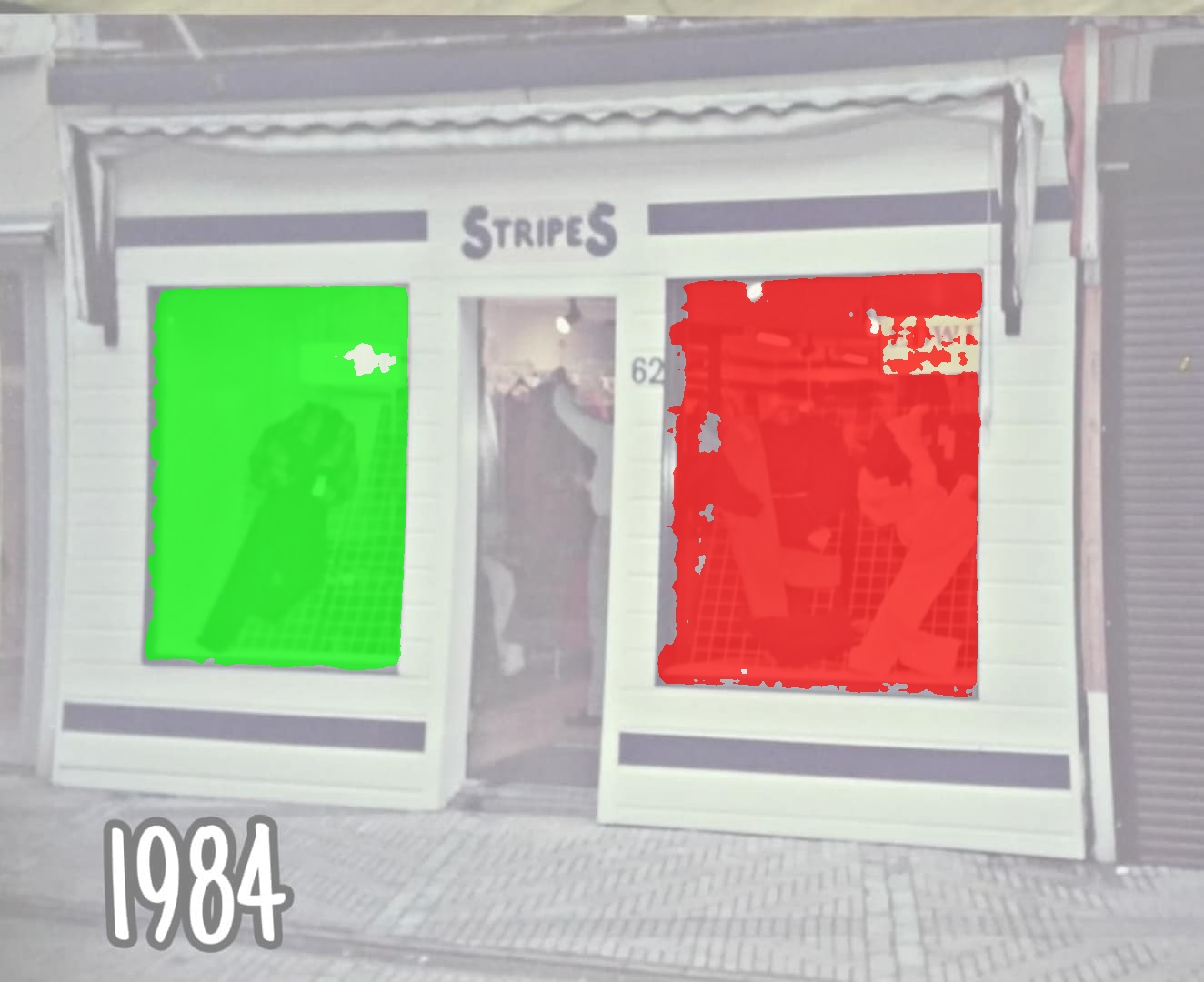}%
  \hspace{5pt}%
  \includegraphics[width=\sacoimgw]{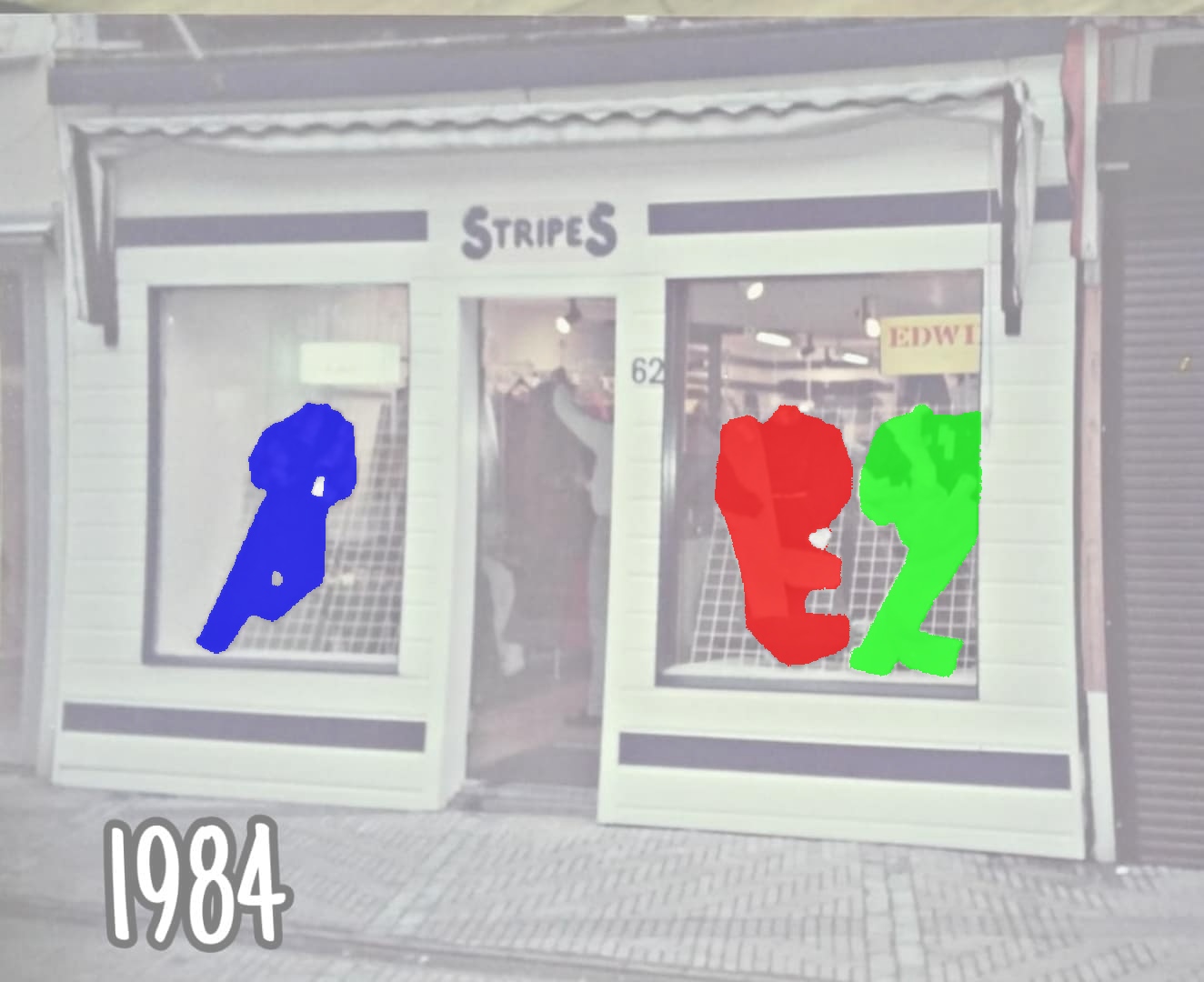}
  \\[-4pt]
  \leftline{\textbf{(\ref*{fig:saco_qualitative_failures}.b)} ``\textit{clothing window display}``%
    \hfill%
    \makebox[\sacoimgw][c]{\tiny F1=0.95 (GT a), 2 masks}%
    \hspace{5pt}%
    \makebox[\sacoimgw][c]{\tiny F1=0.00 (GT a), 3 masks}}
  \includegraphics[width=\sacoimgw]{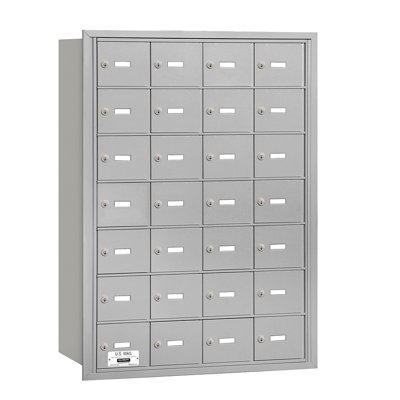}%
  \hspace{5pt}%
  \includegraphics[width=\sacoimgw]{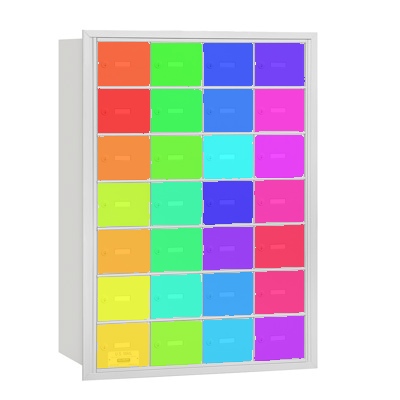}%
  \hspace{1pt}%
  \includegraphics[width=\sacoimgw]{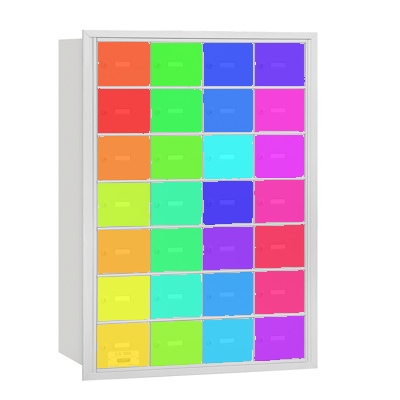}%
  \hspace{1pt}%
  \includegraphics[width=\sacoimgw]{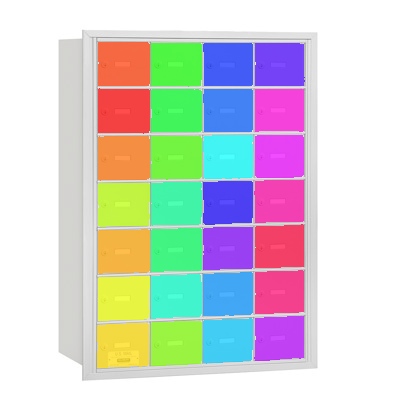}%
  \hspace{5pt}%
  \includegraphics[width=\sacoimgw]{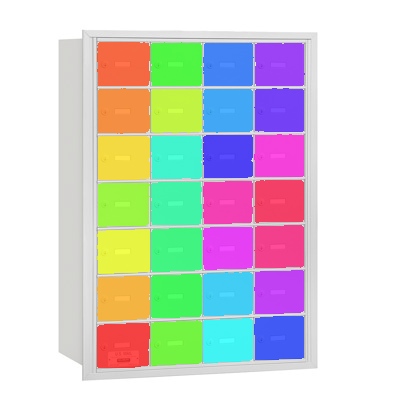}%
  \hspace{5pt}%
  \includegraphics[width=\sacoimgw]{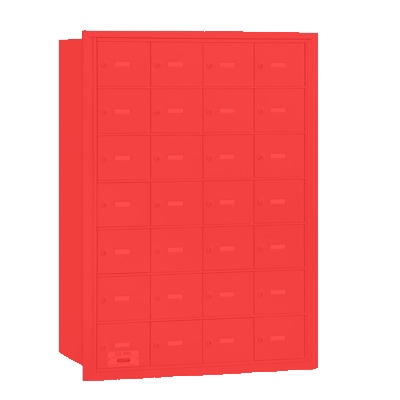}
  \\[-4pt]
  \leftline{\textbf{(\ref*{fig:saco_qualitative_failures}.c)} ``\textit{post office box}``%
    \hfill%
    \makebox[\sacoimgw][c]{\tiny F1=1.00 (GT a), 28 masks}%
    \hspace{5pt}%
    \makebox[\sacoimgw][c]{\tiny F1=0.00 (GT a), 1 mask}}
  \includegraphics[width=\sacoimgw]{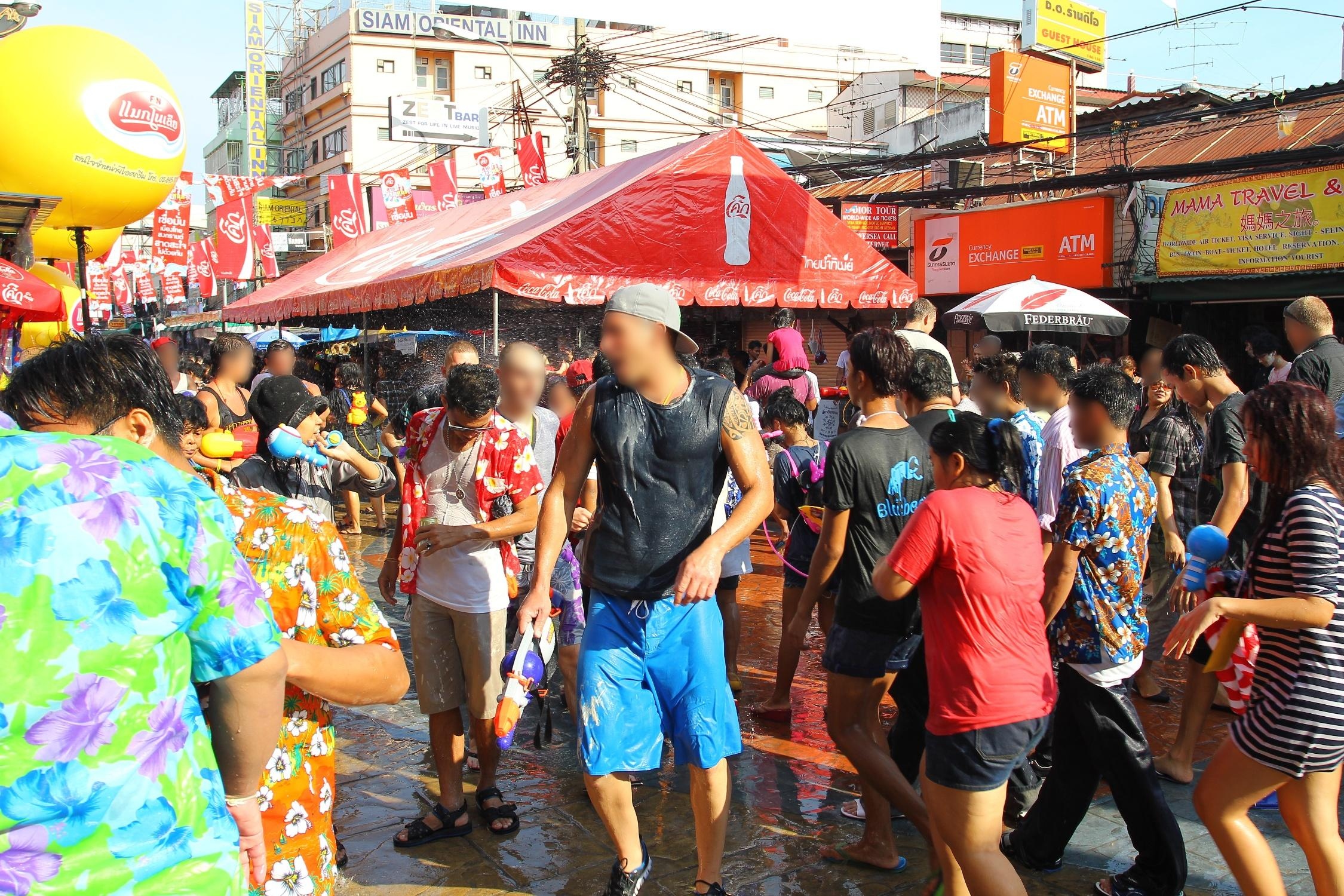}%
  \hspace{5pt}%
  \includegraphics[width=\sacoimgw]{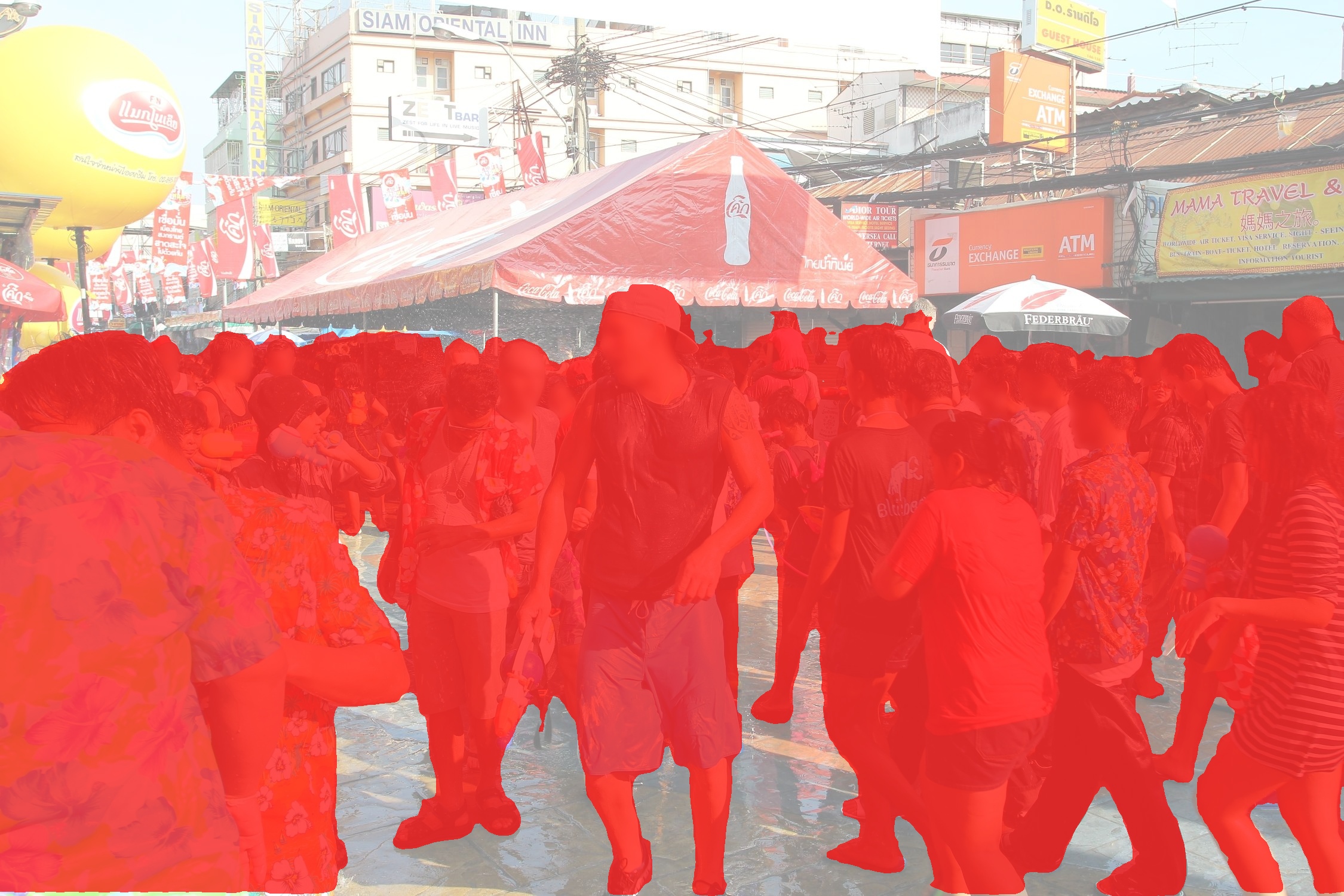}%
  \hspace{1pt}%
  \includegraphics[width=\sacoimgw]{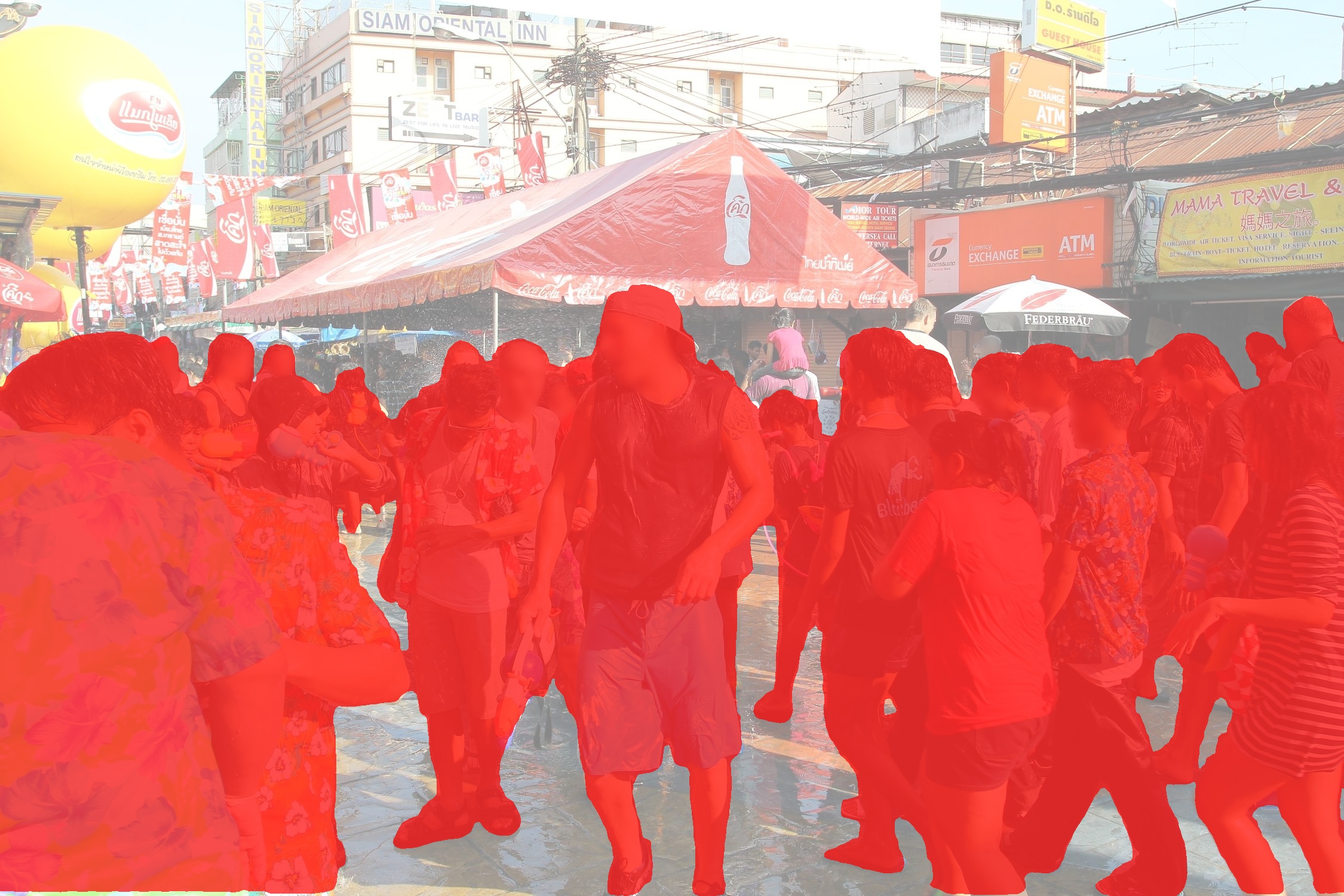}%
  \hspace{1pt}%
  \includegraphics[width=\sacoimgw]{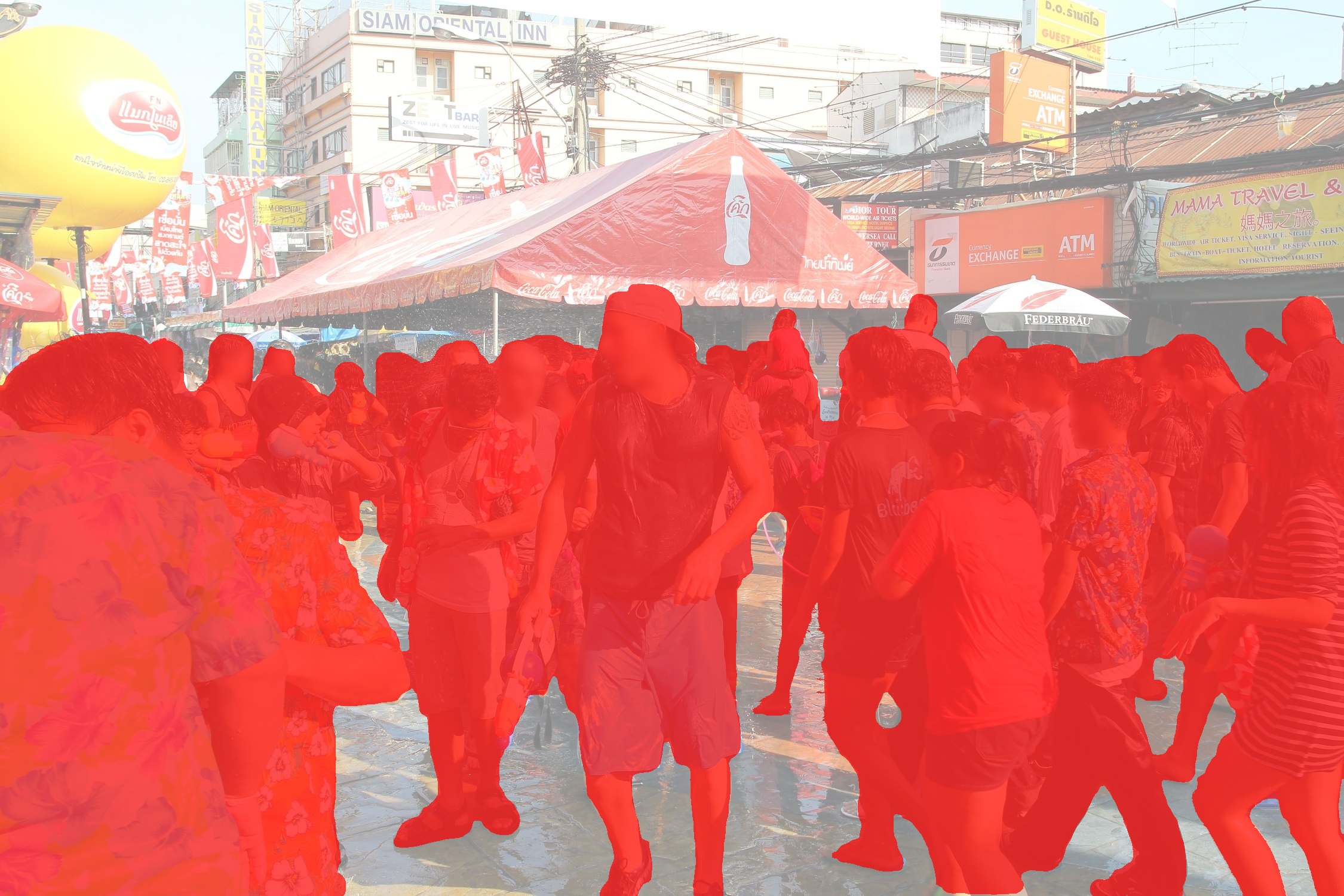}%
  \hspace{5pt}%
  \includegraphics[width=\sacoimgw]{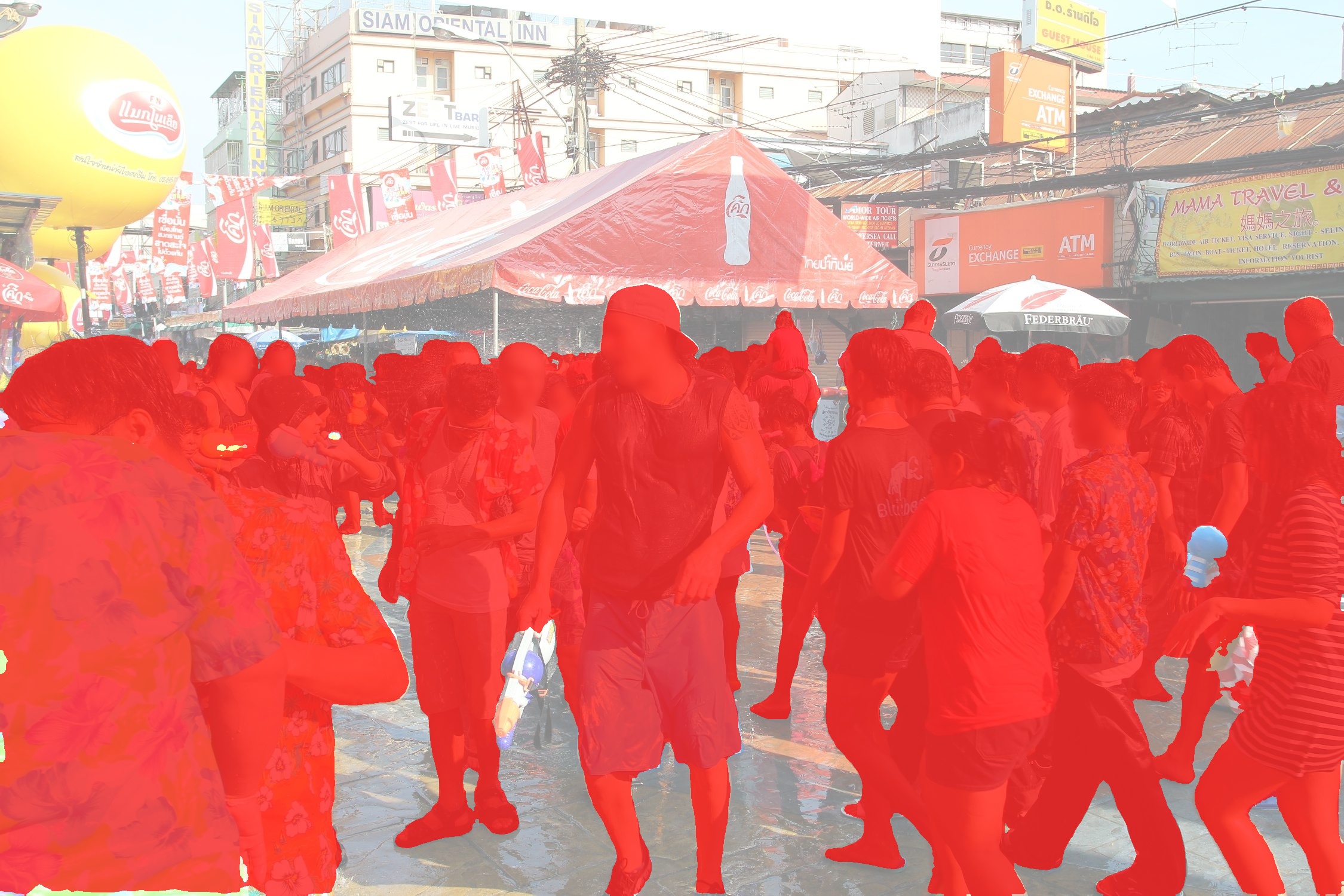}%
  \hspace{5pt}%
  \includegraphics[width=\sacoimgw]{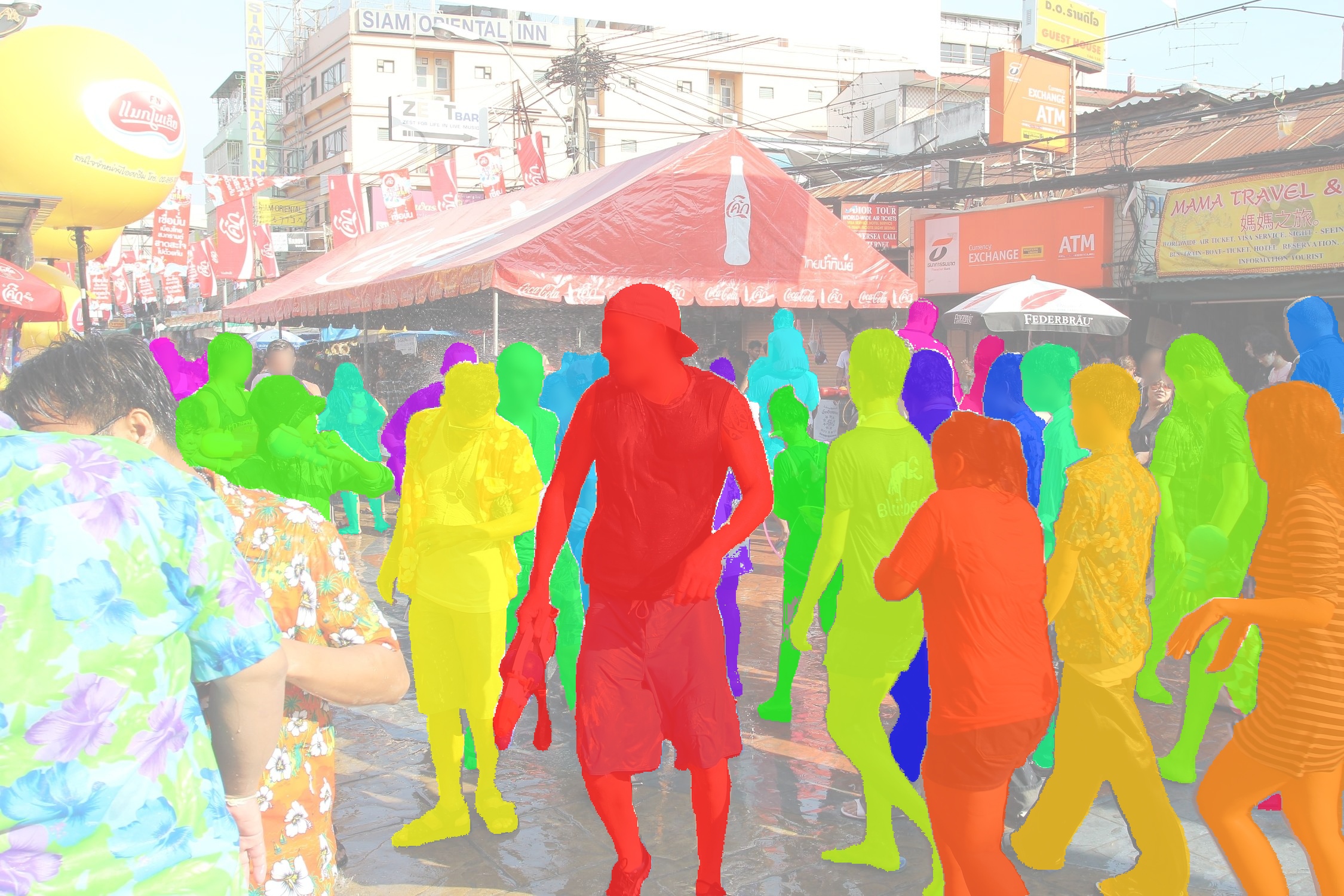}
  \\[-4pt]
  \leftline{\textbf{(\ref*{fig:saco_qualitative_failures}.d)} ``\textit{the crowd}``%
    \hfill%
    \makebox[\sacoimgw][c]{\tiny F1=1.00 (GT c), 1 mask}%
    \hspace{5pt}%
    \makebox[\sacoimgw][c]{\tiny F1=0.00 (GT a), 24 masks}}
  \includegraphics[width=\sacoimgw]{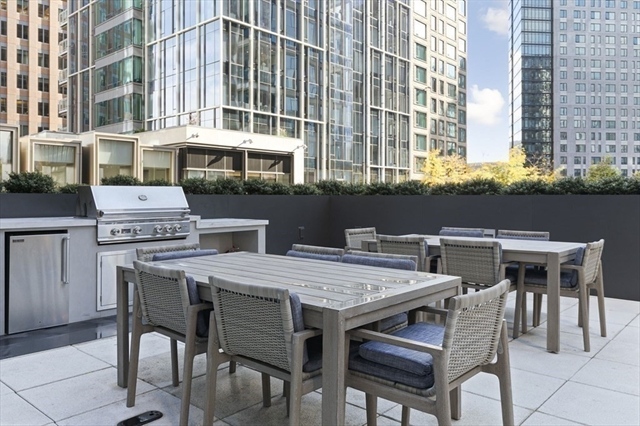}%
  \hspace{5pt}%
  \includegraphics[width=\sacoimgw]{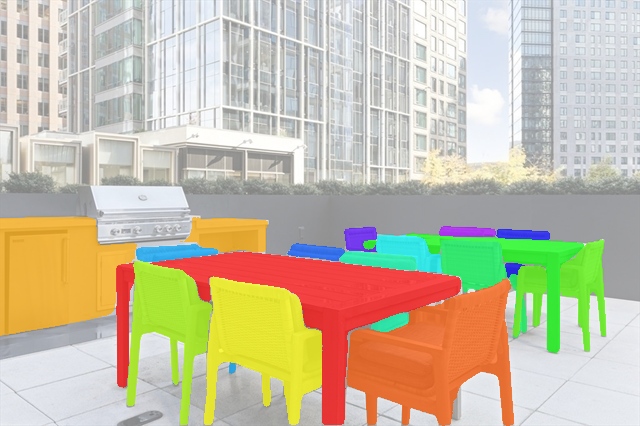}%
  \hspace{1pt}%
  \includegraphics[width=\sacoimgw]{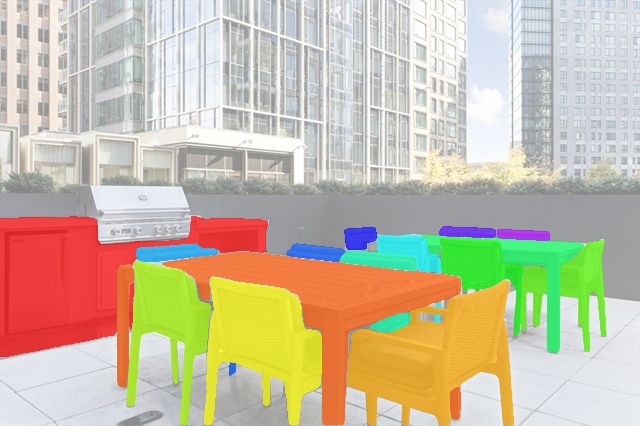}%
  \hspace{1pt}%
  \includegraphics[width=\sacoimgw]{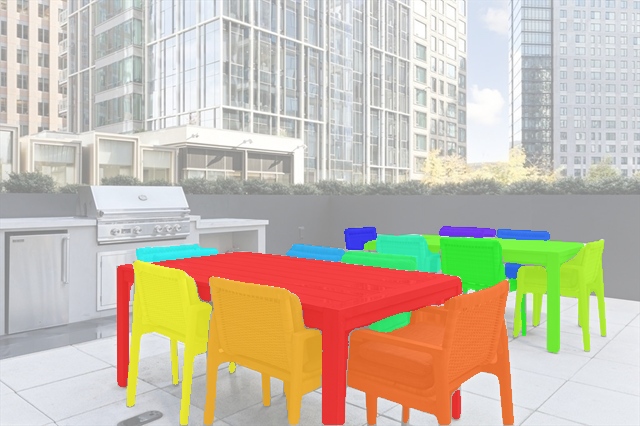}%
  \hspace{5pt}%
  \includegraphics[width=\sacoimgw]{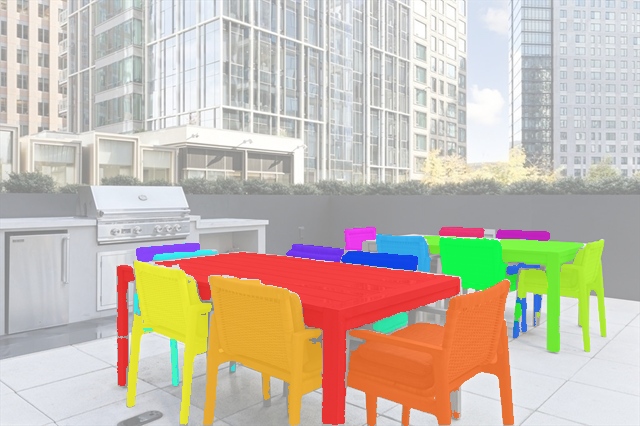}%
  \hspace{5pt}%
  \includegraphics[width=\sacoimgw]{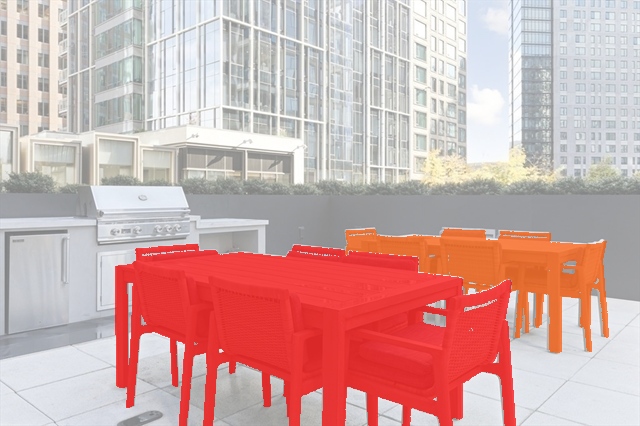}
  \\[-4pt]
  \leftline{\textbf{(\ref*{fig:saco_qualitative_failures}.e)} ``\textit{the furniture}``%
    \hfill%
    \makebox[\sacoimgw][c]{\tiny F1=0.65 (GT c), 18 masks}%
    \hspace{5pt}%
    \makebox[\sacoimgw][c]{\tiny F1=0.00 (GT a), 2 masks}}
  \includegraphics[width=\sacoimgw]{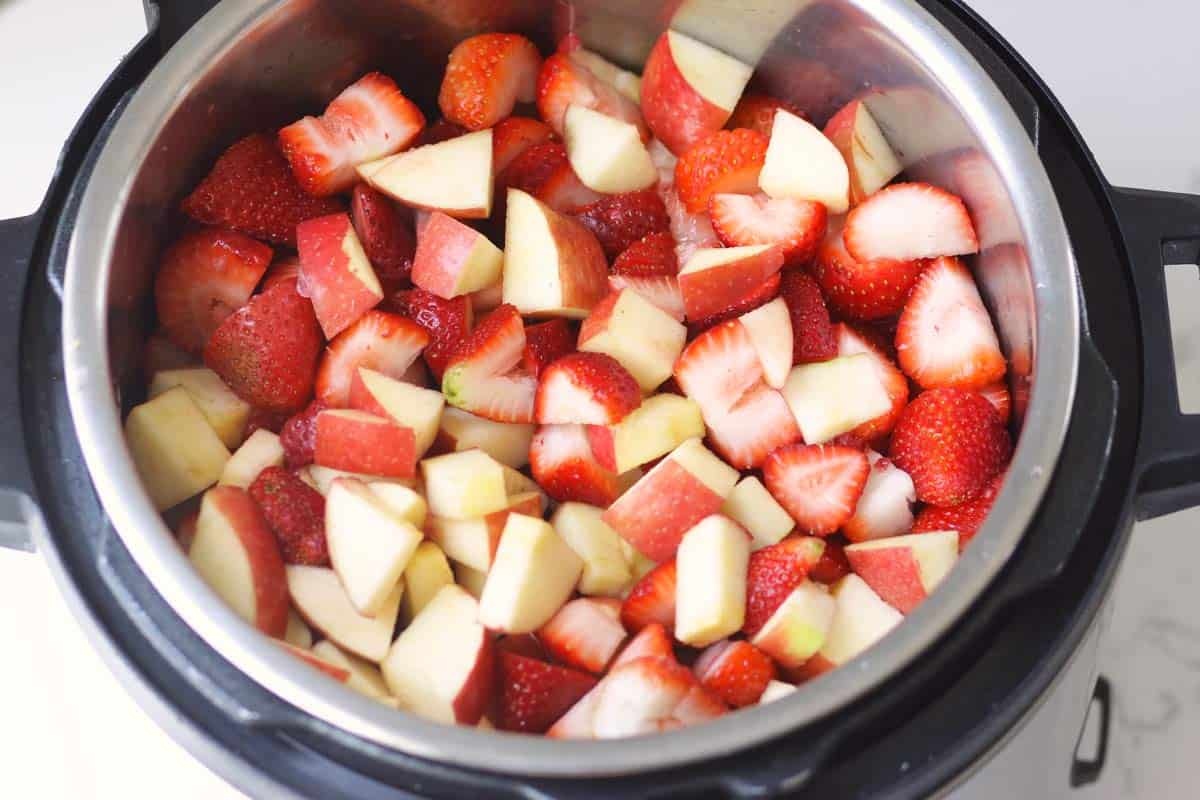}%
  \hspace{5pt}%
  \includegraphics[width=\sacoimgw]{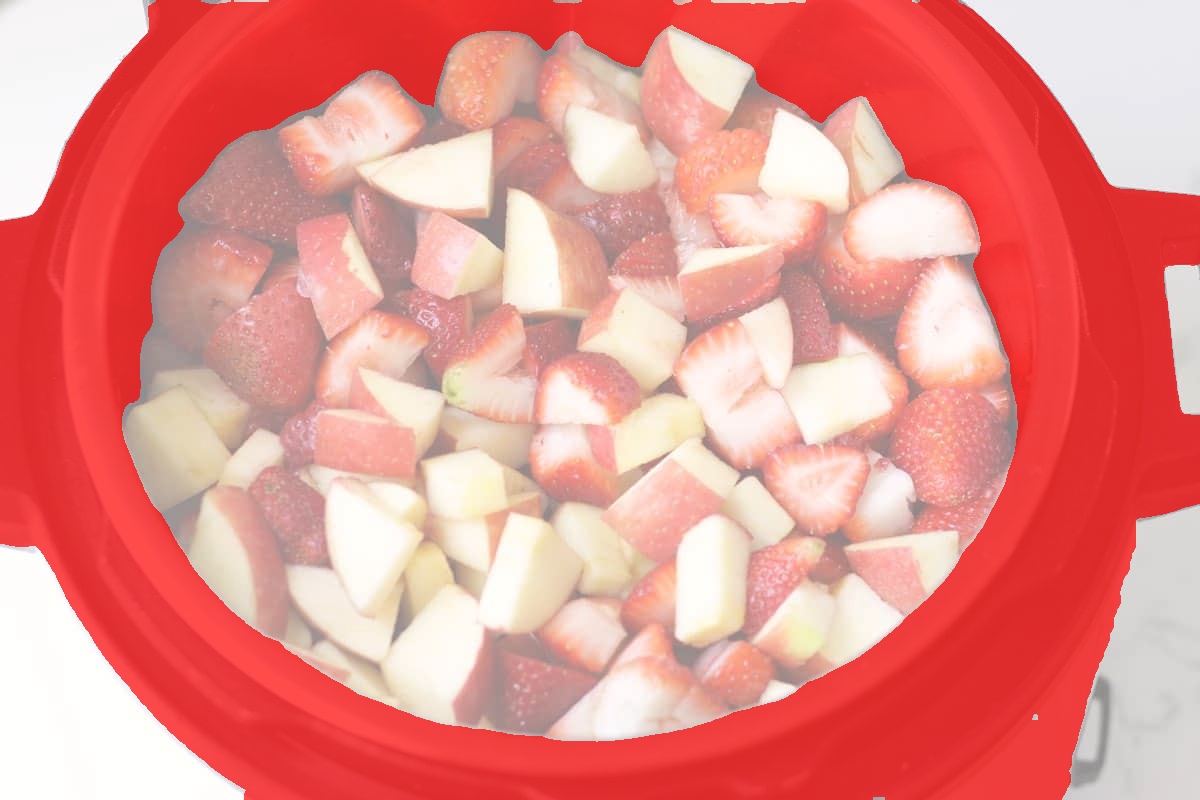}%
  \hspace{1pt}%
  \includegraphics[width=\sacoimgw]{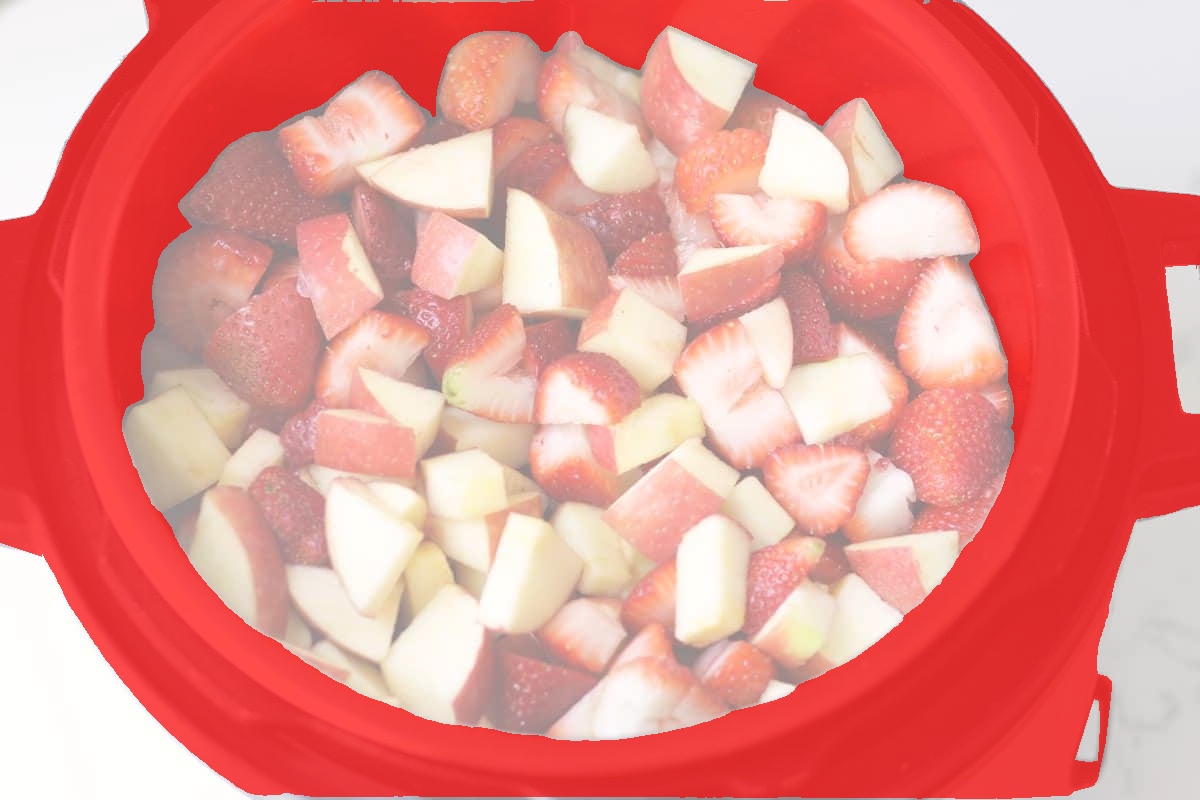}%
  \hspace{1pt}%
  \includegraphics[width=\sacoimgw]{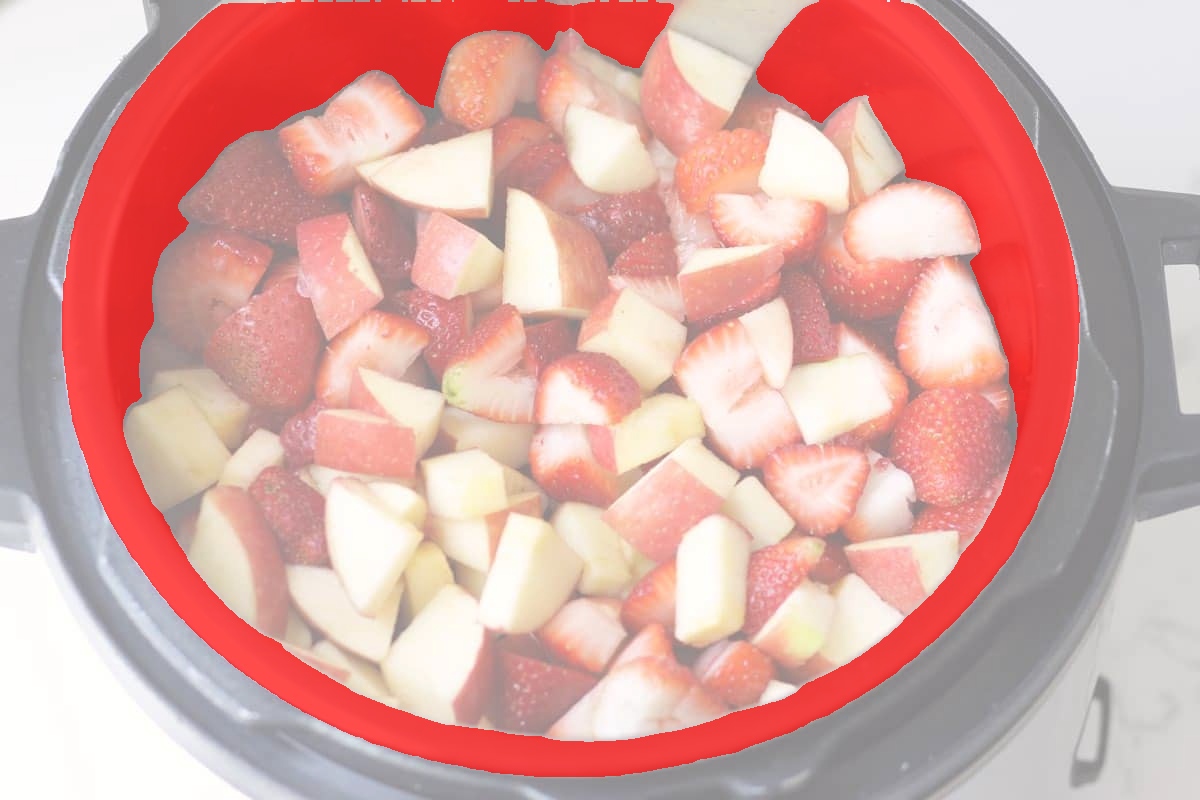}%
  \hspace{5pt}%
  \includegraphics[width=\sacoimgw]{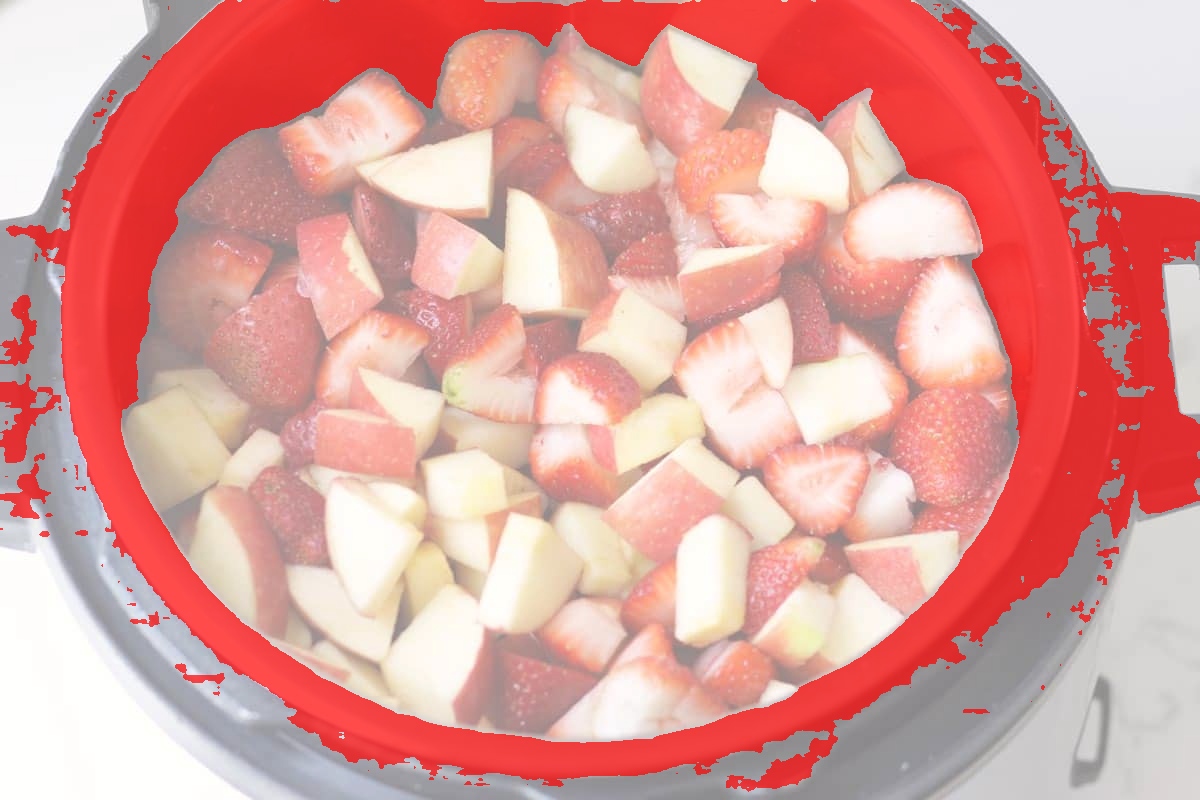}%
  \hspace{5pt}%
  \includegraphics[width=\sacoimgw]{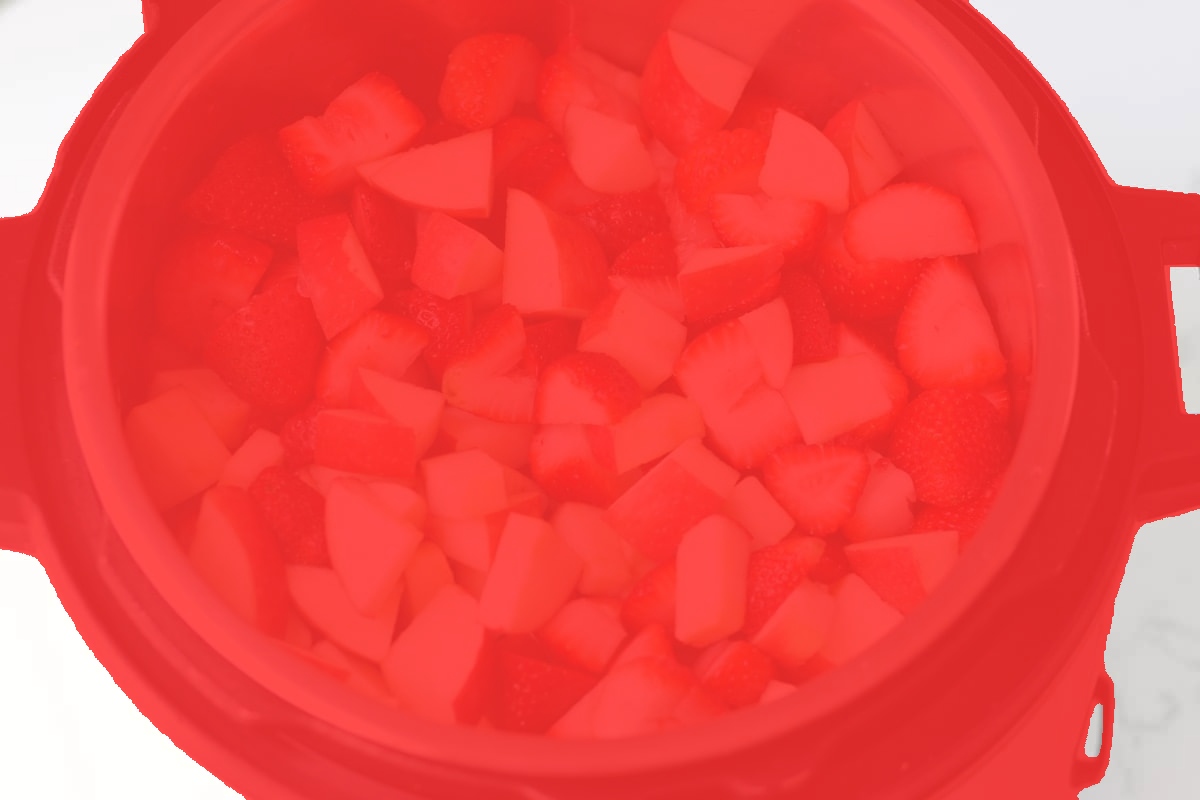}
  \\[-4pt]
  \leftline{\textbf{(\ref*{fig:saco_qualitative_failures}.f)} ``\textit{a pot}``%
    \hfill%
    \makebox[\sacoimgw][c]{\tiny F1=0.40 (GT c), 1 mask}%
    \hspace{5pt}%
    \makebox[\sacoimgw][c]{\tiny F1=0.00 (GT a), 1 mask}}
  \includegraphics[width=\sacoimgw]{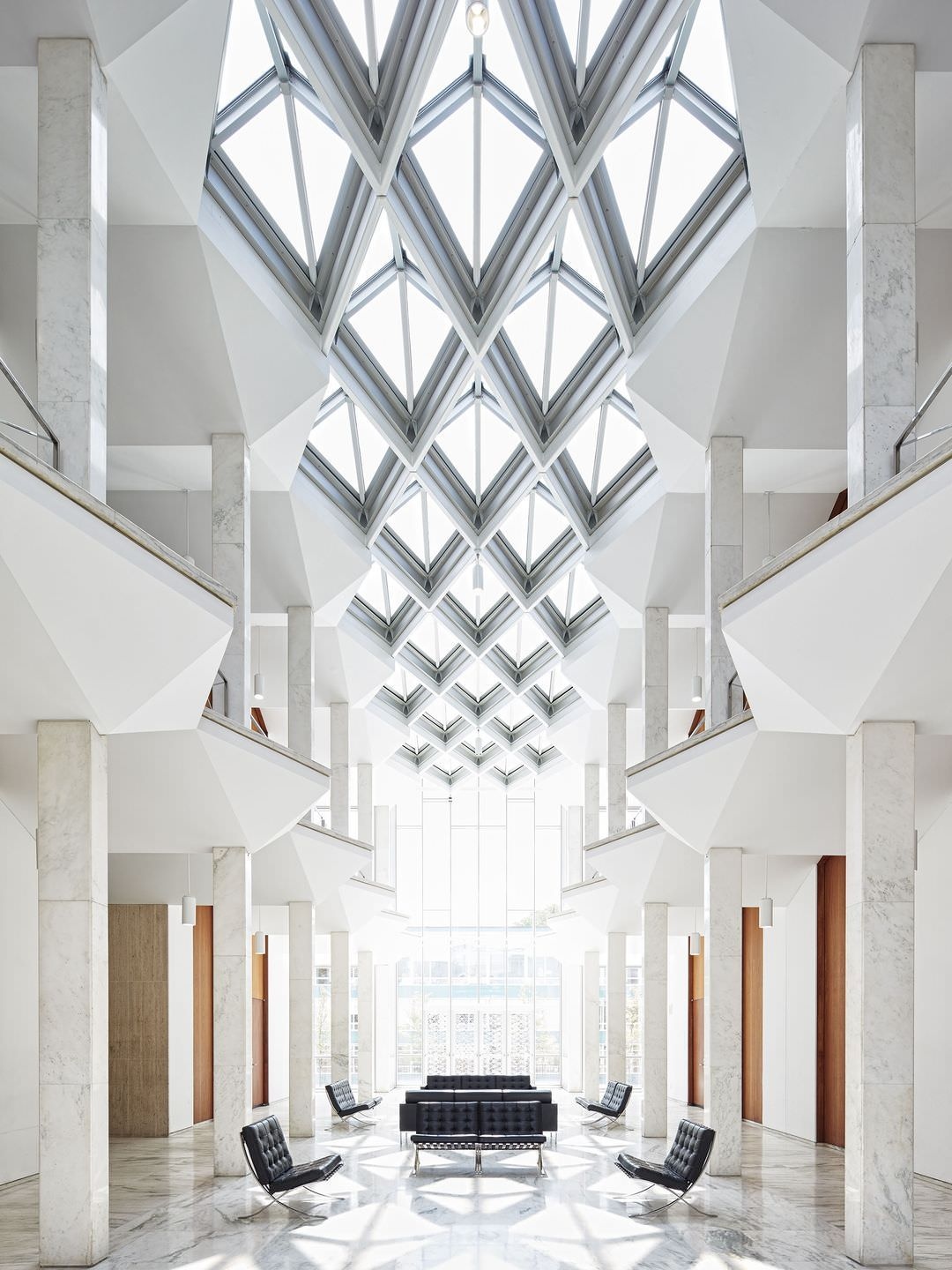}%
  \hspace{5pt}%
  \includegraphics[width=\sacoimgw]{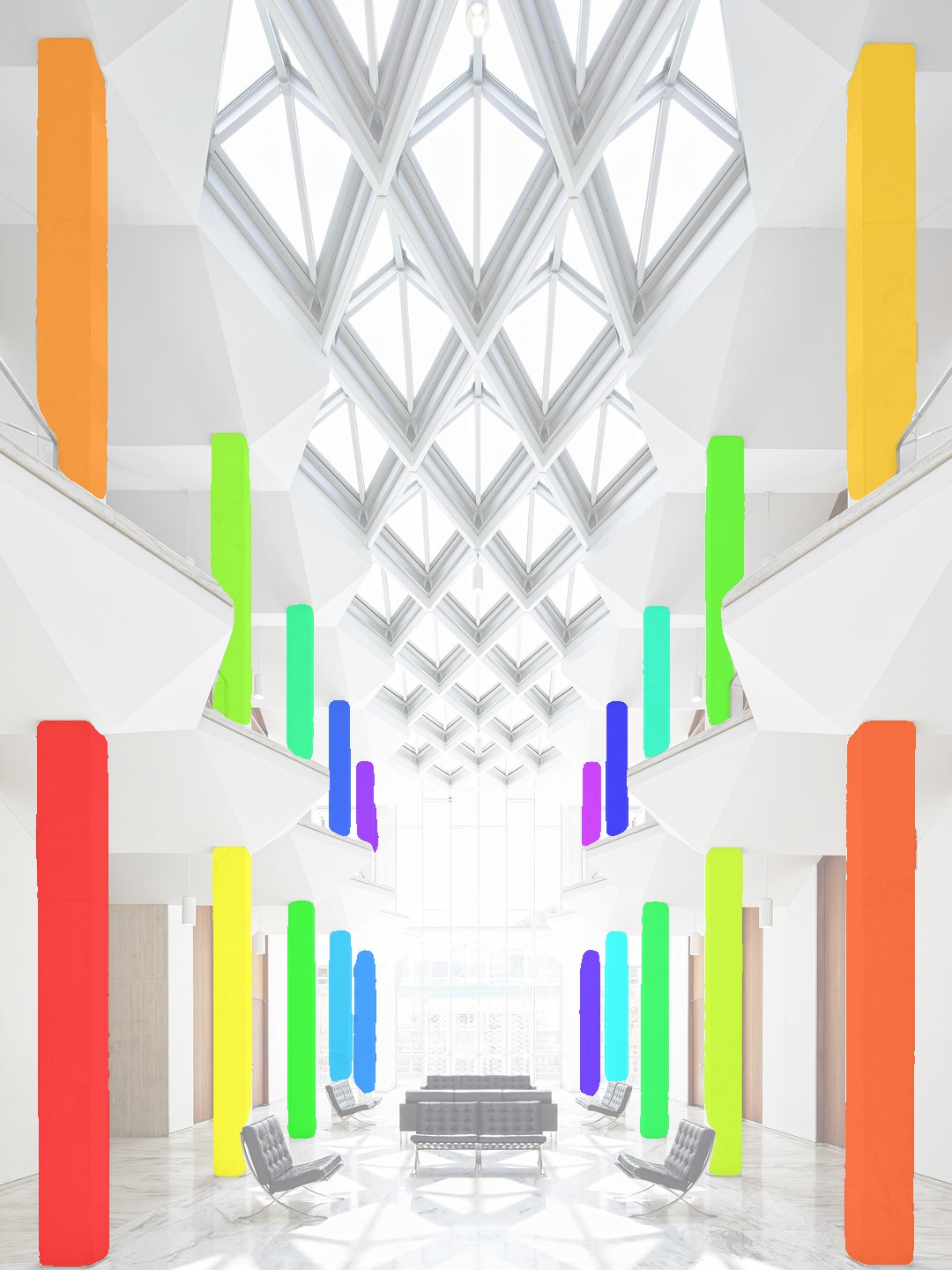}%
  \hspace{1pt}%
  \includegraphics[width=\sacoimgw]{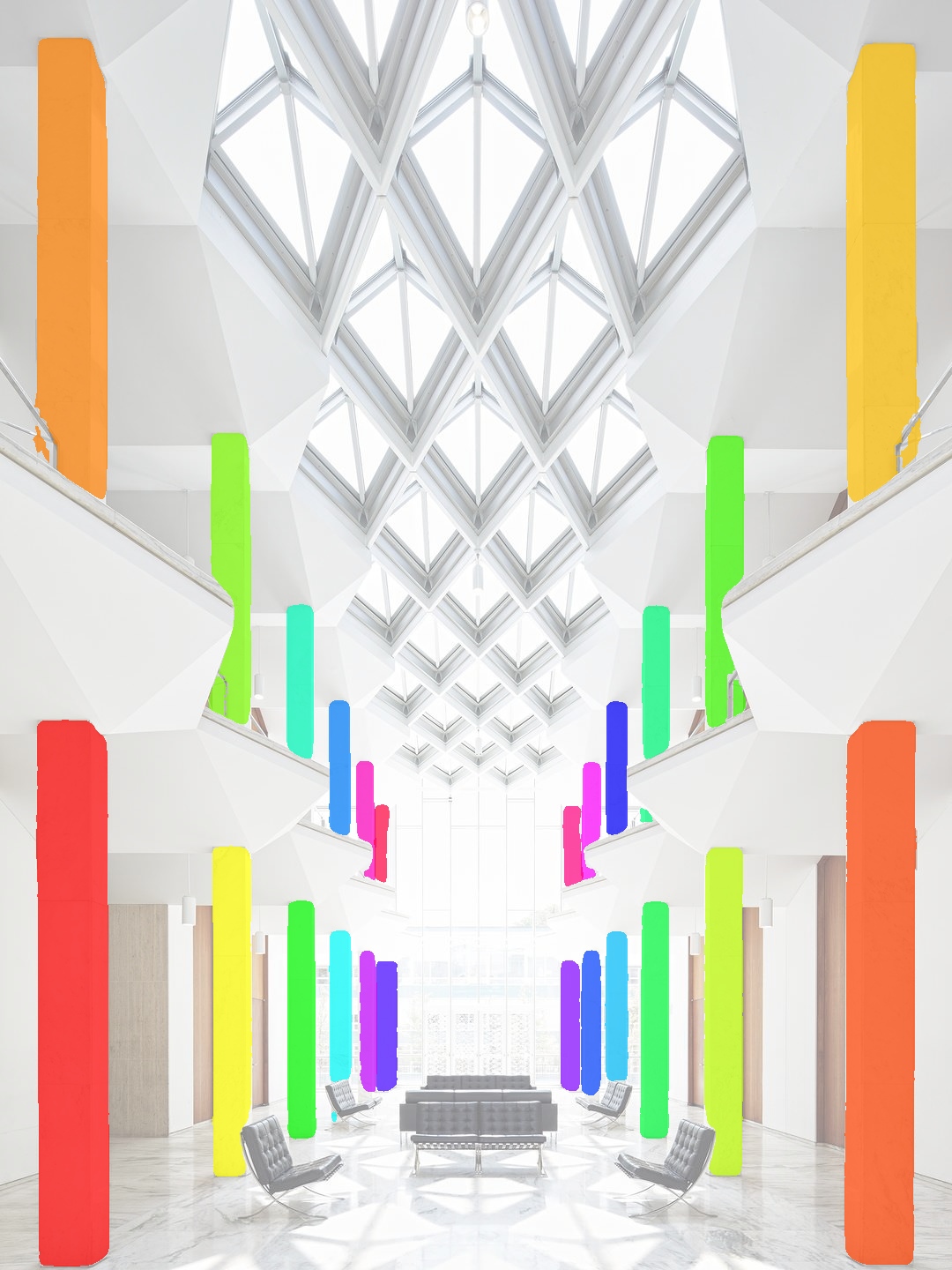}%
  \hspace{1pt}%
  \includegraphics[width=\sacoimgw]{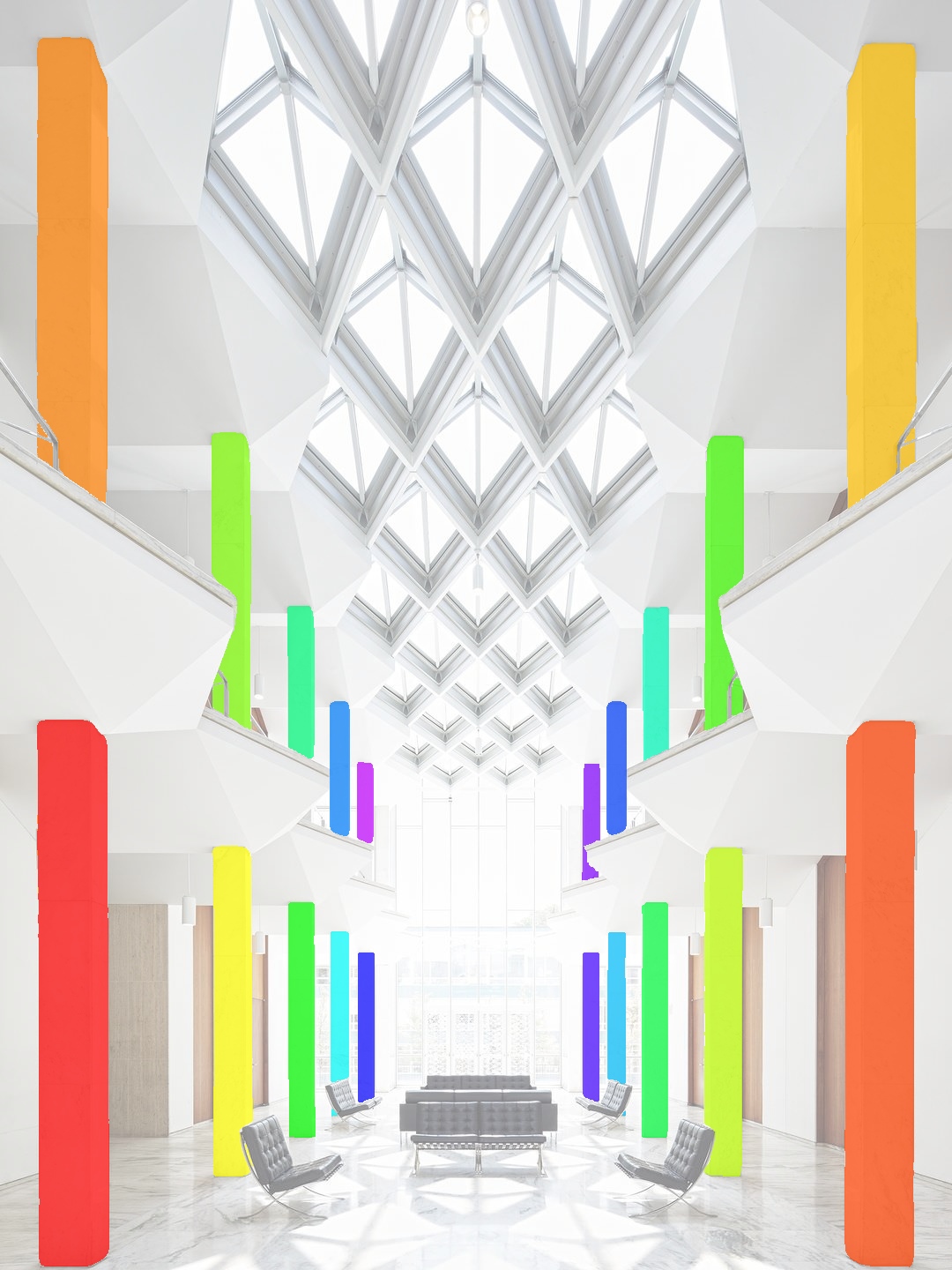}%
  \hspace{5pt}%
  \includegraphics[width=\sacoimgw]{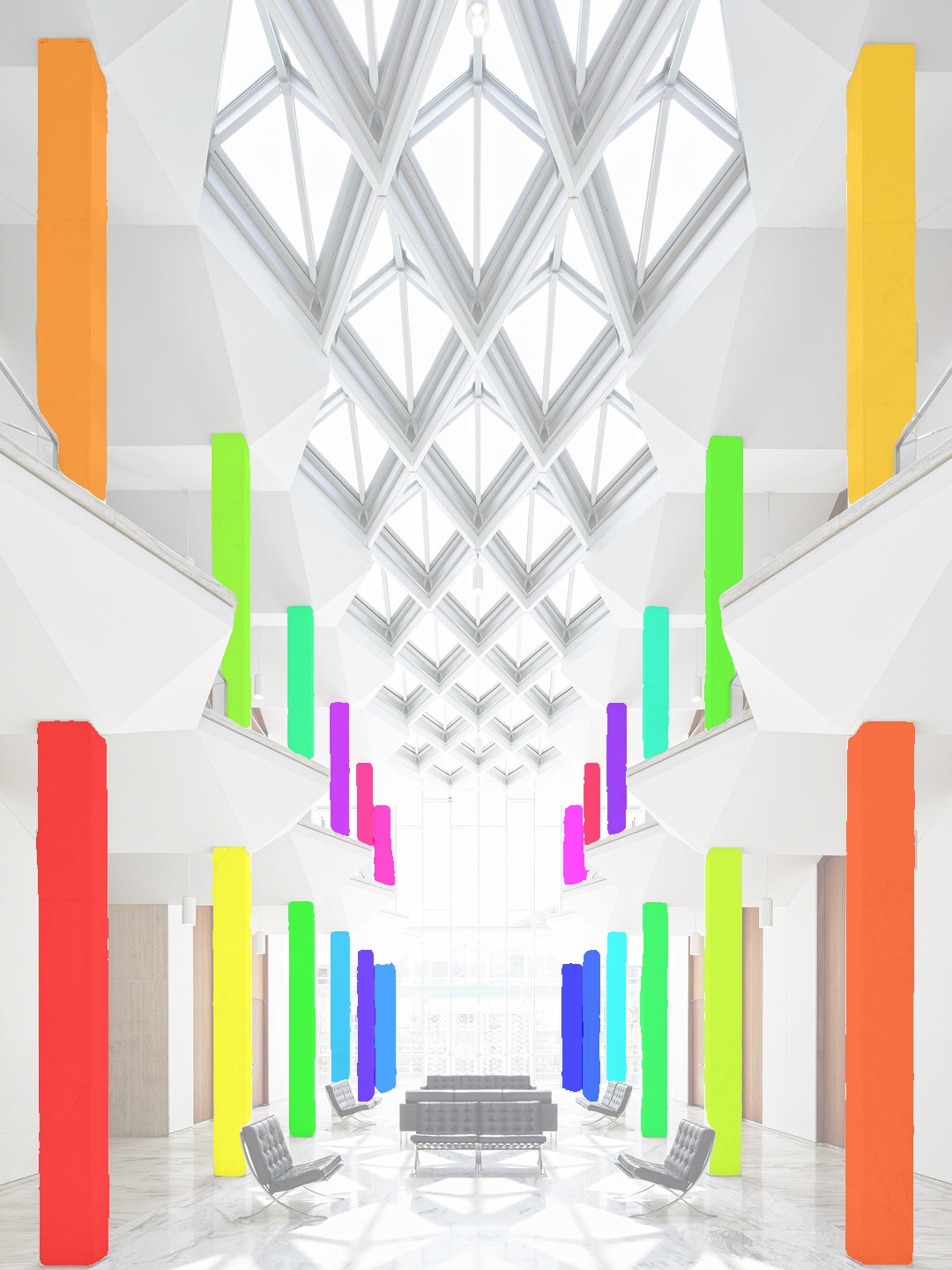}%
  \hspace{5pt}%
  \includegraphics[width=\sacoimgw]{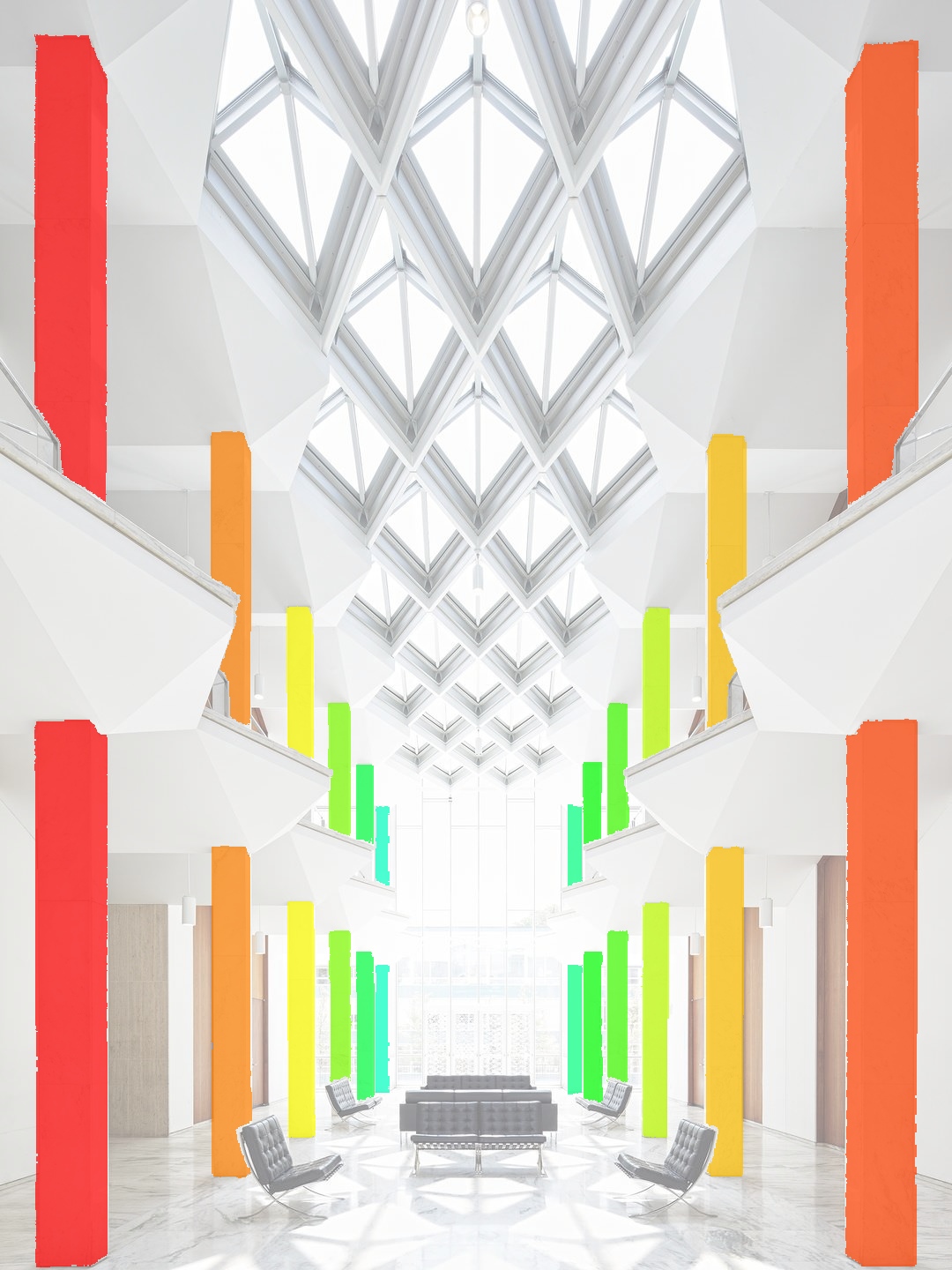}
  \\[-4pt]
  \leftline{\textbf{(\ref*{fig:saco_qualitative_failures}.g)} ``\textit{the column}``%
    \hfill%
    \makebox[\sacoimgw][c]{\tiny F1=0.85 (GT c), 24 masks}%
    \hspace{5pt}%
    \makebox[\sacoimgw][c]{\tiny F1=0.08 (GT c), 12 masks}}
  \caption{\textbf{Failure cases.}
  Qualitative evaluation of instance segmentation results on SA-Co/Gold.
  For each example, we display the noun-phrase query and the original image, followed by visualizations of the three independent ground truth annotations, the prediction from SAM 3, and the prediction from Vision Banana.
  Under each prediction visualization, we report the best F1 score along with its associated GT and the number of masks in the prediction.
  Best seen zoomed-in on screen.
  }
  \label{fig:saco_qualitative_failures}
\end{figure*}

In \cref{fig:saco_qualitative_successes} and \cref{fig:saco_qualitative_failures}, we present a qualitative comparison of instance segmentation masks predicted by the SAM 3 specialist and our zero-shot Vision Banana model, on examples from the SA-Co/Gold \citep{sam3} benchmark.
Below each prediction, we report its best $F_1$ score across the three independent ground-truth annotations, computed as the average of the $F_1$ scores across the ten IoU thresholds defined by the SA-Co benchmark.

We illustrate several successful segmentation results in \cref{fig:saco_qualitative_successes}.
For instance, in examples (\ref{fig:saco_qualitative_successes}.a), (\ref{fig:saco_qualitative_successes}.b), and (\ref{fig:saco_qualitative_successes}.c), Vision Banana correctly segments 17 instances of 'manicured and pedicured nail', 3 instances of 'incandescent lightbulb', and 1 instance of 'hand-drawn design', respectively.
Despite producing visually accurate segmentation masks, our quantitative $F_1$ scores can remain low, falling significantly below those of SAM 3. 
We found out that this can be explained by misalignments with the SAM-annotated ground-truth masks, which heavily penalize Vision Banana performance at high IoU thresholds. 
Specifically, the masks produced by our model are usually slightly smaller (see example (\ref{fig:saco_qualitative_successes}.a)) and finer (see example (\ref{fig:saco_qualitative_successes}.c)) than the SAM-produced masks.

Compared to discriminative models like SAM 3, which suffer from 'mode-averaging' artifacts under spatial ambiguity (see examples (\ref{fig:saco_qualitative_successes}.b), (\ref{fig:saco_qualitative_successes}.c) and (\ref{fig:saco_qualitative_failures}.f)), Vision Banana naturally resolves such ambiguity by committing to a single, coherent mode of the target mask distribution.
Furthermore, unlike SAM 3, our model consistently produces non-overlapping masks and does not leave unnatural gaps between adjacent object boundaries (see example (\ref{fig:saco_qualitative_successes}.f)).

In \cref{fig:saco_qualitative_failures}, we report failure modes of our model.
These primarily manifest as: (i) incorrectly merging instances together (e.g., merging wall segments in (\ref{fig:saco_qualitative_failures}.a), all post boxes in example (\ref{fig:saco_qualitative_failures}.c), furniture in (\ref{fig:saco_qualitative_failures}.e), or top and down columns in (\ref{fig:saco_qualitative_failures}.g)), and (ii) breaking cohesive ``group instances'' into multiple masks (e.g., segmenting the crowd into individual people in example (\ref{fig:saco_qualitative_failures}.d) despite the query requesting ``the crowd'').
Additionally, Vision Banana occasionally misses target objects entirely, such as background walls in example (\ref{fig:saco_qualitative_failures}.a) or the persons on the left in example (\ref{fig:saco_qualitative_failures}.d).

This qualitative analysis reveals that our model still has clear room for improvement, particularly in highly cluttered or crowded scenes.
Nonetheless, a portion of the quantitative performance gap between Vision Banana and SAM 3 is also attributable to the specific annotation biases of the SA-Co dataset, our model being evaluated zero-shot.

\section{Image generation and editing capabilities}
\label{sec:generation_capabilities}
\begin{figure}[!h]
    \centering
    \begin{subfigure}{0.48\textwidth}
        \centering
        \includegraphics[width=\textwidth]{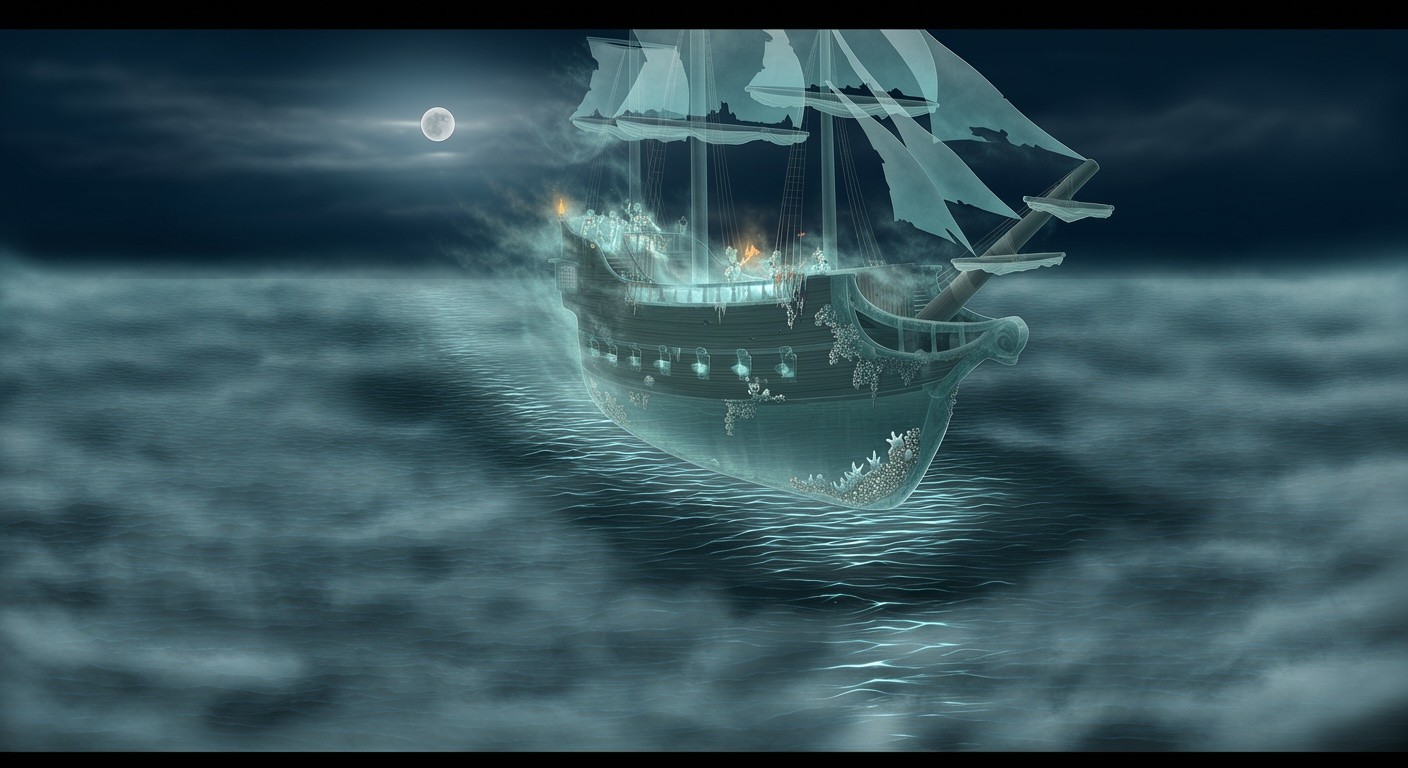}
        \caption*{\textit{A ghostly ship sailing on a fog-shrouded, moonlit sea.}}
    \end{subfigure}
     \begin{subfigure}{0.48\textwidth}
        \centering
        \includegraphics[width=\textwidth]{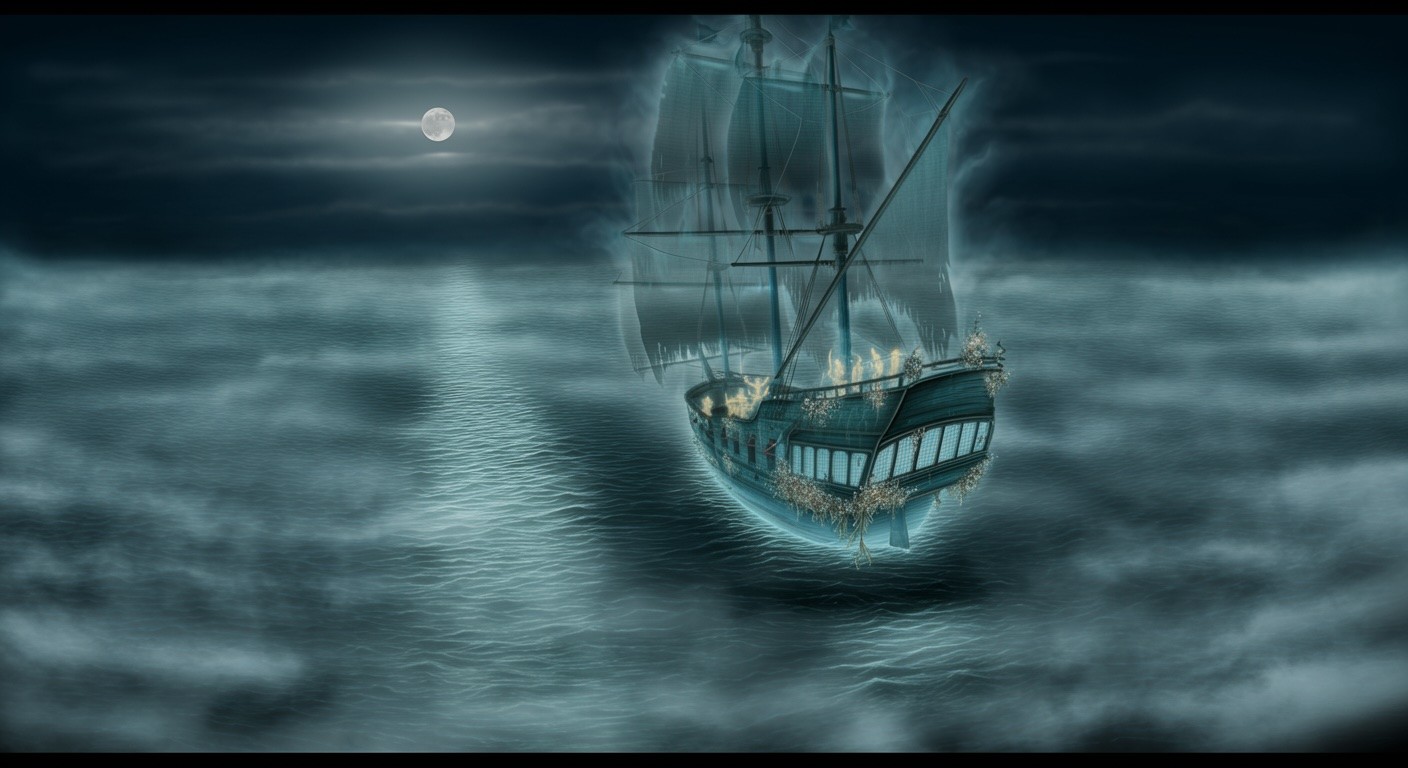}
        \caption*{ }
    \end{subfigure}

    \begin{subfigure}{0.48\textwidth}
        \centering
        \includegraphics[width=\textwidth]{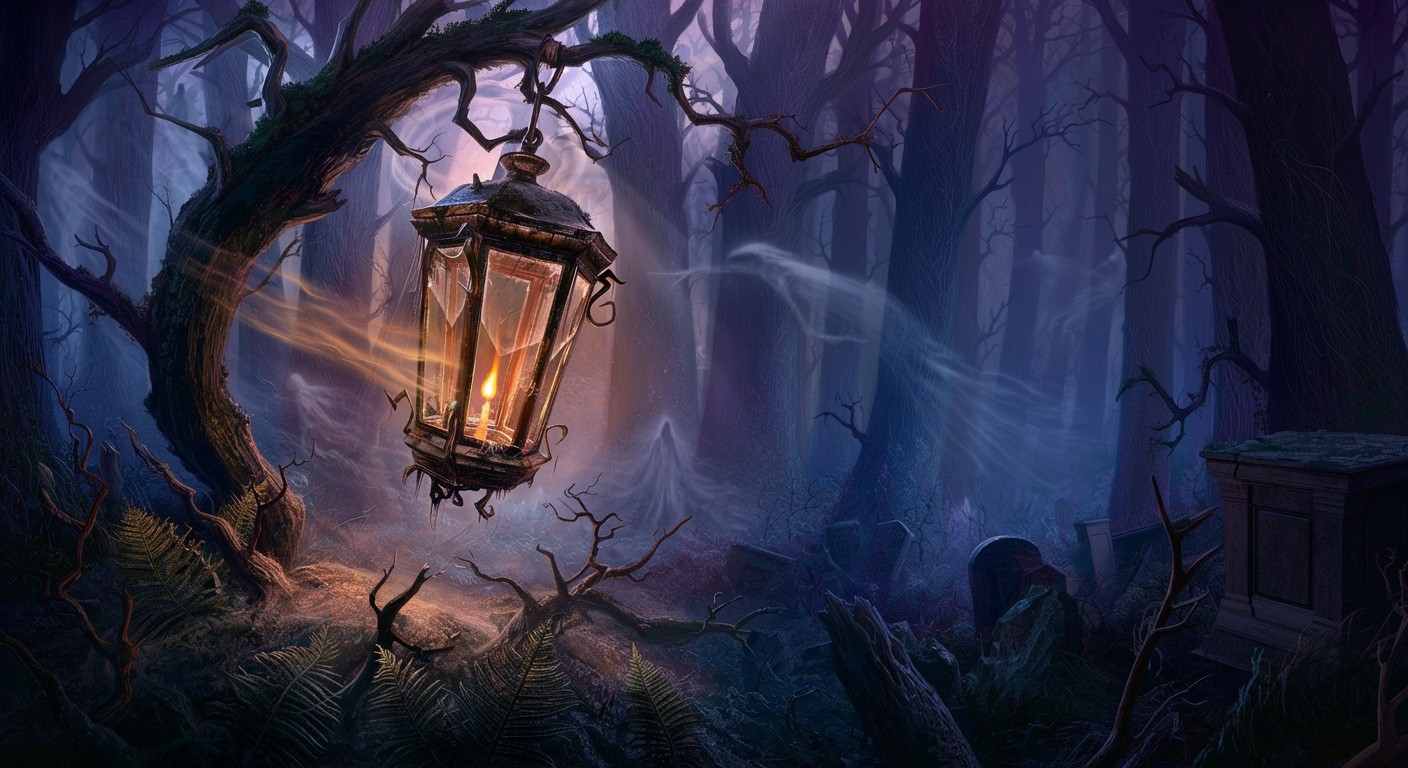}
        \caption*{\textit{A lantern casting dim light in a haunted forest.}}
    \end{subfigure}
     \begin{subfigure}{0.48\textwidth}
        \centering
        \includegraphics[width=\textwidth]{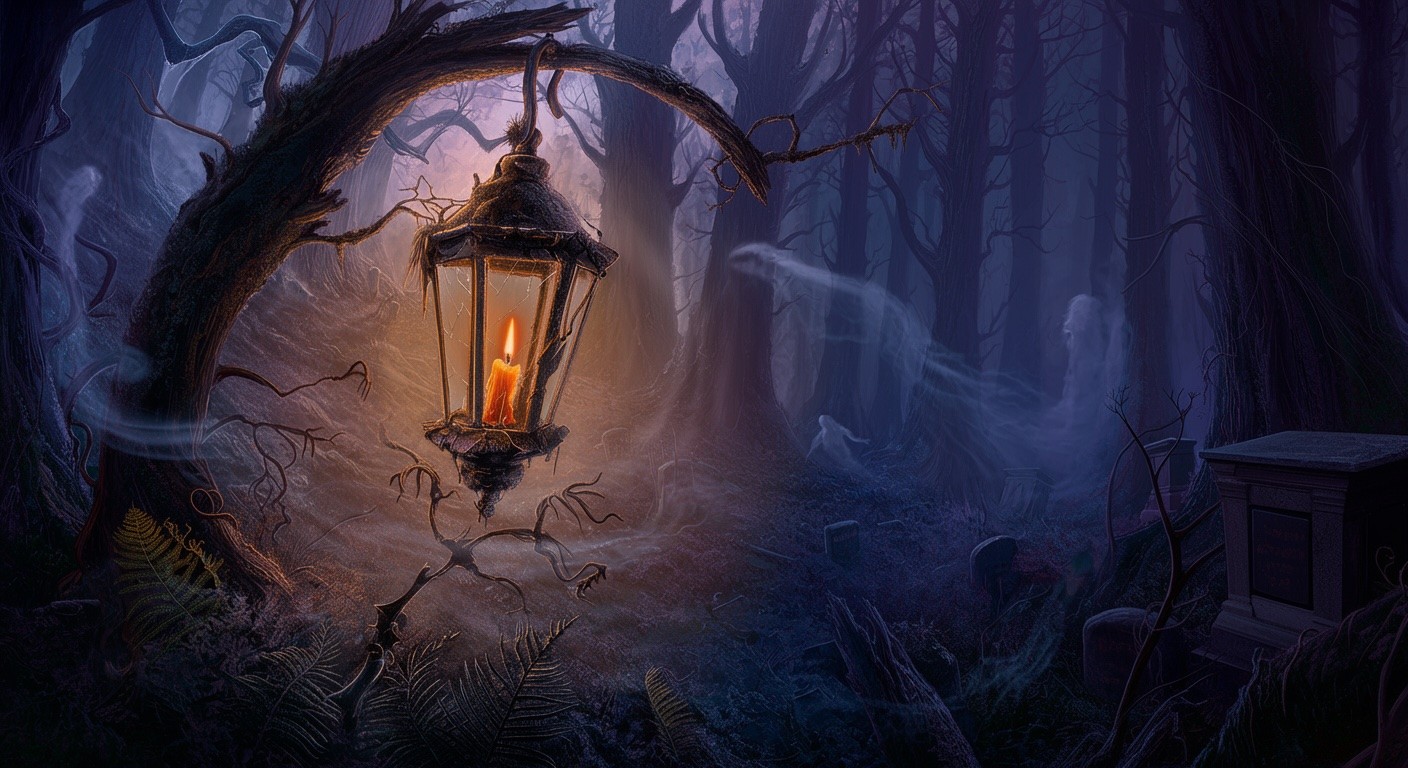}
        \caption*{ }
    \end{subfigure}
    
    \begin{subfigure}{0.48\textwidth}
        \centering
        \includegraphics[width=\textwidth]{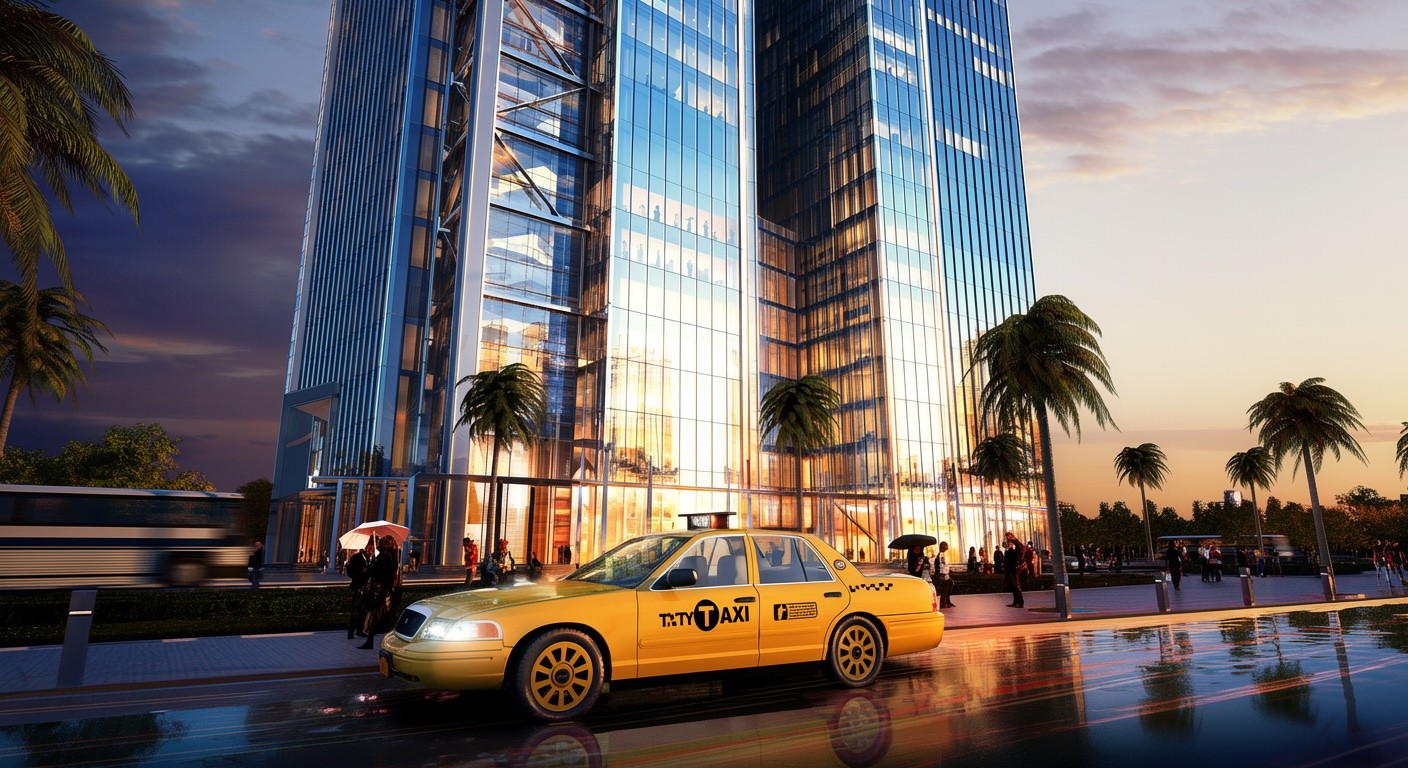}
        \caption*{\textit{A yellow taxi waiting outside a modern glass building.}}
    \end{subfigure}
     \begin{subfigure}{0.48\textwidth}
        \centering
        \includegraphics[width=\textwidth]{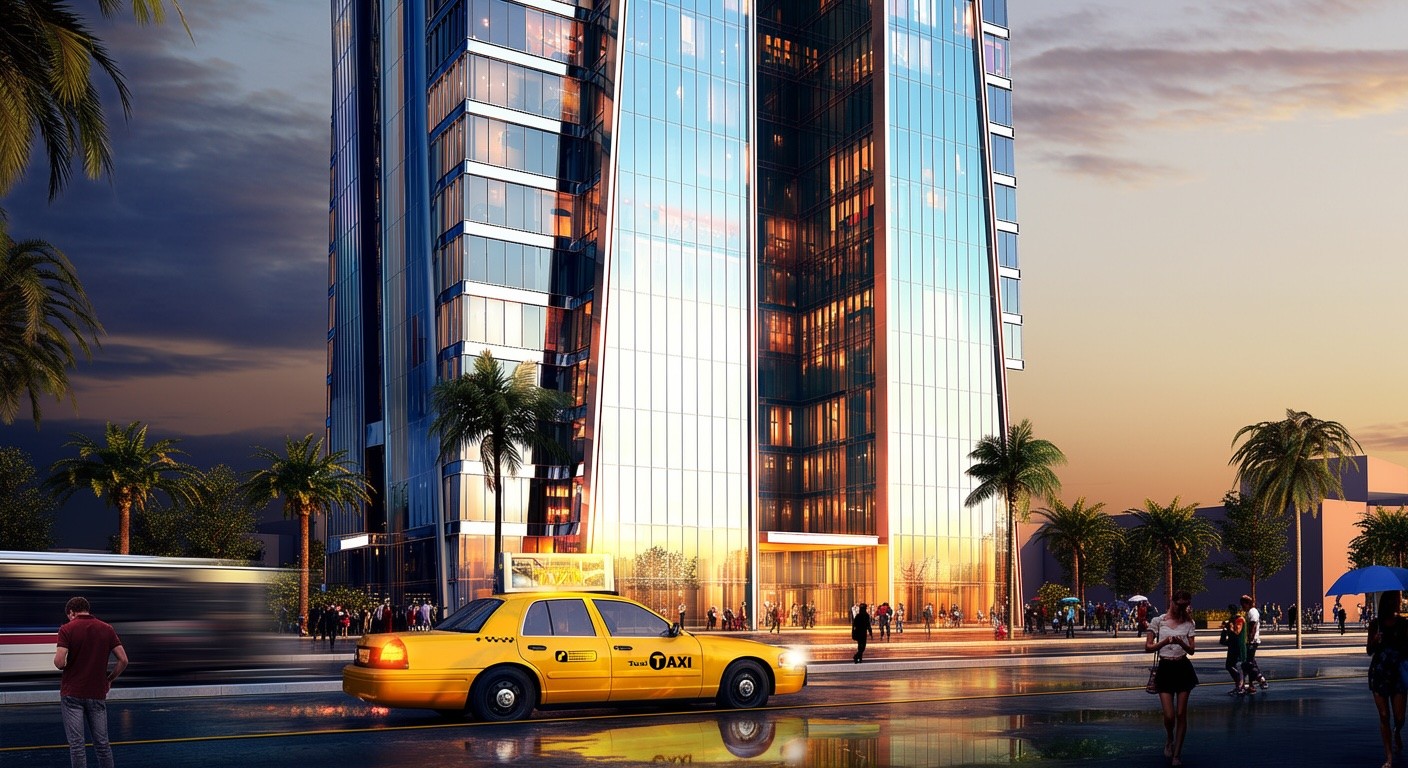}
        \caption*{ }
    \end{subfigure}

    \begin{subfigure}{0.48\textwidth}
        \centering
        \includegraphics[width=\textwidth]{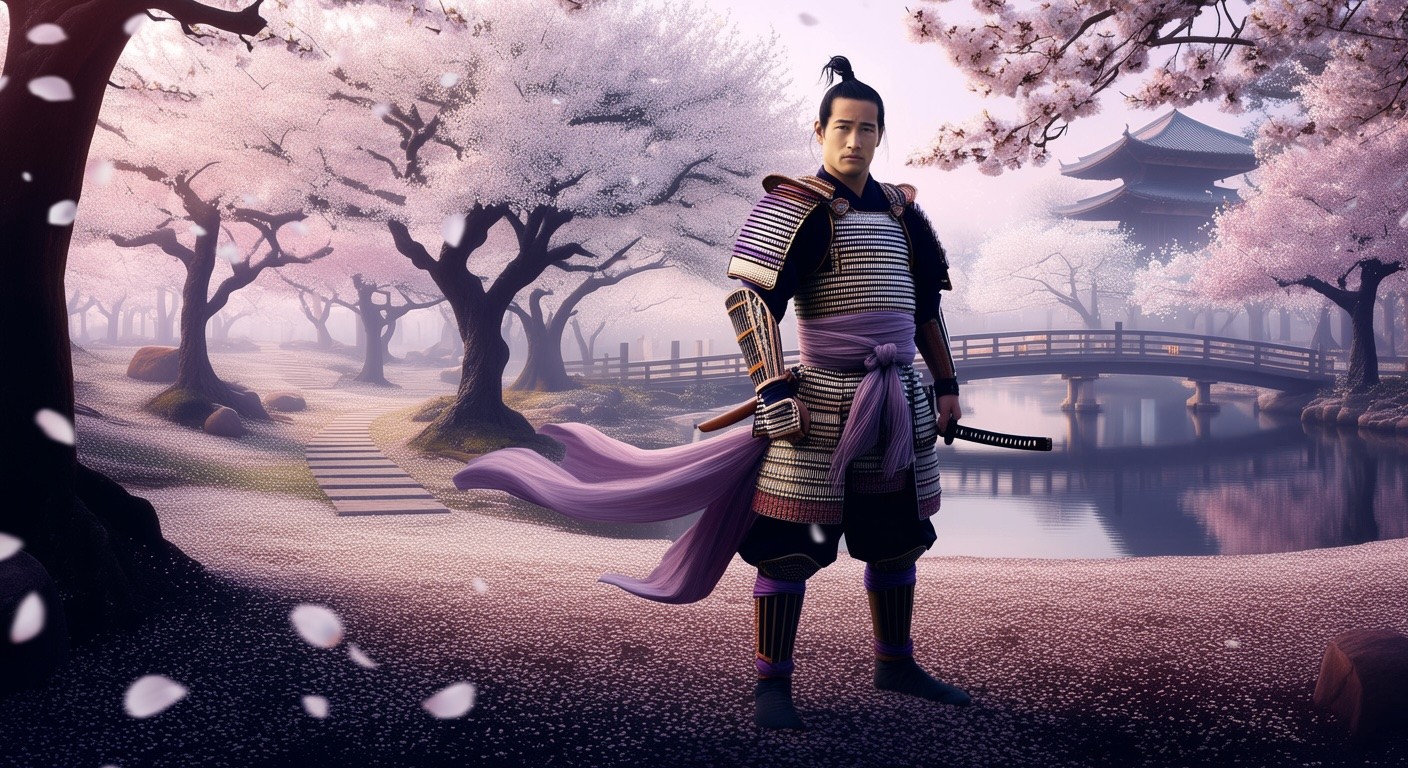}
        \caption*{\textit{A samurai with a silk sash in a cherry blossom garden.}}
    \end{subfigure}
     \begin{subfigure}{0.48\textwidth}
        \centering
        \includegraphics[width=\textwidth]{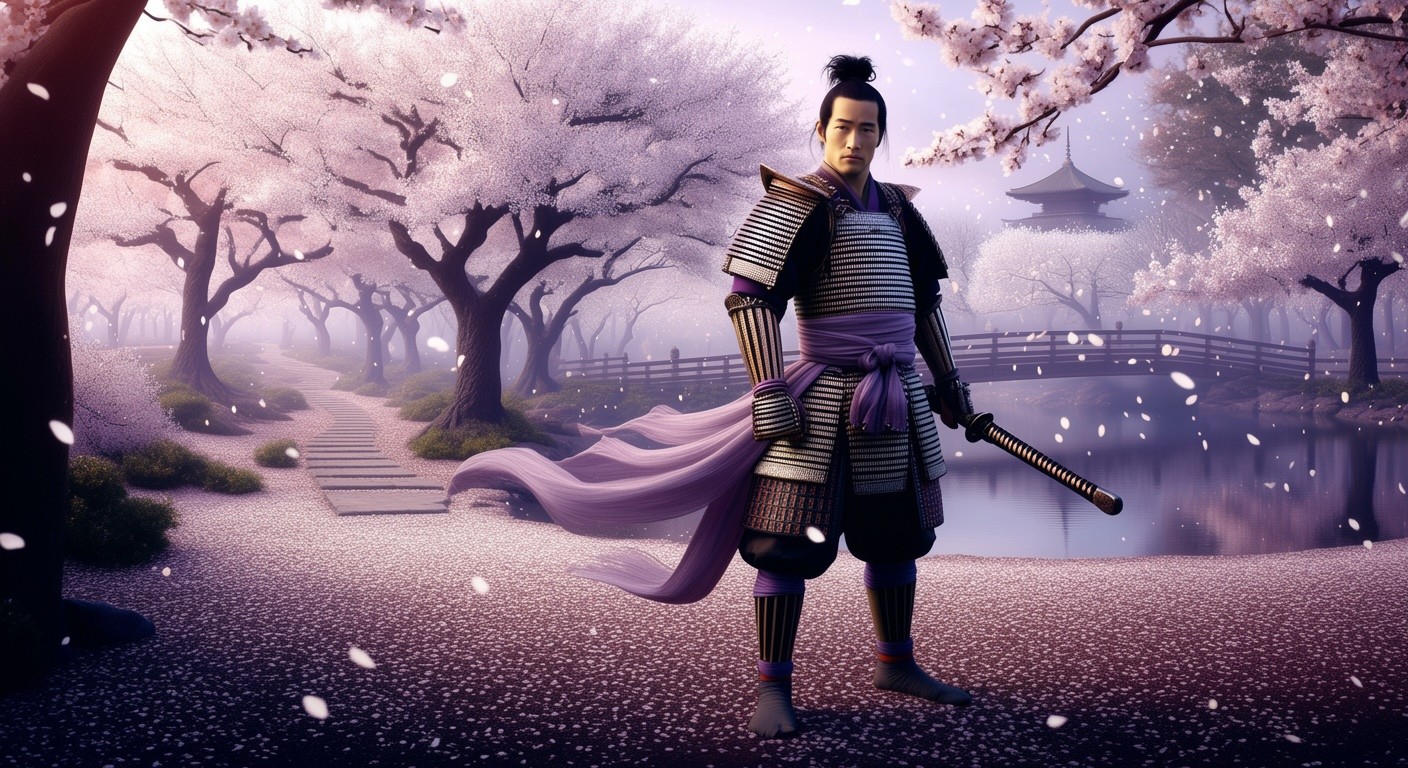}
        \caption*{ }
    \end{subfigure}
    
    \caption{Comparing Vision Banana (left) and Nano Banana Pro (right) on text-to-image generation.
    Prompts are sampled from GenAI-Bench \citep{li2024genaibench}.
    Results verify that Vision Banana does not forget its generative features during the instruction-tuning.
    }
    \label{fig:t2i-demo}
\end{figure}

\begin{figure}[!t]
    \centering
        \begin{tabular}{cccc}

        Original image & Vision Banana \emojivb & Nano Banana Pro \\

        \includegraphics[width=0.32\textwidth]{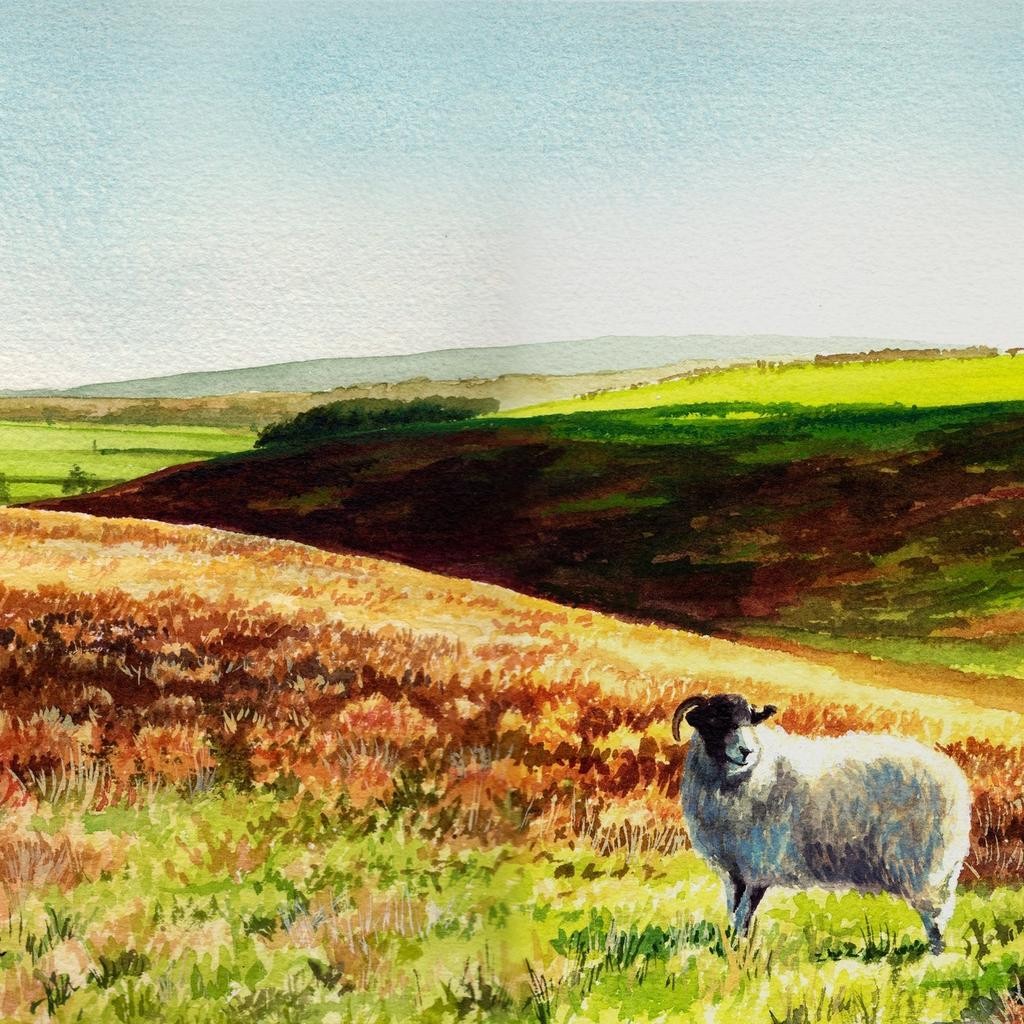} &
        \includegraphics[width=0.32\textwidth]{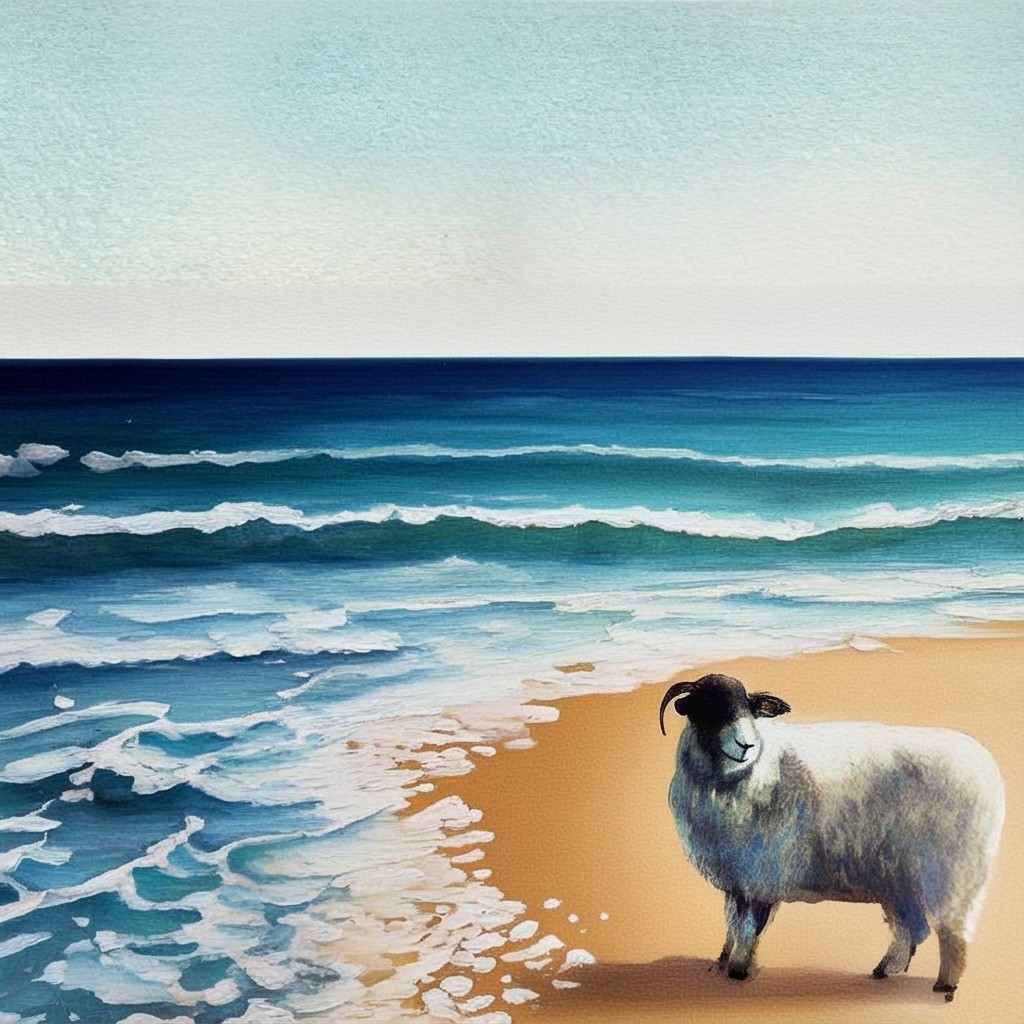} &
        \includegraphics[width=0.32\textwidth]{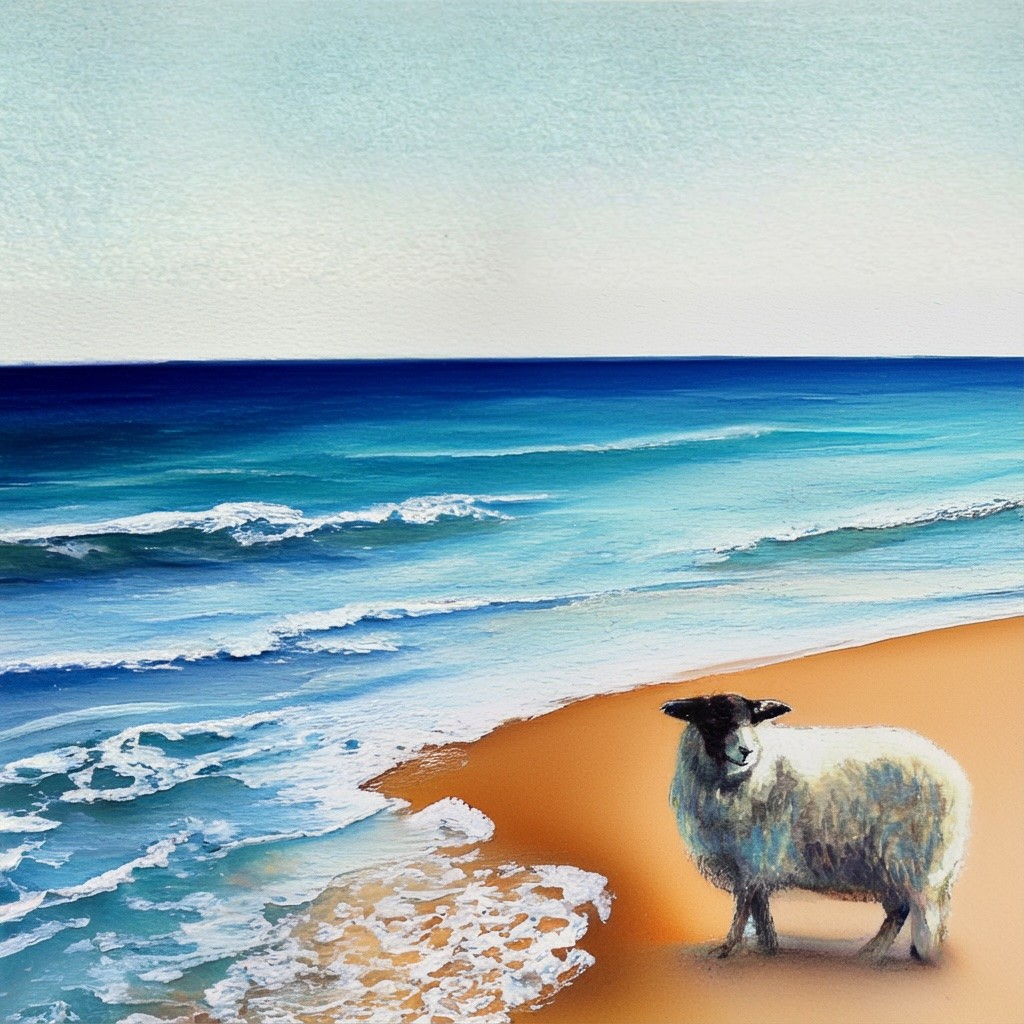} \\
        \multicolumn{3}{l}{\textit{Change the grassy hills in the picture to a beach with ocean waves.}}  \\

        \includegraphics[width=0.32\textwidth]{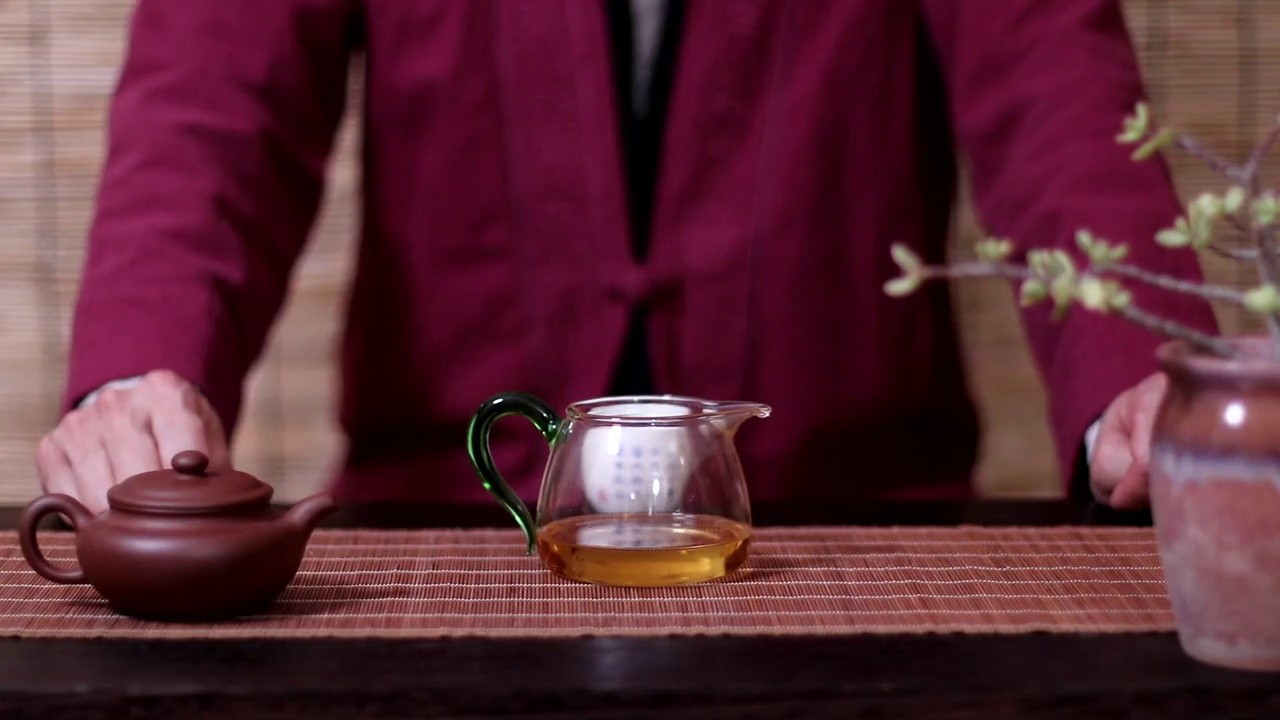} &
        \includegraphics[width=0.32\textwidth]{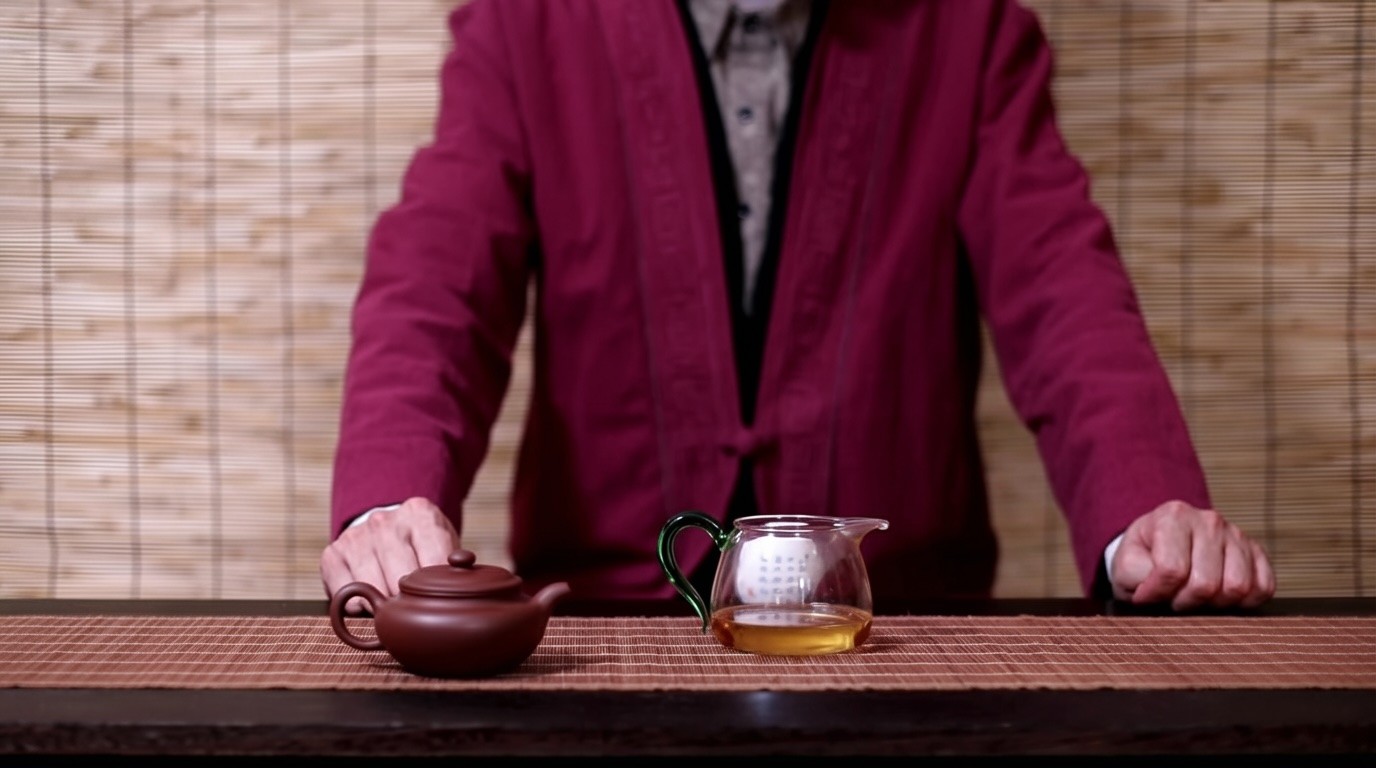} &
        \includegraphics[width=0.32\textwidth]{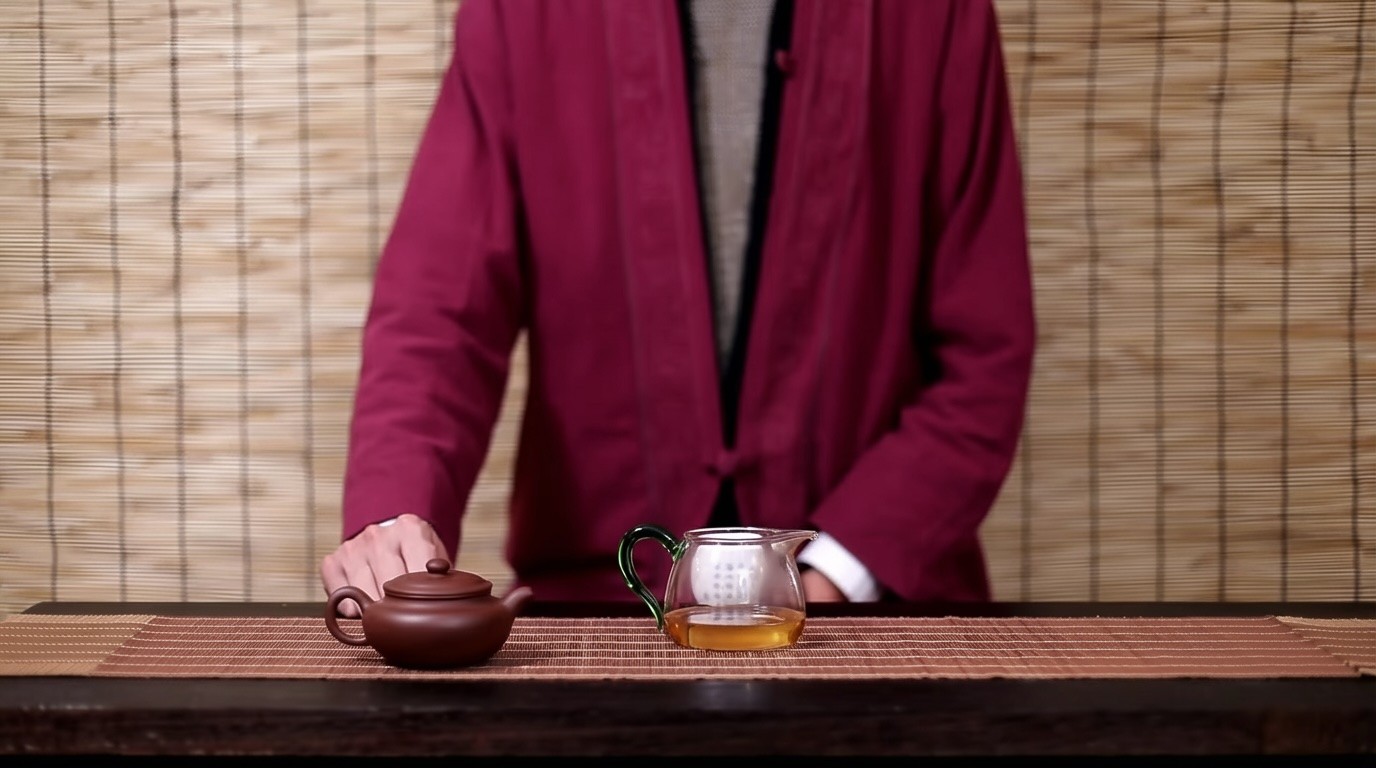} \\
        \multicolumn{3}{l}{\textit{Remove the plant from the shelf, and resize the picture frame to be larger.}}  \\

        \includegraphics[width=0.32\textwidth]{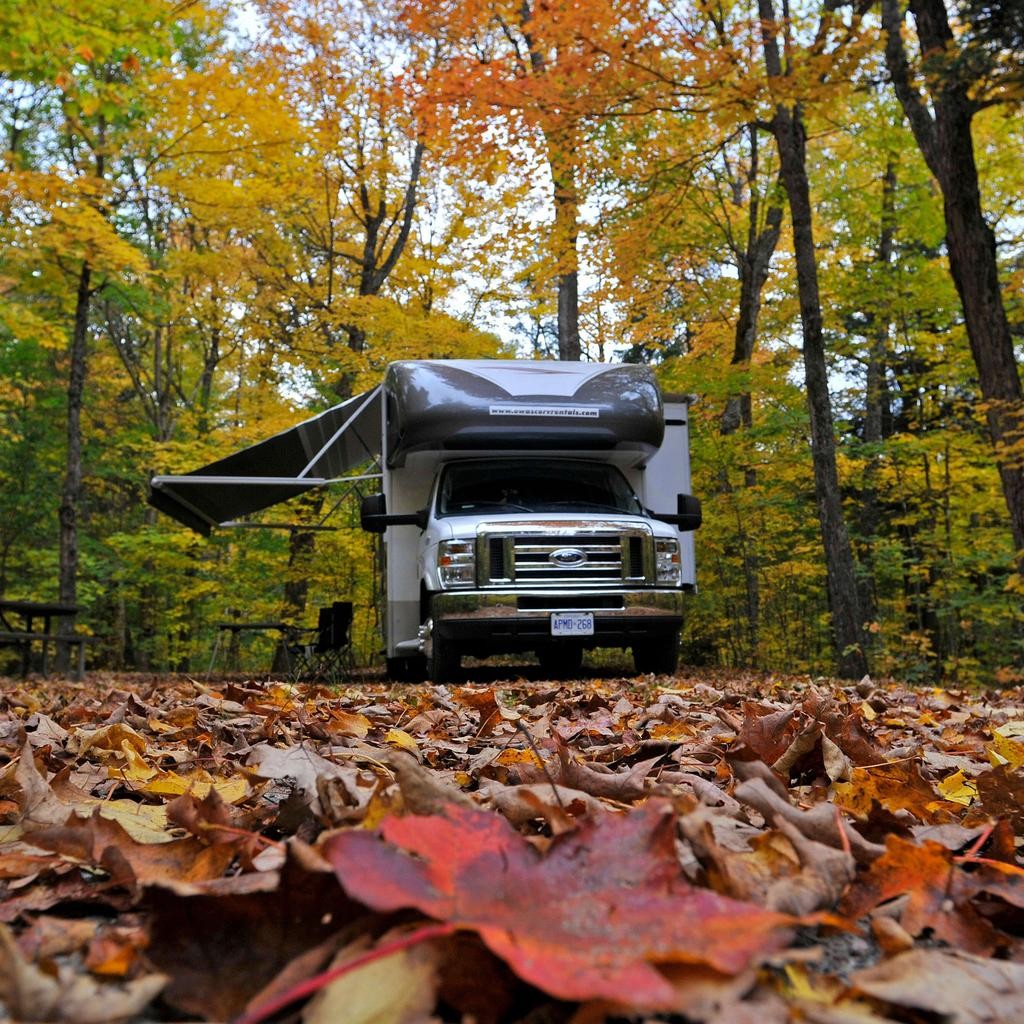} &
        \includegraphics[width=0.32\textwidth]{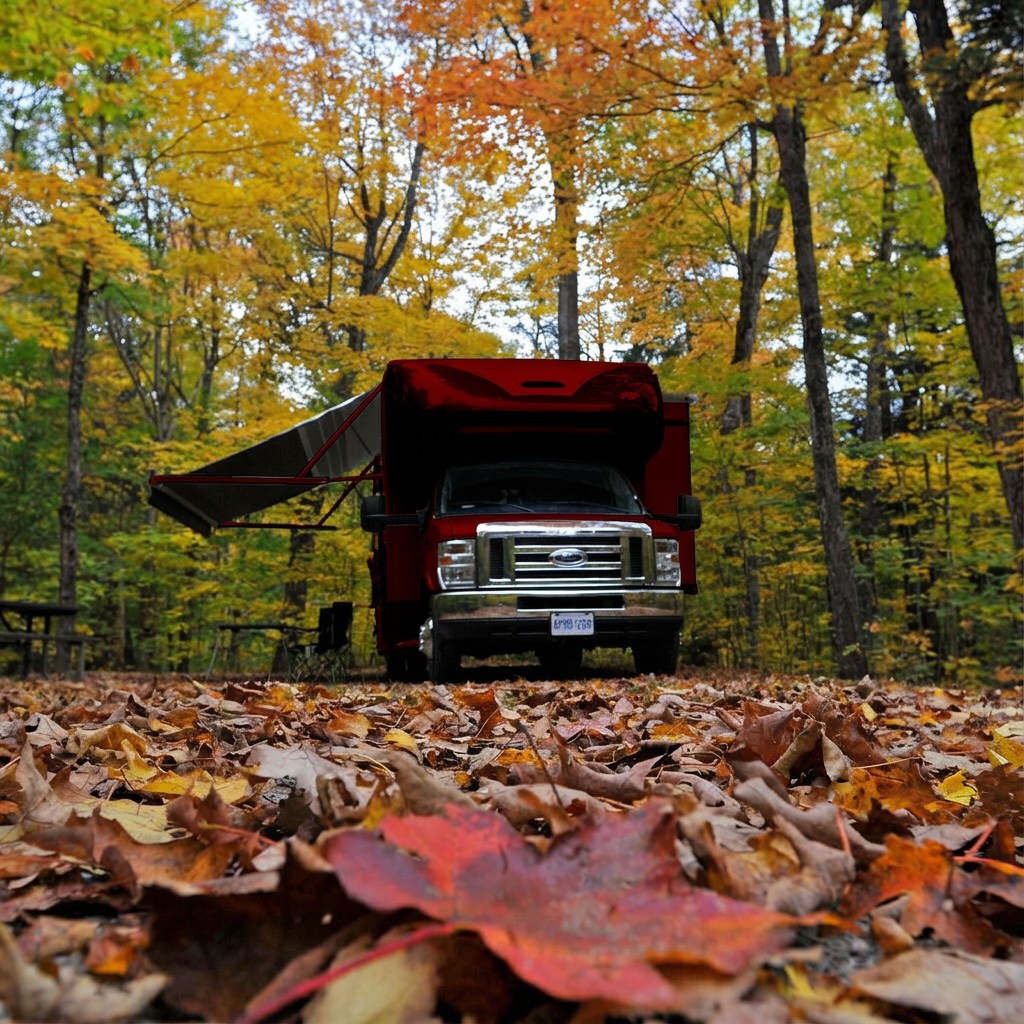} &
        \includegraphics[width=0.32\textwidth]{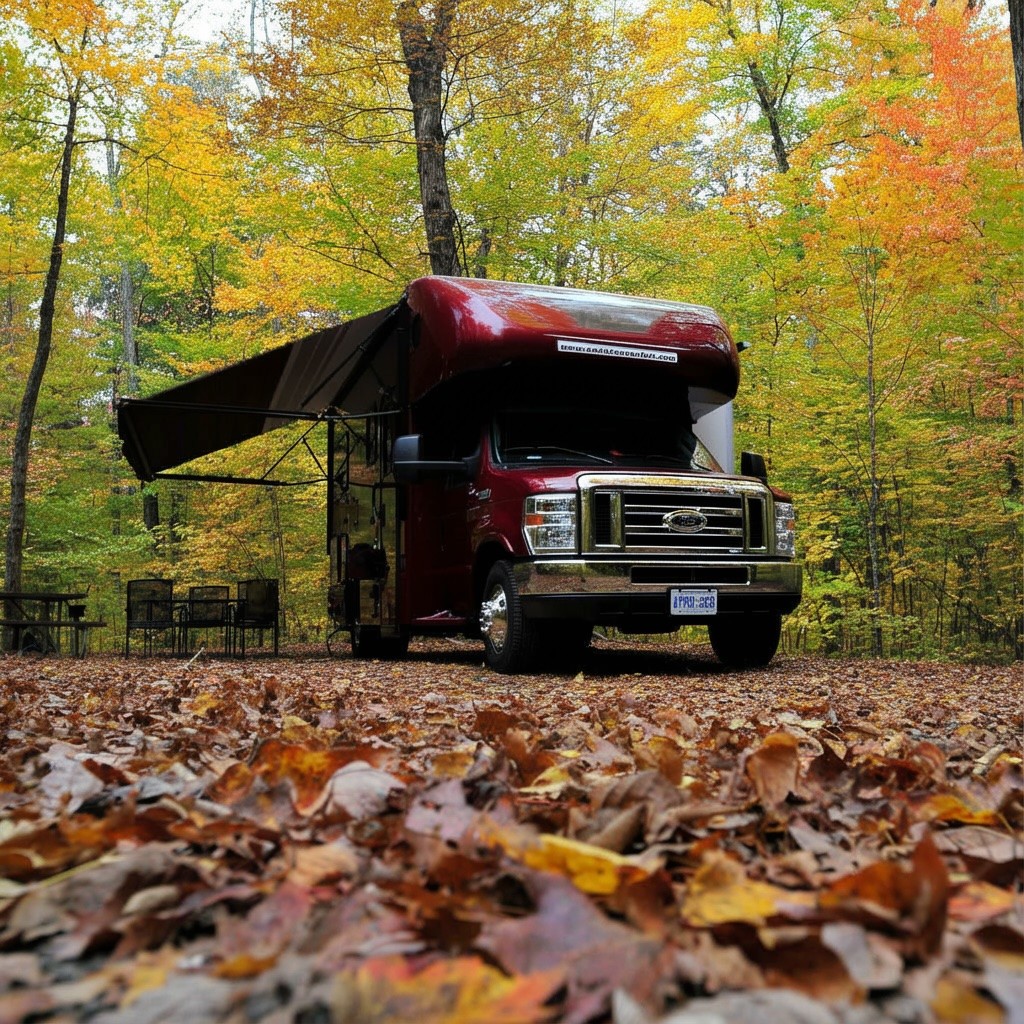} \\
        \multicolumn{3}{l}{\textit{Change the vehicle's color to red.}}  \\

        \includegraphics[width=0.32\textwidth]{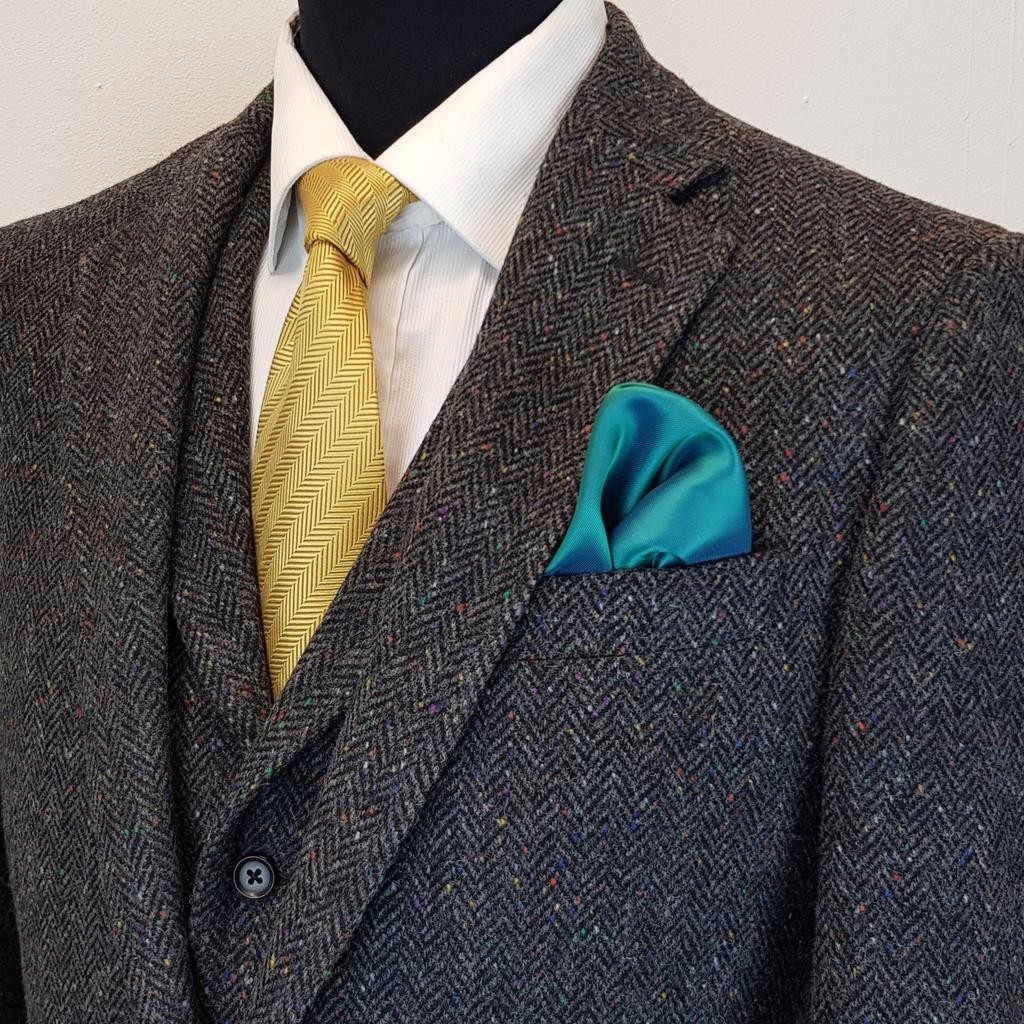} &
        \includegraphics[width=0.32\textwidth]{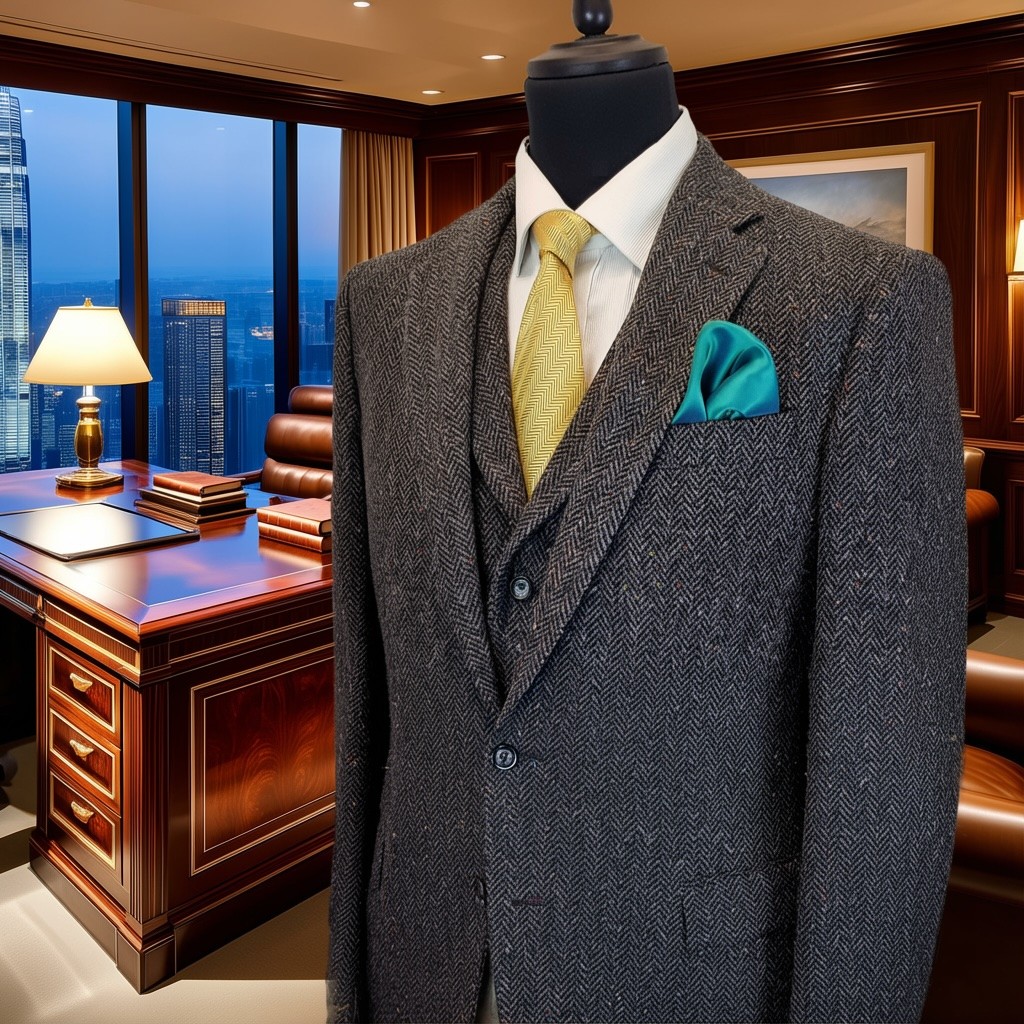} &
        \includegraphics[width=0.32\textwidth]{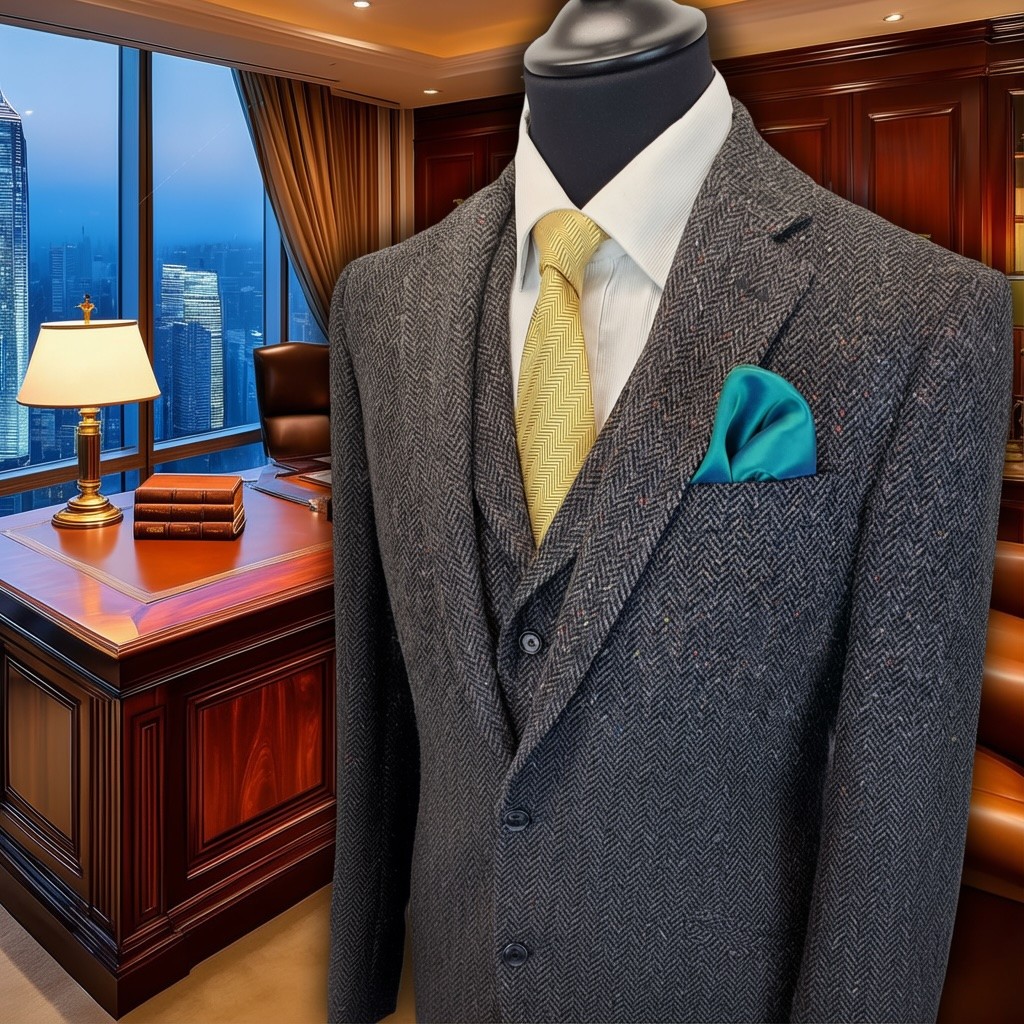} \\
        \multicolumn{3}{l}{\makecell[l]{
        \textit{Change the background of the suit from a blank wall to a luxurious office setting that includes} \\  \textit{a wooden desk and a large window showing a cityscape.}
        }}  \\

    \end{tabular}

    \caption{Comparing Vision Banana (left) and Nano Banana Pro (right) on image-editing.
    Prompts are sampled from ImgEdit \citep{ye2025imgedit}.
    }
    \label{fig:imgedit-demo}
\end{figure}

Finally, we assess the generative capabilities of Vision Banana post-instruction-tuning to ensure that fine-tuning the model did not trigger catastrophic forgetting of its core generative features.
Figures~\ref{fig:t2i-demo} and~\ref{fig:imgedit-demo} compare Vision Banana with its base model, Nano Banana Pro, on text-to-image generation and image-editing tasks.

In Fig.~\ref{fig:t2i-demo}, we present generated outputs for prompts sampled from GenAI-Bench~\citep{li2024genaibench}.
These results demonstrate that Vision Banana continues to produce detailed and contextually accurate images with a similar quality to those of the base generative model.

Furthermore, in Fig.~\ref{fig:imgedit-demo}, we evaluate the two checkpoints on instruction-based image-editing tasks using prompts from ImgEdit~\citep{ye2025imgedit}.
Vision Banana shows robust adherence to detailed instructions, confirming that our instruction-tuning successfully teaches output formats for dense vision tasks without degrading the image editing capabilities learned during pre-training.

\end{document}